\UseRawInputEncoding
\pdfoutput=1

\documentclass{article}

\usepackage{neurips_2024}  

\usepackage[utf8]{inputenc} 
\usepackage[T1]{fontenc}    
\usepackage{hyperref}       
\usepackage{url}            
\usepackage{booktabs}       
\usepackage{amsfonts}       
\usepackage{nicefrac}       
\usepackage{microtype}      
\usepackage{xcolor}         

\usepackage{graphicx} 

\usepackage{pifont}
\newcommand{\cmark}{\ding{51}}%
\newcommand{\xmark}{\ding{55}}%

\def\ie{{\em i.e.}}
\def\eg{{\em e.g.}}
\def\etal{{\em et al.}}
\def\etc{{\em etc}}

\usepackage{tablefootnote}
\newcommand{\tabincell}[2]{\begin{tabular}{@{}#1@{}}#2\end{tabular}}

\usepackage{wrapfig}
\usepackage{makecell}
\usepackage{amsthm,amsmath,amssymb}

\usepackage{mathrsfs}

\usepackage{arydshln} 

\usepackage{booktabs}

\usepackage{multirow}
\usepackage {subfig}
\usepackage {caption}

\makeatletter

\newcommand{\Rmnum}[1]{\expandafter\@slowromancap\romannumeral #1@}
\makeatother

\newcommand{\myPara}[1]{\vspace{-0.1cm}\noindent\textbf{#1}}

\usepackage{amsfonts}

\usepackage{caption}
 \usepackage{threeparttable}
\usepackage{tabularx}
\usepackage{longtable}
\usepackage{lipsum}
\usepackage{supertabular}
\usepackage{geometry}
\usepackage{algorithm}
\usepackage{algorithmic}

\usepackage{float}
\usepackage{subcaption}
\usepackage{enumerate}

\title{WebUOT-1M: Advancing Deep Underwater Object Tracking with A Million-Scale Benchmark}

\author{%
Chunhui~Zhang$^{1,2,3}$,~Li~Liu$^{2}$\thanks{Corresponding author.  E-mail: avrillliu@hkust-gz.edu.cn.},~Guanjie~Huang$^{2}$,~Hao~Wen$^{3}$,~Xi~Zhou$^{3}$,~Yanfeng~Wang$^{1,4}$ \\
  $^{1}$ Cooperative~Medianet~Innovation~Center,~Shanghai~Jiao~Tong University,~Shanghai~200240,~China\\
  $^{2}$ The~Hong~Kong~University~of~Science~and~Technology (Guangzhou),~Guangzhou~511458,~China\\
  $^{3}$ CloudWalk~Technology~Co.,~Ltd,~201203,~China\\
  $^{4}$ Shanghai~AI~Laboratory,~Shanghai~200232,~China\\
}

\begin{document}

\maketitle

\vspace{-0.7cm}

\begin{abstract}
\vspace{-0.3cm}

Underwater object tracking (UOT) is a foundational task for identifying and tracing submerged entities in underwater video sequences. However, current UOT datasets suffer from limitations in scale, diversity of target categories and scenarios covered, hindering the training and evaluation of modern tracking algorithms. To bridge this gap, we take the first step and introduce WebUOT-1M, \ie, the largest public UOT benchmark to date, sourced from complex and realistic underwater environments. It comprises 1.1 million frames across 1,500 video clips filtered from 408 target categories, largely surpassing previous UOT datasets, \eg, UVOT400. Through meticulous manual annotation and verification, we provide high-quality bounding boxes for underwater targets. Additionally, WebUOT-1M includes language prompts for video sequences, expanding its application areas, \eg, underwater vision-language tracking. Most existing trackers are tailored for open-air environments, leading to performance degradation when applied to UOT due to domain gaps. Retraining and fine-tuning these trackers are challenging due to sample imbalances and limited real-world underwater datasets. To tackle these challenges, we propose a novel omni-knowledge distillation framework based on WebUOT-1M, incorporating various strategies to guide the learning of the student Transformer. To the best of our knowledge, this framework is the first to effectively transfer open-air domain knowledge to the UOT model through knowledge distillation, as demonstrated by results on both existing UOT datasets and the newly proposed WebUOT-1M. Furthermore, we comprehensively evaluate WebUOT-1M using 30 deep trackers, showcasing its value as a benchmark for UOT research by presenting new challenges and opportunities for future studies. The complete dataset, codes and tracking results, will be made publicly available at https://github.com/983632847/Awesome-Multimodal-Object-Tracking.


\end{abstract}

\vspace{-0.5cm}
\section{Introduction}
\label{sec:introduction}
\vspace{-0.3cm}

\begin{wrapfigure}{r}{5.0cm}
\vspace{-0.6cm}
  \centering
  \includegraphics[width=1.0\linewidth]{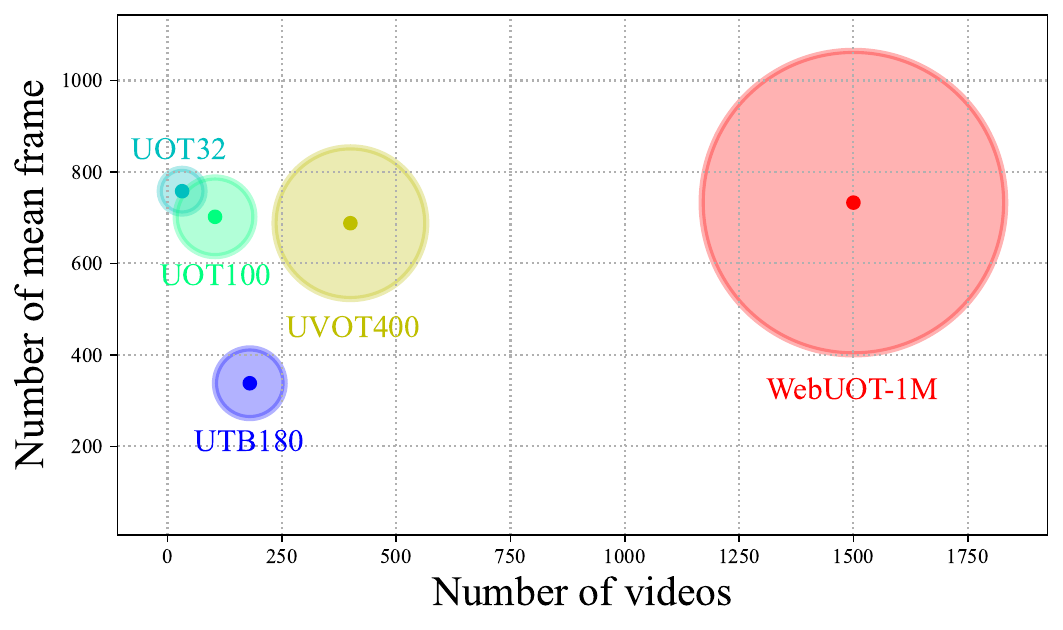}
  \caption{The proposed WebUOT-1M is much larger than existing UOT benchmarks~\cite{kezebou2019underwater,panetta2021comprehensive,alawode2022utb180,alawode2023improving}.}
  \label{fig:motivation}
  \vspace{-0.3cm}
\end{wrapfigure} 
Underwater object tracking (UOT) refers to the task of sequentially locating a submerged instance in an underwater video, given its initial position in the first frame~\cite{kezebou2019underwater,panetta2021comprehensive,cai2023semi,alawode2023improving}. As a fundamental task in computer vision, it has a wide range of applications, such as marine animal conservation~\cite{cai2023semi}, underwater search and rescue~\cite{alawode2023improving}, underwater photography~\cite{li2023underwater}, and homeland and maritime security~\cite{gonzalez2023survey}. The underwater environment usually exhibits uneven lighting conditions, low visibility, low contrast, watercolor variations, similar distractors, camouflage and target blurring, posing distinct challenges for UOT compared to traditional open-air tracking tasks~\cite{zhang2022webuav,fan2019lasot,huang2019got,wang2021towards,muller2018trackingnet}. Despite its significance, UOT has not been thoroughly explored due to the absence of large-scale datasets, benchmarks, and challenges in gathering abundant underwater videos~\cite{alawode2023improving,cai2023semi}.

\begin{figure*}[t!]
  \centering
  \vspace{-0.5cm}
  \includegraphics[width=1.0\linewidth]{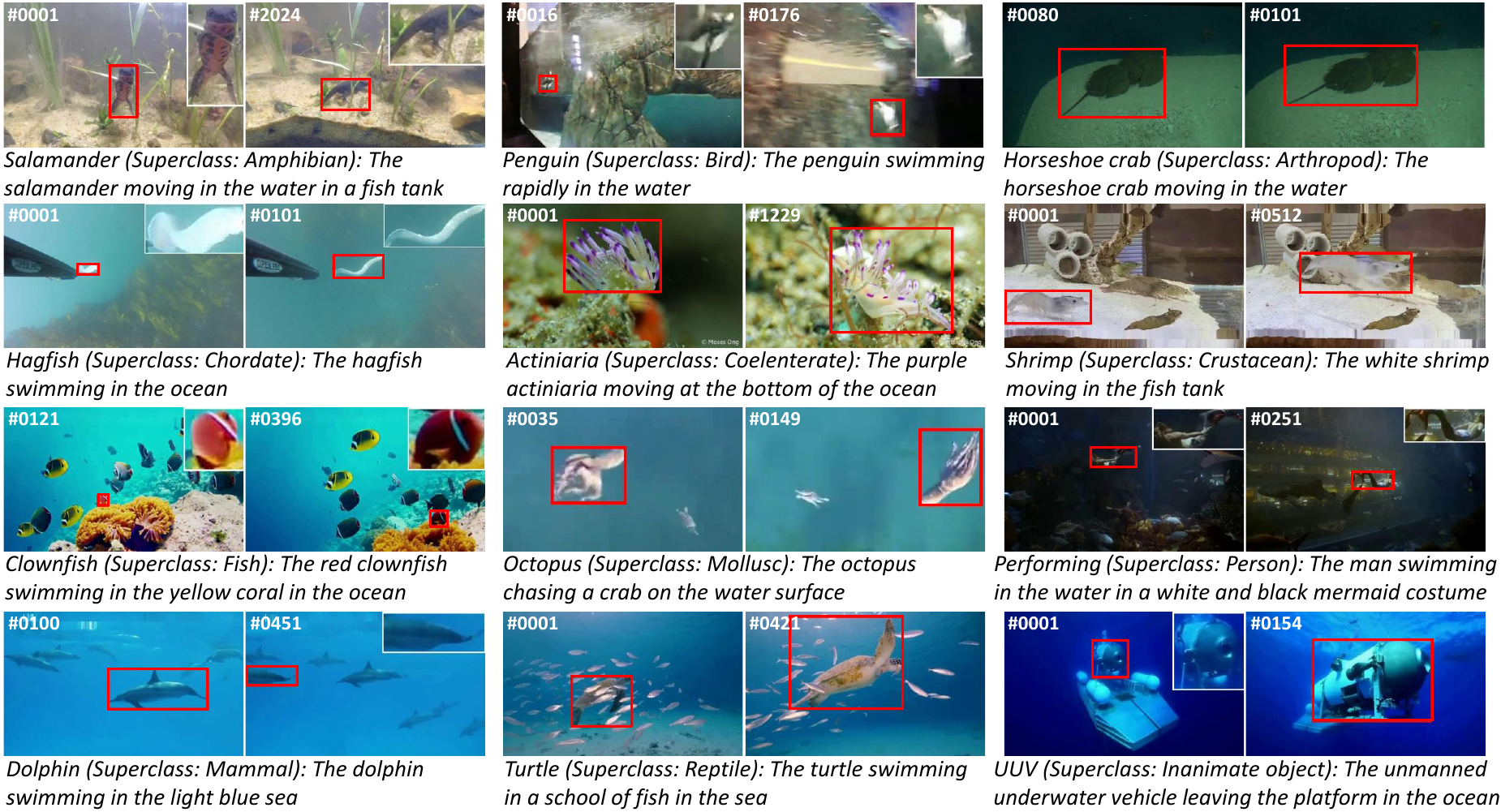}
  \caption{A glance of some video sequences and annotations from the WebUOT-1M dataset. All sequences are divided into 12 superclasses, including \emph{amphibian, arthropod, bird, chordate, coelenterate, crustacean, fish, mollusc, person, mammal (except humans), reptile, and inanimate object}.}
  \label{fig:examples}
  \vspace{-0.5cm}
\end{figure*}

Recently, some efforts have been made to build UOT datasets to promote research in this field. Early works focus on specific underwater tasks (\eg, tracking marine plastic waste~\cite{singh2021marine}), underwater environments (\eg, coral reef~\cite{fisher2016fish4knowledge}), and specific marine species (\eg, zebrafish~\cite{pedersen20203d} and ocean mammal~\cite{karnowski2015dolphin,laplanche2015tracking}). These datasets do help advance research on enhancing the tracking and monitoring of relevant marine species. \textbf{Due to the huge appearance variation and behavioral differences among various marine animals, models trained on these early datasets struggle with unseen species, leading to poor generalization performance.} To further facilitate research on UOT, datasets covering multiple species are proposed, \eg, UTB180~\cite{alawode2022utb180} and UVOT400~\cite{alawode2023improving}. However, these datasets are still limited in terms of their dataset volume, diversity in animal species and scenarios covered due to severe
challenges in underwater video collection and annotation.

To fill this gap, we propose WebUOT-1M, the first million-scale dataset for UOT. As shown in Figs.~\ref{fig:motivation} and~\ref{fig:examples}, the WebUOT-1M is much larger than existing datasets and comprises abundant categories, diverse underwater scenarios, and rich annotations. WebUOT-1M comprises 1.1 million frames with precise bounding box annotations across 1,500 underwater videos and 408 highly diverse target categories (see Tab.~\ref{tab:Comp_WebUOT_1M}, Fig.~\ref{fig:Group_of_object_classes}). These targets are further classed into 12 superclasses with reference to WordNet~\cite{miller1995wordnet} to facilitate the evaluation of the cross-superclass generalization ability of tracking models. Most of the video clips are collected from YouTube\footnotemark[1] and BiliBili\footnotemark[2] with carefully filtering. These video websites contain massive underwater video resources. The videos are captured using different cameras, at various shooting perspectives and distances, and with different camera motion patterns. We assembled a professional labeling team to conduct data annotation. To establish a comprehensive benchmark, we offer 23 tracking attributes, \eg, low resolution, fast motion, similar distractors, underwater visibility, and watercolor variations. To explore the complementary advantages of visual and linguistic modalities, we provide a language prompt for each underwater video, which can facilitate the research of multi-modal UOT. To provide a baseline method for other researchers to compare, we propose a simple yet powerful omni-knowledge distillation tracking algorithm based on knowledge distillation (KD)~\cite{hinton2015distilling} and contrastive learning (CL)~\cite{oord2018representation}.

\footnotetext[1]{https://www.youtube.com/~~~~~$^{2}$https://www.bilibili.com/}


The main contribution of this work is three-fold. 1)  We introduce WebUOT-1M, the first million-scale benchmark dataset featuring diverse underwater video sequences, essential for offering a dedicated platform for the development and evaluation of UOT algorithms. 2) We propose a simple yet strong \textbf{O}mni-\textbf{K}nowledge distillion \textbf{Track}ing approach, termed OKTrack, for UOT. It is the first work to explore knowledge transfer from a teacher Transformer using underwater and enhanced frames to a student Transformer in the UOT aera. 3) We comprehensively benchmark the proposed approach, along with representative tracking algorithms based on CNN, CNN-Transformer, and Transformer on both the newly proposed WebUOT-1M and existing UOT datasets, obtaining some valuable insights. 

\begin{table*}[t!]
\vspace{-0.5cm}
	{\footnotesize
	\renewcommand\arraystretch{1.0}
	\caption{Comparison of WebUOT-1M with popular UOT benchmarks.}
    \vspace{-0.25cm}
	\label{tab:Comp_WebUOT_1M}
	\begin{center}
		\setlength{\tabcolsep}{0.1mm}{
			\scalebox{0.9}{
			\begin{tabular}{lcccccccccccccc}
				\Xhline{0.75pt} 
				Dataset & Year &  Videos & Classes & Attributes  &   \tabincell{c}{ Min \\ frame} & \tabincell{c}{Mean\\ frame} & \tabincell{c}{Max\\frame} & \tabincell{c}{Total\\ frames} &  \tabincell{c}{Total \\duration}  &  \tabincell{c}{Absent\\ label} &\tabincell{c}{ Language \\prompt} 
                & \tabincell{c}{Data \\partition} & \tabincell{c}{Open \\source} \\
				
				\hline
    
	        	\textbf{UOT32}~\cite{kezebou2019underwater} &  2019  & 32 & - & -  & 283 & 758 & 1,573 & 24 K   & 16 min & \xmark  & \xmark & Test & Proprietary\\
	        	
	        	\textbf{UOT100}~\cite{panetta2021comprehensive} & 2022 & 104  & - & 3 &  264 & 702  & 1,764 & 74 K   & 41 min & \xmark & \xmark & Test &  Fully\\
	        	
	        	\textbf{UTB180}~\cite{alawode2022utb180} & 2022 & 180 & - & 10 & 40 & 338 & 1,226 & 58 K  & 32 min & \xmark   & \xmark & Train/Test &  Fully\\

                \textbf{VMAT}~\cite{cai2023semi} & 2023 & 33  & 17 & 13 &  438 & 2,242 &  5,550 & 74 K & 41 min & \xmark & \xmark  & Test &  Fully\\

	        	\textbf{UVOT400}~\cite{alawode2023improving} &  2023 & 400 & 50 & 17 & 40 & 688 & 3,273 & 275 K   & 2.6 hours & \xmark   & \xmark & Train/Test & Partially \\	        	
	        	
				\hline

				\textbf{WebUOT-1M}~ & 2024 &  1,500 & 408 & 23  & 49 & 733 &  9,985 & 1.1 M & 10.5 hours & \cmark  & \cmark  & Train/Test  &  Fully\\
				\Xhline{0.75pt} 
			\end{tabular}
	   }
		}
	\end{center}
    }
    \vspace{-0.5cm}
\end{table*}

\vspace{-0.3cm}
\section{Related Work}
\label{sec:related_work}
\vspace{-0.2cm}

\myPara{Open-air Object Tracking.} Open-air object tracking is an active research field in computer vision, aiming to learn a class-agnostic appearance model to estimate the state of an arbitrary object in open-air videos (\eg, ground~\cite{fan2019lasot,muller2018trackingnet,wang2021towards}, UAV~\cite{zhang2022webuav,huang2023anti,mueller2016benchmark}, and indoor scenes~\cite{wu2015otb,wu2013online}) given an initial bounding box. In the past decade,  significant progress has been made in open-air tracking by embracing deep neural networks. Early deep tracking paradigms include deep discriminative correlation filters~\cite{danelljan2017eco,danelljan2019atom,ge2020cascaded,ge2019distilling,zhang2019robust} and Siamese networks~\cite{bertinetto2016fully,li2019siamrpn++,ChenZLZJ20,guo2020siamcar}. They usually require carefully designed online learning strategies or complex post-processing. Recently, with the development of foundation models~\cite{radford2021learning,devlin2018bert}, many advanced tracking techniques have emerged, such as unified architectures~\cite{ye2022joint,wu2023single}, autoregressive models~\cite{wei2023autoregressive,chen2023seqtrack}, prompt learning~\cite{zhu2023visual}, and diffusion~\cite{xie2024diffusiontrack}. All these modern deep tracking models benefit from public large-scale datasets~\cite{lin2014microsoft,fan2019lasot,huang2019got,muller2018trackingnet}.

\myPara{Underwater Object Tracking.} UOT~\cite{kezebou2019underwater,panetta2021comprehensive,cai2023semi,alawode2023improving,alawode2022utb180} aims to predict the location of objects submerged in various underwater environments. Recently, it has attracted increasing attention from the research community due to the underwater vision understanding and marine animal conservation demands. Yoerger~\etal~\cite{yoerger2021hybrid} propose a Mesobot platform to track slow-moving marine animals. This platform tracks jellyfish and larvaceans by building color segmentation and blob-tracking~\cite{bradski2000opencv} methods. However, these methods can only be used for simple underwater scenarios and a few species. Katija~\etal~\cite{katija2021visual} further propose using tracking-by-detection and deep neural networks to improve tracking robustness and adaptability in more complex underwater environments. Li~\etal~\cite{li2023underwater} introduce underwater images and open-air sequence hybrid training and motion-based post-processing to address the sample imbalance and model drift problems, respectively. To promote the research of UOT, many datasets are established, \eg, UOT32~\cite{kezebou2019underwater}, UOT100~\cite{panetta2021comprehensive}, UTB180~\cite{alawode2022utb180}, VMAT~\cite{cai2023semi}, and UVOT400~\cite{alawode2023improving}. However, these datasets either lack training sets~\cite{kezebou2019underwater,panetta2021comprehensive,cai2023semi} (see Tab.~\ref{tab:Comp_WebUOT_1M}) or are difficult to train models with good generalization capabilities due to limited size and scenarios covered~\cite{alawode2022utb180,alawode2023improving}. To the best of our knowledge, there is still no public million-scale benchmark specifically dedicated to the UOT task. We believe that our benchmark can greatly facilitate the research of UOT.

\myPara{Knowledge Distillation.} KD~\cite{hinton2015distilling,gou2021knowledge,park2019relational}, \ie, efficiently learning a small student network from a large teacher network, is a widely studied area. Its core idea is that the student network imitates the teacher network to obtain competitive or even better performance~\cite{gou2021knowledge}. In recent years, there are many KD-based deep trackers, which include but do not limit to channel distillation~\cite{ge2019distilling}, training-set distillation~\cite{li2020training}, cross-modality distillation~\cite{zhang2023efficient,wang2023event}, correlation filter distillation~\cite{chen2022teacher}, and Siamese network distillation~\cite{shen2021distilled,zhao2022distillation}. Inspired by recent RGB-event distillation method~\cite{wang2023event}, we propose a novel omni-knowledge distillation approach in the UOT area by combining token contrastive representation, similarity matrix, feature embeddings, and response maps distillation losses.

\vspace{-0.3cm}
\section{Dataset}
\label{sec:dataset}
\vspace{-0.3cm}

\subsection{Data Collection and Annotation}
\begin{figure}[t]
\vspace{-0.5cm}
  \centering
  \includegraphics[width=1.0\linewidth]{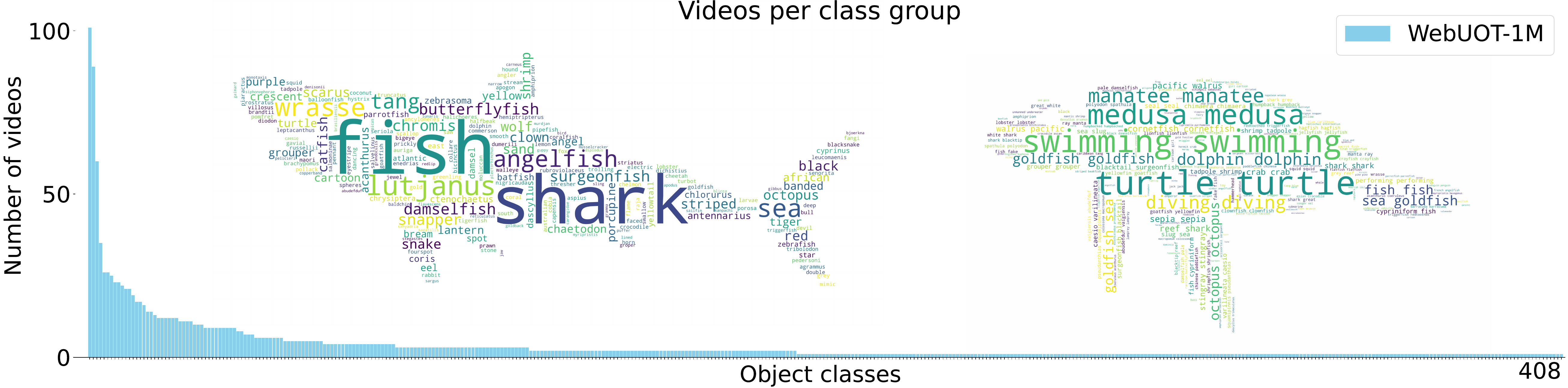}
  \caption{We propose a challenging benchmark containing diverse object classes shown in word clouds, and the number of videos in each class group forms a long-tail distribution.}
  \label{fig:Group_of_object_classes}
\end{figure}

\vspace{-0.3cm}
The goal of WebUOT-1M is to provide a large-scale benchmark for UOT in various real-world underwater scenarios. To this end, we mainly resort to online video platforms, \eg, YouTube and BiliBili, and carefully collect and filter 1,500 video sequences covering 408 different categories from abundant  underwater scenarios, \eg, sea, river, lake, pool, aquarium, fish tank, bowl, and cup. The video platforms contain massive real-world videos captured from different devices/platforms (\eg, diver-based cameras, human-occupied vehicles, underwater robots, and mobile phones), with different shooting angles, distances, and camera movement patterns, greatly reducing the cost of collecting large-scale UOT datasets. Then, we perform data cleaning to discard videos that are not suitable for tracking, \eg, repeated scenes, long-term static targets, and incomplete trajectories. The number of videos in each class group forms a long-tail distribution (see Fig.~\ref{fig:Group_of_object_classes}), which meets real-world situations, making WebUOT-1M more challenging and encouraging the learning of more general UOT algorithms. Moreover, this brings a significant advantage to our dataset, since WebUOT contains a wide range of categories, especially many \emph{rare species} (\eg, flame squid, salamander, and Chinese sturgeon), which can facilitate the visual observations and tracking of these rare species. All videos according to the target are divided into 12 superclasses with reference to WordNet~\cite{miller1995wordnet}, \ie,  \emph{amphibian, arthropod, bird, chordate, coelenterate, crustacean, fish, mollusc, person, mammal (except humans), reptile, and inanimate object} (see Fig.~\ref{fig:examples}). Unlike most previous UOT datasets~\cite{kezebou2019underwater,panetta2021comprehensive,cai2023semi,alawode2022utb180} that only contain marine animals, WebUOT-1M incorporates inanimate objects (\eg, unmanned underwater vehicle, and amphibious drone), resulting in a more comprehensive and versatile benchmark.

\begin{figure*}[t]
\vspace{-0.6cm}
  \centering
  \subfloat{\includegraphics[width =0.26\columnwidth]{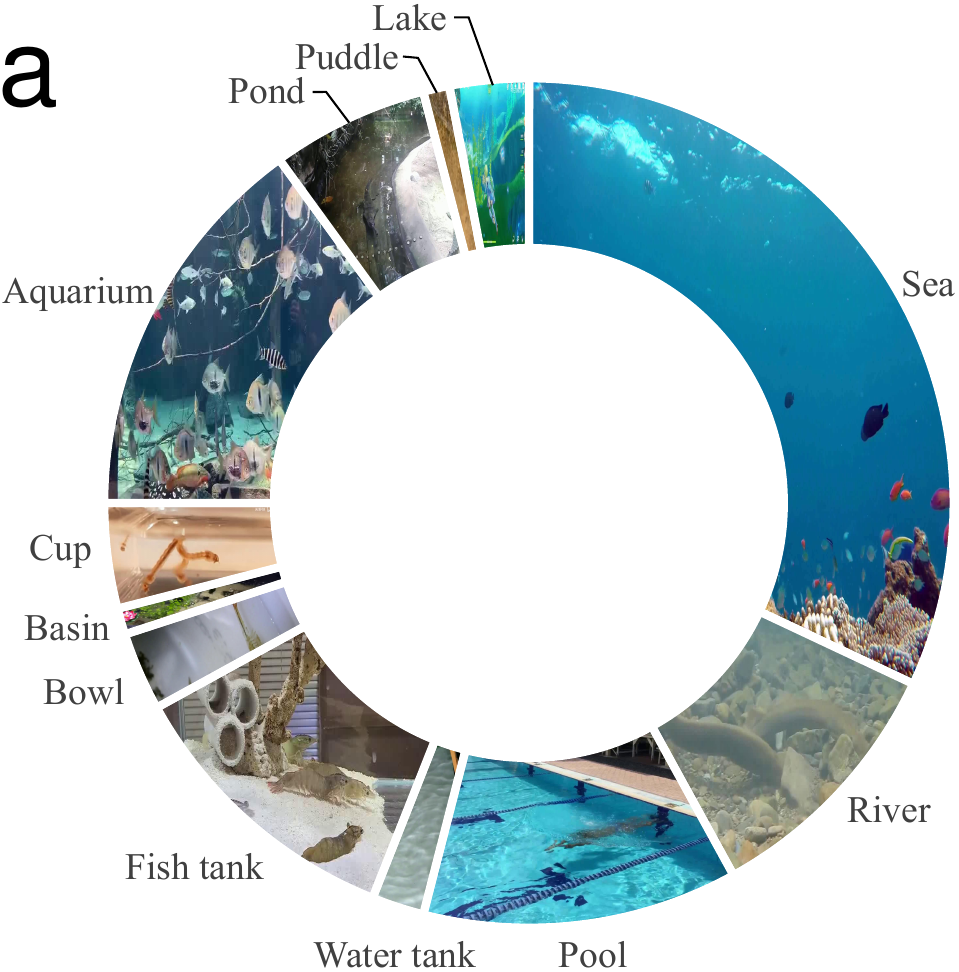}}
~~~~~\subfloat{\includegraphics[width =0.332\columnwidth]{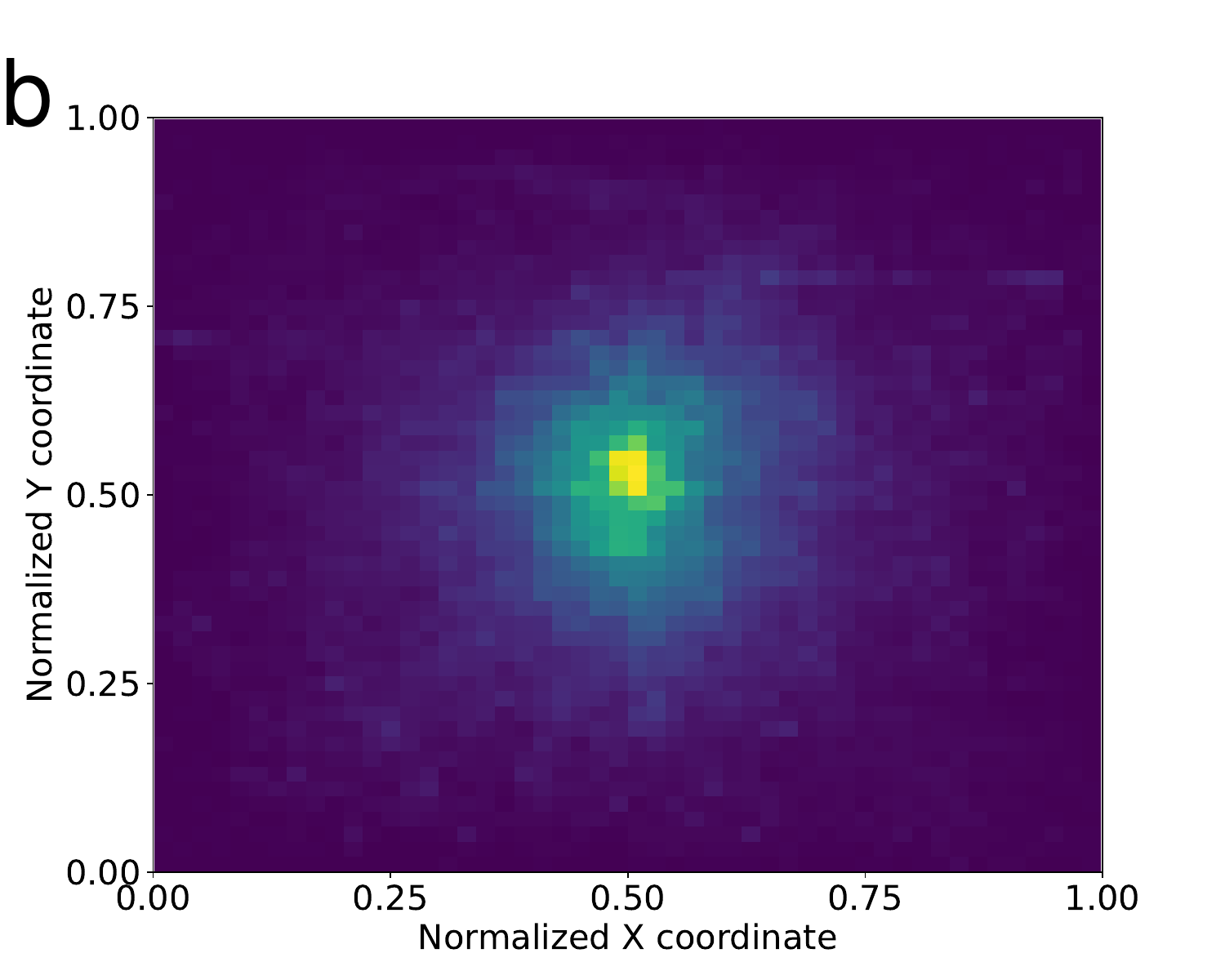}}
~\subfloat{\includegraphics[width =0.29\columnwidth]{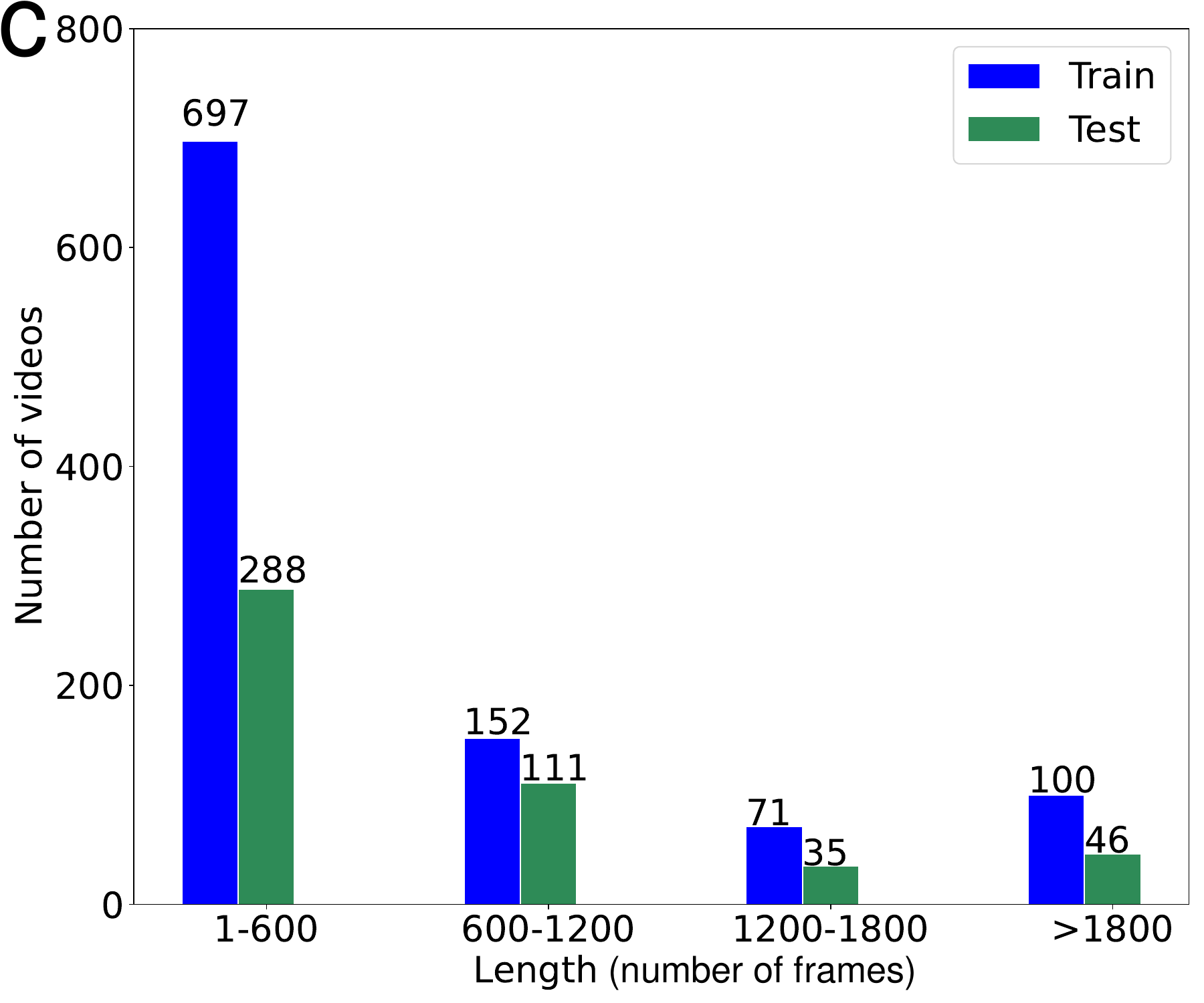}}
  \caption{Statistics of WebUOT-1M. (a) Abundant underwater scenarios. (b) Distribution of normalized target center position. (c) Distribution of video length.}
  \label{fig:statistics}
\vspace{-0.5cm}
\end{figure*}


After the video cleaning, we randomly select moving targets in the videos to ensure the diversity of the dataset (see Fig.~\ref{fig:examples}). Then, a professional data labeling team conducts multiple rounds of manual annotation and correction. \emph{However, directly annotating some underwater videos with severe color deviation and blurring is extremely difficult or even impossible.} To address this issue, we provide the annotators with enhanced videos using a semi-supervised method~\cite{huang2023contrastive}. The author team performs the last data verification to ensure the high quality of the annotations. Specifically, the bounding box $[x, y, w, h]$ is used as the ground-truth of the target in each frame of the video, where $(x, y)$, $w$, $h$ are the left corner point, width and height of the target respectively. Following~\cite{zhang2022webuav,wang2021towards,fan2021lasot}, a sentence of language prompt describing the color, behavior, attributes, and surroundings of the target is given for each video sequence to encourage the exploration of multi-modal UOT. The \emph{absent} label for each frame is also annotated to provide rich information for accurate tracking (see Tab.~\ref{tab:Comp_WebUOT_1M}).

\vspace{-0.3cm}
\subsection{Attributes} 
\vspace{-0.3cm}

To enable comprehensive and in-depth evaluation of trackers~\cite{alawode2023improving,zhang2022webuav}, we label each video sequence with rich attributes. Specifically, we provide 23 attributes, including low resolution (LR), fast motion (FM), scale variations (SV), aspect ratio variations (ARV), camera motion (CM), viewpoint changes (VC), partial occlusion (PO), full occlusion (FO), out-of-view (OV), rotation (ROT), deformation (DEF), similar distractors (SD), illumination variations (IV), motion blur (MB), partial target information (PTI), natural or artificial object (NAO), camouflage (CAM), underwater visibility (UV), watercolor variations (WCV), underwater scenarios (US), shooting perspective (SP), size (SIZ), and length (LEN) of video.  The detailed definition and statistics of attributes are shown in \textbf{Appendices}.

\begin{figure*}[t]
  \centering
  \vspace{-0.5cm}
  \includegraphics[width=1.0\linewidth]{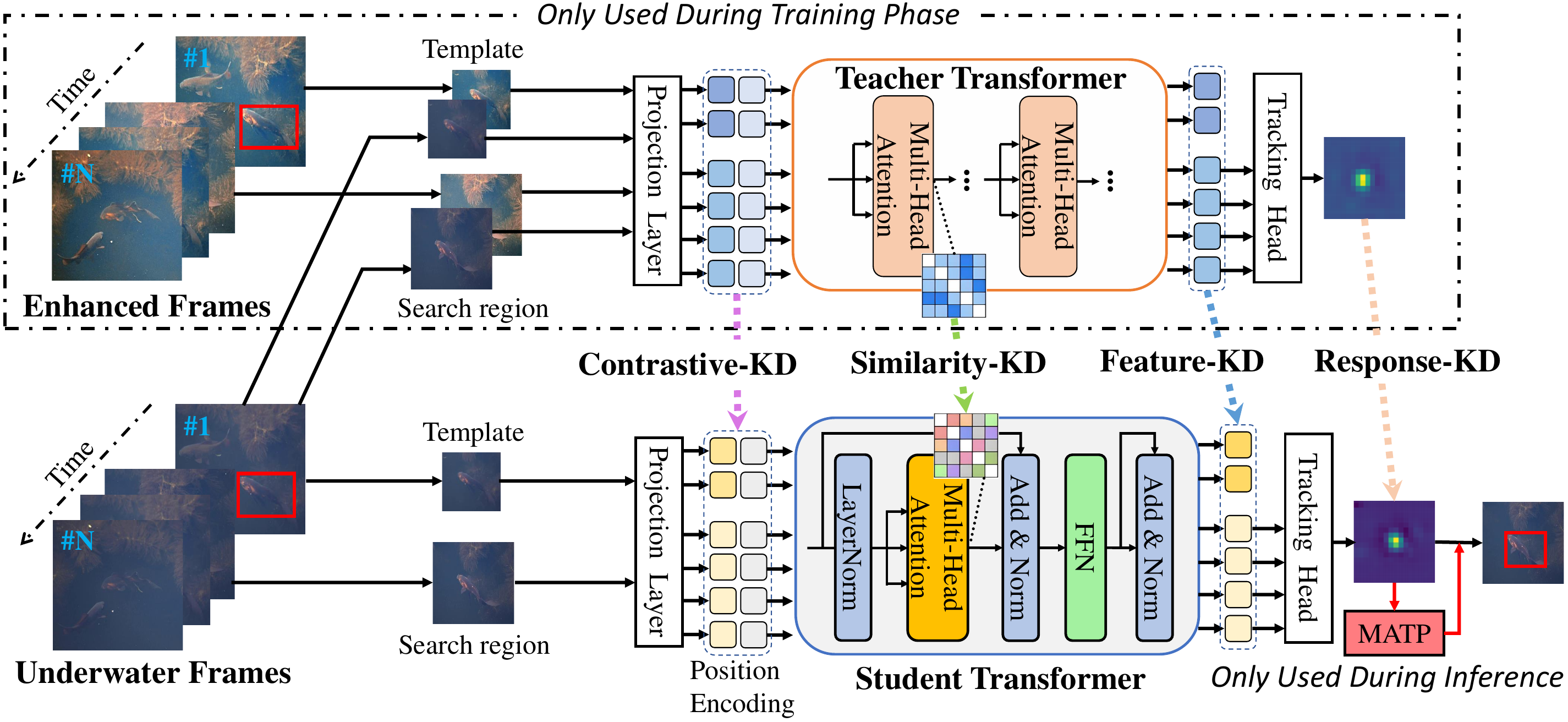}
  \caption{OKTrack overview. During training phase, we adopt four distillation losses (see Sec.~\ref{sec:okd}). A training-free MATP module (see Sec.~\ref{sec:network}) is used to enhance the tracking robustness of inference.}
  \label{fig:method}
  \vspace{-0.5cm}
\end{figure*}

\vspace{-0.3cm}
\subsection{ Statistical Analysis}
\vspace{-0.3cm}

As shown in Fig.~\ref{fig:statistics}(a), WebUOT-1M contains abundant underwater scenarios, including sea, river, lake, pool, fish tank, water tank, basin, bowl, cup, aquarium, pond, and puddle. The normalized target center position distribution presents a center mean Gaussian (see Fig.~\ref{fig:statistics}(b)), indicating the high quality and diversity of the dataset. The distribution of video length is demonstrated in Fig.~\ref{fig:statistics}(c). We can see that WebUOT-1M contains 985, 263, 106, and 146 videos with segments containing 1-600, 600-1200, 1200-1800, and more than 1800 frames, respectively. The various video lengths make our dataset suitable for benchmarking both short-term and long-term tracking algorithms.

\vspace{-0.3cm}
\section{Methodology}
\label{sec:methidology}
\vspace{-0.3cm}

In this section, we present an omni-knowledge distillation framework (see Fig.~\ref{fig:method}). The core insight is to leverage a teacher model pre-trained on the massive open-air data to enhance tracking performance on UOT. KD~\cite{hinton2015distilling,gou2021knowledge} has been proven to efficiently learn a student model from a teacher model. Inspired by recent RGB-event tracking~\cite{wang2023event}, we propose to distill the knowledge from underwater and enhanced frames with the supervision of a pre-trained teacher tracker to a student tracker devoted to handling underwater frames. Due to the limited space, more details are provided in \textbf{Appendices}.

\vspace{-0.3cm}
\subsection{Problem Formulation}
\vspace{-0.3cm}

Given underwater frames $\mathcal{I}=\{I_{i}\}_{i=1}^{N}\in \mathbb{R}^{H\times W\times C}$, we first adopt an off-the-shelf semi-supervised method~\cite{huang2023contrastive} to obtain corresponding enhanced frames $\mathcal{E}=\{e_{i}\}_{i=1}^{N}\in \mathbb{R}^{H\times W\times C}$, where $(H,W)$ denotes the resolution of video frames, $C$ and $N$ represent the number of channels and video frames, respectively. Like Transformer-based trackers~\cite{ye2022joint,chen2023seqtrack}, we can crop a pair of images patches, \ie, the template patch $z\in\mathbb{R}^{H_{z}\times W_{z}\times C}$ and the search region patch $x\in\mathbb{R}^{H_{x}\times W_{x}\times C}$, from underwater frames. Then, the image patches are divided into multiple non-overlapping patches, which will be further transformed into 1D tokens using a projection layer~\cite{ye2022joint}. To preserve position information, learnable positional embeddings~\cite{dosovitskiy2020image} are added to these tokens. Given an underwater video with a pair of template and search region patches $\mathcal{X}_{xz}=\{x,z\}$, and an initial target box $\mathcal{B}_{0}$, the UOT problem can be formulated as $\mathcal{S}:\{\mathcal{X}_{xz},\mathcal{B}_{0}\}\to\mathcal{B}$, where $\mathcal{S}$ is the student tracker, $\mathcal{B}$ is the predicted target box in subsequent frames. Adding enhanced template and search region patches $\tilde{\mathcal{X}}_{xz}=\{\tilde{x},\tilde{z}\}$ and a pre-trained teacher tracker $\mathcal{T}$, the learning of the student tracker can be expressed as optimizing $\mathcal{L}_{OKD}:\{\mathcal{X}_{xz},\tilde{\mathcal{X}}_{xz},\mathcal{T},\mathcal{S}_{0}\}\to\mathcal{S}$, where $\mathcal{L}_{OKD}$ represents the omni-knowledge distillation loss (see Sec.~\ref{sec:okd}), $\mathcal{S}_{0}$ is an initial student model based on the plain ViT backnone~\cite{dosovitskiy2020image}.

\vspace{-0.3cm}
\subsection{Network Architecture}
\vspace{-0.2cm}
\label{sec:network}

\myPara{Multi-view Teacher Network.} We adopt a modified version of unified backbone~\cite{zhang2023all} as the teacher Transformer, which consists of multiple Transformer layers. The teacher network can use multi-view modalities simultaneously, \ie, underwater and enhanced frames. Specifically, tokens from underwater and enhanced frames are concatenated and fed into the teacher Transformer. Then, the extracted feature embeddings are used for target prediction by a tracking head.

\myPara{Unimodal Student Network.} To realize efficient and low-latency UOT, the student network~\cite{ye2022joint} only uses underwater video streams. As shown in Fig.~\ref{fig:method}, the student Transformer is a plain ViT architecture~\cite{dosovitskiy2020image}. We argue that through the omni-knowledge distillation, an accomplished teacher can effectively transfer the knowledge obtained from handling underwater and enhanced frames to the student network, significantly enhancing the tracking performance of the student network.

\myPara{Motion-aware Target Prediction.} The underwater targets (\eg, fish and sharks) are often surrounded by similar distractors, leading to model drift~\cite{li2023underwater}. To tackle this issue, we design a \emph{training-free} motion-aware target prediction (MATP) (see Fig.~\ref{fig:method}) based on Kalman filtering~\cite{kalman1960new}. It involves two primary stages: \emph{prediction}, which estimates the current state using the previous state, and \emph{correction}, which combines the estimated state with current observations to determine the optimal state.

\subsection{Omni-Knowledge Distillation}
\label{sec:okd}
\vspace{-0.2cm}

\myPara{Token-based Contrastive Distillation (CKD).} The CKD is employed to explicitly align underwater and enhanced tokens by CL~\cite{oord2018representation}, aiming to mitigate the distribution discrepancies between multi-view modalities (\ie, underwater and enhanced frames). We define underwater tokens for student network as $\mathbf{t}^{s}\in \mathbb{R}^{320}$, underwater and enhanced tokens for teacher network as $\mathbf{t}^{t}\in \mathbb{R}^{320}$ and $\tilde{\mathbf{t}}^{t}\in \mathbb{R}^{320}$, respectively. Formally, the CKD losses between teacher and student networks are defined as:
\vspace{-0.2cm}
\begin{small} 
\begin{equation}
    \mathcal{L}_{u2e}\!=\!-\frac{1}{K}\!\sum_{i=1}^{K}\!\log \frac{{\exp(sim(\mathbf{t}^{s}_{i}},\tilde{\mathbf{t}}^{t}_{i})/\tau)}{\sum_{j=1}^{K}\exp(sim(\mathbf{t}^{s}_{i},{\tilde{\mathbf{t}}}^{t}_{j})/\tau)}\rbrack,~~~
    \mathcal{L}_{e2u}\!=\!-\frac{1}{K}\!\sum_{i=1}^{K}\!\log \frac{{\exp(sim(\tilde{\mathbf{t}}^{t}_{i},\mathbf{t}^{s}_{i}})/\tau)}{\sum_{j=1}^{K}\exp(sim({\tilde{\mathbf{t}}}^{t}_{i},\mathbf{t}^{s}_{j})/\tau)}\rbrack.
    \label{eq:ckd_loss}
\end{equation}
\end{small}

\vspace{-0.35cm}
where $K$ is the batch size, $\tau$ is a temperature parameter, $sim(\cdot)$ denotes cosine similarity function. The CKD losses, \ie, $\mathcal{L}'_{u2e}$ and $\mathcal{L}'_{e2u}$, with the teacher network can be calculated similarly. The total CKD loss is $\mathcal{L}_{CKD} = \mathcal{L}_{u2e}+\mathcal{L}_{e2u}+\mathcal{L}'_{u2e}+\mathcal{L}'_{e2u}$.

\myPara{Similarity-based Distillation (SKD).} The similarity matrix (\ie, dot product of query and key) in the multi-head attention can capture rich long-range dependencies and cross-modal information~\cite{wang2023event,dosovitskiy2020image}. Given the similarity matrices (\ie, $S_{i}^{t}$ and $S_{i}^{s}$) of the $i^{th}$ layer of teacher and student Transformers. We define the SKD loss as $\mathcal{L}_{SKD}\!=\!\sum_{i=1}^{L}(S_{i}^{t}-S_{i}^{s})^{2}$, where $L$ is the number of Transformer layers.

\myPara{Feature-based Distillation (FKD).} The advanced feature embeddings contain rich semantic information. The FKD loss between teacher and student networks can be formalized as $\mathcal{L}_{FKD}\!=\!||F^{t}-F^{s}||^{2}$, where $F^{t}$ and $F^{s}$ are feature representations of them, respectively.

\myPara{Response-based Distillation (RKD).} In general, directly mimicking the response map $R^{t}$ of the teacher network used for target localization enables the student network to achieve better tracking accuracy~\cite{shen2021distilled}. Following~\cite{wang2023event}, we adopt the Gaussian weighted focal loss function $\mathcal{L}_{GWF}(\cdot)$ to define the RKD loss as $\mathcal{L}_{RKD}\!=\!\mathcal{L}_{GWF}(R^{t}/\mu,R^{s}/\mu)$, where $\mu$ is a scale factor.

Therefore, the omni-knowledge distillation loss is formulated as $\mathcal{L}_{OKD}\!=\!\mathcal{L}_{CKD}\!+\!\mathcal{L}_{SKD}\!+\!\mathcal{L}_{FKD}\!+\!\mathcal{L}_{RKD}$. We also borrow the tracking loss function used in~\cite{ye2022joint,li2023citetracker} (\ie, GIoU loss $\mathcal{L}_{GIoU}$, focal loss $\mathcal{L}_{focal}$, and $L_{1}$ loss $\mathcal{L}_{L1}$) to enhance the convergence of training. Finally, the total loss can be written as $\mathcal{L}_{total}=\mathcal{L}_{OKD}\!+\!\lambda_{1}\mathcal{L}_{GIoU}+\lambda_{2}\mathcal{L}_{focal}+\lambda_{3}\mathcal{L}_{L1}$, where $\lambda_{1},\lambda_{2},\lambda_{3}$ are balance factors.


\vspace{-0.3cm}
\section{Experiments}
\label{sec:experiments}
\vspace{-0.3cm}

\subsection{Implementation Details}
\vspace{-0.3cm}

\begin{wraptable}{r}{6.0cm}    
\footnotesize
\vspace{-1.6cm}
  \caption{Summary of open-air and underwater tracking algorithms. ``Trans.'' denotes Transformer. ``B'' represents base model.}
  \vspace{-0.2cm}
  \label{tab:summary_of_trackers}
  \centering
  \setlength{\tabcolsep}{0.2mm}{
  \scalebox{0.7}{
  \begin{threeparttable}
  \begin{tabular}{lcccc}
    \Xhline{0.7pt} 
    \textbf{Tracker}     & \textbf{Source }    & \textbf{Backbone }& \textbf{FPS}~ & ~\textbf{UOT}\\
    \hline
    
    SiamFC~\cite{bertinetto2016fully} & ECCVW16  &   AlexNet  & 86    & \xmark \\
    
    ECO~\cite{danelljan2017eco}   &  CVPR17 &     VGG-M  & 8       & \xmark \\
    
    VITAL~\cite{song2018vital} &     CVPR18  &     VGG-M   &  1.5  & \xmark \\
    
    ATOM~\cite{danelljan2019atom} &     CVPR19  &     ResNet-18   & 30   & \xmark \\
    
    SiamRPN++~\cite{li2019siamrpn++} &     CVPR19  &      ResNet-50   & 35   & \xmark \\


    SiamBAN~\cite{ChenZLZJ20} &    CVPR20  &     ResNet-50   & 40   & \xmark \\
    
    SiamCAR~\cite{guo2020siamcar} &    CVPR20  &     ResNet-50   & 52  & \xmark \\
    
    Ocean~\cite{zhang2020ocean} &     ECCV20  &      ResNet-50   & 58   & \xmark \\

    PrDiMP~\cite{danelljan2020probabilistic} &    CVPR20  &      ResNet-50   & 30   & \xmark \\
    
    TrDiMP~\cite{wang2021transformer} &     CVPR21  &      ResNet-50, Trans.   & 26   & \xmark \\
    
    TransT~\cite{chen2021transformer} &    CVPR21  &     ResNet-50, Trans.   & 50   & \xmark \\
    
    STARK-ST50~\cite{yan2021learning} &     ICCV21  &     ResNet-50, Trans.   & 40   & \xmark \\

    KeepTrack~\cite{mayer2021learning} &    ICCV21  &     ResNet-50   & 18  & \xmark \\
    
    AutoMatch~\cite{zhang2021learn} &     ICCV21  &      ResNet-50  & 50   & \xmark \\

    TCTrack~\cite{cao2022tctrack} &    CVPR22  &     AlexNet   & 126   & \xmark \\
    
    ToMP-101~\cite{mayer2022transforming} &     CVPR22  &     ResNet-101, Trans.   & 25   & \xmark \\

    AiATrack~\cite{gao2022aiatrack} &    ECCV22  &     ResNet-50, Trans.   & 38   & \xmark \\
        
    SimTrack-B32~\cite{chen2022backbone} &     ECCV22  &     ViT-B   & 30   & \xmark \\
    
    OSTrack~\cite{ye2022joint} &     ECCV22  &     ViT-B   & 105   & \xmark \\
    
    MixFormerV2-B~\cite{cui2023mixformerv2} &    NeurIPS23  &     ViT-B   & 165   & \xmark \\

    GRM~\cite{gao2023generalized} &    CVPR23  &     ViT-B   & 45  & \xmark \\
    
    SeqTrack-B256~\cite{chen2023seqtrack} &    CVPR23  &     ViT-B   & 40   & \xmark \\
    
    \hdashline
    
    VLT$_{\rm SCAR}$~\cite{guo2022divert} &     NeurIPS22  &     ResNet-50, Bert-B  & 43  & \xmark \\
    
     VLT$_{\rm TT}$~\cite{guo2022divert} &     NeurIPS22  &      ResNet-50, Bert-B   & 35   & \xmark \\
     
    JointNLT~\cite{zhou2023joint} &     CVPR23  &      Swin-B, Bert-B    & 39   & \xmark \\
    
    CiteTracker-256~\cite{li2023citetracker} &     ICCV23  &     ViT-B, CLIP   & 24   & \xmark \\
    
    All-in-One~\cite{zhang2023all} &    ACM MM23  &     ViT-B, Bert-B   & 60  & \xmark \\

    UVLTrack~\cite{ma2024unifying} &    AAAI24  &     ViT-B, Bert-B   & 57   & \xmark \\
    
    \hdashline
    
    UOSTrack$^{1}$~\cite{li2023underwater} &    arXiv23  &    ViT-B   & 110   & \cmark \\
    
    OKTrack &    Ours  &     ViT-B   & 115   & \cmark \\
    \Xhline{0.75pt} 
  \end{tabular}
  \begin{tablenotes}
\footnotesize
\item[1] For a fair comparison, we fine-tune this tracker on WebUOT-1M.
\end{tablenotes}
  \end{threeparttable}
  }}
\vspace{-1.0cm}
  \end{wraptable}  
We adopt the unified tracking model~\cite{zhang2023all} as the teacher network. The student network~\cite{ye2022joint} is based on a plain ViT-base backbone, stacked by $L$ (\ie, 12) transformer encoder layers. We use an AdamW optimizer~\cite{loshchilov2017decoupled} with initial learning rate $4\!\times\!10^{-4}$. The weight decay factor is $1\!\times\!10^{-4}$ after 240 epochs. The batch size and total epoch are 32 and 300. The temperature parameter $\tau$ is 0.5, and scale factor $\mu$ is empirically set to 2. Following~\cite{wang2023event}, the balance factors $\lambda_{1},\lambda_{2},\lambda_{3}$ are set to 1, 1, and 14, respectively. For proper and fast verification, models are trained for 50 epochs in ablation experiments. Our experiment platform is an Ubuntu server with 8 NVIDIA A6000 GPUs.

\vspace{-0.3cm}
\subsection{Metrics and Protocols}
\vspace{-0.3cm}

Following~\cite{zhang2022webuav,fan2021lasot}, we perform the one-pass evaluation (OPE) and measure trackers using five evaluation metrics (\ie, percision (Pre), normalized precision (nPre), success rate (AUC), complete success rate (cAUC), and mean accuracy (mACC)) under two protocols.

\myPara{Protocol \Rmnum{1}.} In this protocol, we conduct a \emph{cross-domain evaluation} of existing tracking models trained on \emph{open-air} tracking datasets. We report the results of different trackers on the WebUOT-1M test set. Cross-domain evaluation helps ascertain how well a tracker can adapt to new and unseen data distributions, providing insights into its robustness.

\begin{figure*}[ht]
  \centering
\includegraphics[width=1.0\linewidth]{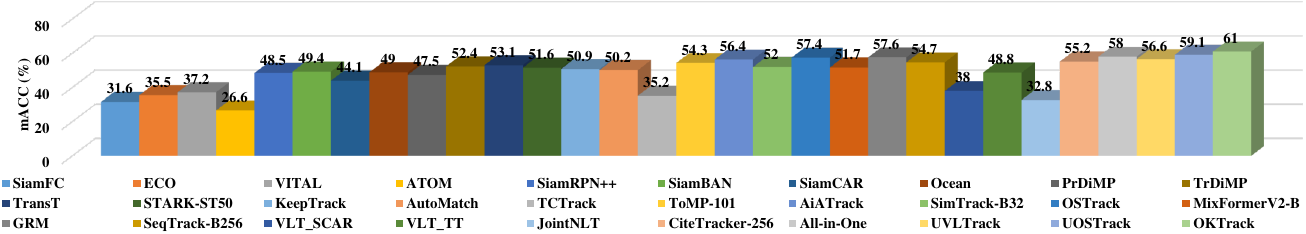}
  \caption{Overall performance on WebUOT-1M using mACC score. Best viewed by zooming in.}
  \label{fig:mACC_results}
  \vspace{-0.5cm}
\end{figure*}

\begin{figure*}[t]
\begin{minipage}[c]{1.0\textwidth}
  \centering
  \subfloat{\includegraphics[width =0.25\columnwidth]{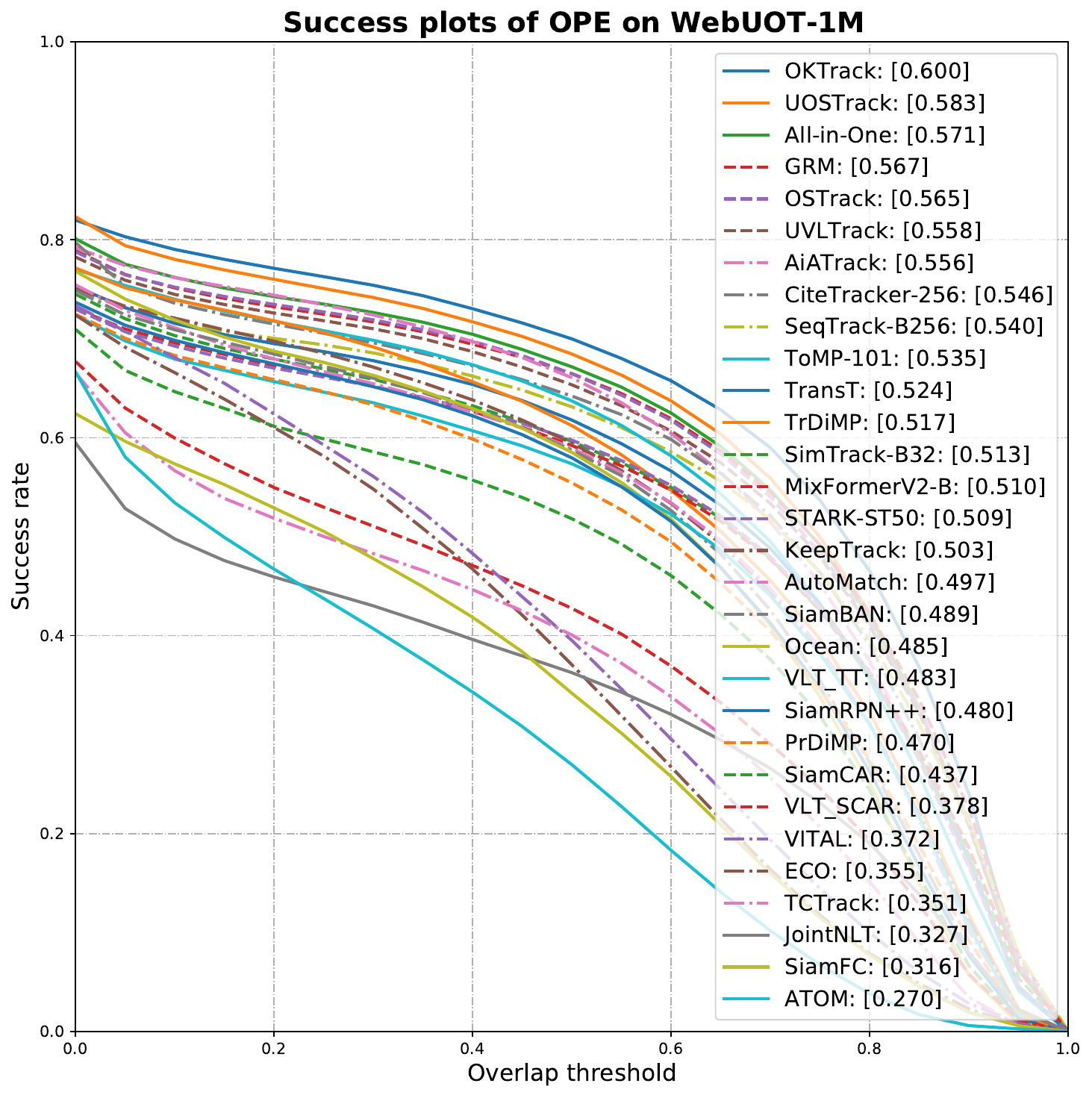}}
\subfloat{\includegraphics[width =0.25\columnwidth]{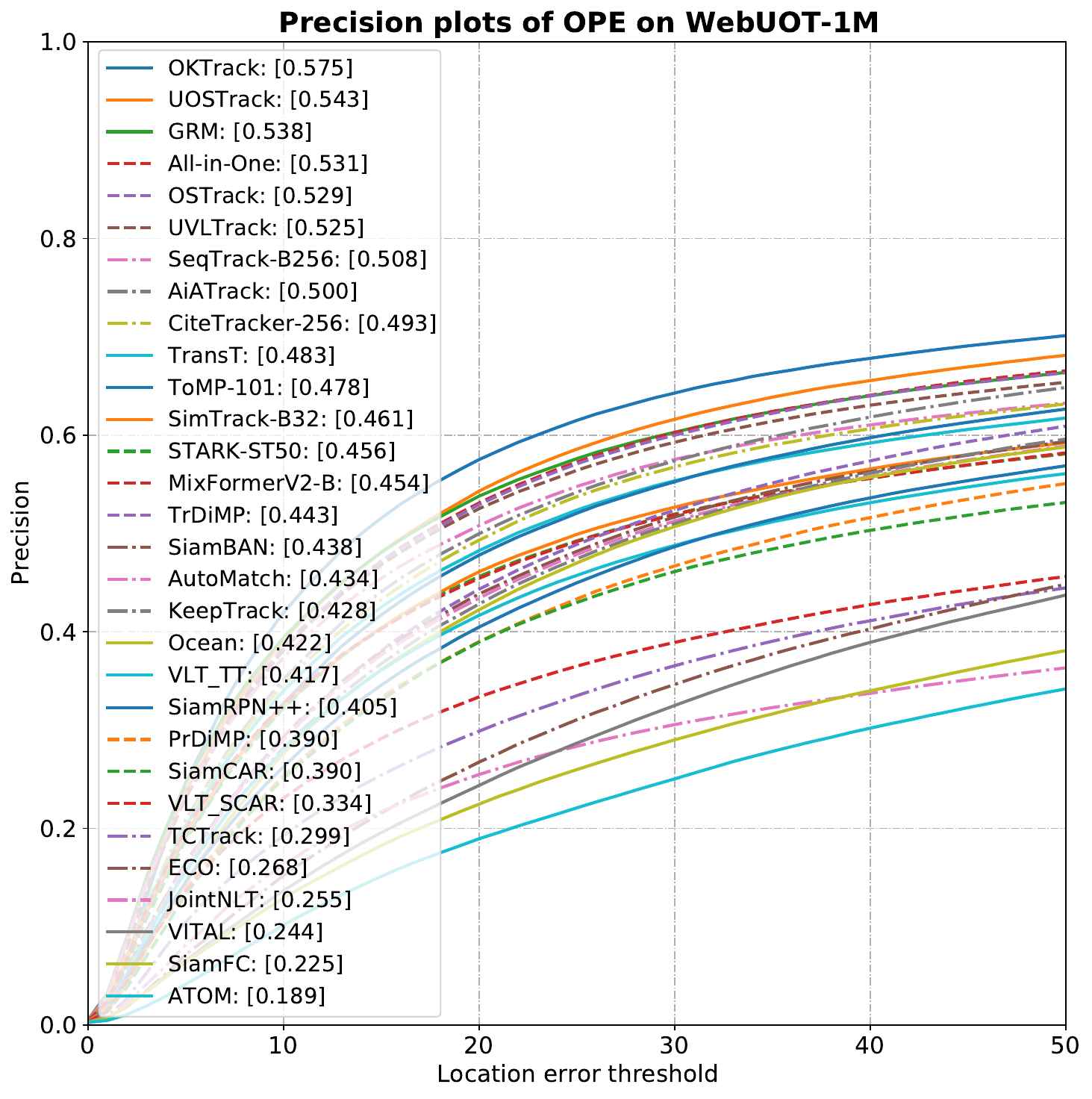}}
\subfloat{\includegraphics[width =0.25\columnwidth]{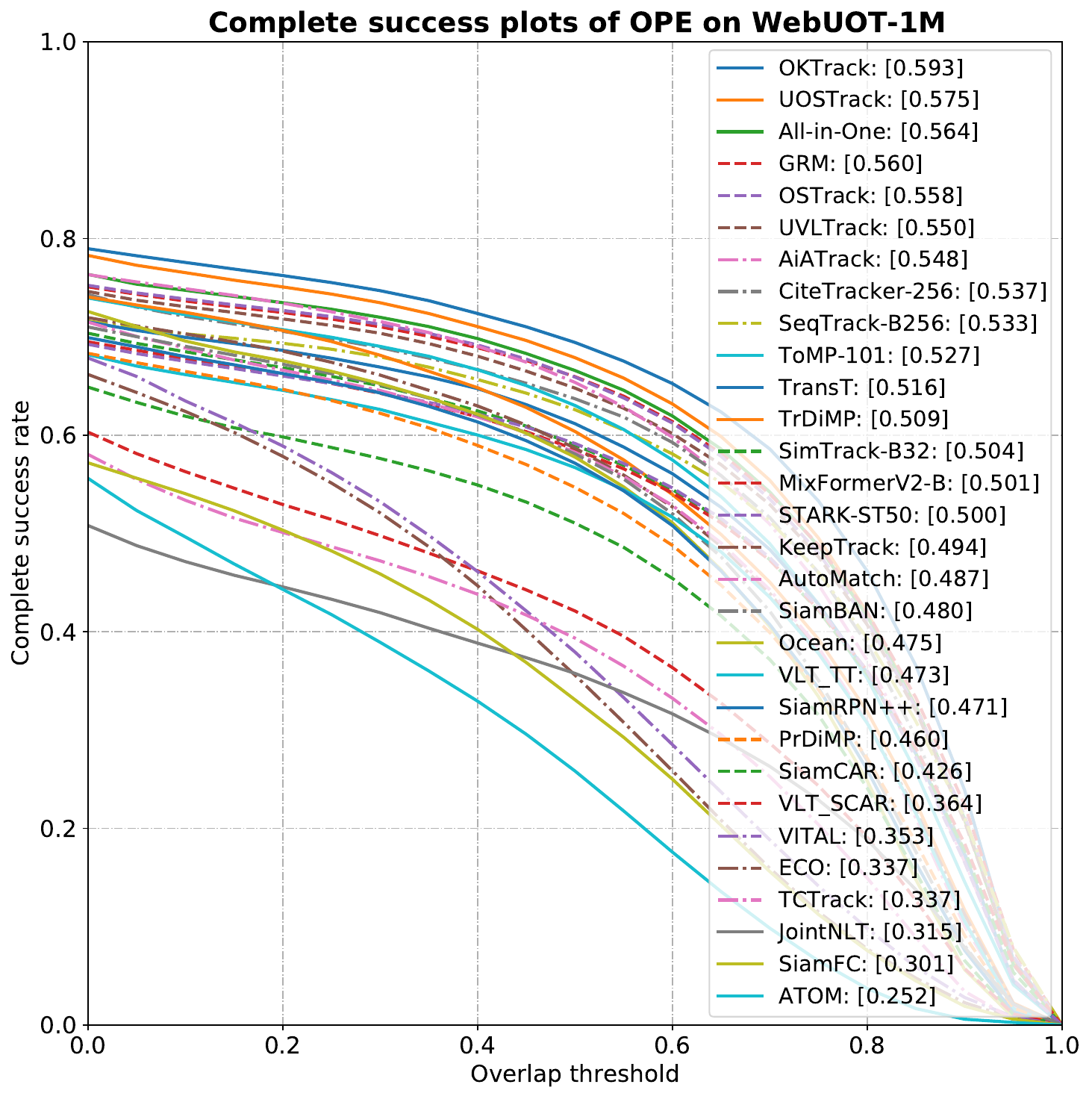}}
\subfloat{\includegraphics[width =0.25\columnwidth]{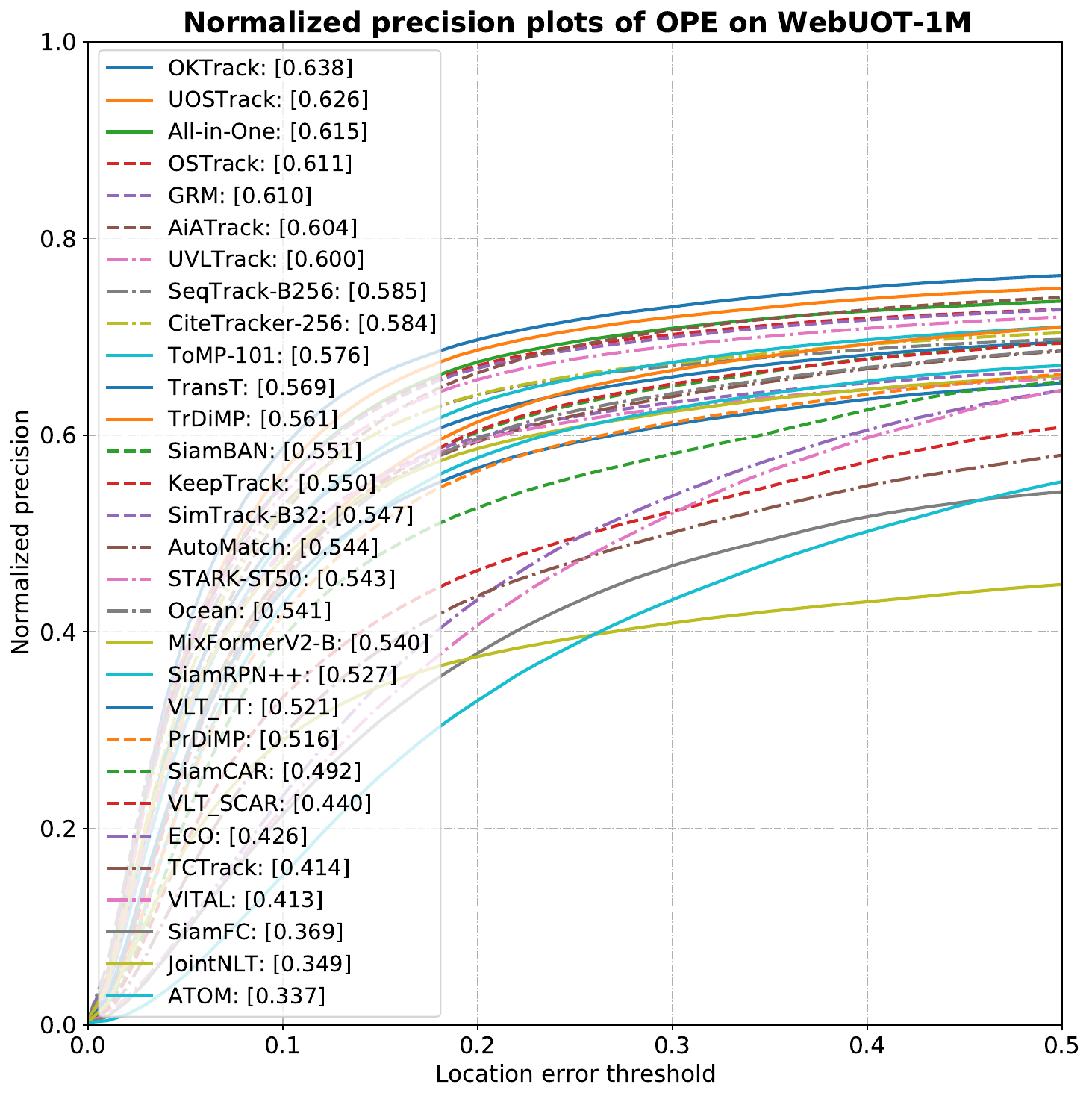}}
  \caption{Overall performance on WebUOT-1M using AUC, Pre, cAUC, and nPre scores.}
  \label{fig:results_on_webuot1m}
\end{minipage}
\begin{minipage}[c]{1.0\textwidth}
  \centering
  \subfloat{\includegraphics[width =0.25\columnwidth]{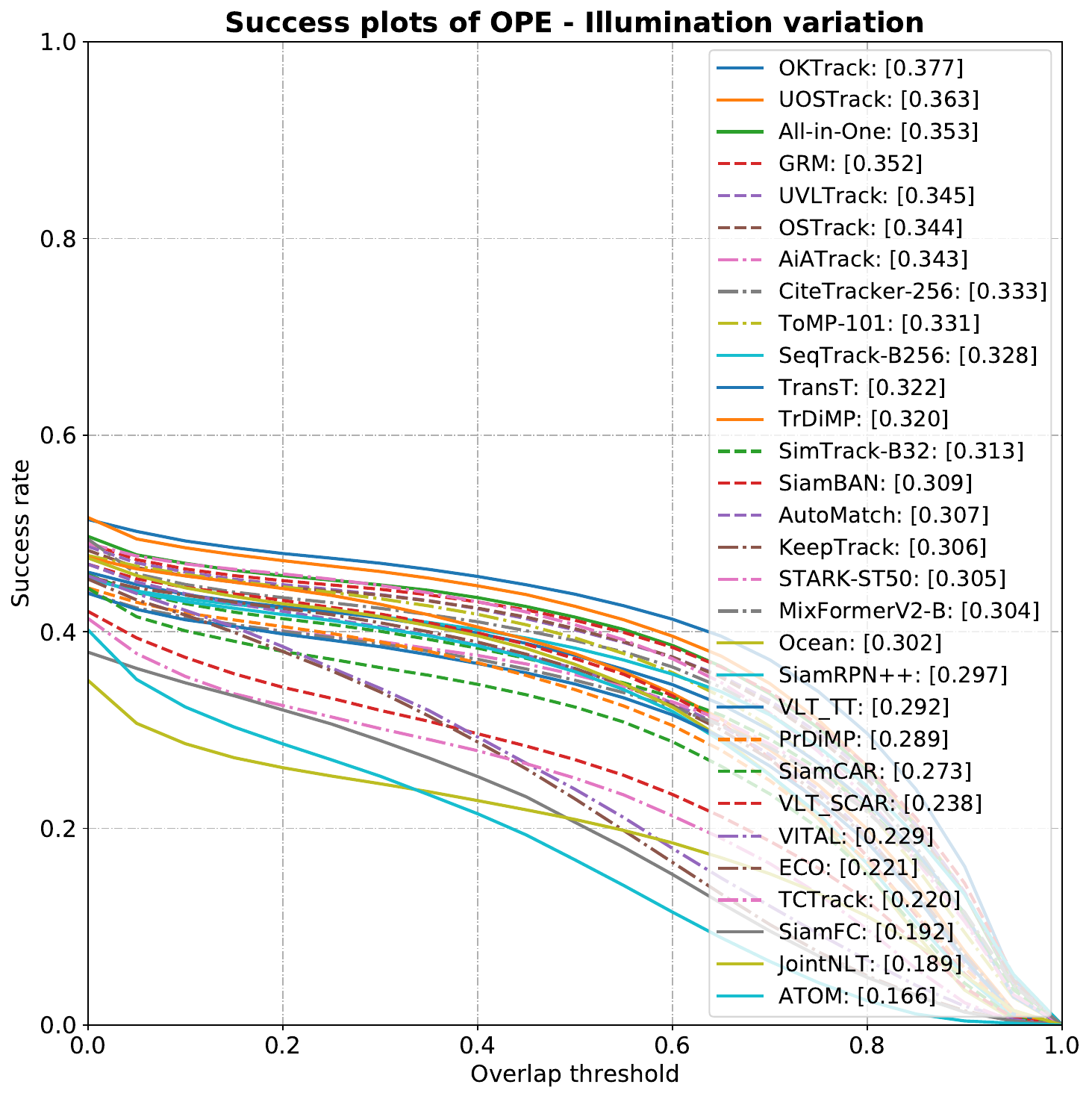}}
\subfloat{\includegraphics[width =0.25\columnwidth]{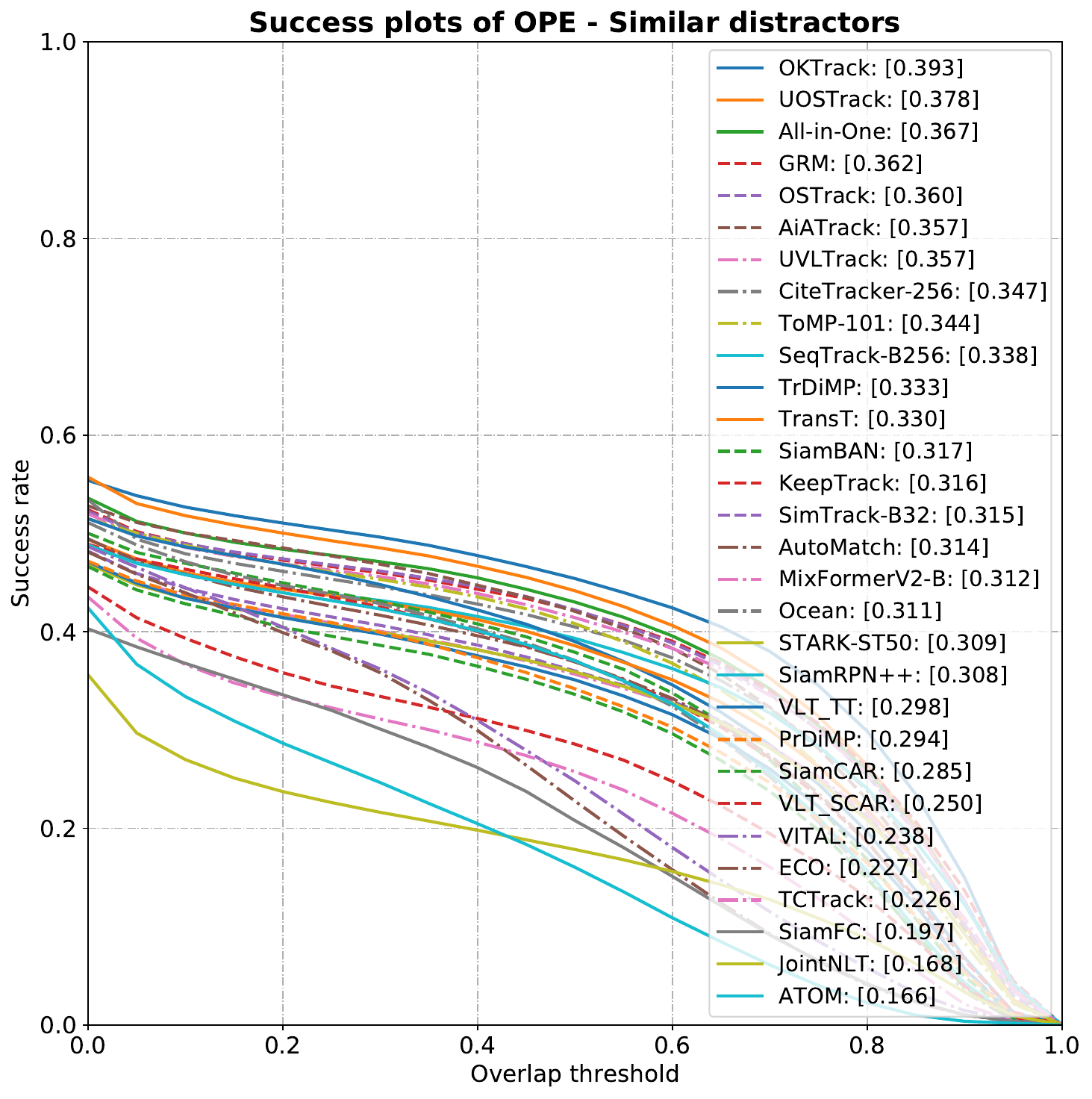}}
\subfloat{\includegraphics[width =0.25\columnwidth]{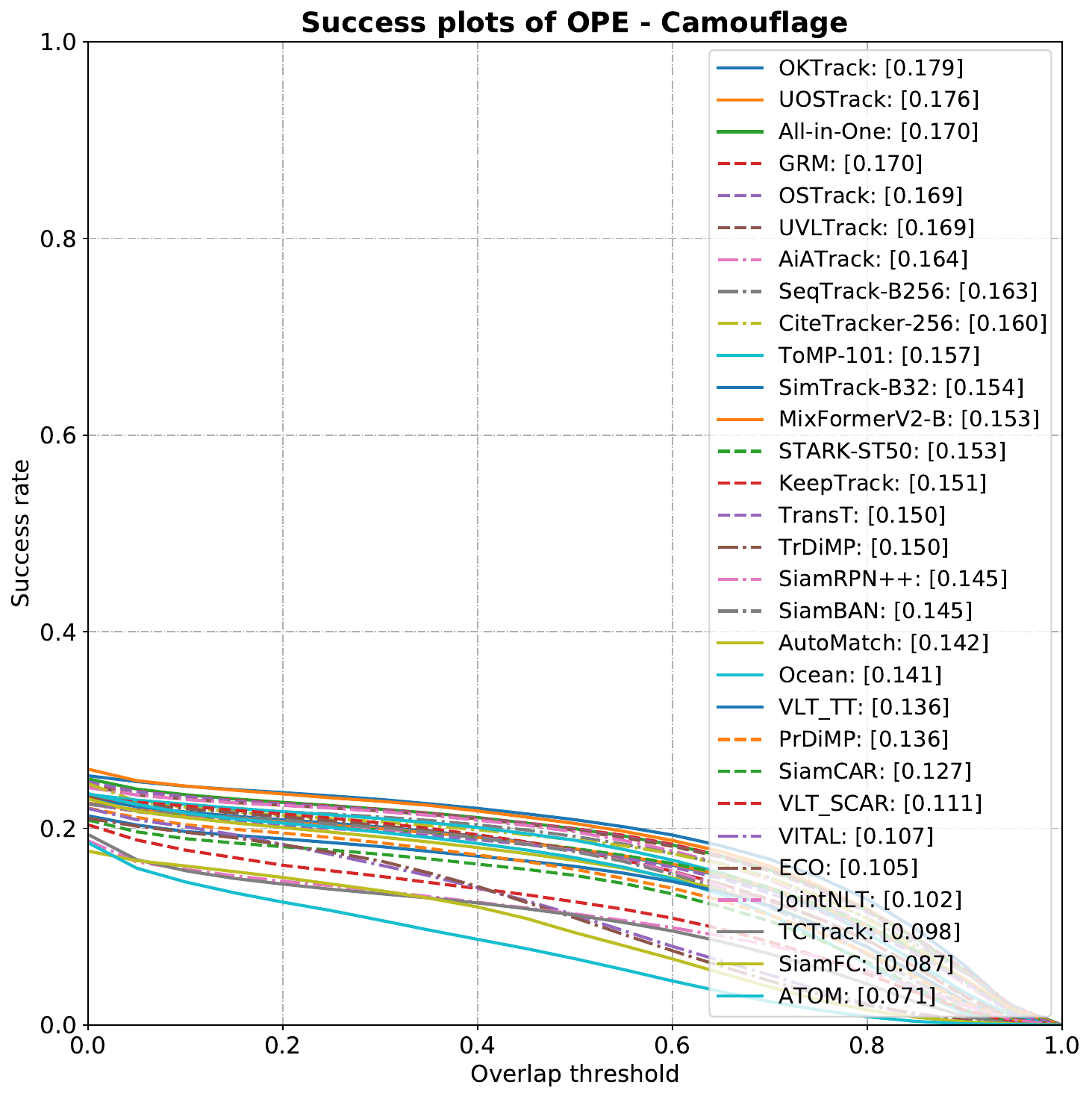}}
\subfloat{\includegraphics[width =0.25\columnwidth]{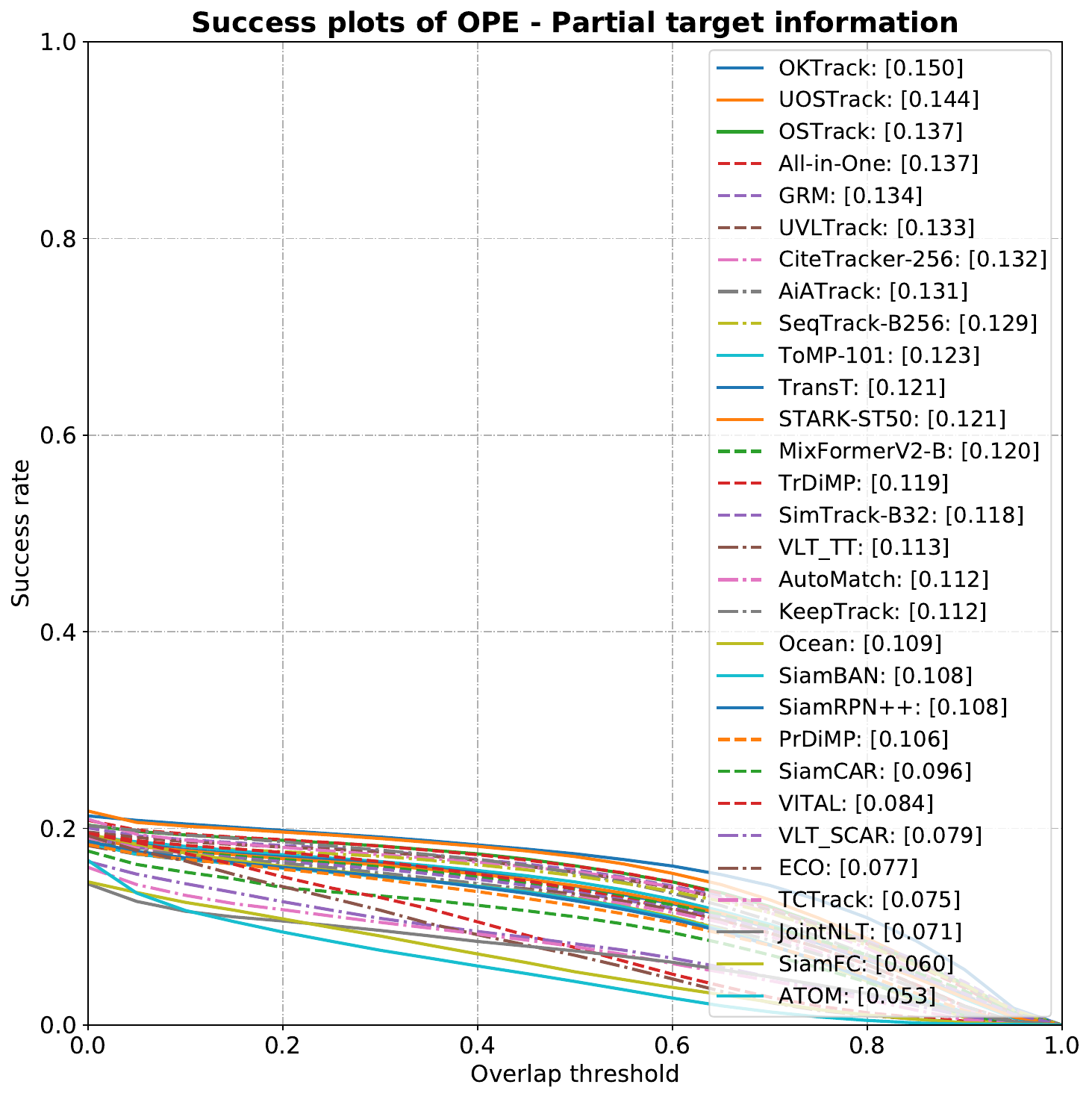}}
  \caption{Evaluation results of four tracking attributes on WebUOT-1M using AUC score.}
  \label{fig:attribute_results_on_webuot1m}
  \vspace{-0.5cm}
\end{minipage}

\end{figure*}

\myPara{Protocol \Rmnum{2}.} In this protocol, we perform \emph{within-domain evaluation} of open-resoure trackers on the WebUOT-1M dataset. Concretely, we retrain different trackers on the training set and evaluate them on the test set. Protocol \Rmnum{2} aims to provide benchmark results for the underwater tracking community to train and evaluate trackers using a large number of underwater videos.

\vspace{-0.3cm}
\subsection{Evaluated Trackers}
\vspace{-0.3cm}

To provide baseline results for future research, we extensively evaluate 30 deep trackers, including \emph{CNN-based} (\eg, SiamFC~\cite{bertinetto2016fully}, ECO~\cite{danelljan2017eco}, VITAL~\cite{song2018vital}, ATOM~\cite{danelljan2019atom}, SiamRPN++~\cite{li2019siamrpn++}, Ocean~\cite{zhang2020ocean}), \emph{CNN-Transformer-based} (\eg, TrDiMP~\cite{wang2021transformer}, TransT~\cite{chen2021transformer}, STARK-ST50~\cite{yan2021learning}, ToMP-101~\cite{mayer2022transforming}), \emph{Transformer-based} methods (\eg, OSTrack~\cite{ye2022joint}, SimTrack-B32~\cite{chen2022backbone}, MixFormerV2-B~\cite{cui2023mixformerv2}, SeqTrack-B256~\cite{chen2023seqtrack}), vision-language trackers (\eg, VLT$_{\rm TT}$~\cite{guo2022divert}, JointNLT~\cite{zhou2023joint}, CiteTracker-256~\cite{li2023citetracker}, All-in-One~\cite{zhang2023all}), and UOT trackers (UOSTrack~\cite{li2023underwater}, OKTrack), as shown on Tab.~\ref{tab:summary_of_trackers}.

\vspace{-0.3cm}
\subsection{Evaluation Results}
\vspace{-0.2cm}

\myPara{Overall Performance.} Figs.~\ref{fig:mACC_results} and~\ref{fig:results_on_webuot1m} demonstrate the cross-domain evaluation results of 30 deep trackers on WebUOT-1M. We have the following observations. \textbf{1)} The top-5 trackers (\ie, OKTrack, UOSTrack, All-in-One, GRM, OSTrack) are based on Transformer~\cite{dosovitskiy2020image}, indicating that exploring advanced architectures is still a promising direction for tracking~\cite{peng2024vasttrack}. \textbf{2)} The UOT trackers (\ie, OKTrack, UOSTrack) using the plain ViT backbone surpasses state-of-the-art (SOTA) trackers~\cite{li2023citetracker,chen2023seqtrack} for open-air tracking. The possible reason is that there is a huge domain gap between underwater and open-air environments. \textbf{3)} The vision-language tracker (\ie, All-in-One) achieves the best results among open-air trackers, demonstrating that using the additional language modality can enhance tracking performance.

\myPara{Attribute-based Performance.} To comprehensively analyze the trackers facing different tracking attributes, we conduct an attribute-based evaluation using 23 attributes. The results show that the SOTA trackers still have significant room for improvement in various challenging attributes, \eg,  IV, SD, CAM, and PTI (see Fig.~\ref{fig:attribute_results_on_webuot1m}). More attribute-based results are shown in \textbf{Appendices}.

\myPara{Retraining Experiments.} In Tab.~\ref{tab:retraining_results}, we retrain four representative deep trackers (\ie, SiamFC, ATOM, OSTrack, CiteTracker-256) on the WebUOT-1M. The results indicate that compared with the original models, the retraining models can effectively reduce the domain gap between underwater and open-air environments. This reveals the great value of the proposed WebUOT-1M dataset for developing more powerful deep UOT algorithms.

\myPara{Underwater Vision-Language Tracking.} Previous UOT datasets lack language prompt annotations~\cite{kezebou2019underwater,panetta2021comprehensive,alawode2022utb180,alawode2023improving}. In this work, we perform a pioneering exploration of \emph{underwater vision-language tracking} through carefully annotated language prompts. From Tab.~\ref{tab:vlt_results}, we make the following observations. \textbf{1)} The usage of more cues (\eg, language prompt and bounding box) can significantly boost tracking performance. \textbf{2)} The language prompt-only methods~\cite{ma2024unifying,zhou2023joint} achieve poor results on WebUOT-1M, similar to existing multi-modal open-air tracking datasets~\cite{wang2021towards,zhang2022webuav,zhang2024awesome}, indicating that multi-modal tracking is far from being explored. We expect that the proposed WebUOT-1M dataset can inspire the community to develop multi-modal underwater tracking algorithms.

\vspace{-0.4cm}
\subsection{Ablation Study}
\vspace{-0.2cm}

\myPara{Component Analysis.} The impact of four distillation strategies (\ie, CKD, SKD, FKD, and RKD) is shown in Tab.~\ref{tab:compent_analysis}. Each distillation strategy brings performance improvements compared to the baseline model on WebUOT-1M and UTB180. We can observe that the RKD strategy offers a greater improvement compared to the other three distillation strategies because it directly allows the student model to mimic the response map of the teacher model for target localization.

\begin{table}[t]
\begin{minipage}[c]{0.5\textwidth}
\centering
\vspace{-2.1cm}
\scriptsize
  \caption{\textbf{Retraining with WebUOT-1M.} Pre/\\AUC scores are reported.}
  \label{tab:retraining_results}
  \centering
  \setlength{\tabcolsep}{0.2mm}{
  \scalebox{1.0}{
  \begin{tabular}{lcccc}
    \Xhline{0.75pt} 
      \multicolumn{1}{l}{\multirow{2}[1]{*}{Method}} &  \multicolumn{2}{c}{ UTB180}  & \multicolumn{2}{c}{  WebUOT-1M} \\
   
   \cmidrule(r){2-3} \cmidrule(r){4-5}  &  Original  &  Retraining & Original  &  Retraining \\
    \hline
          
    ATOM~\cite{danelljan2019atom}  & 19.3/31.4 &   25.7/36.5  &   18.9/27.0  & 21.4/32.6    \\
   SiamFC~\cite{bertinetto2016fully}  & 22.3/35.1 &  27.5/39.1    &  22.5/31.6 &   25.8/38.2  \\
   
    OSTrack~\cite{ye2022joint} & 57.0/62.9 &   60.8/63.2  &  52.9/56.5 &   55.1/57.0  \\

    
     CiteTracker-256~\cite{li2023citetracker} & 54.5/61.7 &   61.6/66.3  &  49.3/54.6 &   54.2/57.7  \\

     OKTrack (Ours)  & -/- &   67.3/69.7  &   -/-  & 57.5/60.0    \\
    
    \Xhline{0.75pt} 

  \end{tabular}
  }}~~

\end{minipage}%
\begin{minipage}[c]{0.5\textwidth}
\centering
\scriptsize
\vspace{-1.1cm}
  \caption{\textbf{Vision-language tracking.} SOTA vision-language trackers are compared on WebUOT-1M.}
  \label{tab:vlt_results}
  \centering
  \setlength{\tabcolsep}{1.5mm}{
  \scalebox{1.0}{
  \begin{tabular}{lccccc}
    \Xhline{0.75pt} 

   Method &  Pre  &  nPre & AUC  &  cAUC  &  mACC \\
    \hline
    
  \multicolumn{6}{c}{Language prompt} \\
    \hline

    JointNLT~\cite{zhou2023joint}  & 22.4 &   32.2  &  31.2  &   29.8  &   31.2 \\
         
    UVLTrack~\cite{ma2024unifying} & 22.5 &   33.8  &  31.2  &   30.1  &   31.3 \\
    \hline
    
    \multicolumn{6}{c}{Language prompt + bounding box} \\
    \hline

    JointNLT~\cite{zhou2023joint} & 25.5 &   34.9  &  32.7  &   31.5  &   32.8 \\
    
    VLT$_{\rm SCAR}$~\cite{guo2022divert}  & 33.4 &   44.0  &  37.8  &   36.4  &   38.0 \\

    VLT$_{\rm TT}$~\cite{guo2022divert}  & 41.7 &   52.1  &  48.3  &   47.3  &   48.8 \\

     CiteTracker-256~\cite{li2023citetracker}  & 49.3 &   58.4  &  54.6  &   53.7  &   55.2 \\

    UVLTrack~\cite{ma2024unifying} & 52.5 &   60.0  &  55.8  &  55.0 &   56.6 \\
        
     All-in-One~\cite{zhang2023all}  & 53.1 &   61.5  &  57.1  &   56.4  &   58.0 \\
    \Xhline{0.75pt} 

  \end{tabular}
  }}
\end{minipage}
\begin{minipage}[c]{0.5\textwidth}
\vspace{-1.1cm}
\centering
\scriptsize
  \caption{\textbf{Component analysis.} Pre/AUC scores are reported on UTB180 and WebUOT-1M.}
  \label{tab:compent_analysis}
  \centering
  \setlength{\tabcolsep}{0.7mm}{
  \scalebox{1.0}{
  \begin{tabular}{cccccccc}
    \Xhline{0.75pt} 
    {Base}    &  {CKD}   & {SKD} & {FKD}  & {RKD}  & {MATP}  &  {UTB180} & {WebUOT-1M} \\
    
    \hline
    \cmark &   &     &   &     &     &   62.3/66.6 &    52.0/56.5   \\ 
    
     \cmark &  \cmark &     &   &     &     &   63.6/67.9 &    53.9/57.8   \\ 
     
     \cmark &   &  \cmark   &   &     &     &   63.2/67.4 &    53.5/57.2   \\ 
    
    \cmark &   &     &  \cmark &     &     &   63.2/66.9 &    53.2/57.2   \\  
    
     \cmark &   &     &   &   \cmark  &     &   65.0/68.1 &    54.8/57.9   \\ 
     
     \cmark &  \cmark &  \cmark   &  \cmark &   \cmark  &    &   65.4/68.3 &    55.2/58.3   \\ 

    \cmark &  \cmark &  \cmark   &  \cmark &   \cmark  &  \cmark   &   66.0/68.5 &   56.1/58.9   \\
    \Xhline{0.75pt} 
\vspace{-0.28cm}
  \end{tabular}
  }}
\end{minipage}%
\begin{minipage}[c]{0.5\textwidth}
\scriptsize
  \caption{\textbf{Architecture of student Transformer.} Pre/AUC scores are reported.}
  \label{tab:architecture_of_student}
  \centering
  \setlength{\tabcolsep}{0.9mm}{
  \scalebox{1.0}{
  \begin{tabular}{lccccc}
    \Xhline{0.75pt} 
    {\#Layers}  & {\#Params} & {FLOPs} & {FPS}  &  {UTB180} & {WebUOT-1M} \\
    \hline
     4 layers & 35.4 M  & 10.6 G & 229 &  47.2/57.4 &   39.3/48.0  \\
    
     8 layers & 63.8 M & 16.8 G &  154 &  56.2/63.2 &  47.9/54.0  \\
     
     12 layers  & 92.1 M & 21.5 G &  115  &  66.0/68.5 &   56.1/58.9  \\
    \Xhline{0.75pt} 

  \end{tabular}
  }}
\end{minipage}
\end{table}

\myPara{Analysis on Motion-aware Target Prediction.} The MATP module was introduced to mitigate model drift, as underwater targets (such as fish) are often surrounded by similar distractors. We conduct experiments shown in Tab.~\ref{tab:compent_analysis}. It can be observed that MATP brought gains on UTB180 and WebUOT-1M, verifying the effectiveness of MATP.

\myPara{Architecture of Student Transformer.} The main architecture of the student model is the ViT network. We conduct experiments, shown in Tab.~\ref{tab:architecture_of_student}, to explore the impact of different Transformer layers. We can find that increasing the number of Transformer layers (from 4 to 12) significantly improves performance, but it also increases the number of parameters, model complexity, and reduces speed. To balance performance and cost, we use a student model with 12 Transformer layers.

\begin{figure*}[t]
\begin{minipage}[c]{0.255\textwidth}
\vspace{-0.12cm}
  \centering
  \subfloat{\includegraphics[width =1.0\columnwidth]{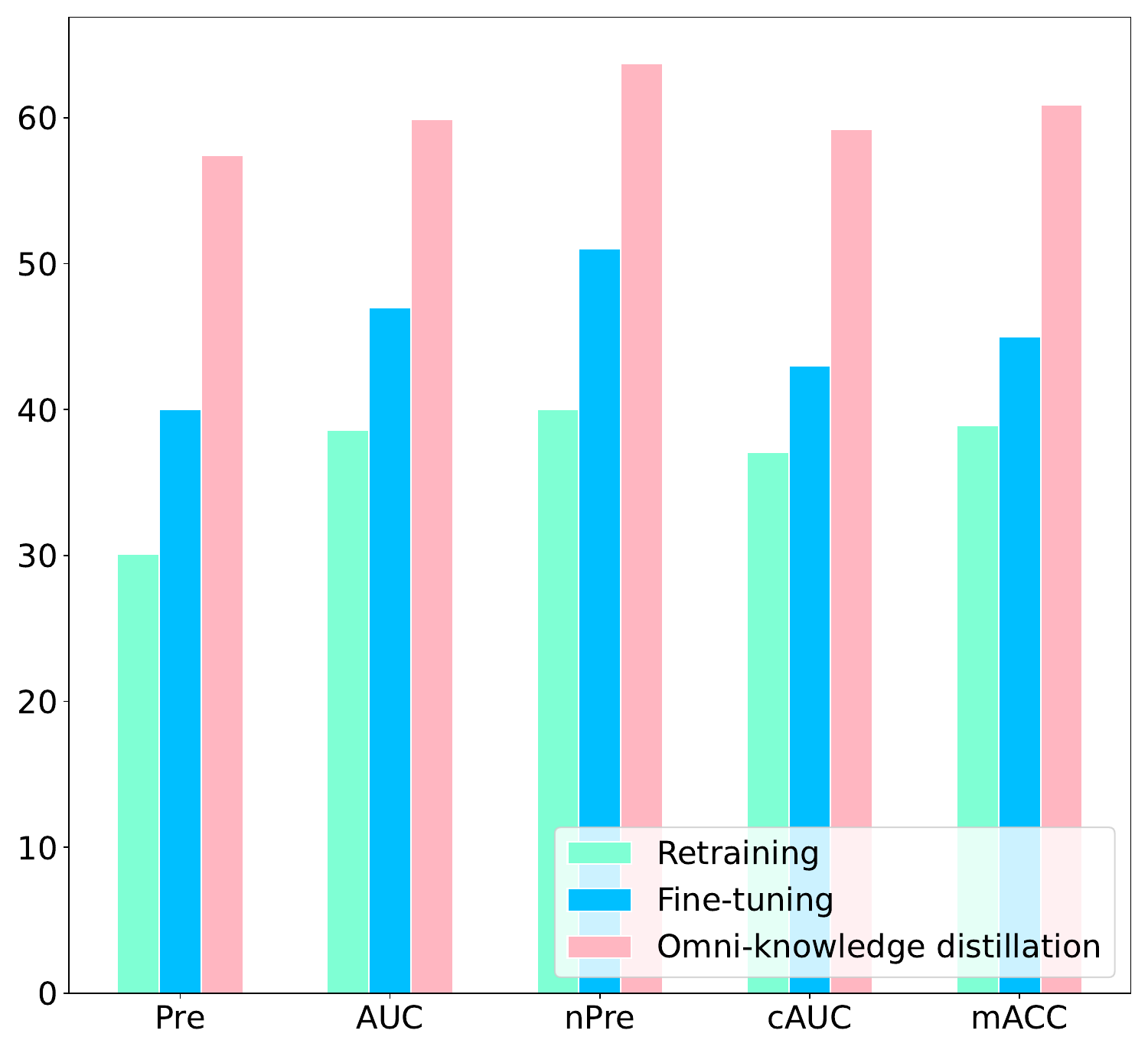}}
  \caption{\textbf{Comparison of three training settings.}}
\label{fig:three_training_settings}
\end{minipage}~
\begin{minipage}[c]{0.75\textwidth}
\vspace{-0.5cm}
  \centering
  \subfloat{\includegraphics[width =0.31\columnwidth]{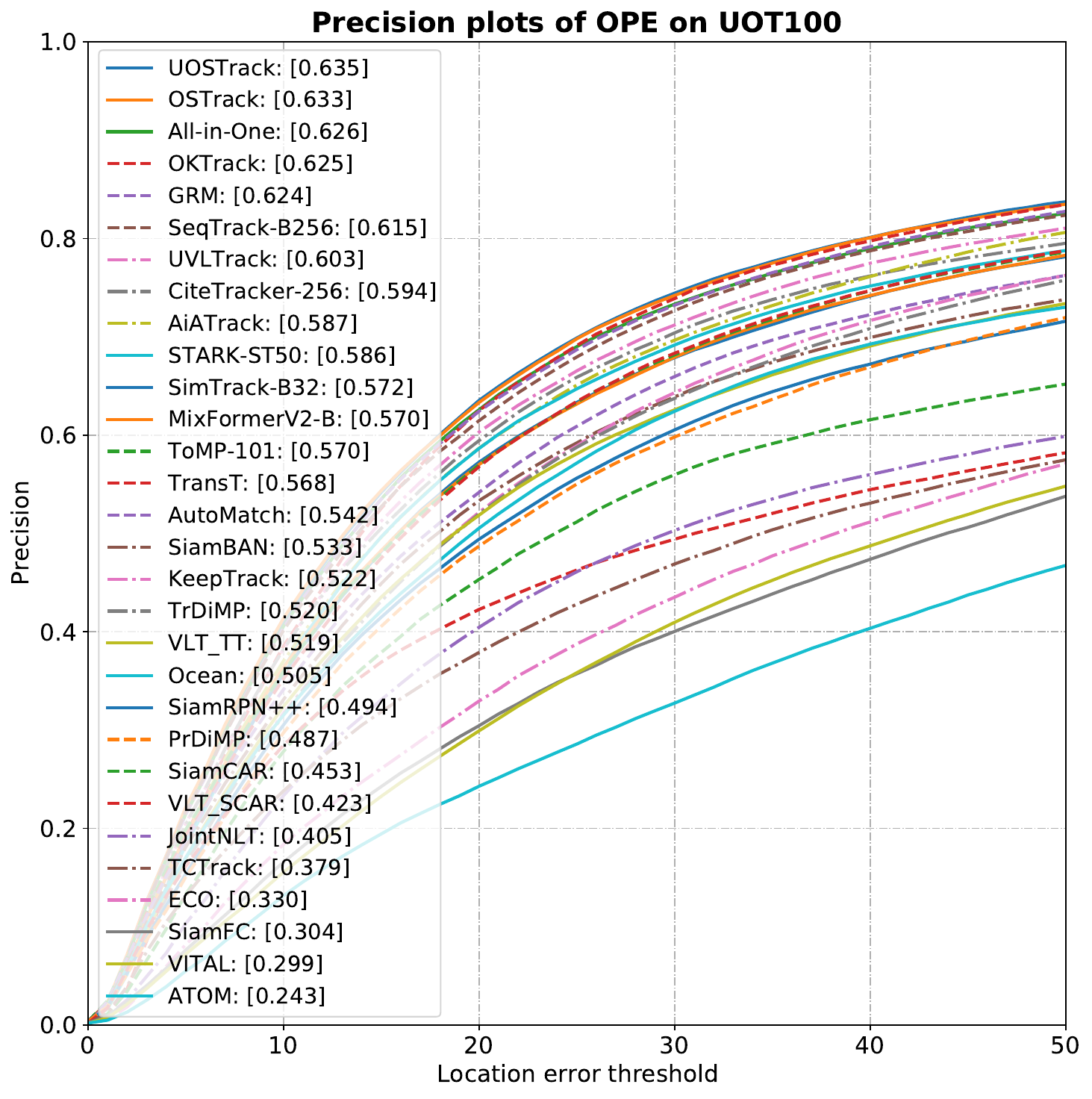}}~
\subfloat{\includegraphics[width =0.31\columnwidth]{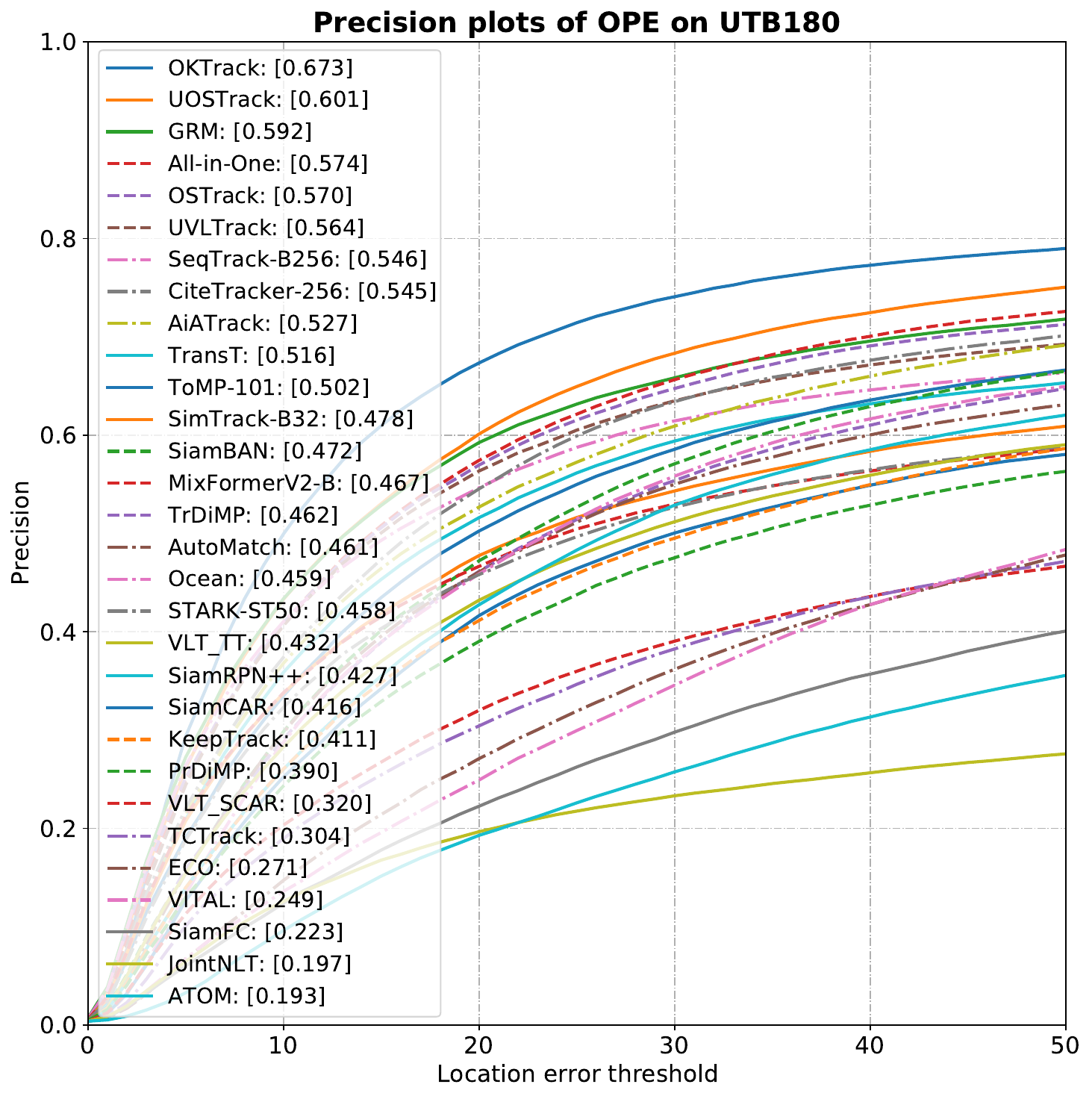}}~
\subfloat{\includegraphics[width =0.31\columnwidth]{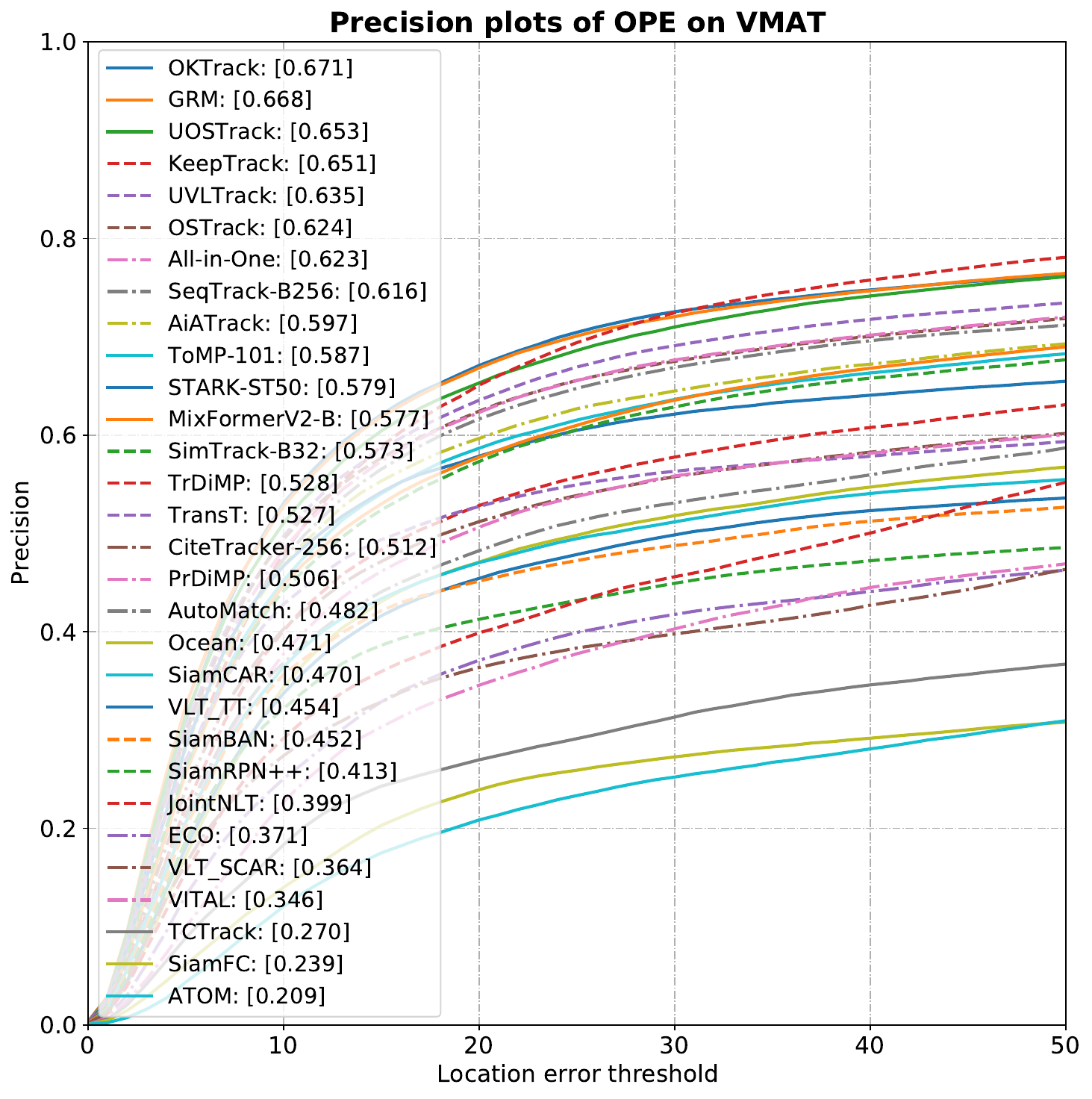}}
  \caption{\textbf{Evaluation on existing UOT benchmarks.} Pre scores are reported on three open-source datasets, \ie, UOT100, UTB180, and VMAT. }
  \label{fig:comparison_to_other_UOT}
\end{minipage}
\vspace{-0.5cm}
\end{figure*}

\vspace{-0.4cm}
\subsection{Empirical Discussions}
\vspace{-0.2cm}

\myPara{Retraining \emph{vs.} Fine-tuning \emph{vs.} Omni-Knowledge Distillation.} In Fig.~\ref{fig:three_training_settings}, we compare three different training settings: \emph{retraining} the student tracker using WebUOT-1M and open-air tracking datasets, \emph{fine-tuning} the student tracker, and adopting \emph{omni-knowledge distillation} for the student tracker on WebUOT-1M. The omni-knowledge distillation achieves the best performance. Fine-tuning the model is preferable to retraining it, as the former can mitigate the issue of insufficient data to some extent, while the latter is limited by the sample imbalance between underwater and open-air objects.

\myPara{Comparison to Other UOT Benchmarks.} We experimentally compare WebUOT-1M with three open-source UOT datasets~\cite{panetta2021comprehensive,alawode2022utb180,cai2023semi}. From Figs.~\ref{fig:results_on_webuot1m} and~\ref{fig:comparison_to_other_UOT}, we obtain some valuable insights. \textbf{1)} OKTrack achieves the best results on UTB180 and VMAT, and a comparable result on UOT100. The possible reason, as noted in~\cite{alawode2022utb180}, is that UOT100 contains a large amount of annotation errors. \textbf{2)} Compared with existing UOT datasets, WebUOT-1M is a more challenging and comprehensive benchmark suitable for both short-term tracking~\cite{panetta2021comprehensive,alawode2022utb180} and long-term tracking~\cite{cai2023semi}. \textbf{3)} The relatively poor result on the long-term tracking dataset VMAT (see Fig.~\ref{fig:comparison_to_other_UOT}) indicates that long-term tracking is still challenging. One solution is to utilize the rich temporal information in video sequences.

\begin{figure*}[t]
\begin{minipage}[c]{0.5\textwidth}
\vspace{-1.55cm}
  \centering
  \subfloat{\includegraphics[width =0.5\columnwidth]{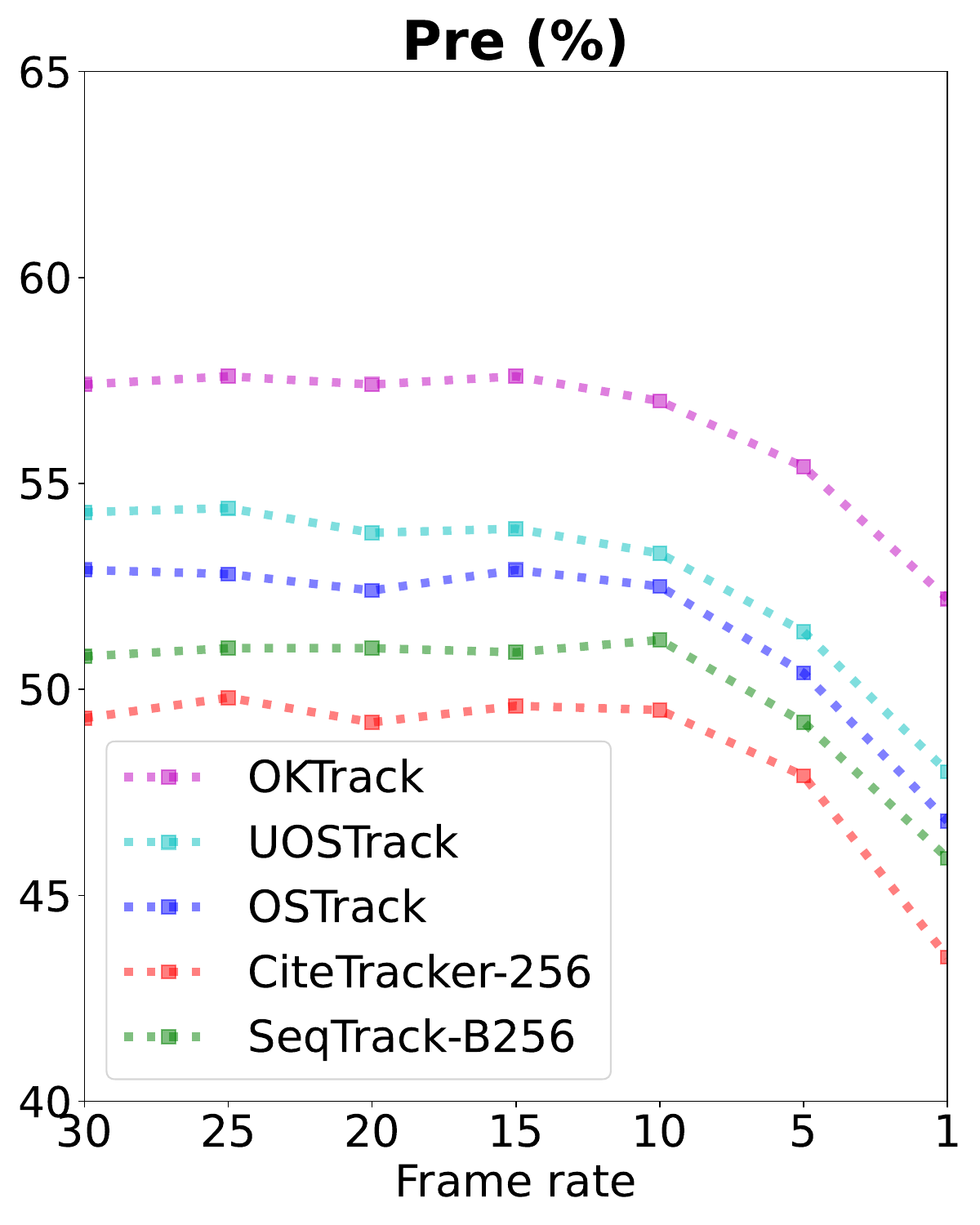}}
\subfloat{\includegraphics[width =0.5\columnwidth]{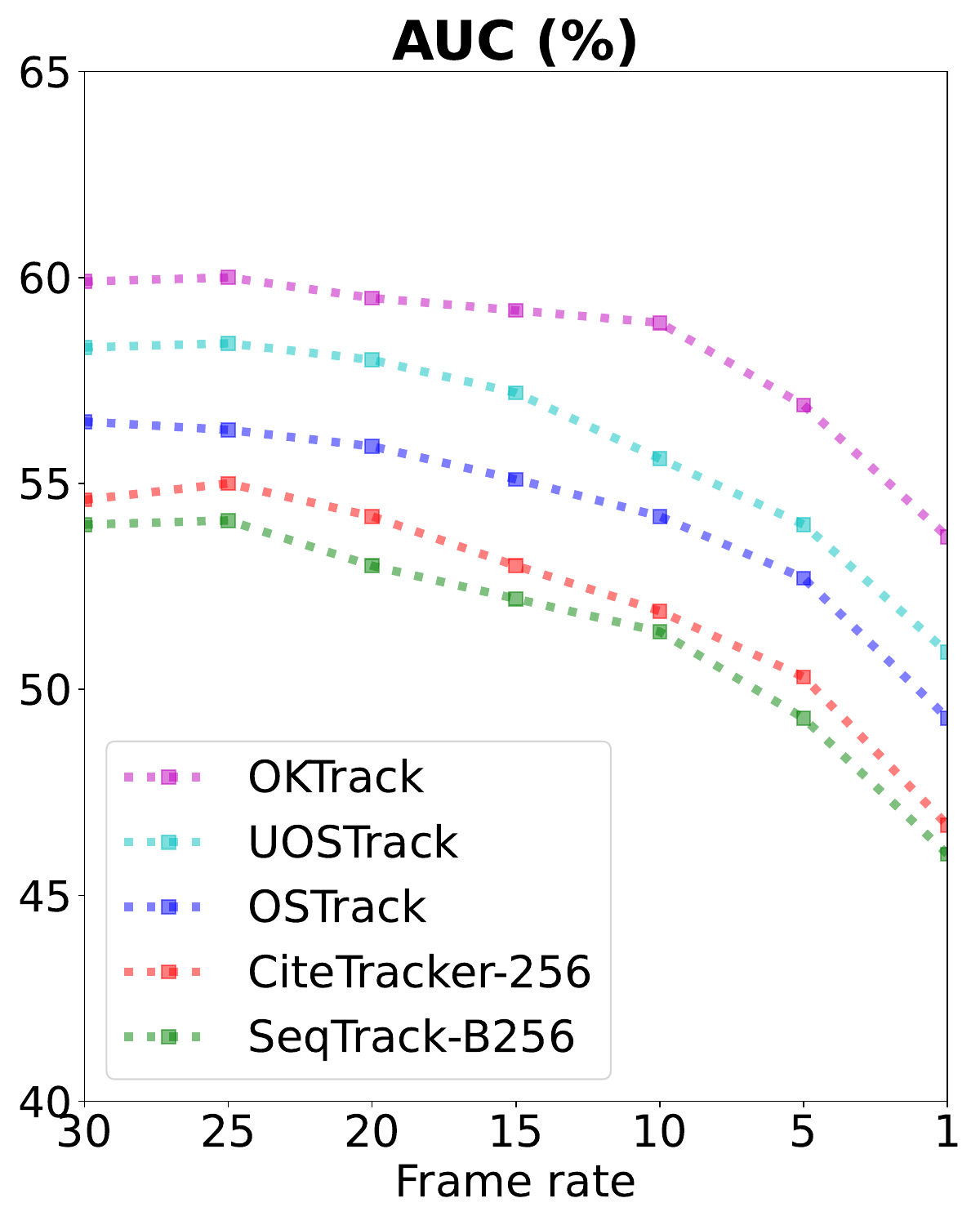}}
    \caption{Stability comparison against frame rate reduction between SOTA methods on WebUOT-1M. Better stability indicates better robustness in various underwater environments.}
  \label{fig:stability_comparison}
  \vspace{-0.5cm}
\end{minipage}~~~
\begin{minipage}[c]{0.43\textwidth}
\vspace{-0.5cm}
  \centering
\subfloat{\includegraphics[width=1.0\linewidth]{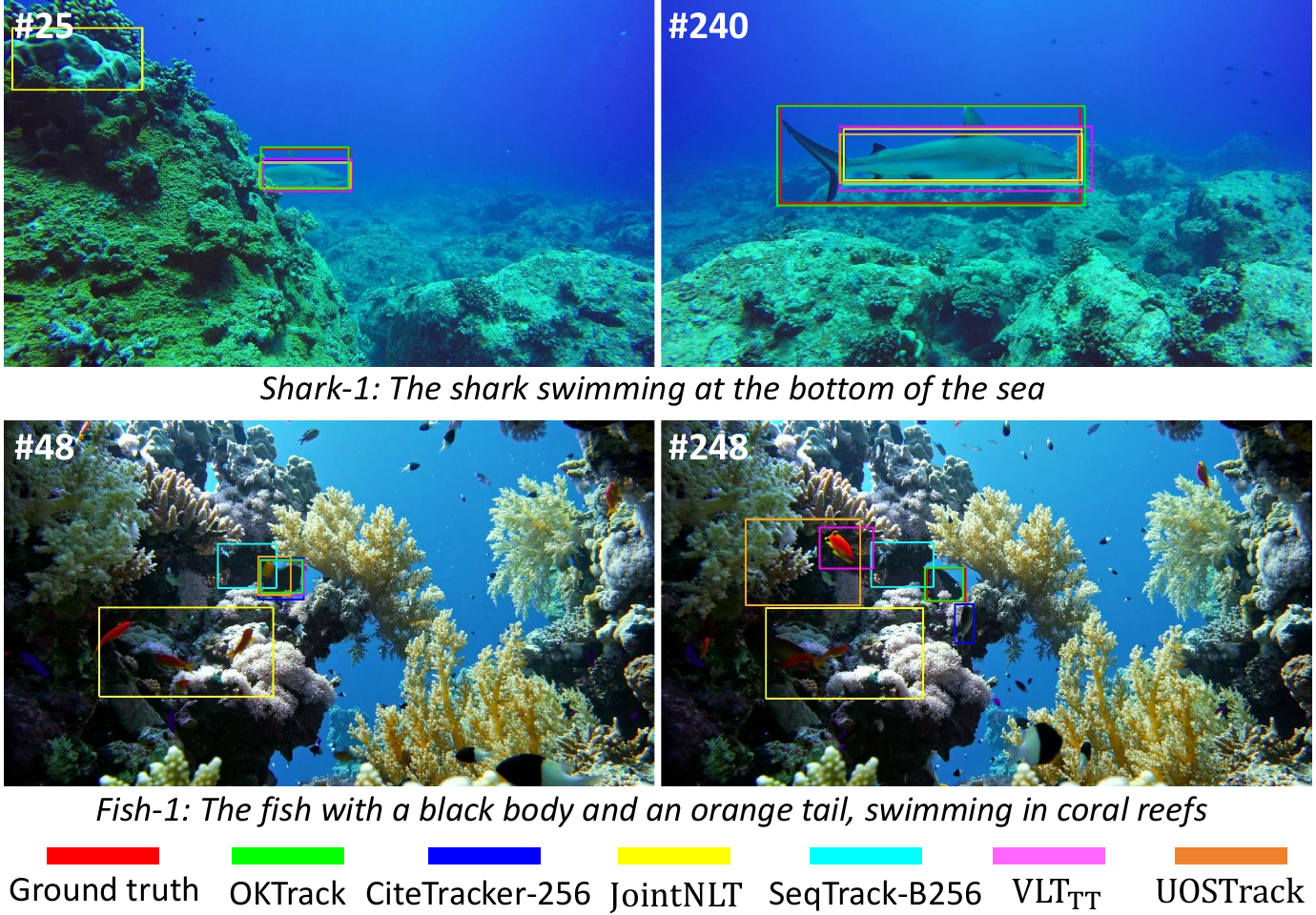}}
  \caption{Qualitative results of six representative deep trackers on WebUOT-1M. OKTrack demonstrates high tracking quality in complex underwater scenarios.}
  \label{fig:Qualitative_results}
\end{minipage}
\vspace{-0.5cm}
\end{figure*}

\myPara{Stability Against Frame Rate Reduction.} In practical applications of UOT, especially in platforms of underwater unmanned robots, the need to save energy or reduce computational load often results in low frame rates~\cite{zheng2024nettrack}, significantly exacerbating the challenges posed by watercolor deviation, blurring, and dynamic targets. To simulate frame rate reduction, we randomly discard some video frames and evaluate the tracking performance of different trackers on the remaining video frames. Fig.\ref{fig:stability_comparison} demonstrates the tracking performance on WebUOT-1M of five deep trackers (\ie, OSTrack, CiteTracker-256, SeqTrack-B256, UOSTrack, OKTrack) with reduced frame rates, from the default frame rate (30 FPS) to the extreme thirtieth (1 FPS). We can observe that the proposed OKTrack exhibits better tracking stability in the face of video frame rate degradation.

\myPara{Tracking in Complex Underwater Scenarios.} As mentioned earlier, compared with open-air tracking, UOT presents many distinct challenges, especially watercolor variations, low underwater visibility, dense similar distractors and camouflage that often appear simultaneously in underwater scenarios. Fig.~\ref{fig:Qualitative_results} shows that the proposed OKTrack can achieve more accurate tracking in complex underwater scenarios, \eg, partial target information and low underwater visibility in \emph{shark-1}, similar distractors and occlusion in \emph{fish-1}, compared to the other five SOTA methods~\cite{zhou2023joint,li2023citetracker,li2023underwater,guo2022divert,chen2023seqtrack}. This is thanks to OKTrack gaining the ability to address multi-view modalities from the teacher model, so it can achieve better performance in underwater scenarios. In addition, the MATP module makes OKTrack more robust to similar distractors, appearance changes, \etc.

\vspace{-0.3cm}
\section{Conclusion and Future Research}
\label{sec:conclusion}
\vspace{-0.3cm}

\myPara{Conclusion.} In this paper, we establish WebUOT-1M, \ie, the first million-scale UOT dataset to facilitate the development of more powerful and versatile tracking systems. It is substantially larger and more diverse than existing UOT datasets, encompassing 1,500 video sequences across 408 object categories. The dataset covers various underwater scenarios and provides rich attributes and language prompts for comprehensive evaluation. Furthermore, a simple yet strong omni-knowledge distillation approach called OKTrack is proposed to boost the research of UOT. Evaluation of 30 deep trackers on WebUOT-1M reveals that Transformer-based and UOT-specific methods perform well. By providing a large-scale dataset, WebUOT-1M not only facilitates the evaluation and comparison of existing tracking algorithms but also paves the way for the development of new methodologies.

\myPara{Future Research.} Although WebUOT-1M significantly surpasses existing UOT datasets in terms of video sequences, target categories and underwater scenarios covered, our data size is still small compared to the latest multi-modal datasets, \eg, LAION-5B~\cite{schuhmann2022laion} and InternVid~\cite{wang2023internvid}. In the future, we consider collecting more underwater videos, and building underwater datasets with more modalities, \eg, depth and audio. By releasing the large-scale WebUOT-1M dataset, we hope it can inspire the community to develop large foundation models for universal object tracking and broader fields, and broaden their application prospects.


{
\small 
\bibliography{WebUOT1M}
\bibliographystyle{plain}
}

\clearpage
\section*{Checklist}



\begin{enumerate}

\item For all authors...
\begin{enumerate}
  \item Do the main claims made in the abstract and introduction accurately reflect the paper's contributions and scope?
    \answerYes{}
  \item Did you describe the limitations of your work?
    \answerYes{Please refer to our appendices.}
  \item Did you discuss any potential negative societal impacts of your work?
    \answerYes{Please refer to our appendices.}
  \item Have you read the ethics review guidelines and ensured that your paper conforms to them?
    \answerYes{}
\end{enumerate}

\item If you are including theoretical results...
\begin{enumerate}
  \item Did you state the full set of assumptions of all theoretical results?
    \answerNA{}
	\item Did you include complete proofs of all theoretical results?
    \answerNA{}
\end{enumerate}

\item If you ran experiments (e.g. for benchmarks)...
\begin{enumerate}
  \item Did you include the code, data, and instructions needed to reproduce the main experimental results (either in the supplemental material or as a URL)?
    \answerYes{Please refer to Sec.~\ref{sec:experiments} and our appendices.}
  \item Did you specify all the training details (e.g., data splits, hyperparameters, how they were chosen)?
    \answerYes{Please refer to the implementation details in Sec.~\ref{sec:experiments} and our appendices.}
	\item Did you report error bars (e.g., with respect to the random seed after running experiments multiple times)?
    \answerYes{Please refer to the error ranges in our appendices.}
	\item Did you include the total amount of compute and the type of resources used (e.g., type of GPUs, internal cluster, or cloud provider)?
    \answerYes{Please refer to the implementation details in Sec.~\ref{sec:experiments} and our appendices.}
\end{enumerate}

    \item If you are using existing assets (e.g., code, data, models) or curating/releasing new assets...
    \begin{enumerate}
    \item If your work uses existing assets, did you cite the creators?
    \answerYes{}
    
    \item Did you mention the license of the assets?
    \answerYes{Please refer to our appendices. The constructed dataset under Creative Commons licenses is intended solely for academic research purposes.}
    
  \item Did you include any new assets either in the supplemental material or as a URL?
    \answerYes{The complete dataset, codes and tracking results, will be made publicly available at {\color{magenta}https://github.com/983632847/Awesome-Multimodal-Object-Tracking}.}
  \item Did you discuss whether and how consent was obtained from people whose data you're using/curating?
    \answerNA{We use publicly available data to establish the WebUOT-1M.}
  \item Did you discuss whether the data you are using/curating contains personally identifiable information or offensive content?
    \answerNA{The WebUOT-1M dataset is specifically designed for underwater object tracking and is meticulously curated to ensure it does not include any identifiable information or offensive content.}
\end{enumerate}

\item If you used crowdsourcing or conducted research with human subjects...
\begin{enumerate}
  \item Did you include the full text of instructions given to participants and screenshots, if applicable?
    \answerNA{}
  \item Did you describe any potential participant risks, with links to Institutional Review Board (IRB) approvals, if applicable?
    \answerNA{}
  \item Did you include the estimated hourly wage paid to participants and the total amount spent on participant compensation?
    \answerNA{}
\end{enumerate}

\end{enumerate}

\clearpage
\appendix
\section*{\Large Appendices}

The appendices present additional details, discussions, and experiment results of our dataset and approach as follows.

\begin{itemize}
    \item \textbf{Appendix~\ref{sec:social_impact} Social Impact.} We present the potential social impacts of our work.
    
    \item \textbf{Appendix~\ref{sec:limitations} Limitations.} We discuss the limitations of our work.
    
    \item \textbf{Appendix~\ref{sec:MoreStatisticsaboutWebUOT1M} More Statistics about WebUOT-1M.} We offer more statistical results and dataset splits of WebUOT-1M.

    \item \textbf{Appendix~\ref{sec:details_of_attributes} Details of Attributes.} We present the definitions and distributions of 23 tracking attributes.
    
    \item \textbf{Appendix~\ref{sec:details_of_Method} Details of Method.} We present more details about the proposed MATP module.

    \item \textbf{Appendix~\ref{sec:additional_discussions} Additional Discussions.}  We perform extensive discussions and analyses on the sample imbalance, inference settings, the role of open-air domain knowledge, the advantages of the OKTrack method, and the differences between our work and existing works.
    
    \item \textbf{Appendix~\ref{sec:experiment_details} Experiment Details.} We present more details of implementation and metrics.
    
    \item \textbf{Appendix~\ref{sec:more_results} More Results.} We demonstrate the error ranges, results on UVOT400, and attribute-based performance on WebUOT-1M.

    \item \textbf{Appendix~\ref{sec:Datasheet} Datasheet.} We provide the datasheet for WebUOT-1M.
    
\end{itemize}

\section{Social Impact} 
\label{sec:social_impact}

The proposed WebUOT-1M dataset\footnote{\url{https://github.com/983632847/Awesome-Multimodal-Object-Tracking}} can promote the research of UOT, which is beneficial for underwater vision understanding, marine environmental monitoring, marine animal conservation, \etc. Despite our best efforts to collect as many target categories as possible, due to the vast diversity of underwater targets in the real world, we still need to be careful about whether models trained on WebUOT-1M can generalize well to \emph{unseen} rare underwater targets. The constructed WebUOT-1M dataset under Creative Commons licenses\footnote{\url{https://creativecommons.org/licenses/}} is intended solely for academic research purposes.

\section{Limitations}
\label{sec:limitations}
One limitation of the proposed method is that it relies on the ViT backbone, which is inherently constrained by the quadratic computational complexity of the self-attention mechanism~\cite{dosovitskiy2020image}. It is interesting to explore more advanced architectures with linear computational complexity, \eg, state space models~\cite{gu2023mamba,zhu2024vision}.

\section{More Statistics about WebUOT-1M}
\label{sec:MoreStatisticsaboutWebUOT1M}

\begin{figure*}[ht]

  \centering
  \subfloat{\includegraphics[width =0.3\columnwidth]{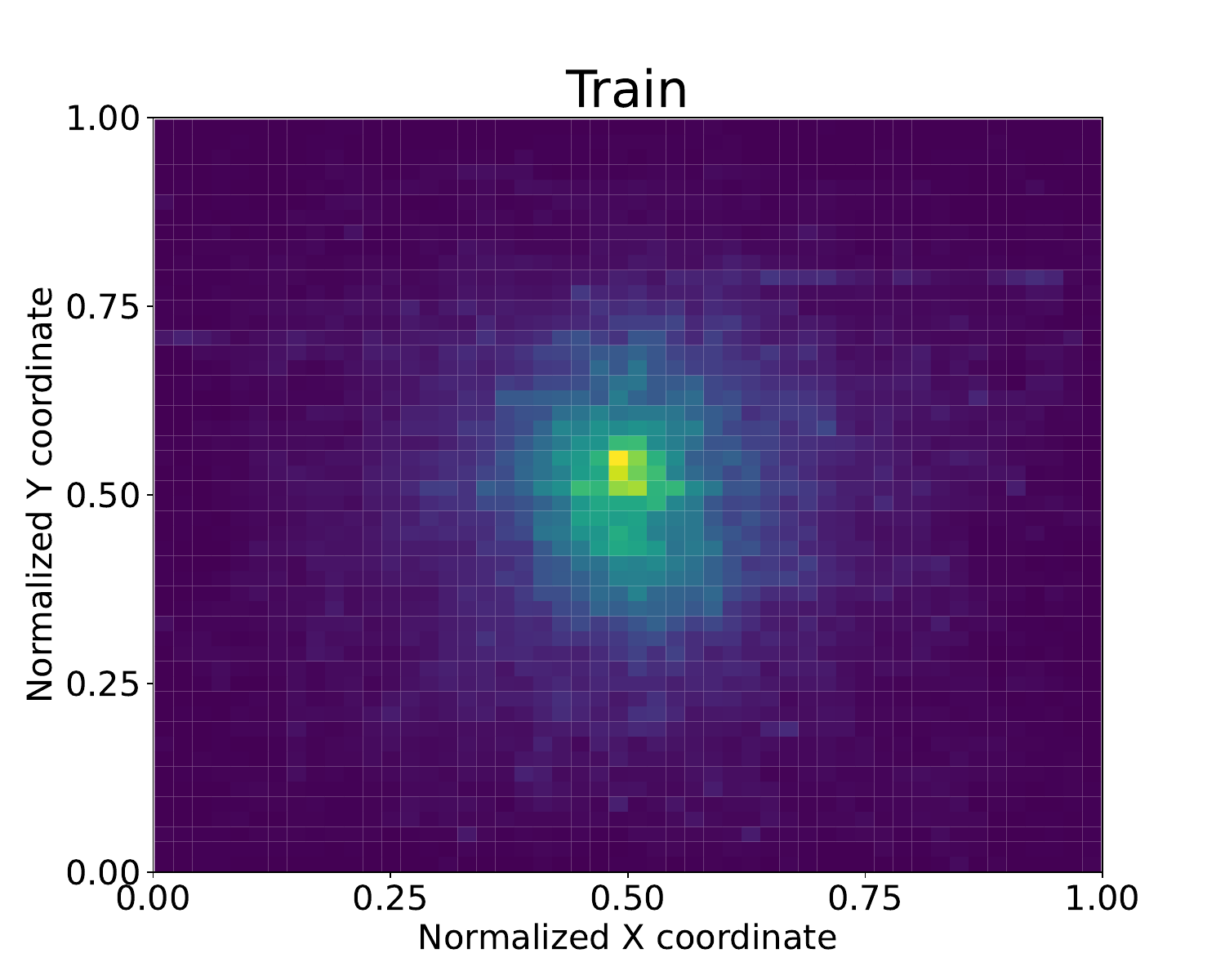}}
~~~~~\subfloat{\includegraphics[width =0.3\columnwidth]{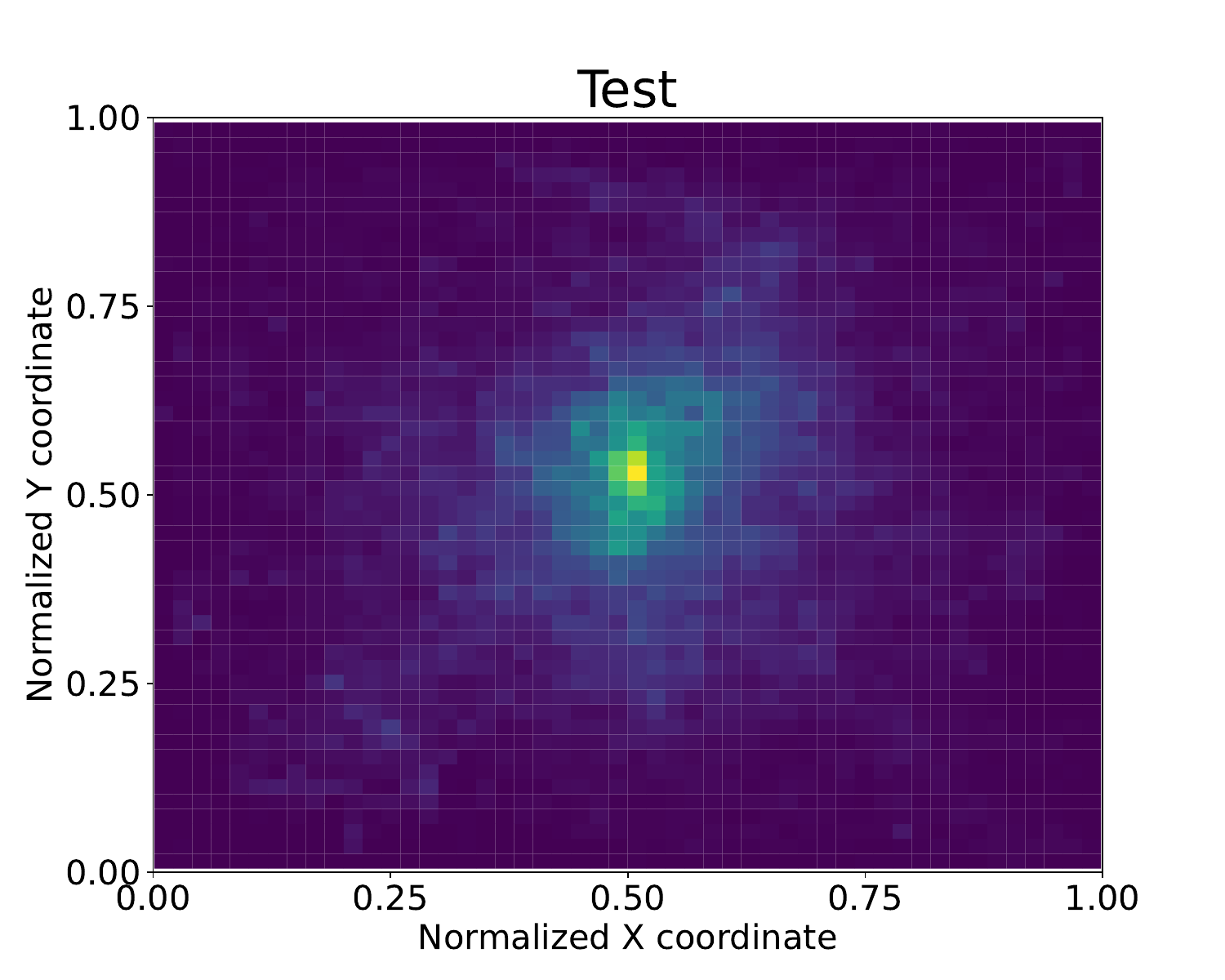}}
~\subfloat{\includegraphics[width =0.355\columnwidth]{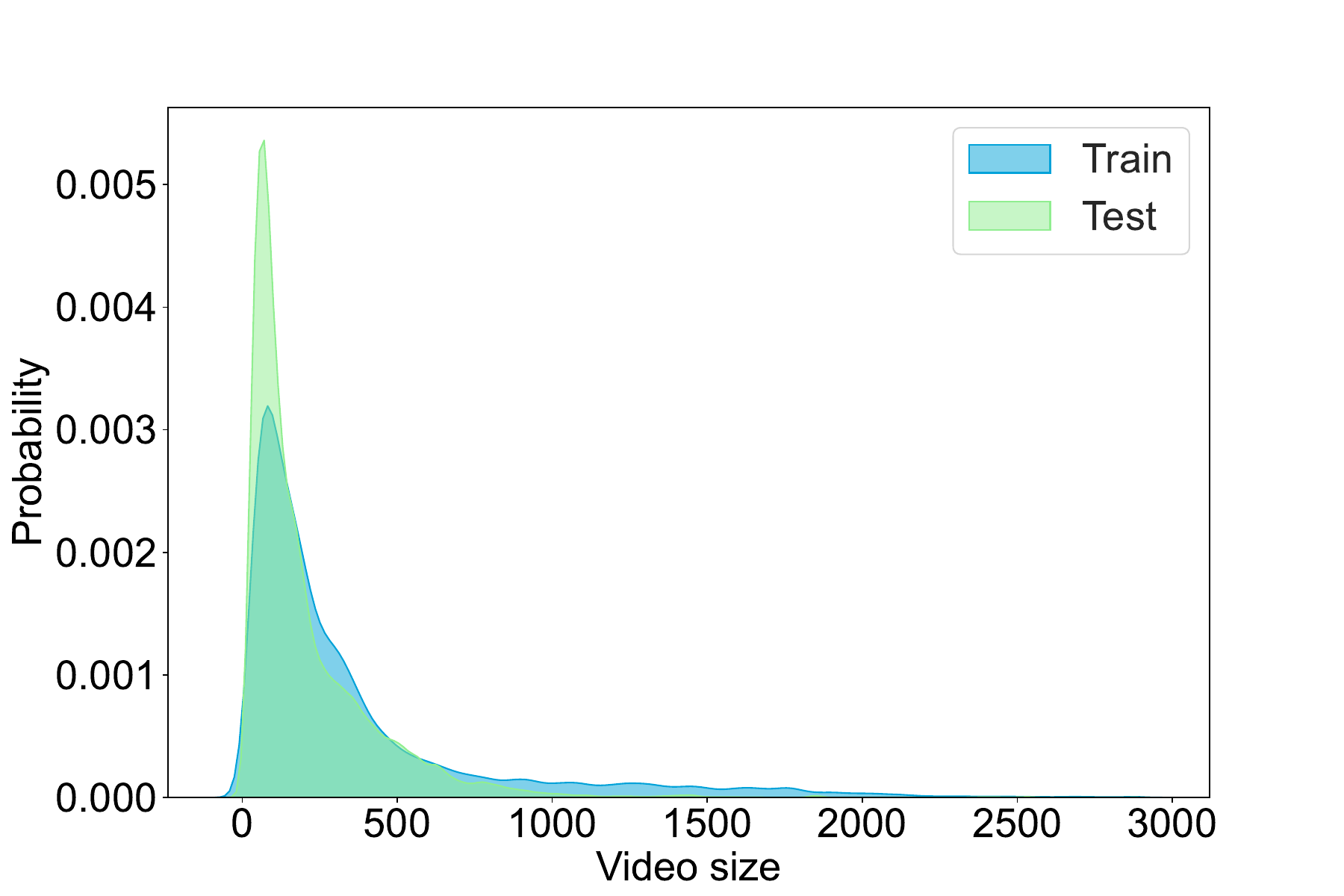}}
  \caption{More statistics of WebUOT-1M. (a) Distribution of target center position of the training set. (b) Distribution of target center position of the test set. (c) Distribution of video size.}
  \label{fig:more_statistics}

\end{figure*}

We provide more statistics to help researchers fully understand and better use the proposed WebUOT-1M dataset. Figs.~\ref{fig:more_statistics}(a) and (b) present the distributes of the target position of training and test sets. The distribution of video size is demonstrated in Fig.~\ref{fig:more_statistics}(c). In Tab.~\ref{tab:training_test_sets}, we compare the training and test sets of WebUOT-1M from multiple aspects, \eg, the number of videos, the number of target categories, the total frames, and the total duration.

\begin{table*}[ht]
	{\footnotesize
	\renewcommand\arraystretch{1.0}
	\caption{Comparison between training and test sets of WebUOT-1M.}
    \vspace{-0.25cm}
	\label{tab:training_test_sets}
	\begin{center}
		\setlength{\tabcolsep}{0.8mm}{
			\scalebox{0.9}{
			\begin{tabular}{lcccccccccccc}
				\Xhline{0.75pt} 
				  &  Videos & Classes  & Superclasses &   \tabincell{c}{ Min  frame}   & \tabincell{c}{Mean frame} & \tabincell{c}{Max frame} & \tabincell{c}{Total frames} &  \tabincell{c}{Total  duration}   
                \\
				
				\hline
                WebUOT-1M test & 480 & 202 & 12  & 49   & 849 & 8,000 & 407 K   & 3.86 hours  \\

				WebUOT-1M training &  1,020 & 287  & 12  & 49  & 680 &  9,985 & 693 K & 6.64 hours \\

                WebUOT-1M &  1,500 & 408  & 12  & 49  & 733 &  9,985 & 1.1 M & 10.5 hours \\
				\Xhline{0.75pt} 
			\end{tabular}
	   }
		}
	\end{center}
    }
\end{table*}

\section{Details of Attributes}
\label{sec:details_of_attributes}

\begin{figure*}[ht]
  \centering
  \includegraphics[width=1.0\linewidth]{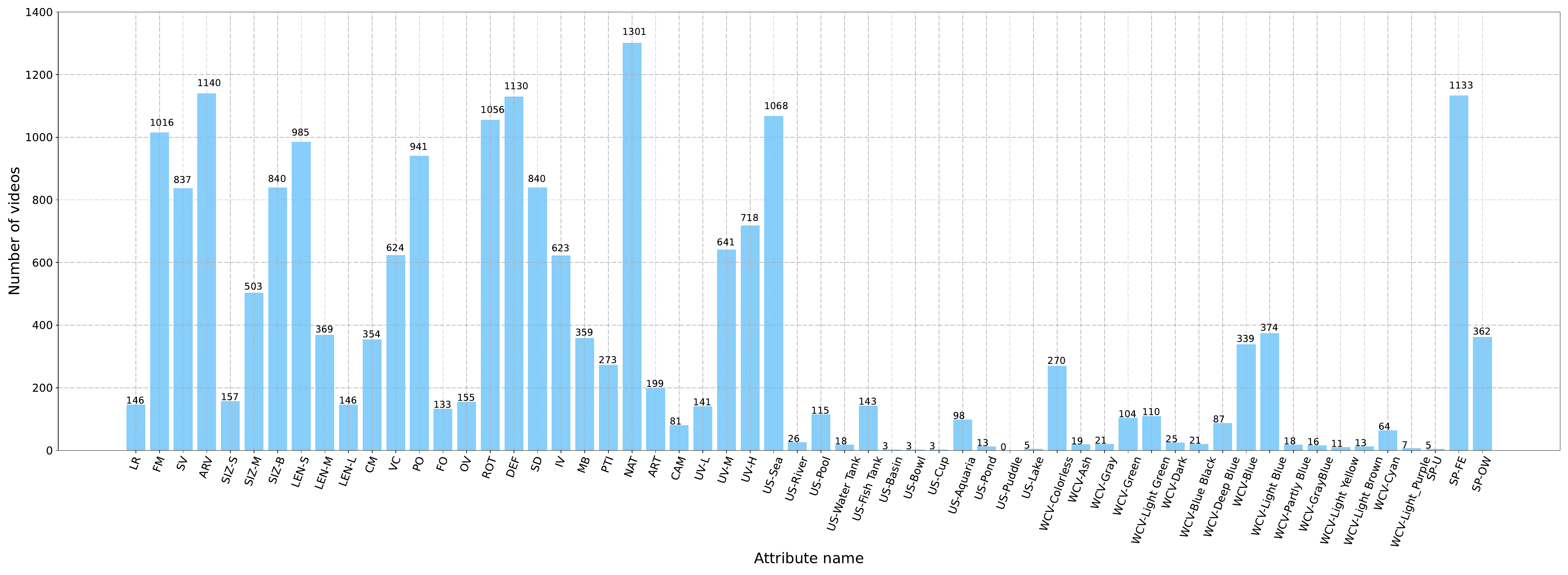}
  \caption{Distribution of videos in each attribute in WebUOT-1M. Best viewed by zooming in.}
  \label{fig:Attribute_Distribution}
\end{figure*}

Tab.~\ref{tab:attribute_definition} demonstrates the definitions of 23 tracking attributes (low resolution (LR), fast motion (FM), scale variations (SV), aspect ratio variations (ARV), camera motion (CM), viewpoint changes (VC), partial occlusion (PO), full occlusion (FO), out-of-view (OV), rotation (ROT), deformation (DEF), similar distractors (SD), illumination variations (IV), motion blur (MB), partial target information (PTI), natural or artificial object (NAO), camouflage (CAM), underwater visibility (UV), watercolor variations (WCV), underwater scenarios (US), shooting perspective (SP), size (SIZ), and length (LEN) of video). We annotate 12 underwater scenarios, including sea, river, lake, pool, water tank, fish tank, basin, bowl, cup, aquarium, pond, and puddle. The WCV contains 16 watercolors (colorless, ash, green, light blue, gray, light green, deep blue, dark, gray-blue, partly blue, light yellow, light brown, blue, cyan, light purple, and blue-black). The distribution of attributes is shown in Fig.~\ref{fig:Attribute_Distribution}.

\begin{table*}[ht]
\caption{Descriptions of the 23 tracking attributes in WebUOT-1M.} \label{tab:attribute_definition} 

  \centering
  \setlength{\tabcolsep}{2.0mm}{
  \scalebox{1.0}{
  \begin{tabular}{ll}
  \Xhline{0.75pt} 

%
%
%
%

\textbf{Attribute} & \textbf{Definition}\\
\hline

\textbf{01. LR} & \makecell[l]{If the size of the bounding box of the target in one frame is less than 400 pixels. }  \\

\textbf{02. FM} & The center position of the target in two consecutive frames exceeds 20 pixels. \\

\textbf{03. SV} & The ratio of the target bounding box is not within the range $[0.5, 2]$.  \\

\textbf{04. ARV} & The aspect ratio of the target bounding box is not in the range $[0.5,2]$. \\

\textbf{05. CM} & There is severe camera movement in the video frame. \\

\textbf{06. VC} & Viewpoint changes significantly affect the appearance of the target. \\

\textbf{07. PO} & If the target appears partially occluded in one frame. \\

\textbf{08. FO} & As long as the target is completely occluded in one frame. \\

\textbf{09. OV} & There is one frame where the target completely leaves the video frame. \\
\textbf{10. ROT} & The target rotates in the video frame. \\

\textbf{11. DEF} & The target appears deformation in the video frame. \\

\textbf{12. SD} & Similarity interference appears around the target. \\

\textbf{13. IV} & The illumination of the target area changes significantly. \\

\textbf{14. MB} & The target area becomes blurred due to target motion or camera motion. \\

\textbf{15. PTI} & In the initial frame only partial information about the target is visible. \\

\textbf{16. NAO} & The target belongs to a natural or artificial object. \\

\textbf{17. CAM} & The target is camouflaging in the video frame. \\

\textbf{18. UV} & The underwater visibility of the target area (low, medium, or high visibility). \\

\textbf{19. WCV} & The color of the water of the target area. \\

\textbf{20. US} & Different underwater scenarios where the target is located.  \\

\textbf{21. SP} & Different shooting perspectives (underwater, outside-water, and fish-eye views). \\

\textbf{22. SIZ} & \makecell[l]{The size $s=\sqrt{w \times h}$ of the video is small ($s<\sqrt{640 \times 480}$),\\ medium ($\sqrt{640 \times 480}\le s < \sqrt{1280 \times 720}$), or large ($s \ge \sqrt{1280 \times 720}$).} \\

\textbf{23. LEN} & \makecell[l]{The length $l$ of the video is short ($l \le 600$  frames),\\ medium ($ 600$ frames $<l \le 1800$ frames), or long ($l > 1800$ frames).} \\ 

    \Xhline{0.75pt} 

  \end{tabular}
  }}
\end{table*}

\section{Details of Method}
\label{sec:details_of_Method}

\subsection{Motion-aware Target Prediction}

The Kalman filtering-based motion-aware target prediction (MATP)~\cite{kalman1960new} is adopted to address tracking drift when the tracker incorrectly locates similar objects. For fast deployment, we borrowed the implementation from SORT~\cite{bewley2016simple} and UOSTrack~\cite{li2023underwater}. Readers are strongly recommended to refer to~\cite{bewley2016simple} and~\cite{li2023underwater} for more details. We summarize the workflow of MATP as follows:
\begin{itemize}
    \item \textbf{Candidate Set Extraction}: The search region is divided into $n\times n$ patches and the top-$N$ patches with the highest similarity scores to the template are extracted as a candidate set $C_{t}$.
    \item \textbf{Trajectory Prediction}: A Kalman filter ($kf$) is utilized to predict the target's position in the current frame, generating the estimation box $b^{E}$.
    \item \textbf{Location Score Calculation}: The location score between $b^{E}$ and each candidate box in $C_{t}$ is calculated using a combination of similarity score and IoU between the boxes.
    \item \textbf{Match Processing}: The tracker first predicts the target location using detection-based post-processing. If the IoU between the predicted box and $b^{E}$ is below a threshold, the tracker employs motion-based match processing. It calculates the location scores between $b^{E}$ and each candidate box in $C_{t}$ and outputs the candidate box with the highest score as the tracked target.
\end{itemize}

In summary, the MATP leverages trajectory prediction and matching to relocate the target hidden in candidate regions when tracking drift occurs. It effectively utilizes motion information and candidate boxes in each frame, providing a new solution to improve tracking performance on similar object challenges for UOT and beyond. Note that MATP does not require training and is used directly during the inference phase. In Algorithm~\ref{alg:alg1}, we provide the pseudo-codes of the tracking model to conduct inference with MATP.

\begin{algorithm}[ht]
	\renewcommand{\algorithmicrequire}{\textbf{Input:}}
	\renewcommand{\algorithmicensure}{\textbf{Output:}}
	\caption{Inference with MATP}
	\label{alg:alg1}
        \begin{algorithmic}[1]
            \REQUIRE Kalman filter $kf$, first\_frame, initial\_box, response map, candidate set $C_{t}$, maximum response set $C'_{t}$, scores list $scores$, estimation box $b^{E}$, match result box $b^{M}$, IoU threshold $conf = 0.6$, response map threshold $threshold = 0.8$, IoU threshold $iou\_threshold = 0.5$, match state $match\_state = False$, \etc
            
            \ENSURE Target boxes ${B}$
            \STATE $kf$.init(first\_frame)
            \STATE $B=[initial\_box]$
            
            \FOR{$i=2,3,...,T$ frames}
                    \STATE $C_{t} \leftarrow$ extract\_candidates(response\_map, $threshold$)
                    \STATE $C'_{t} \leftarrow$ NMS($C_{t}$)
                    \STATE $b^{E} \leftarrow kf$.predict()
                    \STATE $scores \leftarrow$ compute\_scores($b^{E}$, $C'_{t}$)
            
                    \IF{iou\_of(max\_response\_box, $b^{E}$) < $conf$}
                        \STATE $match\_state \leftarrow True$
                    \ELSE
                        \STATE $match\_state \leftarrow False$
                    \ENDIF
            
                    \IF{$match\_state$}
                        \STATE $b^{M} \leftarrow$ argmax($scores$)
                    \ELSE
                        \STATE $b^{M} \leftarrow$ max\_response\_box
                    \ENDIF
                    
                \STATE $B$.append($b^{M}$)
                \STATE $kf$.update($b^{M}$)

            \ENDFOR
            \RETURN $B$
        \end{algorithmic}
\end{algorithm}

\section{Additional Discussions}
\label{sec:additional_discussions}

\subsection{Why is There A Sample Imbalance Between Underwater and Open-air Objects?}

In the field of visual object tracking, commonly used open-air training data consists of approximately 20 M frames, including TrackingNet (14.43 M)~\cite{muller2018trackingnet}, LaSOT (3.52 M)~\cite{fan2019lasot}, GOT-10k (1.5 M)~\cite{huang2019got}, and COCO (118 K)~\cite{lin2014microsoft}. However, the previous largest underwater object tracking (UOT) dataset, \ie, UVOT400, contains only 275 K frames. Considering the ratio of total frames between underwater and open-air datasets is \textbf{1:71}, we argue that there is a significant imbalance between underwater and open-air objects.

\subsection{Why Only Use Underwater Frames for Inference?}

For underwater platforms, \eg, unmanned underwater vehicles, the cameras typically deployed do not have image enhancement capabilities. Adding underwater image enhancement would result in additional energy consumption and latency for these low-power devices. Therefore, a reasonable and low-latency solution is to perform object tracking using only the underwater frames. Moreover, this also follows the evaluation of many existing UOT datasets~\cite{kezebou2019underwater,panetta2021comprehensive,cai2023semi,alawode2023improving,alawode2022utb180}. As shown in Tab.~\ref{tab:underwater_vs_enhanced_frames}, it is not surprising that tracking on enhanced frames can further improve performance. Therefore, developing lightweight and more effective underwater image enhancement algorithms is also a promising direction.

\begin{table*}[ht]
  \caption{Tracking using underwater frames \emph{vs} enhanced frames of OKTrack on WebUOT-1M.}
  \label{tab:underwater_vs_enhanced_frames}
  \centering
  \setlength{\tabcolsep}{2.8mm}{
  \scalebox{1.0}{
  \begin{tabular}{lccccc}
    \Xhline{0.75pt} 
    & Pre (\%) & AUC (\%) & nPre (\%)  &  cAUC (\%) & mACC (\%) \\
    \hline
 
     Underwater frames & 57.5  & 60.0 &  63.8  &  59.3  &   61.0 \\

     Enhanced frames & 58.1 ($\uparrow$0.6)  & 60.4 ($\uparrow$0.4)  &  64.5 ($\uparrow$0.7)   &  59.7 ($\uparrow$0.4)   &  61.3 ($\uparrow$0.3)   \\

    \Xhline{0.75pt} 

  \end{tabular}
  }}
\end{table*}

\subsection{Why is Open-air Domain Knowledge Useful for the UOT Task?}

In our experiments, we used a teacher model pre-trained on large-scale open-air tracking datasets~\cite{fan2019lasot,huang2019got,muller2018trackingnet,lin2014microsoft,li2017tracking,wang2021towards,zhang2022webuav,krishna2017visual} to guide the learning of the student tracker. For the learning of the student model, we utilized the proposed large-scale UOT dataset (WebUOT-1M). We argue that the student model needs to learn two primary abilities to achieve high performance in UOT: \textbf{general feature representation capability} and \textbf{domain-specific (\ie, underwater environment) adaptive capability}. The WebUOT-1M dataset, with its rich variety of categories and comprehensive scene coverage, endows the student model with strong domain adaptation capabilities for the UOT task. Additionally, the open-air domain knowledge possessed by the teacher model (learned from large-scale open-air tracking datasets) is effectively imparted to the student model through the proposed omni-knowledge distillation. Given that current large-scale open-air tracking datasets have a more extensive data scale and cover more target categories and scenes compared to underwater datasets, they are beneficial for the student model to learn general feature representation capabilities. Therefore, open-air domain knowledge is very useful for the UOT task. In the future, we plan to continuously expand the scale of the established WebUOT-1M dataset and use more open-air datasets to further enhance the performance of UOT models.

\begin{table*}[ht]
  \caption{Tracking using different prompts. We compare SOTA trackers using language prompt, language prompt+bounding box prompt, and bounding box prompt on WebUOT-1M. $^{*}$ denotes retraining on the WebUOT-1M training set.}
  \label{tab:L_VL_V_results}
  \centering
  \setlength{\tabcolsep}{3.3mm}{
  \scalebox{1.0}{
  \begin{tabular}{lccccc}
    \Xhline{0.75pt} 

   Method &  Pre (\%)  &  nPre (\%) & AUC (\%) &  cAUC (\%)  &  mACC (\%) \\
    \hline
    
  \multicolumn{6}{c}{Language prompt} \\
    \hline

    JointNLT~\cite{zhou2023joint}  & 22.4 &   32.2  &  31.2  &   29.8  &   31.2 \\
         
    UVLTrack~\cite{ma2024unifying} & 22.5 &   33.8  &  31.2  &   30.1  &   31.3 \\
    \hline
    
    \multicolumn{6}{c}{Language prompt + bounding box} \\
    \hline

    JointNLT~\cite{zhou2023joint} & 25.5 &   34.9  &  32.7  &   31.5  &   32.8 \\
    
    VLT$_{\rm SCAR}$~\cite{guo2022divert}  & 33.4 &   44.0  &  37.8  &   36.4  &   38.0 \\

    VLT$_{\rm TT}$~\cite{guo2022divert}  & 41.7 &   52.1  &  48.3  &   47.3  &   48.8 \\

    UVLTrack~\cite{ma2024unifying} & 52.5 &   60.0  &  55.8  &  55.0 &   56.6 \\
        
    All-in-One~\cite{zhang2023all}  & 53.1 &   61.5  &  57.1  &   56.4  &   58.0 \\ 
    
    CiteTracker-256~\cite{li2023citetracker}  & 49.3 &   58.4  &  54.6  &   53.7  &   55.2 \\
     
    CiteTracker-256$^{*}$~\cite{li2023citetracker}  & 54.2  &   61.6  &  57.7  &   56.9  &   58.5 \\    
     \hline

    \multicolumn{6}{c}{Bounding box} \\  
     \hline

    SeqTrack-B256~\cite{chen2023seqtrack} &  50.8  &    58.5  &  54.0 & 53.3 & 54.7 \\
    
    MixFormerV2-B~\cite{cui2023mixformerv2} &  45.4  &   54.0   &  51.0 & 50.1 & 51.7  \\

    OSTrack~\cite{ye2022joint} &  52.9  &   61.1   &  56.5 & 55.8 & 57.4  \\

    OSTrack$^{*}$~\cite{ye2022joint} &  55.1  &   61.3   &  57.0 & 56.2 & 57.8  \\

    OKTrack (Ours) & 57.5  & 63.8  & 60.0   &  59.3  &   61.0 \\
    \Xhline{0.75pt} 

  \end{tabular}
  }}
\end{table*}

\subsection{Why is Our Purely Visual Approach OKTrack Superior to the Current SOAT Visual Trackers and Vision-Language Tracking Methods?}

In our experiments (see the main paper and Tab.~\ref{tab:L_VL_V_results}), we were astonished to find that the proposed vision-based approach OKTrack surpassed SOTA visual trackers (\eg, OSTrack~\cite{ye2022joint}, SeqTrack-B256~\cite{chen2023seqtrack}, and MixFormerV2-B~\cite{cui2023mixformerv2}) even vision-language trackers (\eg, CiteTracker-256~\cite{li2023citetracker}, All-in-One~\cite{zhang2023all}, and UVLTrack~\cite{ma2024unifying}). We speculate that this is due to several factors:

\begin{enumerate}[(1)]
\item \textbf{There is a significant gap between underwater and open-air domains.} Thus, directly applying existing open-air trackers (including visual trackers and vision-language trackers) to underwater environments leads to performance degradation. To quickly verify our hypothesis, we retrained the current SOTA visual tracker (OSTrack) and vision-language tracker (CiteTracker-256) using our WebUOT-1M dataset. The results (see the main paper and Tab.~\ref{tab:L_VL_V_results}) show that retraining these open-air trackers on our underwater dataset indeed enhances their tracking performance. This is due to retraining reducing the gap between underwater and open domains. However, the proposed OKTrack still outperforms these retrained open-air trackers (both visual and vision-language trackers). \emph{This can be attributed to the effectiveness of the proposed omni-knowledge distillation and MAPT.}

\item \textbf{Limited vision-language tracking datasets.} The existing vision-language tracking datasets (including our WebUOT-1M) are still relatively small (in terms of language annotations), making it challenging to train or fine-tune large visual encoders. \emph{It is a promising direction to construct larger-scale vision-language tracking datasets and benchmarks.}

\item \textbf{Ambiguous language annotations.} Most language annotations in existing vision-language tracking datasets primarily describe the target’s state in the initial frame. In long videos and some complex situations, these descriptions fail to accurately convey the target's current appearance changes. Therefore, training and testing models using the existing language annotations may mislead the models, resulting in decreased tracking performance. \emph{A possible solution is to use existing multi-modal large models to generate more accurate language descriptions for the current vision-language tracking datasets, even to produce multi-granularity language descriptions (\eg, concise and detailed descriptions~\cite{li2024dtllm}) for a single video sequence.}
\end{enumerate}

\subsection{Comparison with Previous Works}

In this work, our primary contribution is the introduction of WebUOT-1M, \ie, the largest and most diverse underwater tracking dataset in terms of target categories and underwater scenarios. Our dataset covers major underwater scenarios, target categories, and underwater instances in existing UOT datasets~\cite{kezebou2019underwater,panetta2021comprehensive,alawode2022utb180,alawode2023improving} and open-air object tracking datasets~\cite{fan2019lasot,fan2021lasot,huang2019got}. The annotated tracking attributes include common attributes (\eg, low resolution, fast motion, and illumination variations) as well as those specific to underwater scenes (\eg, underwater visibility, watercolor variations, and camouflage). Based on the established WebUOT-1M dataset, we further propose a simple yet effective omni-knowledge distillation tracking framework, called OKTrack, for the community. The differences between our approach and existing works (\eg, UVOT400~\cite{alawode2023improving}, HDETrack~\cite{wang2023event}, UOSTrack~\cite{li2023underwater}) are listed as follows:

\begin{itemize}
    \item UVOT400~\cite{alawode2023improving} introduced an underwater tracking dataset consisting of 400 video sequences, 275 K frames, and 17 different tracking attributes and target objects spanning 50 different categories. In comparison, our WebUOT-1M includes 1,500 video sequences, 1.1 million frames, 23 attributes, and 408 target categories. UVOT400 is a partially publicly available dataset, where only the annotations for the first frame of the test set are visible. The complete benchmark, source codes, and tracking results of our work, will be made publicly available.

    \item HDETrack~\cite{wang2023event} proposed a multi-modal knowledge distillation strategy for event-based tracking based on RGB frames and event streams. Drawing inspiration from this method, we present an omni-knowledge distillation framework for underwater tracking. In comparison to HDETrack, our approach exhibits several notable differences, such as the addition of token-based contrastive distillation loss and a motion-aware target prediction module.

    \item UOSTrack~\cite{li2023underwater} presented a hybrid training strategy using both underwater images and open-air sequences to address sample imbalance, alongside employing motion-based post-processing to mitigate the influence of similar targets. We draw inspiration from its motion-based post-processing to address model drift caused by similar distractors, proposing a MATP module. Compared to UOSTrack, we contribute a new million-scale UOT dataset and demonstrate that training on such a dataset significantly enhances tracking performance, which will benefit the entire UOT community. Additionally, we introduce an omni-knowledge distillation framework for UOT.

\end{itemize}

\section{Experiment Details}
\label{sec:experiment_details}

\subsection{More Implementation Details}

In our experimental evaluation, we only consider methods for which the code and model weights are publicly available. For fair comparisons, we use the code, weights, and default parameters provided by the original authors for evaluation. The experimental platform is an Ubuntu 20.04 server with two Intel(R) Xeon(R) Gold 6226R CPUs @ 2.90GHz, 8 NVIDIA A6000 GPUs and 512G Memory. Python 3.8.0 and PyTorch 2.0.1 are mainly used in our experiments.

For the proposed OKTrack, the template and the search region are $2^{2}$ times and $4^{2}$ times of the target bounding box, and then resized to $128\times128$ and $256\times256$, respectively. In underwater scenarios, using a larger template and search region may enhance tracking performance further, but it also entails increased computational costs. Additionally, due to the presence of dense similar distractors (\eg, schools of fish) around target objects in underwater environments, enlarging the search region also increases the risk of model drift. Therefore, to reduce computational costs in underwater scenarios and achieve a fair comparison, for some SOTA trackers, we only consider their versions with a search region of $256\times256$, \eg, OSTrack~\cite{ye2022joint}, SimTrack-B32~\cite{chen2022backbone}, MixFormerV2-B~\cite{cui2023mixformerv2}, SeqTrack-B256~\cite{chen2023seqtrack}, and CiteTracker-256~\cite{li2023citetracker}. The teacher tracker~\cite{zhang2023all} was trained using eight commonly used open-air tracking datasets (LaSOT~\cite{fan2019lasot}, GOT-10k~\cite{huang2019got}, TrackingNet~\cite{muller2018trackingnet}, COCO~\cite{lin2014microsoft}, OTB99-L~\cite{li2017tracking}, TNL2K~\cite{wang2021towards}, WebUAV-3M~\cite{zhang2022webuav}, and VisualGenome~\cite{krishna2017visual}). In our experiments, we directly utilize the pre-trained weights of the teacher tracker. Following UOSTrack~\cite{li2023underwater}, two underwater object detection datasets (\ie, RUOD~\cite{fu2023rethinking}, FishExtend~\cite{li2023underwater}) are used to enhance the generalization of the tracking models.

\subsection{Metrics Details}

Tab.~\ref{tab:metric_details} provides some descriptions of the adopted five evaluation metrics, \ie, precision (Pre), normalized precision (nPre), success rate (AUC), complete success rate (cAUC), and mean accuracy (mACC). Readers are referred to~\cite{muller2018trackingnet,fan2019lasot,jiang2021anti,zhang2022webuav} for more details on each metric.

\begin{table*}[ht]
  \caption{Descriptions of five evaluation metrics.}
  \label{tab:metric_details}
  \centering
  \setlength{\tabcolsep}{2.0mm}{
  \scalebox{0.96}{
  \begin{tabular}{ll}
    \Xhline{0.75pt} 
    \textbf{Metric}  & \textbf{Description} \\
    \hline
     \textbf{01. Pre} & \makecell[l]{The Pre is used to measure the percentage of frames where the center position error \\falls within a predefined threshold. Trackers are ranked based on this metric using a\\ given precision score (\eg, obtained when the threshold = 20 pixels). } \\

    \textbf{02. nPre} & \makecell[l]{As Pre is sensitive to target size and image resolution, nPre is introduced in~\cite{muller2018trackingnet}, \\which normalizes each precision score over the size of the ground truth bounding box. } \\

    \textbf{03. AUC} & \makecell[l]{The AUC indicates the percentage of frames with overlap scores higher than a given\\ threshold. Trackers are ranked based on this metric using the area under the curve \\(between 0 and 1) of each success plot. } \\

    \textbf{04. mACC} & \makecell[l]{The mACC measure proposed in~\cite{jiang2021anti}, encourages trackers to provide reliable\\ predictions for the target object even when it disappears. } \\

    \textbf{05. cAUC} & \makecell[l]{The above four metrics only measure center-point distance or overlap area and do \\not reflect the aspect ratio of the target object. To address this,~\cite{zhang2022webuav} introduced the \\cAUC evaluation metric. Like the AUC, the cAUC is defined as the proportion of \\frames where the complete overlap score exceeds a specified threshold. } \\
    \Xhline{0.75pt} 

  \end{tabular}
  }}
\end{table*}

\section{More Results}
\label{sec:more_results}

\subsection{Error Ranges} 

Following popular UOT and open-air tracking benchmarks~\cite{zhang2022webuav,fan2019lasot,fan2021lasot,wang2021towards,kezebou2019underwater,panetta2021comprehensive,alawode2022utb180,alawode2023improving,zhang2024awesome}, we perform the one-pass evaluation (OPE) for different tracking algorithms. To further verify the stability of different tracking algorithms, we conduct multiple tests using the Top-5 algorithms (\ie, OKTrack, UOSTrack, All-in-One, GRM, OSTrack) on WebUOT-1M to obtain their error ranges. The results are shown in Tab.~\ref{tab:Multiple_results}. We can see that these SOTA methods have very small fluctuations in performance across multiple tests.

\begin{table*}[ht]
  \caption{Error range of Top-5 trackers on WebUOT-1M.}
  \label{tab:Multiple_results}
  \centering
  \setlength{\tabcolsep}{4.0mm}{
  \scalebox{1.0}{
  \begin{tabular}{lccccc}
    \Xhline{0.75pt} 
    {Method}  & Pre (\%) & AUC (\%) & nPre (\%)  &  cAUC (\%) & mACC (\%) \\
    \hline
     OSTrack & 52.9$_{\pm 0.4}$   & 56.5$_{\pm 0.4}$ &  61.1$_{\pm 0.7}$  &  55.8$_{\pm 0.4}$  &  57.4$_{\pm 0.4}$\\

     GRM & 53.8$_{\pm 0.3}$   & 56.7$_{\pm 0.1}$ &  61.0$_{\pm 0.2}$  & 56.0$_{\pm 0.2}$ &  57.6$_{\pm 0.2}$ \\

     All-in-One & 53.1$_{\pm 0.2}$   & 57.1$_{\pm 0.2}$ &  61.5$_{\pm 0.1}$  &  56.4$_{\pm 0.1}$ &   58.0$_{\pm 0.2}$  \\

     UOSTrack & 54.3$_{\pm 0.3}$   & 58.3$_{\pm 0.2}$ &  62.6$_{\pm 0.1}$  &  57.5$_{\pm 0.2}$ &   59.1$_{\pm 0.2}$  \\

     OKTrack & 57.5$_{\pm 0.1}$   & 60.0$_{\pm 0.1}$ &  63.8$_{\pm 0.1}$  &  59.3$_{\pm 0.2}$  &   61.0$_{\pm 0.2}$  \\

    \Xhline{0.75pt} 

  \end{tabular}
  }}
\end{table*}

\subsection{Results on UVOT400} 

To further validate the effectiveness of the proposed OKTrack, we show the tracking performance on the UVOT400 test set~\cite{alawode2023improving} in Tab.~\ref{tab:UVOT400_results}. Since the annotations for the UVOT400 test set are not visible, we submit the tracking results of UOSTrack and OKTrack to the official evaluation server to obtain AUC, nPre, and Pre scores. Results indicate that OKTrack achieves the best performance, with $63.2\%$ in terms of AUC, $66.4\%$ in terms of nPre, and $58.4\%$ in terms of Pre. Compared to the previous best UOT tracker (UOSTrack), the gains of the proposed OKTrack are $1.8\%$, $1.6\%$, and $2.8\%$ in terms of AUC, nPre, and Pre scores, respectively.

\begin{table*}[ht]
  \caption{Evaluation on UVOT400 test set. The reported results come from~\cite{alawode2023improving} or from our submissions to the official evaluation server.}
  \label{tab:UVOT400_results}
  \centering
  \setlength{\tabcolsep}{6.0mm}{
  \scalebox{1.0}{
  \begin{tabular}{lccclccc}
    \Xhline{0.75pt} 
    {Method}   & AUC (\%) & nPre (\%)   & Pre (\%)   \\
    \hline
    SiamFC~\cite{bertinetto2016fully}  &   29.6   &   36.2  &  24.8    \\

    SiamCAR~\cite{guo2020siamcar}  &   41.6   &   50.7  &  40.6  \\

    PrDiMP~\cite{danelljan2020probabilistic}  &   42.0   &   50.0  &  36.6  \\

    ATOM~\cite{danelljan2019atom}  &   43.3   &   51.7  &  37.6\\     

    STARK~\cite{yan2021learning}  &   43.4   &   49.9  &  40.4   \\

    AutoMatch~\cite{zhang2021learn}  &   48.6   &   59.8  &  47.0 \\

    KeepTrack~\cite{mayer2021learning}   &   49.4   &   59.0  &  44.1   \\

    SiamBAN~\cite{ChenZLZJ20}  &   49.8   &   61.2  &  47.6 \\
    
    TransT~\cite{chen2021transformer}   &   51.4   &   60.1  &  49.4 \\
    
    TrDiMP~\cite{wang2021transformer}  &   52.2   &   61.9  &  47.3 \\

    ToMP-101~\cite{mayer2022transforming}   &   53.9   &   63.7  &  51.4  \\
     
    \hline
    UOSTrack~\cite{li2023underwater} &  61.4   &  64.8   &   55.6  \\
    
    OKTrack (Ours) &   63.2 &  66.4  &   58.4  \\

    \Xhline{0.75pt} 

  \end{tabular}
  }}
\end{table*}




\subsection{Detailed Attribute-based Performance on WebUOT-1M}

\begin{figure*}[t]
\vspace{-0.6cm}
\centering

\subfloat{\includegraphics[width =0.25\columnwidth]{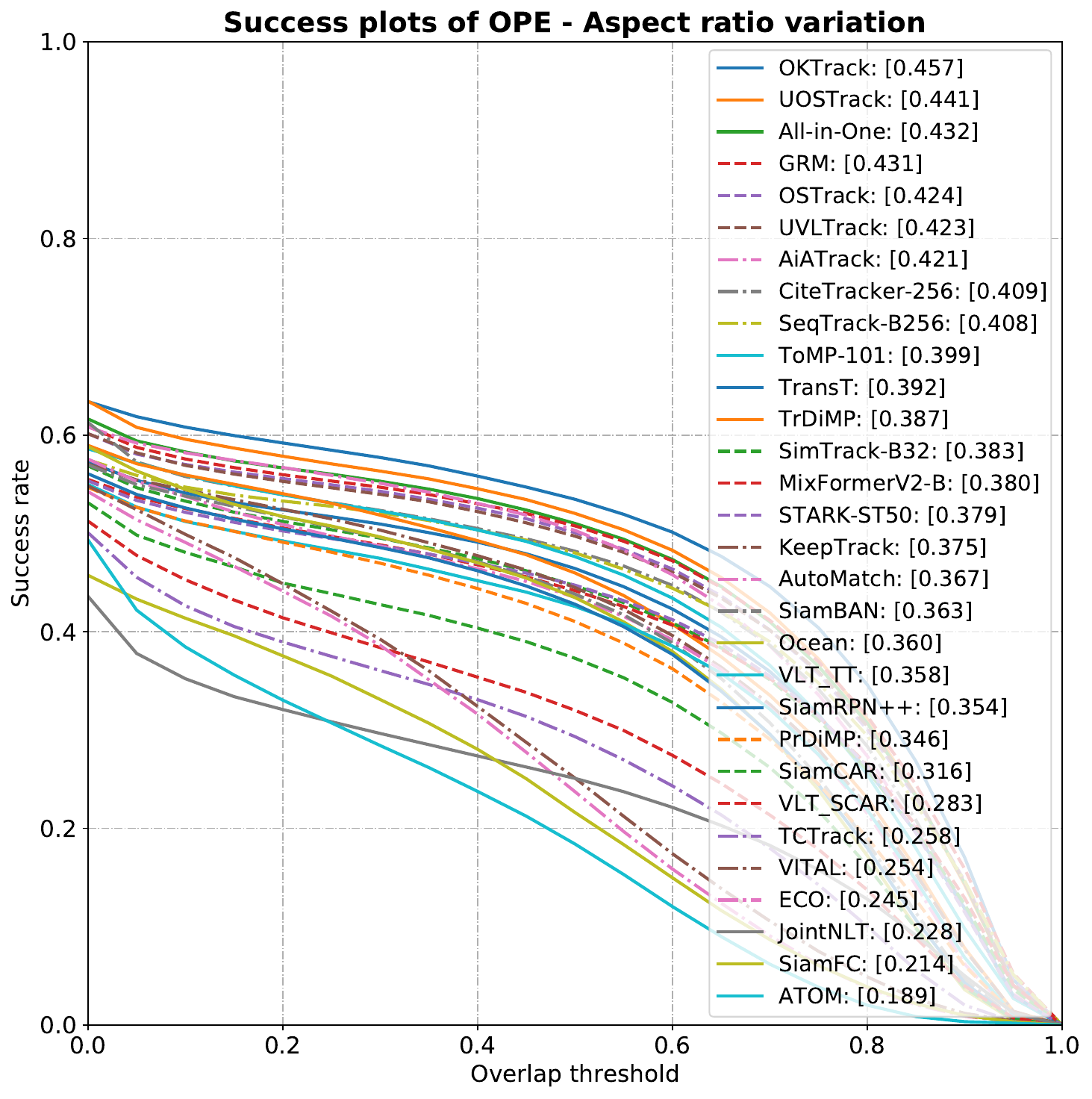}}
\subfloat{\includegraphics[width =0.25\columnwidth]{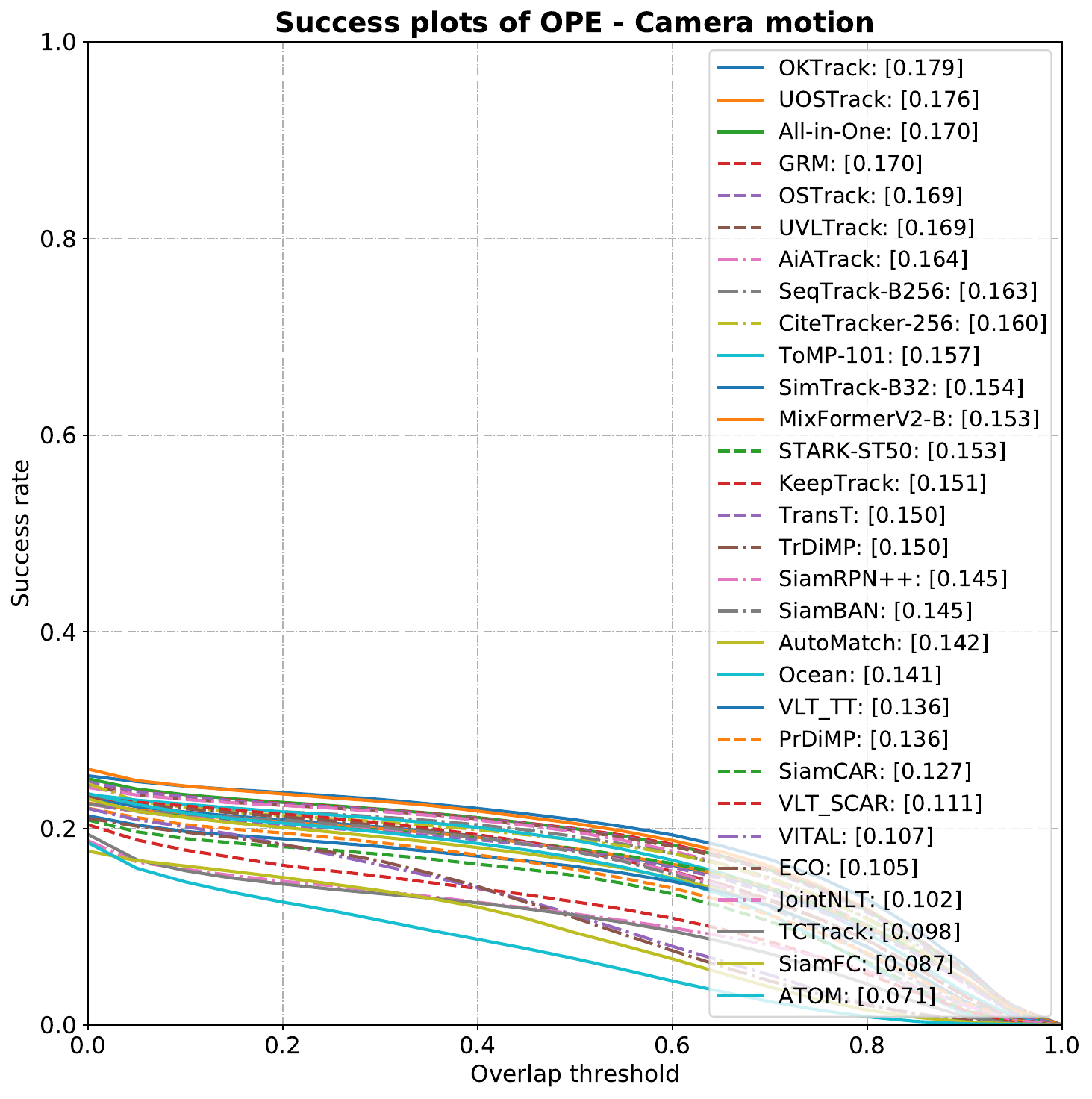}}
\subfloat{\includegraphics[width =0.25\columnwidth]{figs/Attributes-AUC/success_plots_Camouflage.pdf}}
\subfloat{\includegraphics[width =0.25\columnwidth]{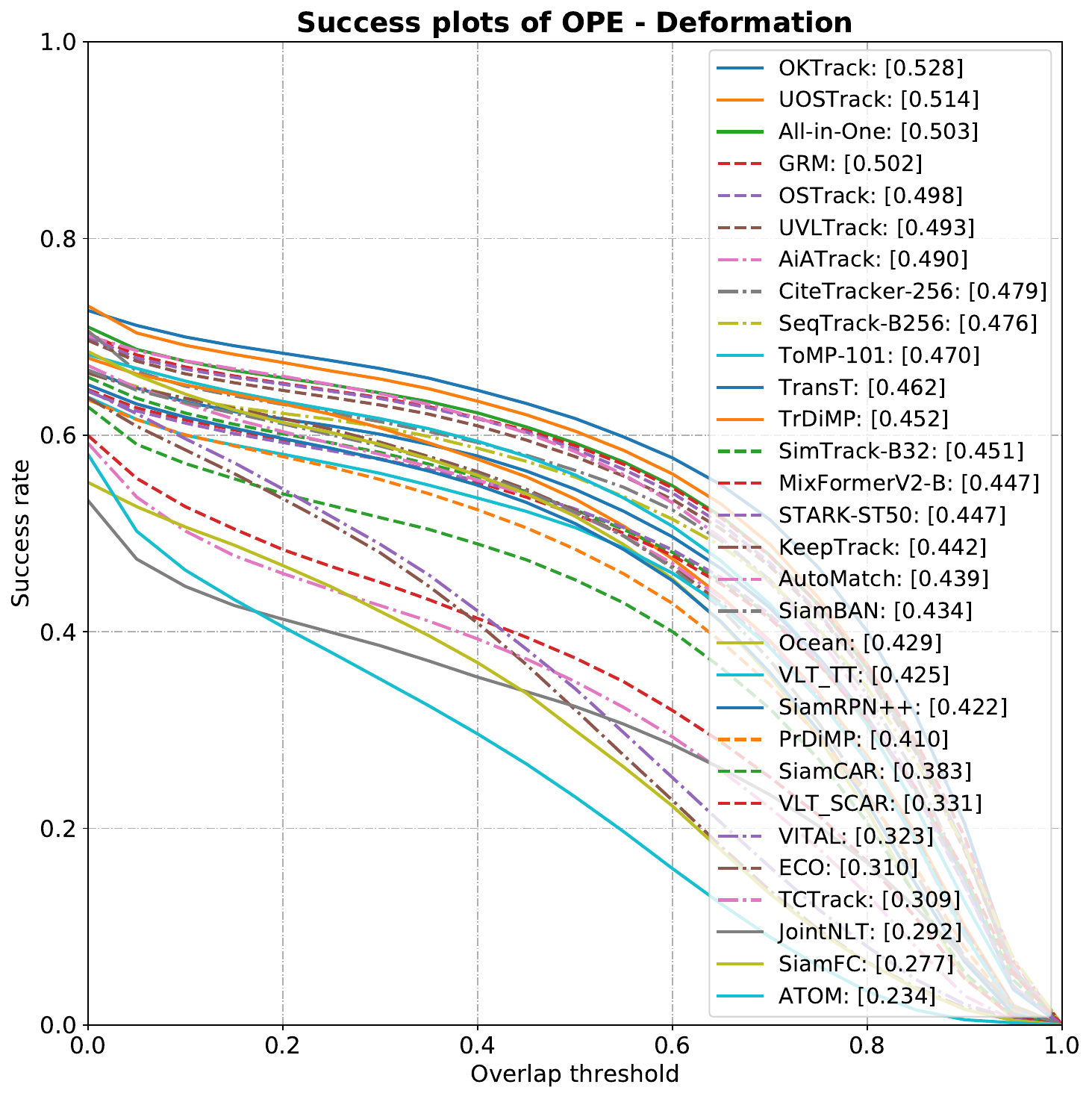}}

\subfloat{\includegraphics[width =0.25\columnwidth]{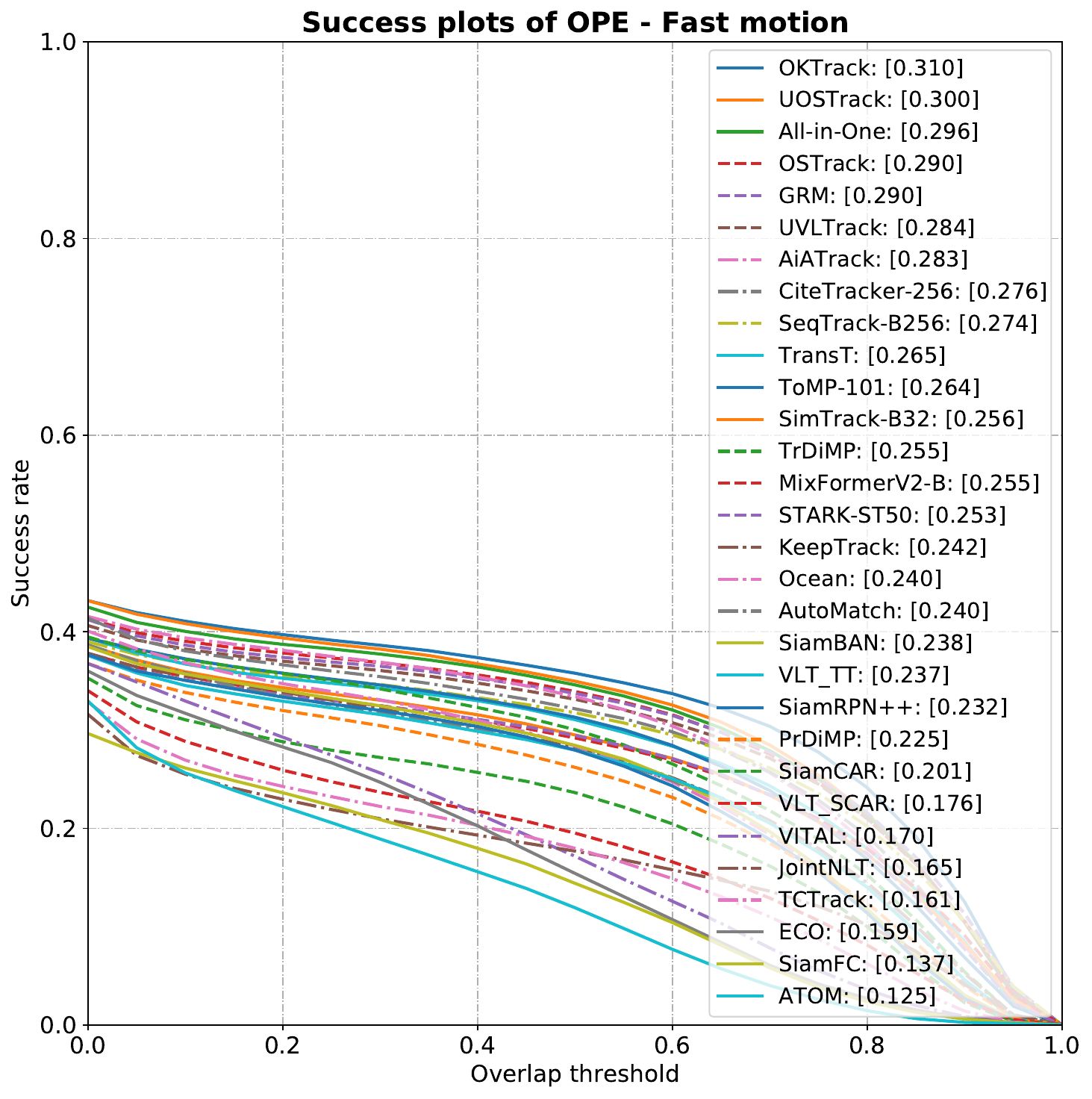}}
\subfloat{\includegraphics[width =0.25\columnwidth]{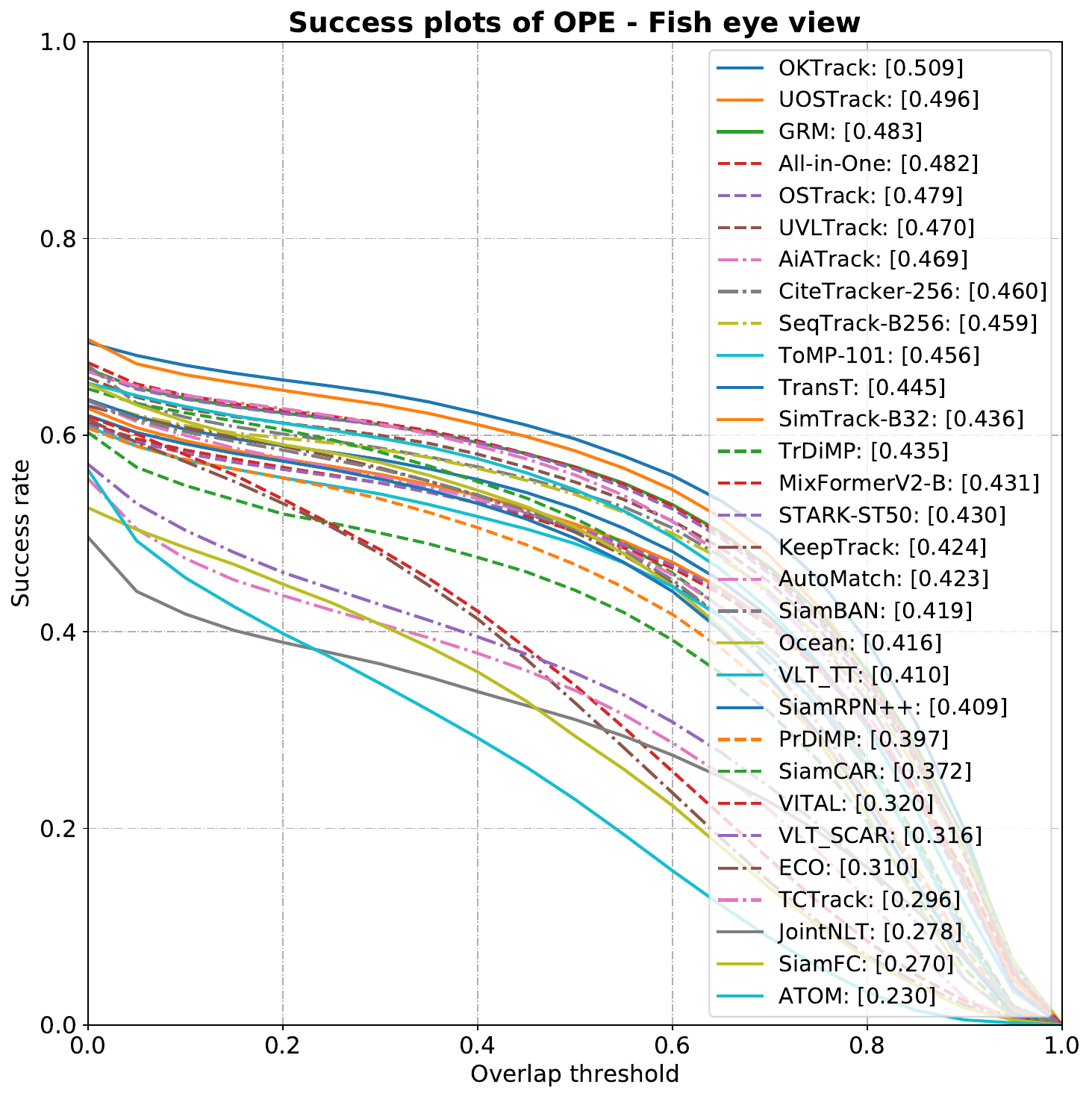}}
\subfloat{\includegraphics[width =0.25\columnwidth]{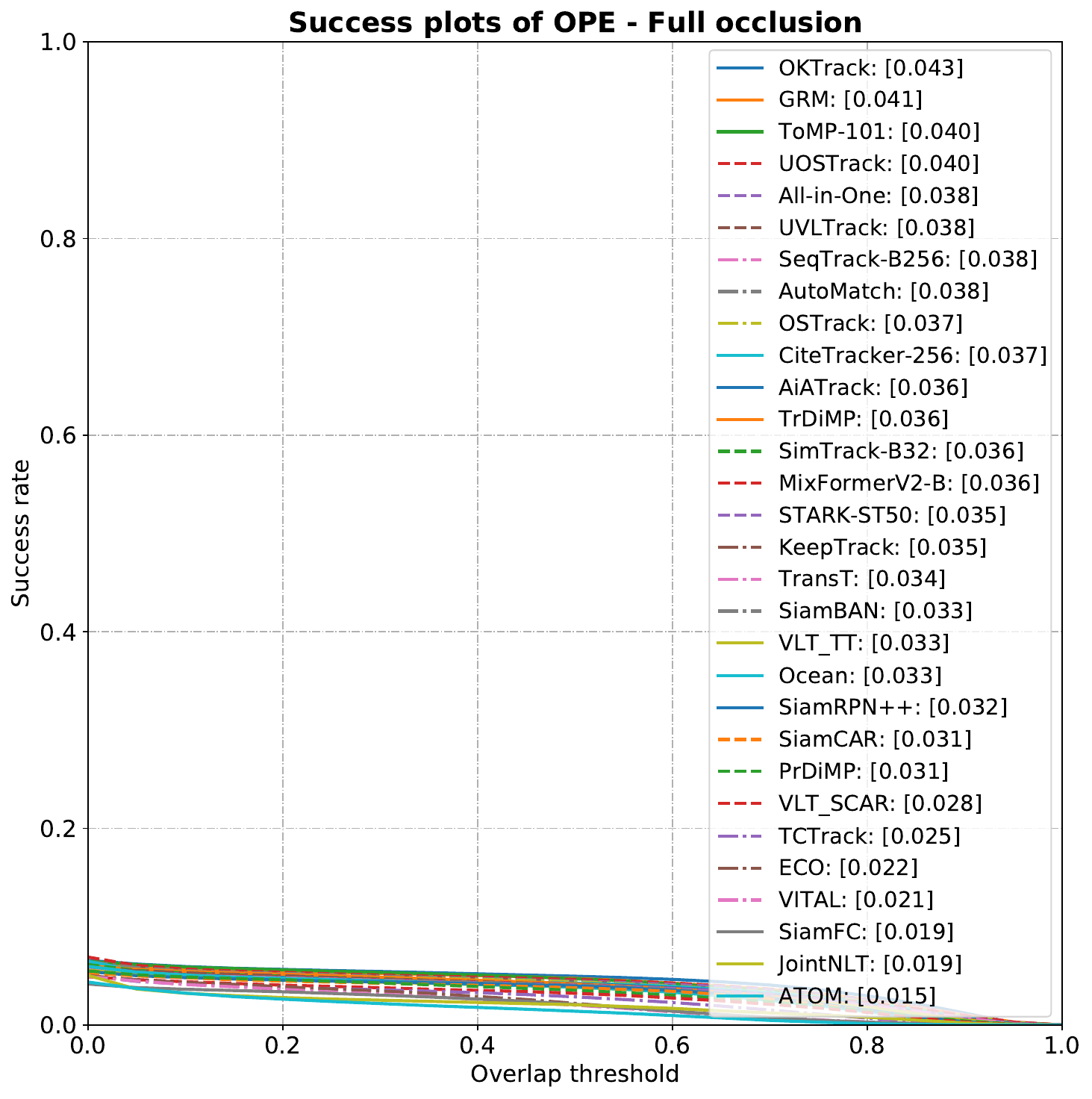}}
\subfloat{\includegraphics[width =0.25\columnwidth]{figs/Attributes-AUC/success_plots_Illumination_variation.pdf}}

\subfloat{\includegraphics[width =0.25\columnwidth]{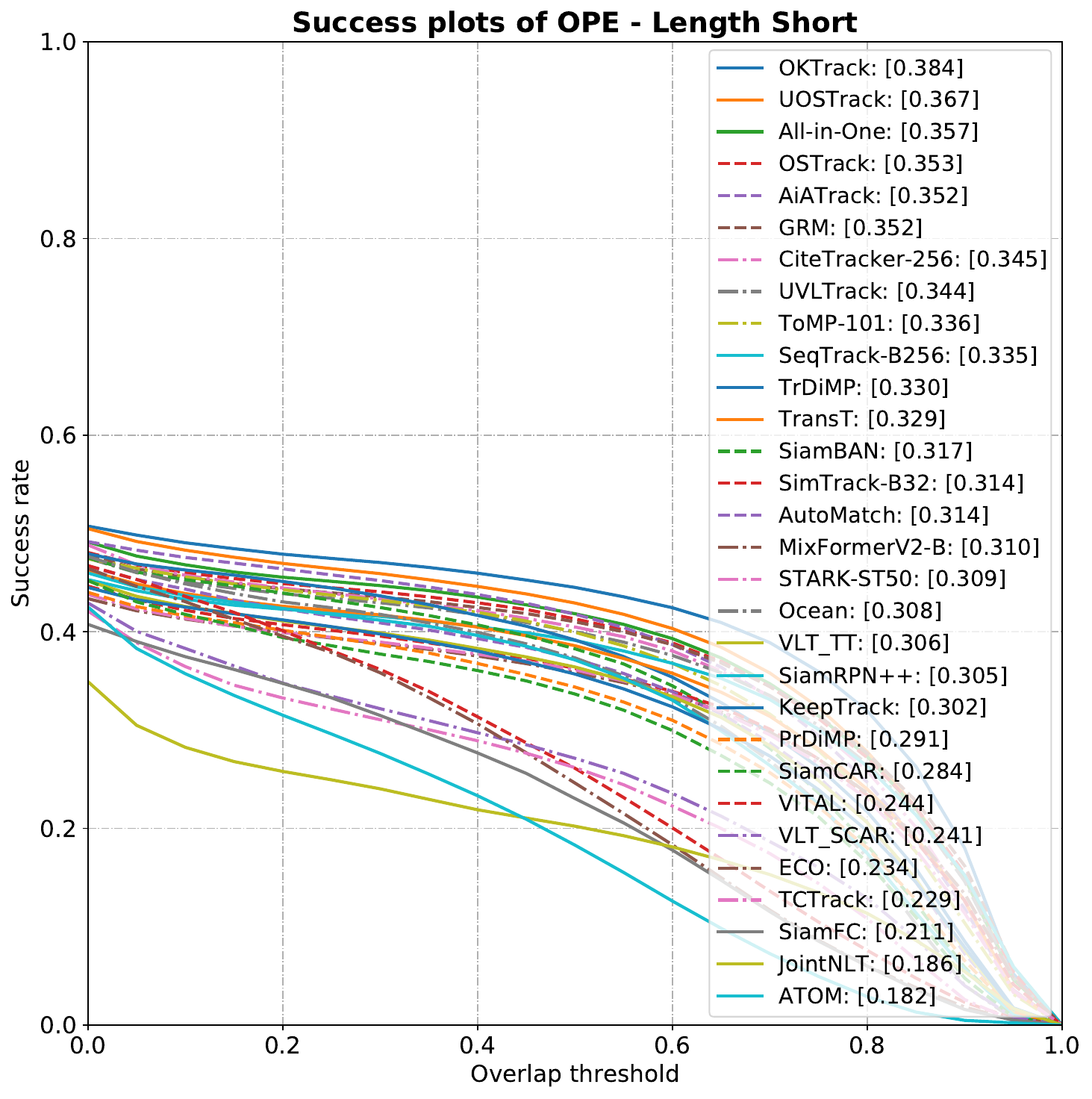}}
\subfloat{\includegraphics[width =0.25\columnwidth]{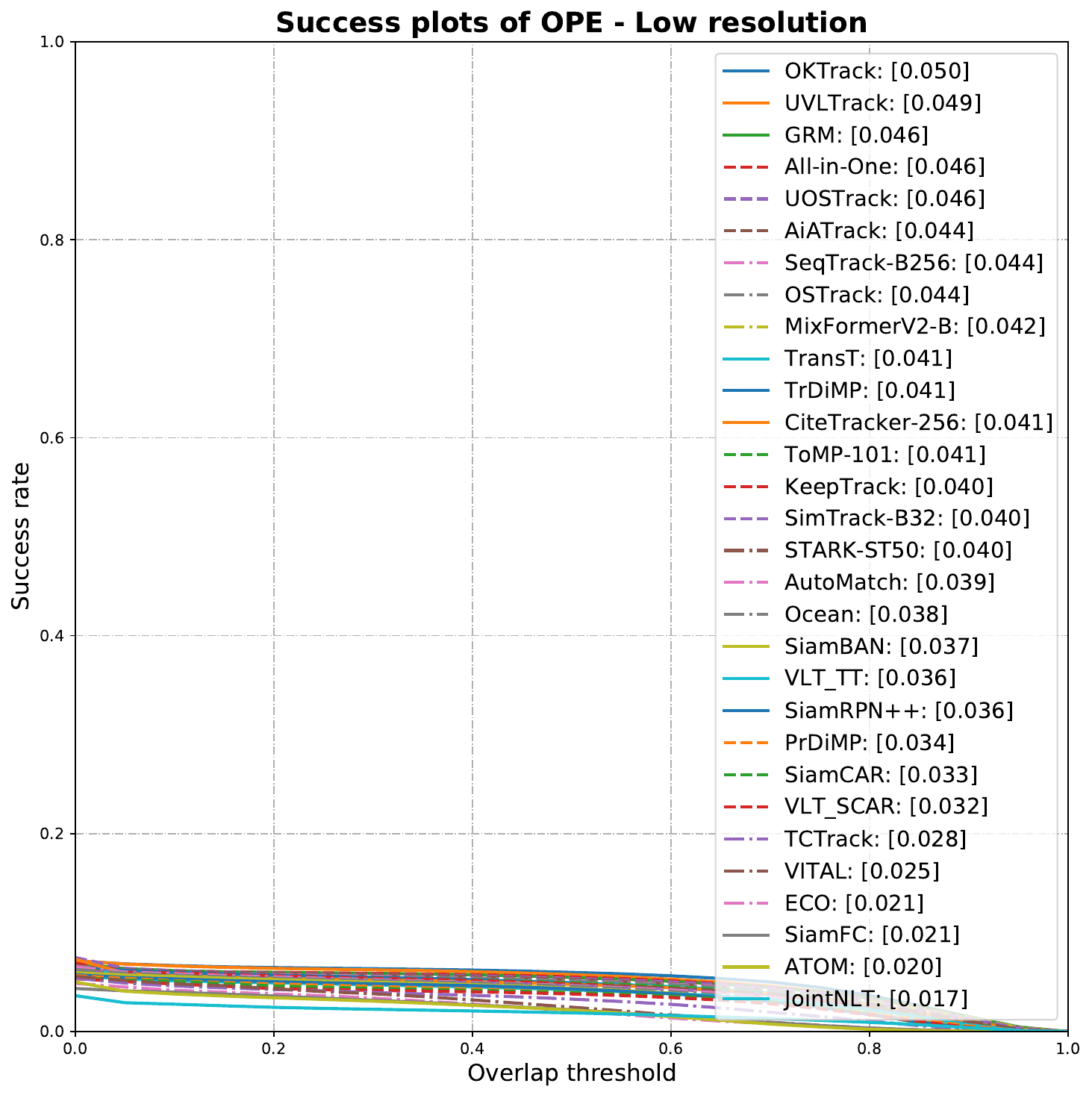}}
\subfloat{\includegraphics[width =0.25\columnwidth]{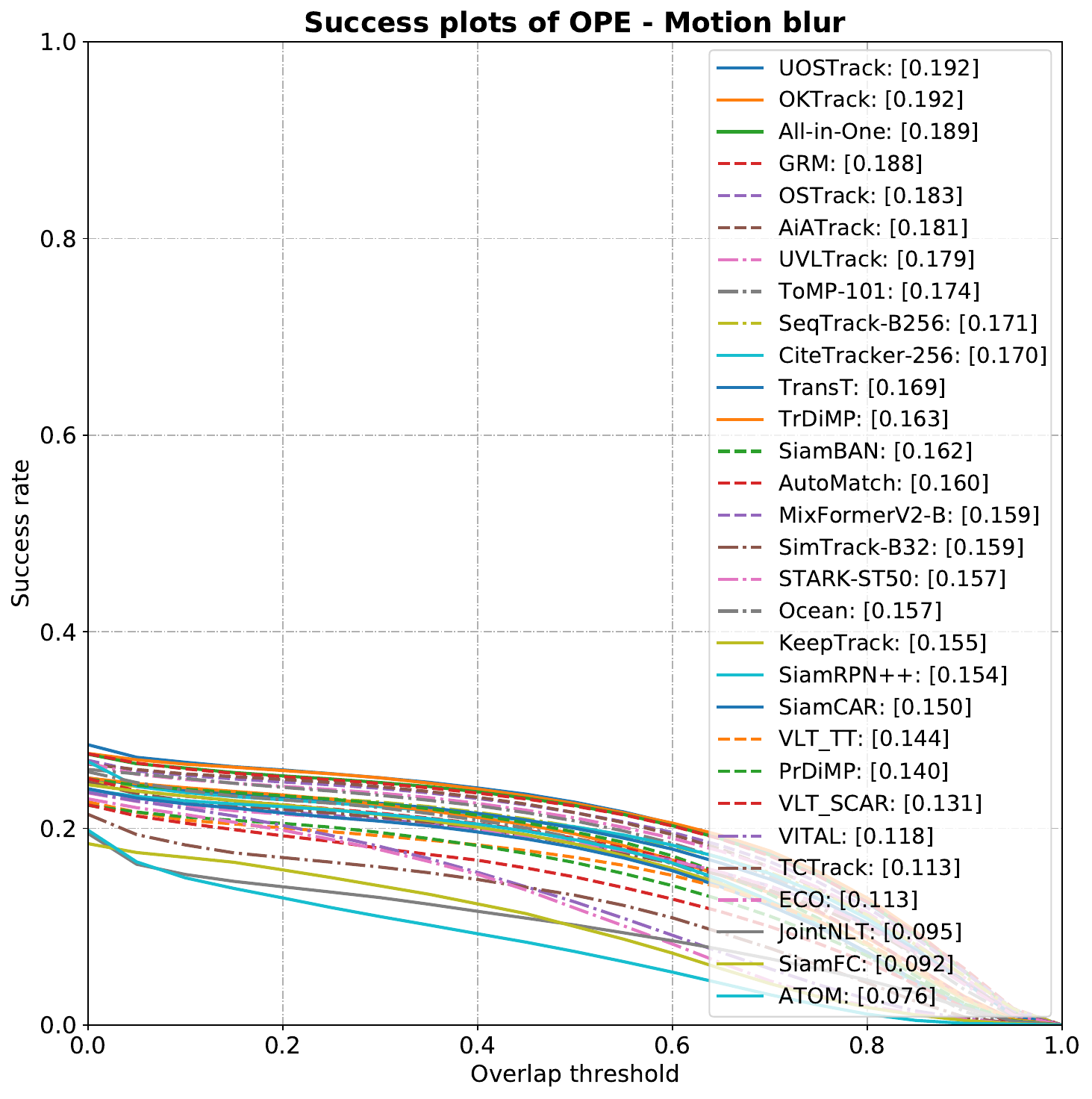}}
\subfloat{\includegraphics[width =0.25\columnwidth]{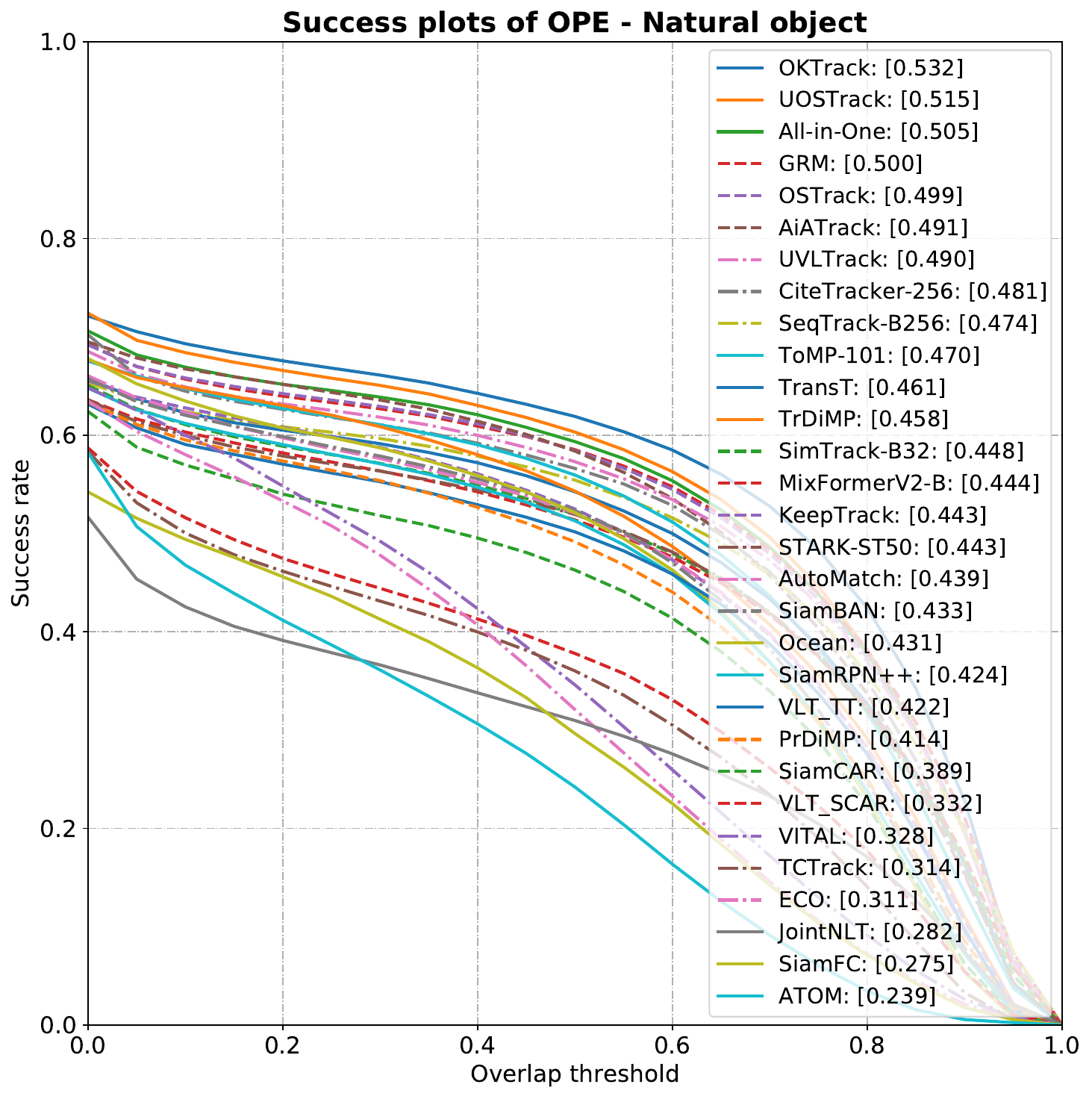}}

\subfloat{\includegraphics[width =0.25\columnwidth]{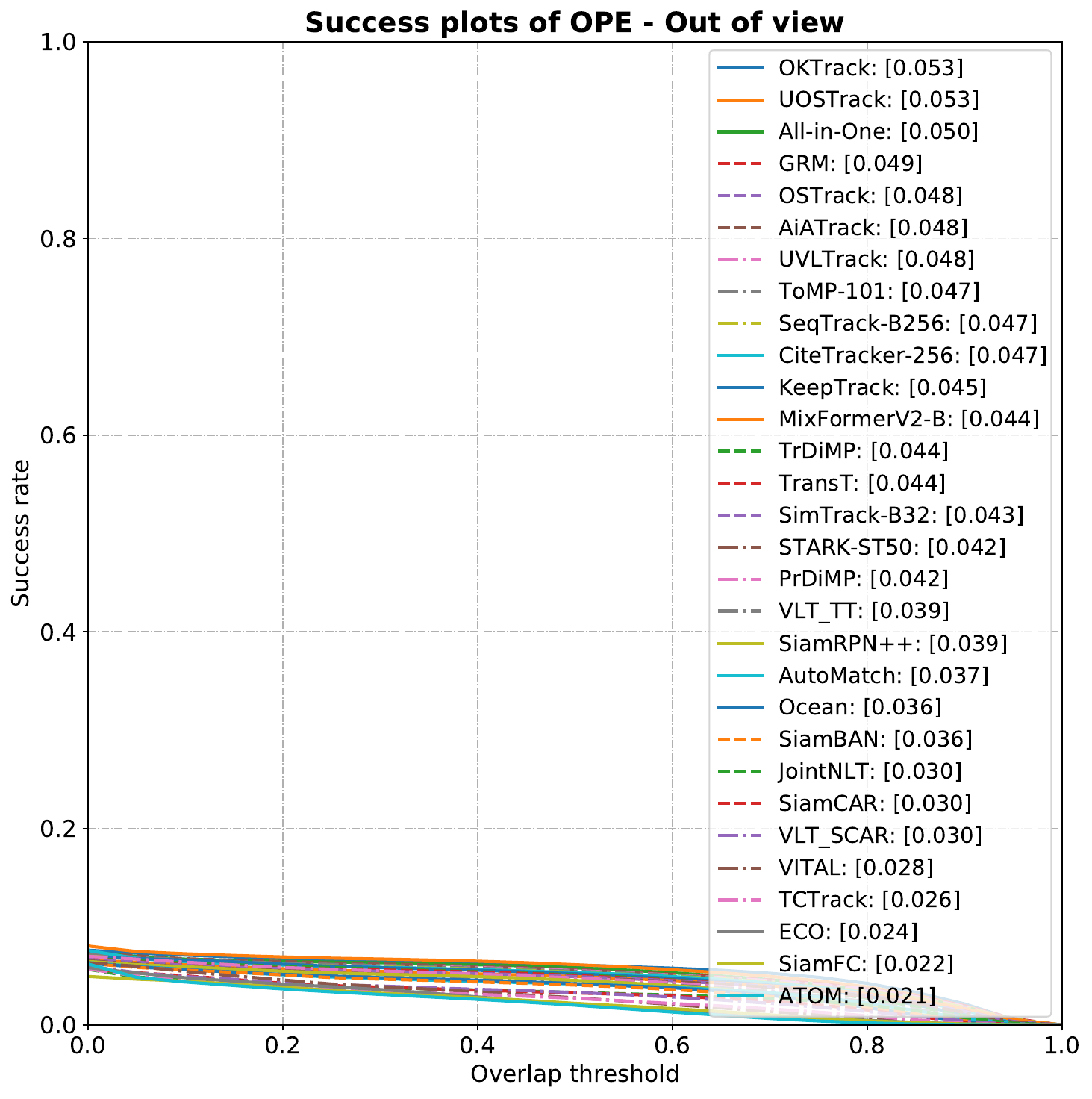}}
\subfloat{\includegraphics[width =0.25\columnwidth]{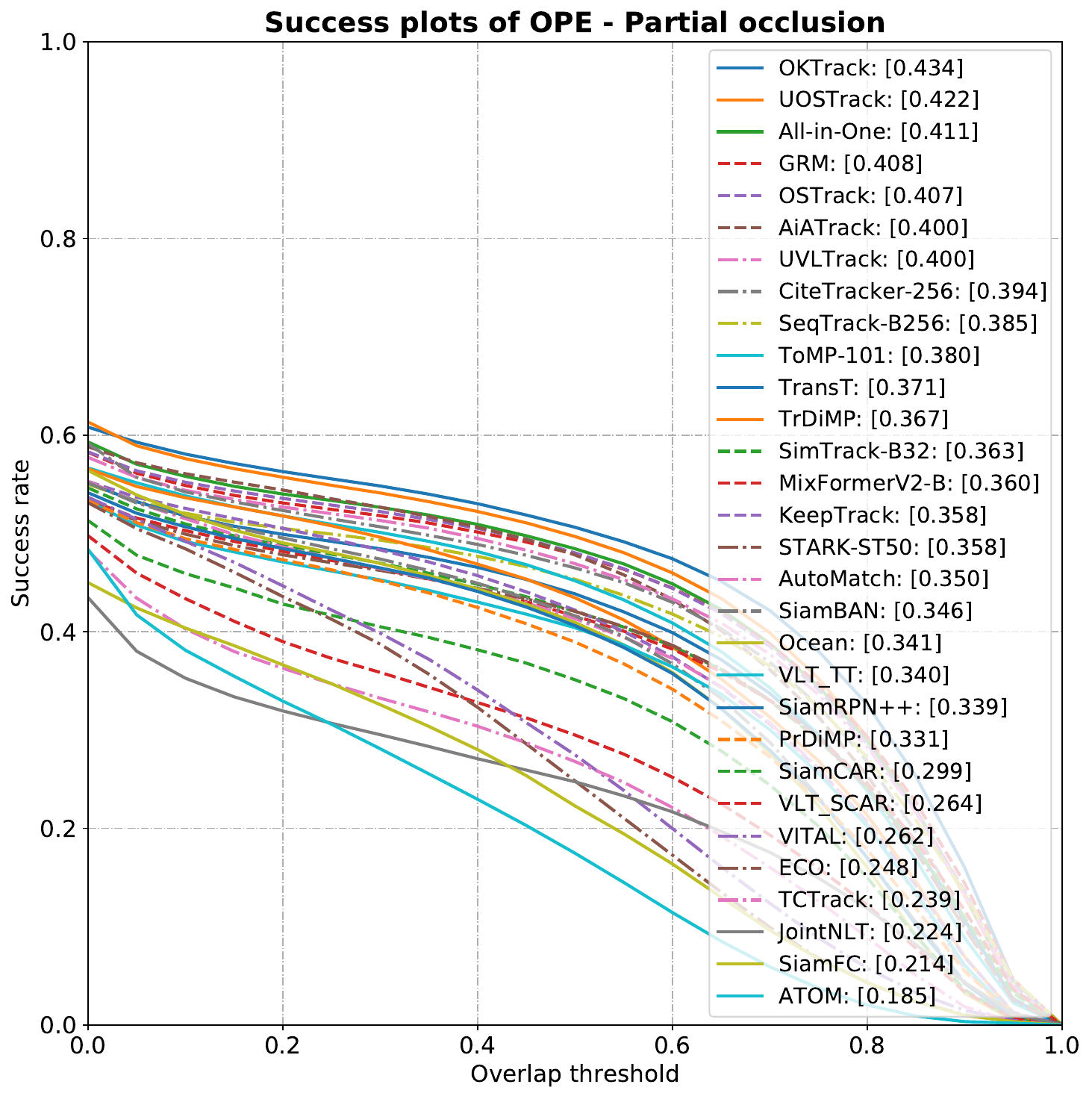}}
\subfloat{\includegraphics[width =0.25\columnwidth]{figs/Attributes-AUC/success_plots_Partial_target_information.pdf}}
\subfloat{\includegraphics[width =0.25\columnwidth]{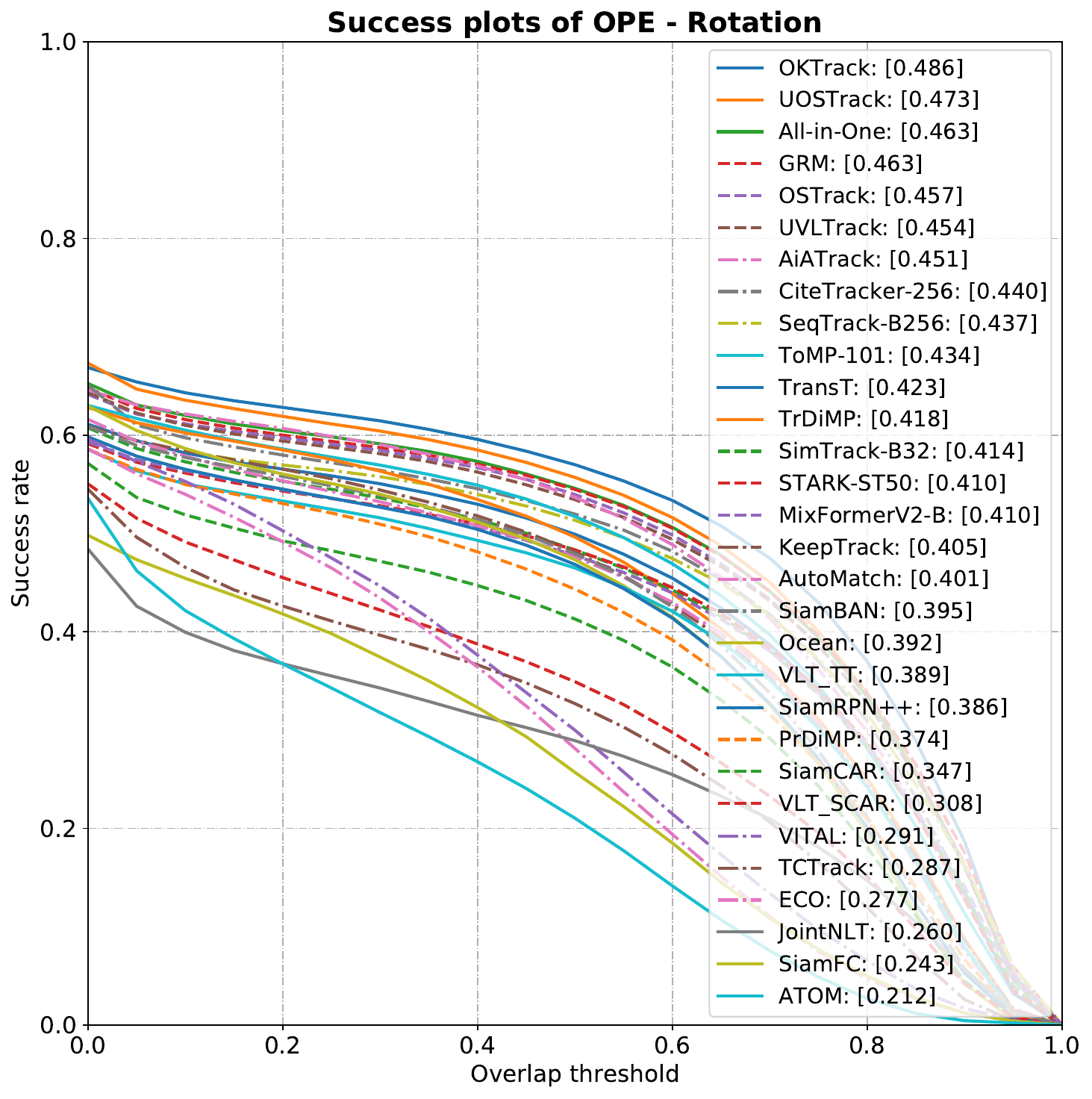}}

\subfloat{\includegraphics[width =0.25\columnwidth]{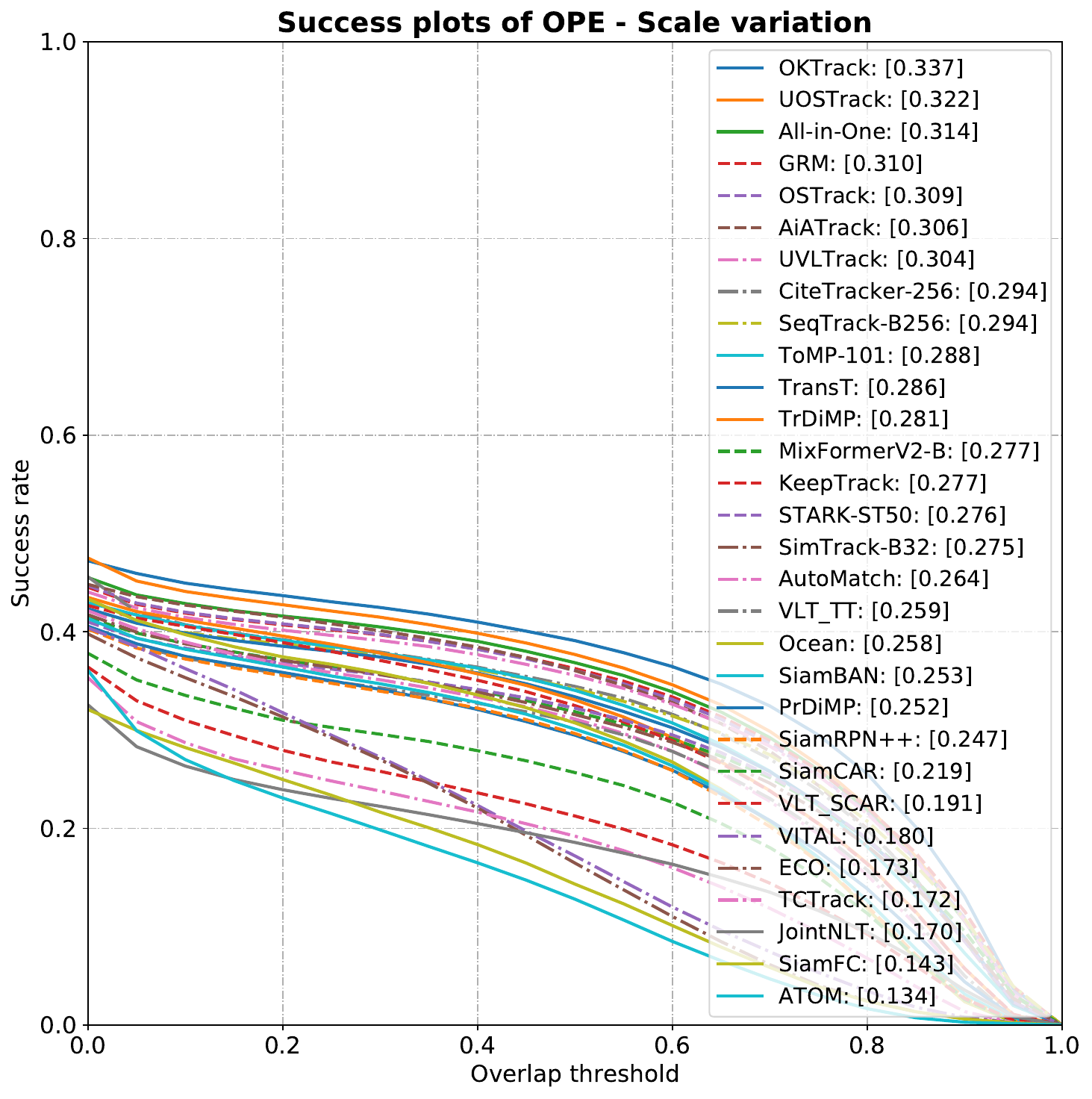}}
\subfloat{\includegraphics[width =0.25\columnwidth]{figs/Attributes-AUC/success_plots_Similar_distractors.pdf}}
\subfloat{\includegraphics[width =0.25\columnwidth]{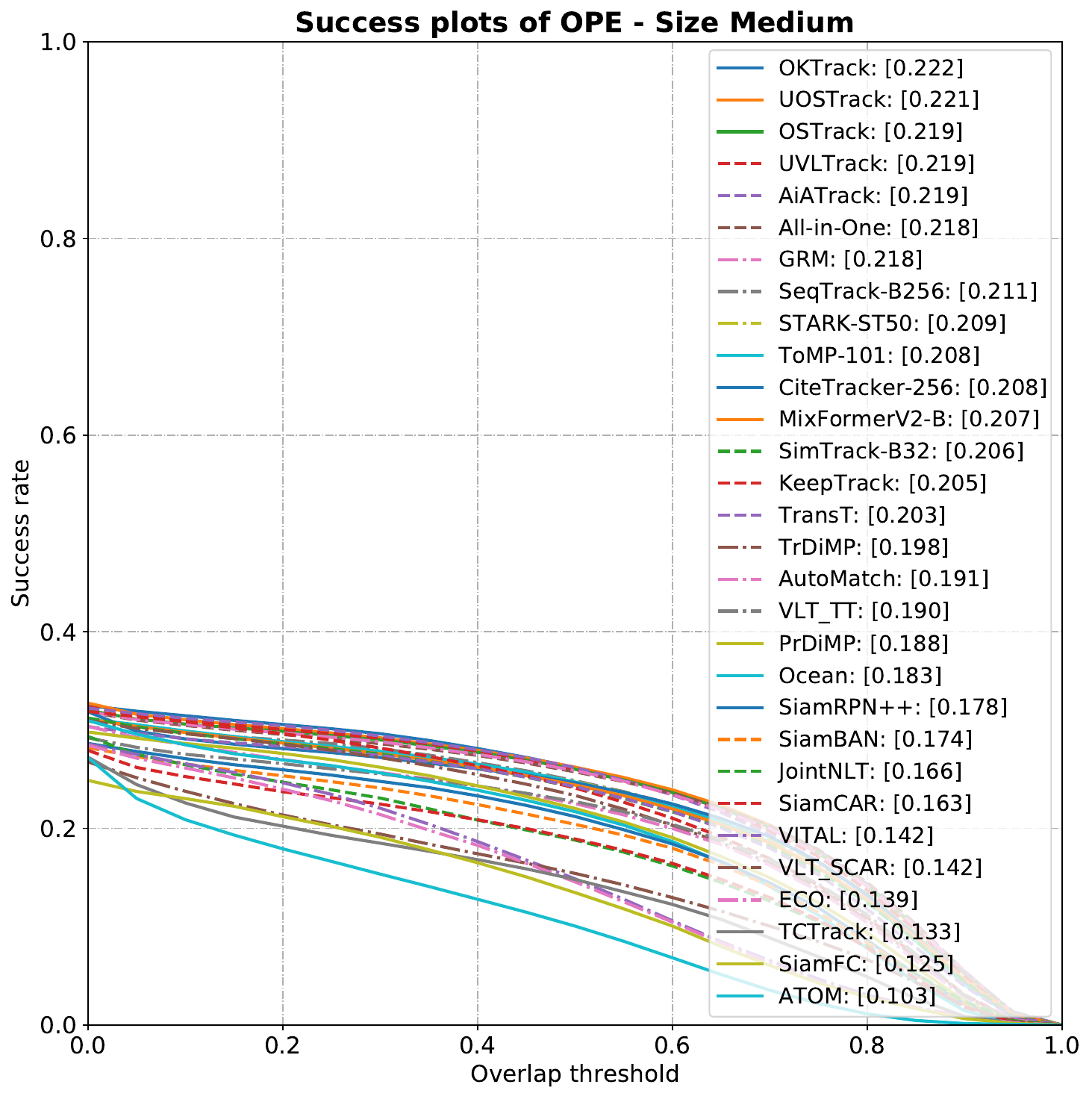}}
\subfloat{\includegraphics[width =0.25\columnwidth]{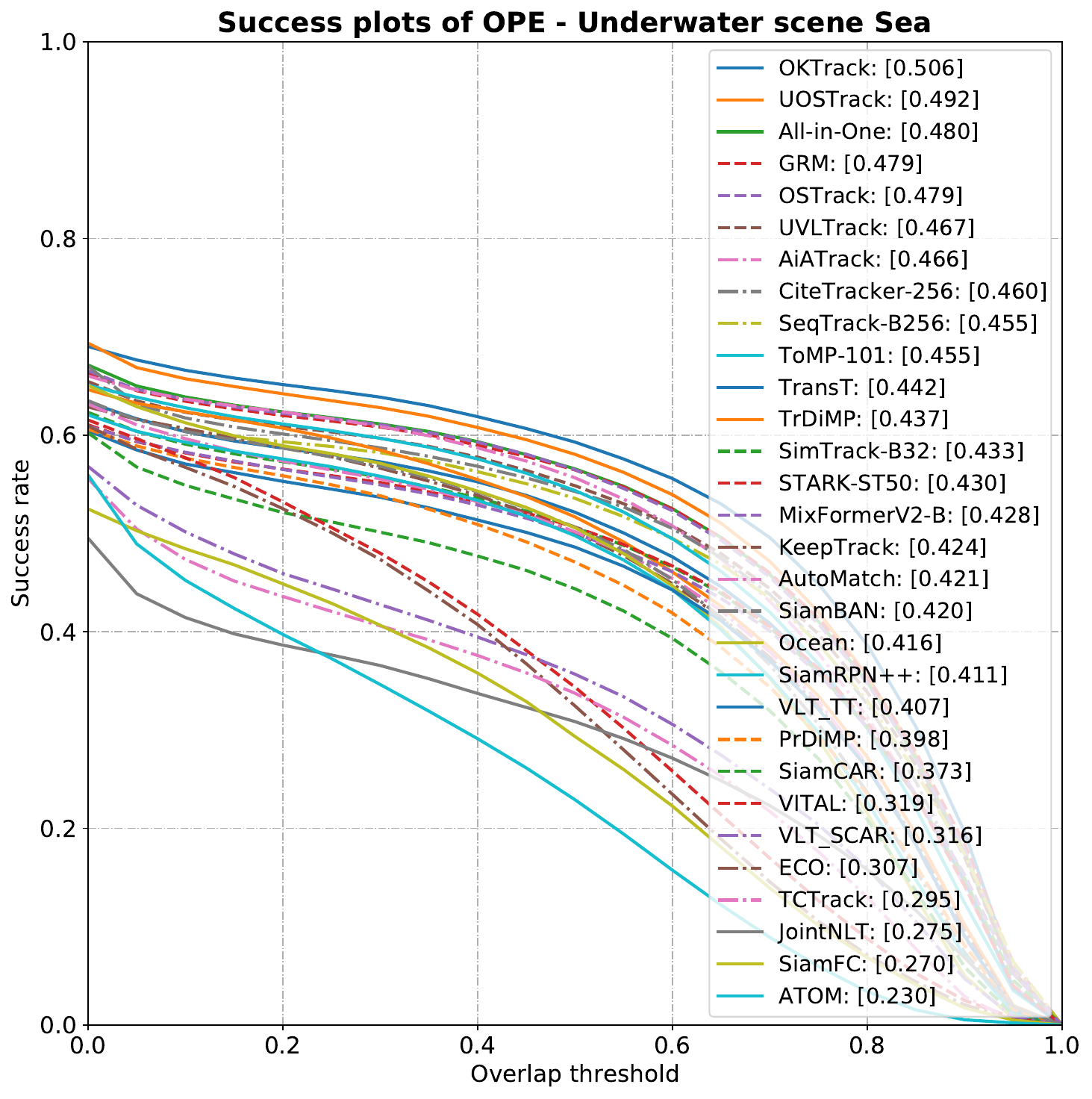}}

\subfloat{\includegraphics[width =0.25\columnwidth]{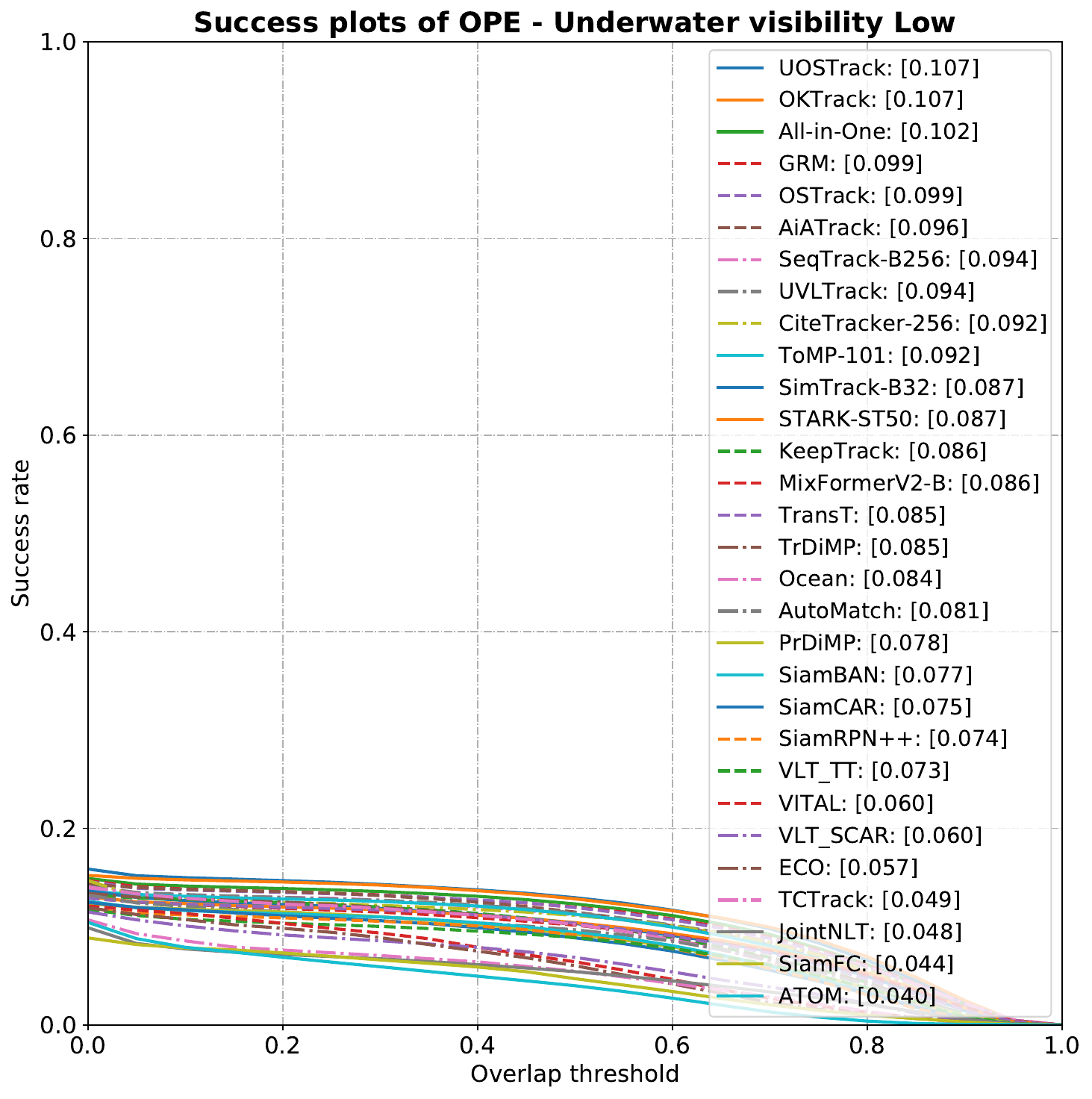}}
\subfloat{\includegraphics[width =0.25\columnwidth]{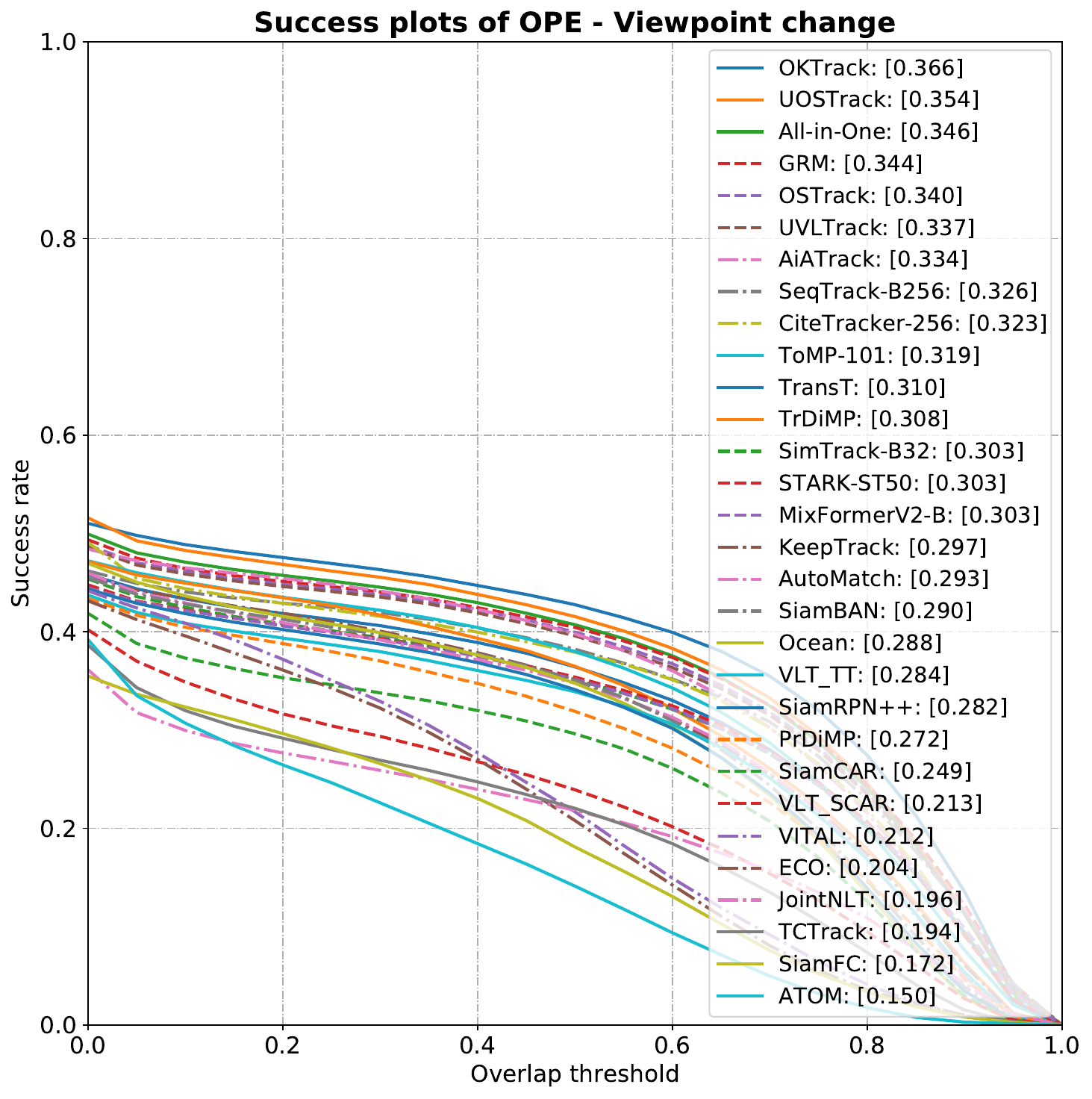}}
\subfloat{\includegraphics[width =0.25\columnwidth]{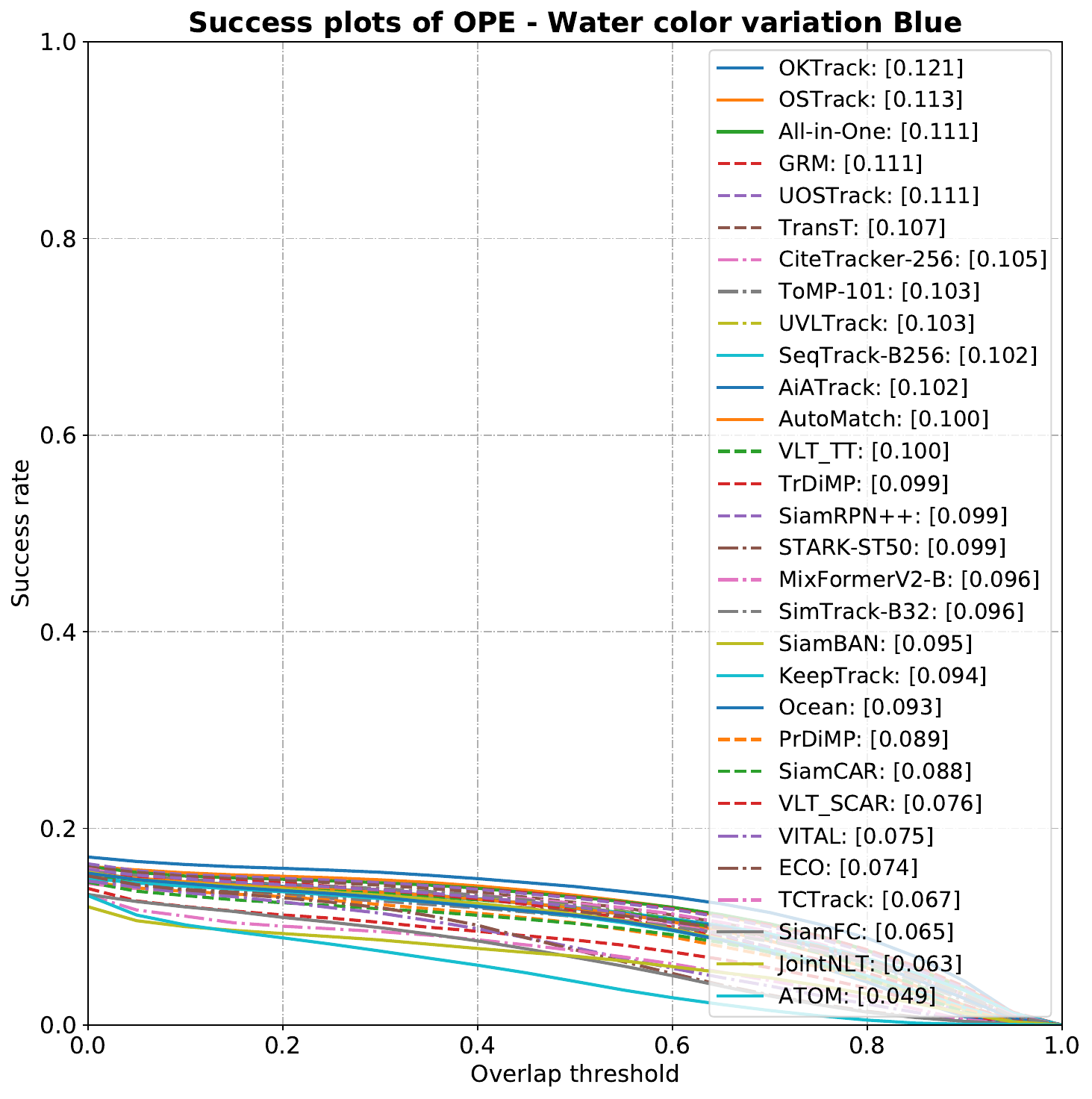}}
\subfloat{\includegraphics[width =0.25\columnwidth]{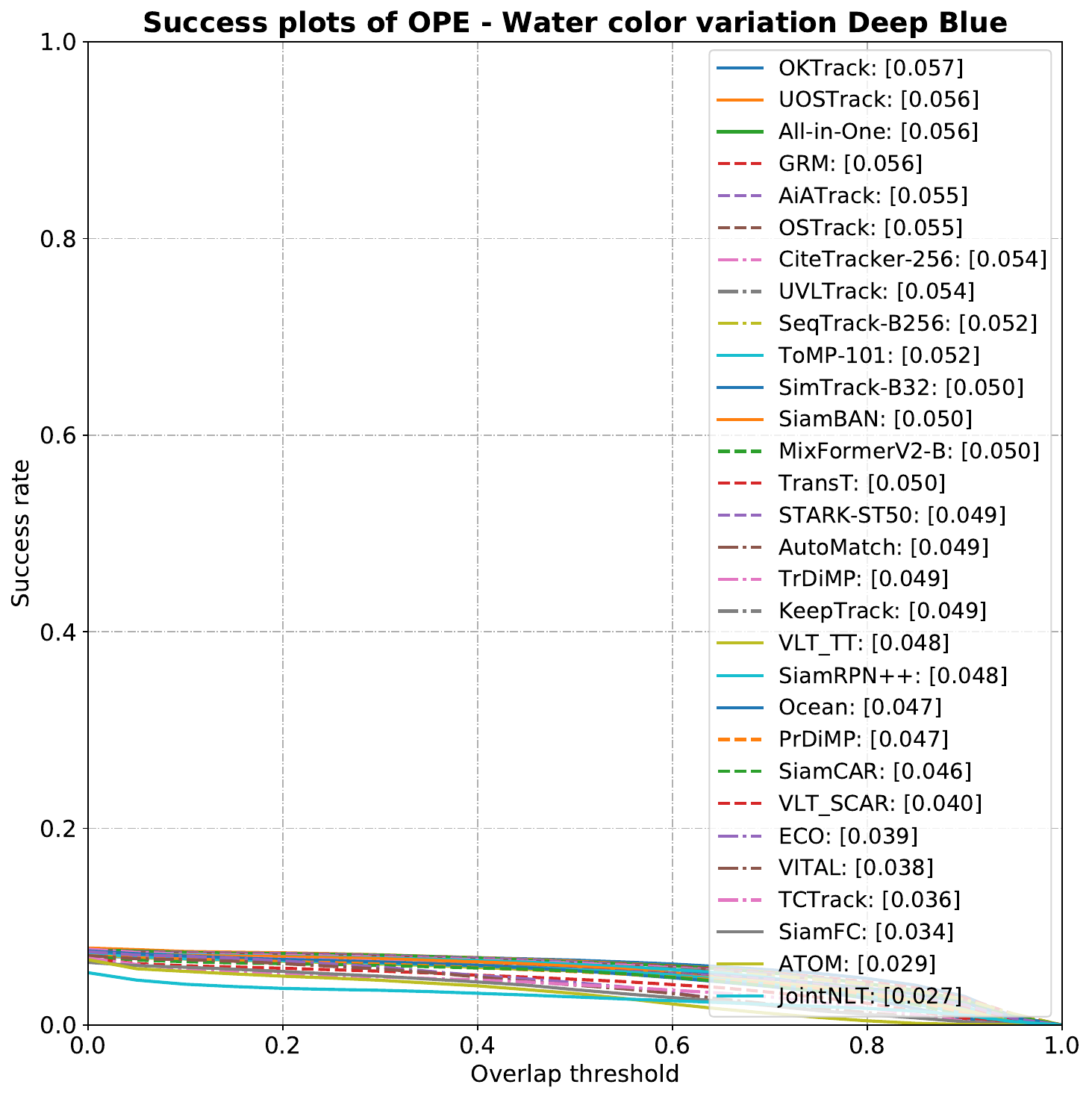}}

\caption{Performances of baseline trackers on the WebOUT-1M test set of different attributes using \textbf{AUC} scores. Best viewed by zooming in.}
  \label{fig:attribute_results_AUC}
\end{figure*}

\begin{figure*}[t]
\vspace{-0.6cm}
  \centering
  
\subfloat{\includegraphics[width =0.25\columnwidth]{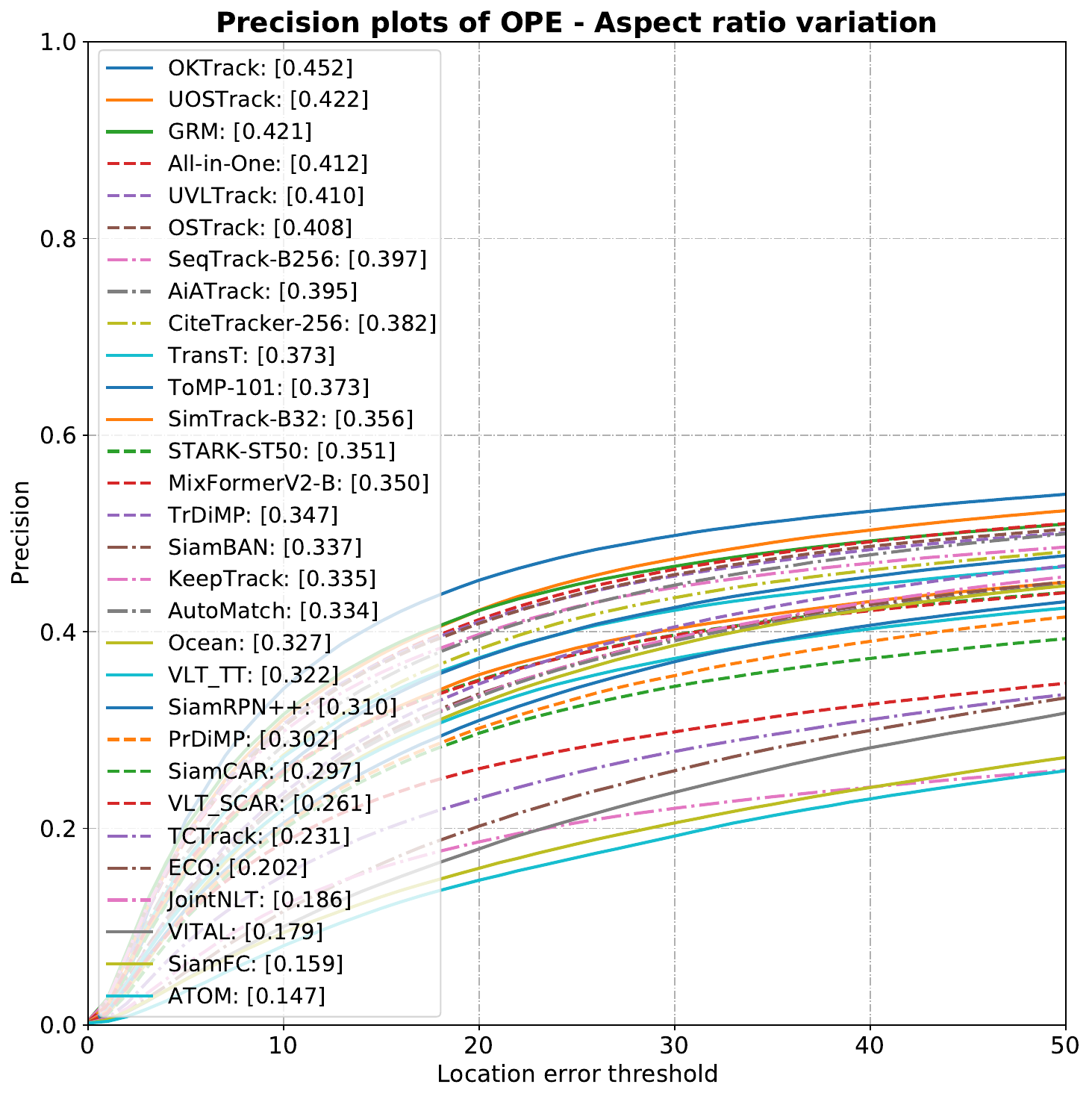}}
\subfloat{\includegraphics[width =0.25\columnwidth]{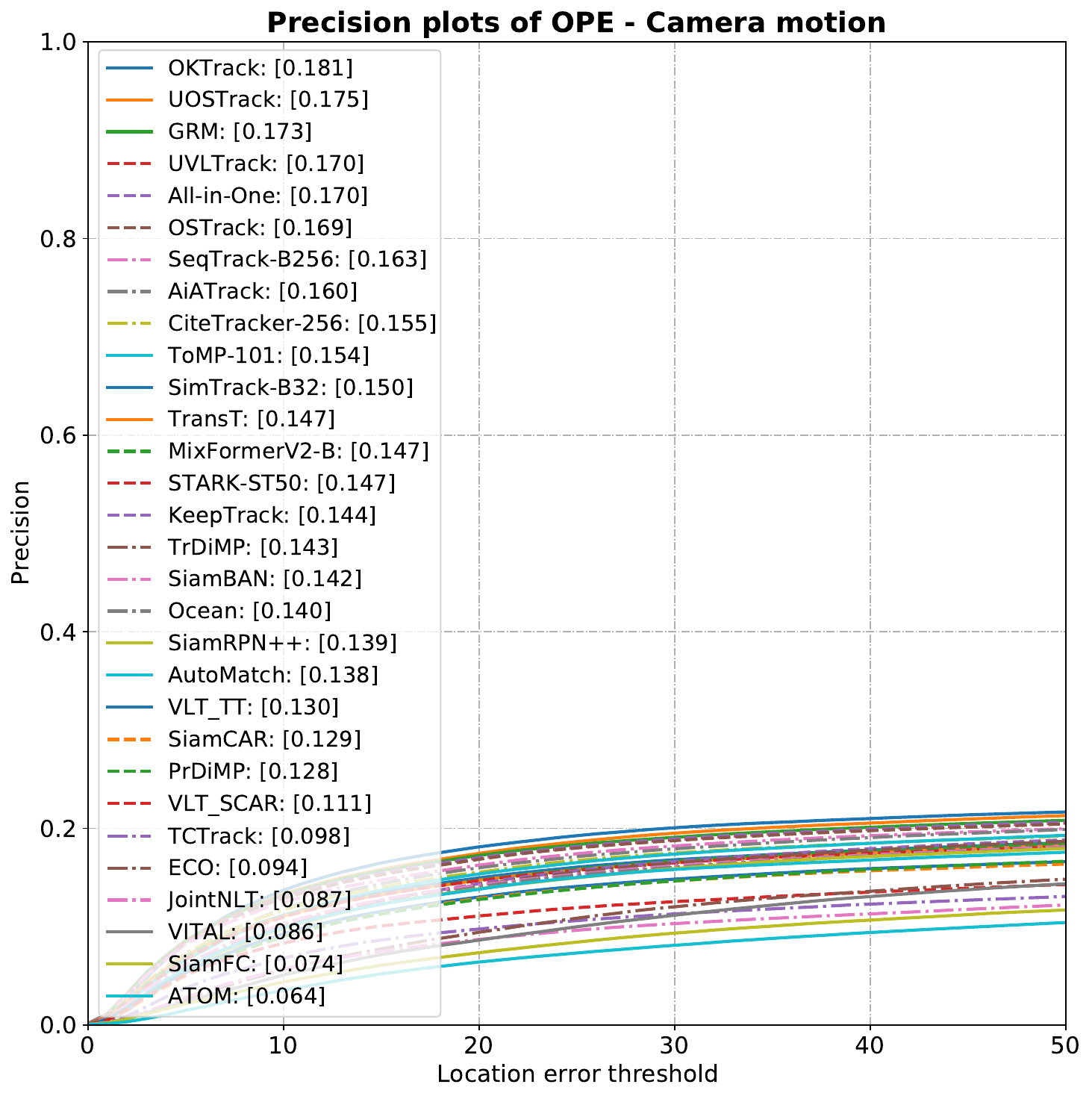}}
\subfloat{\includegraphics[width =0.25\columnwidth]{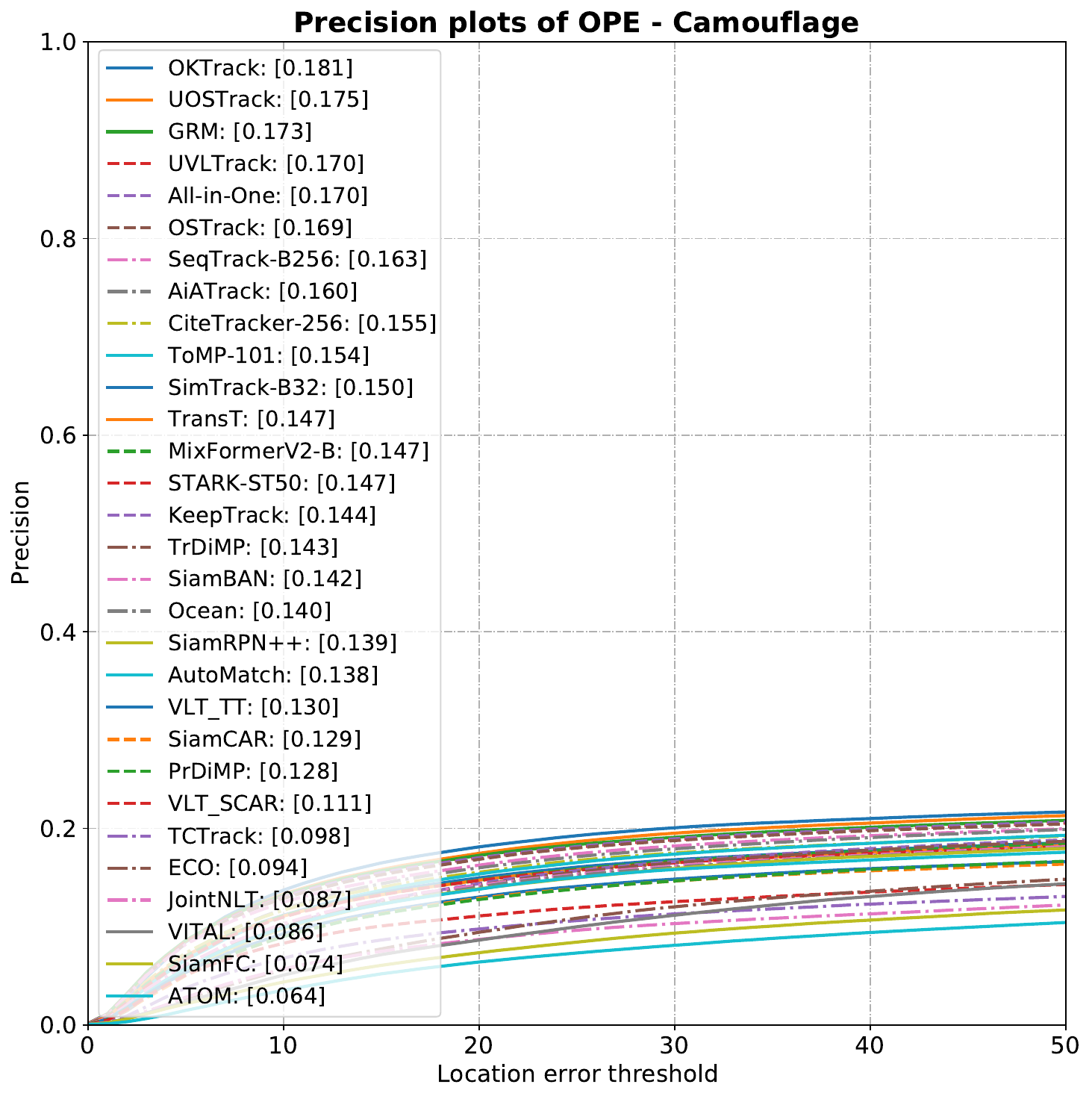}}
\subfloat{\includegraphics[width =0.25\columnwidth]{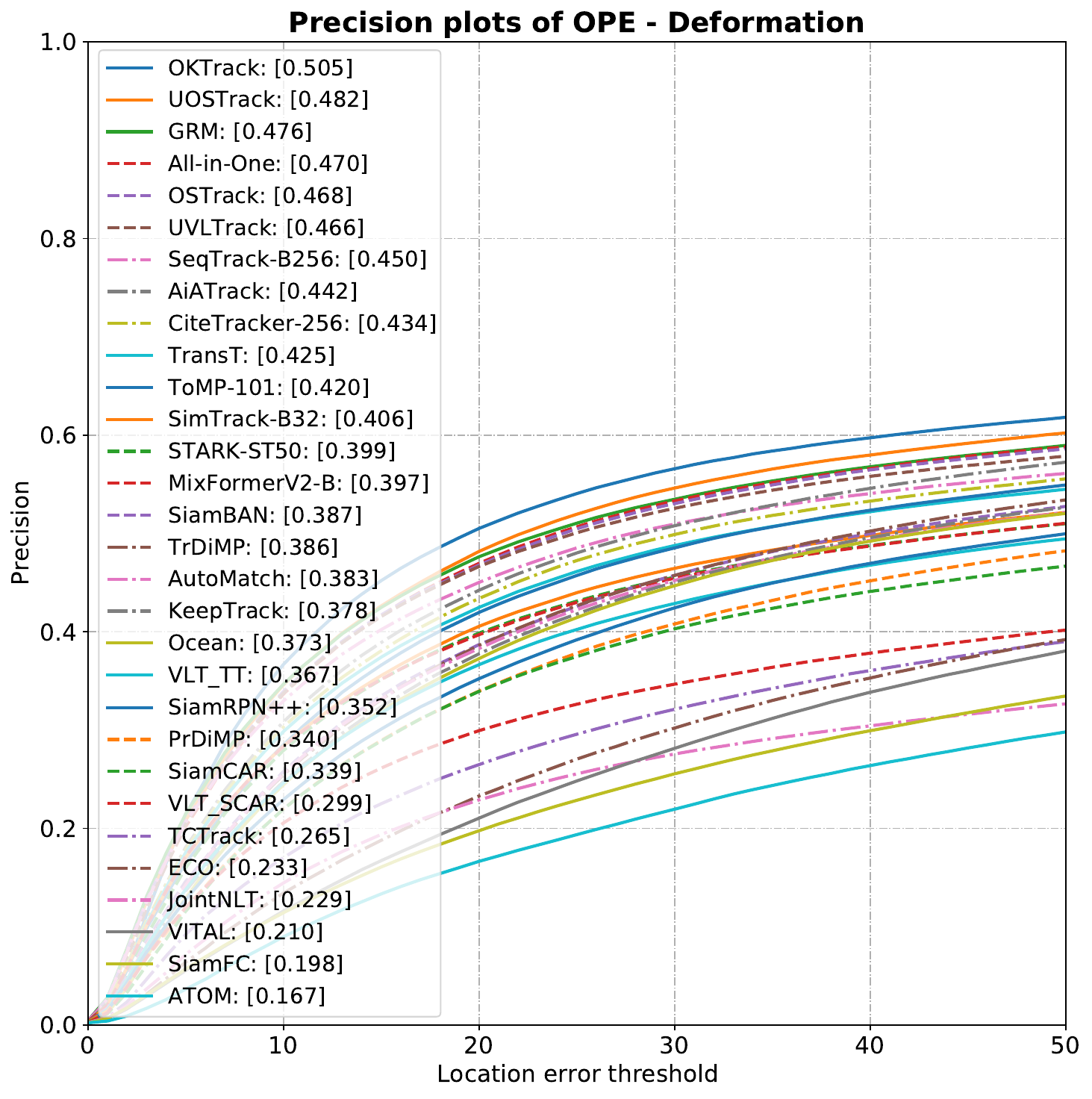}}

\subfloat{\includegraphics[width =0.25\columnwidth]{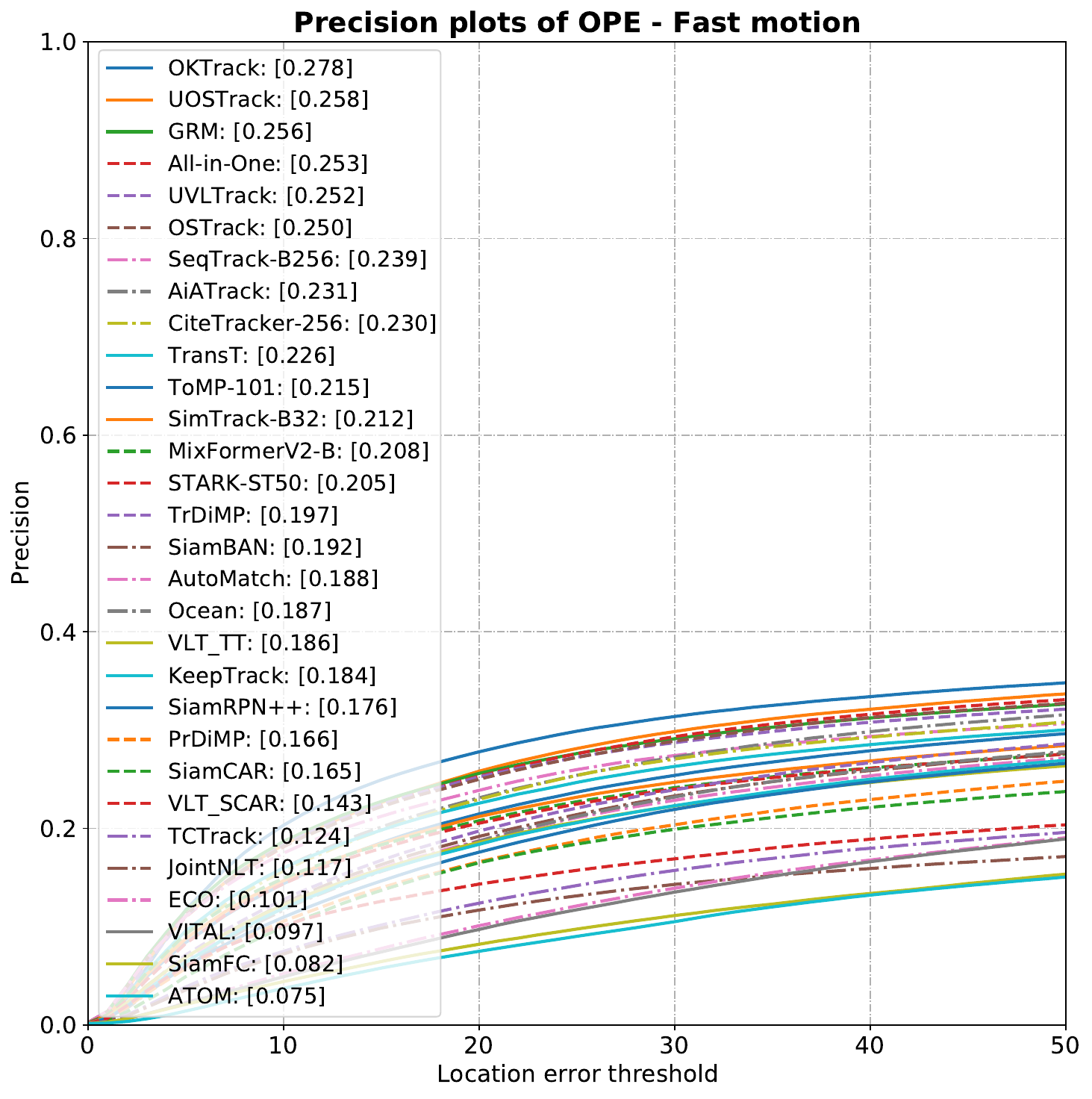}}
\subfloat{\includegraphics[width =0.25\columnwidth]{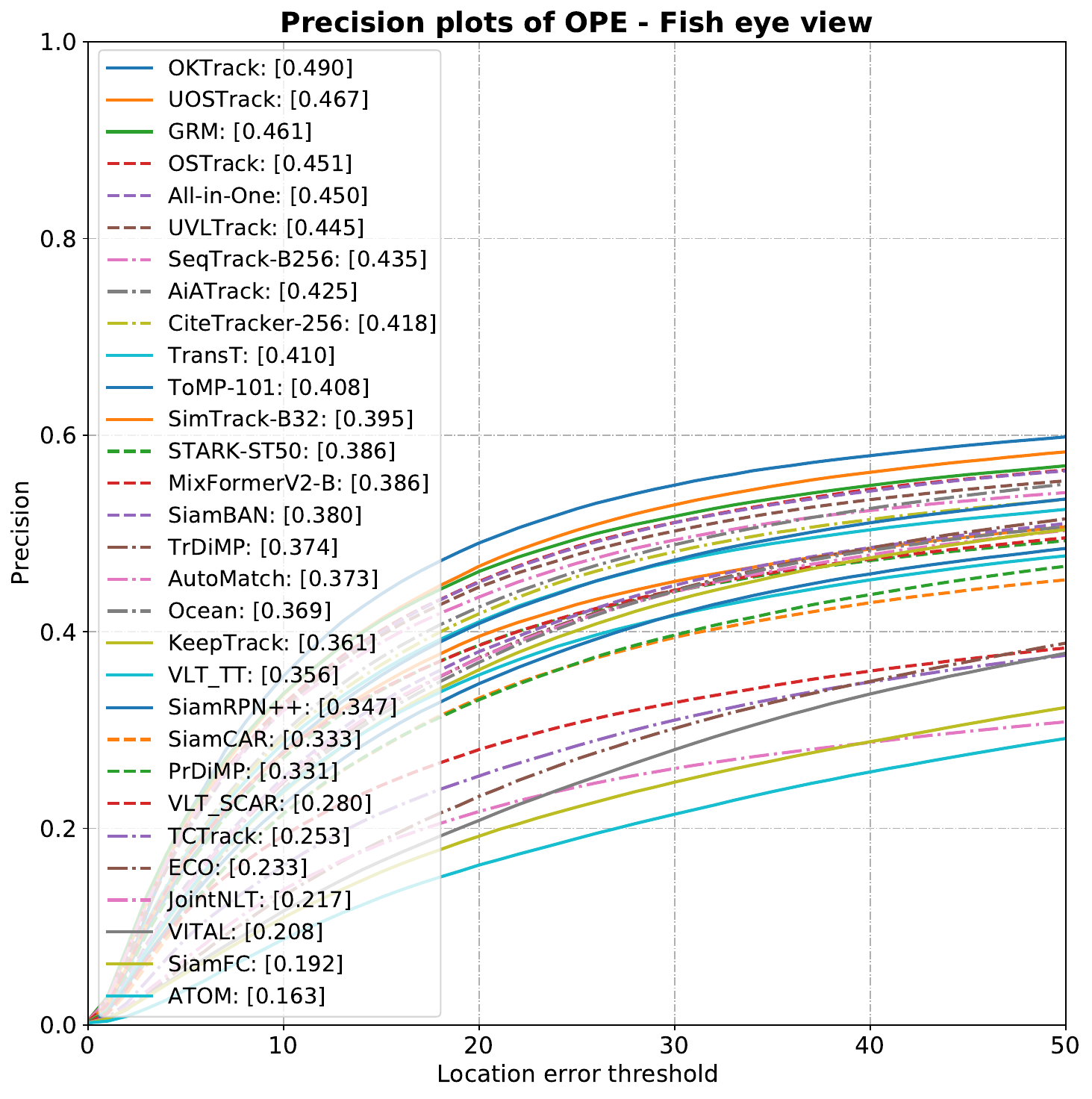}}
\subfloat{\includegraphics[width =0.25\columnwidth]{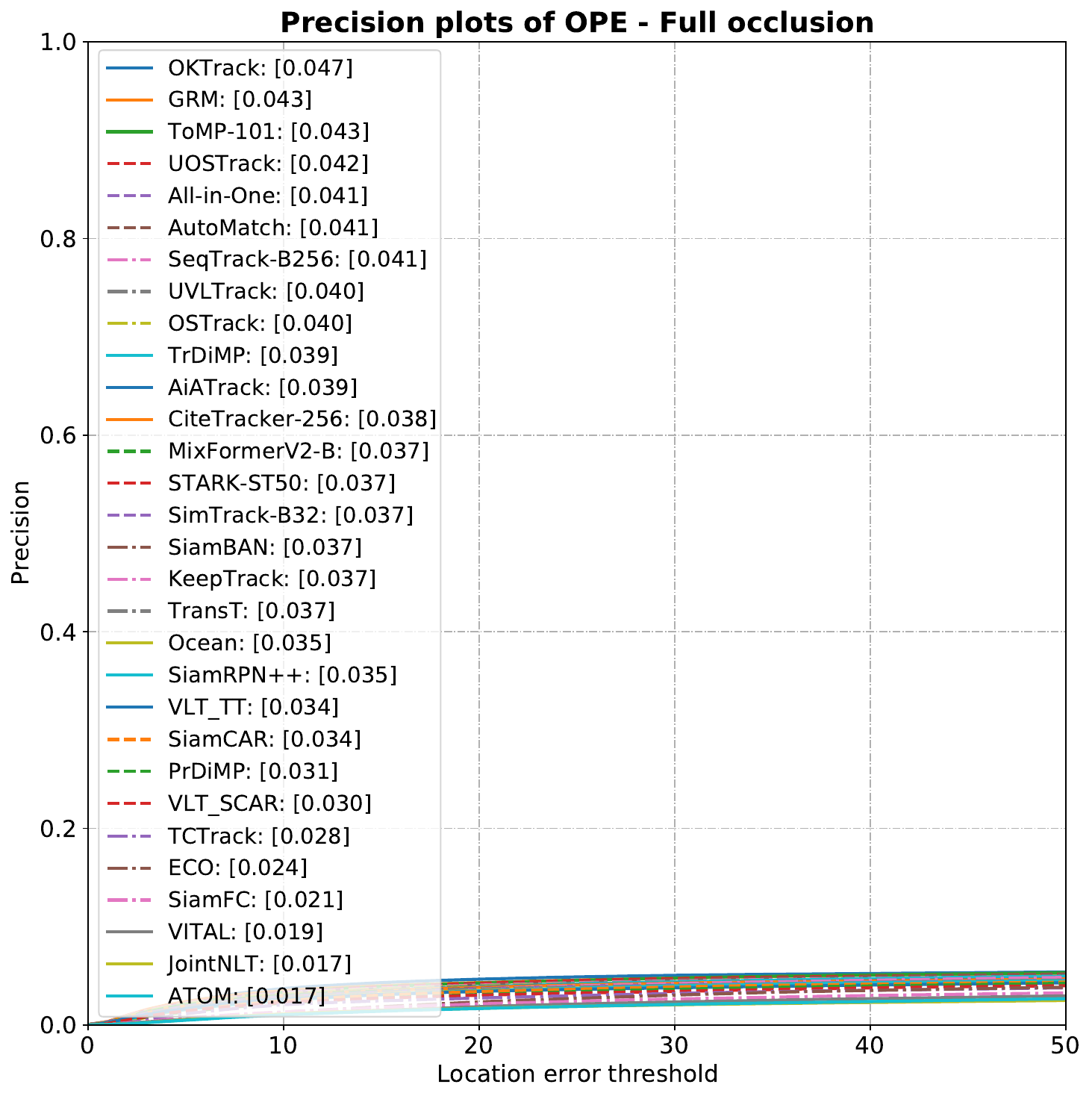}}
\subfloat{\includegraphics[width =0.25\columnwidth]{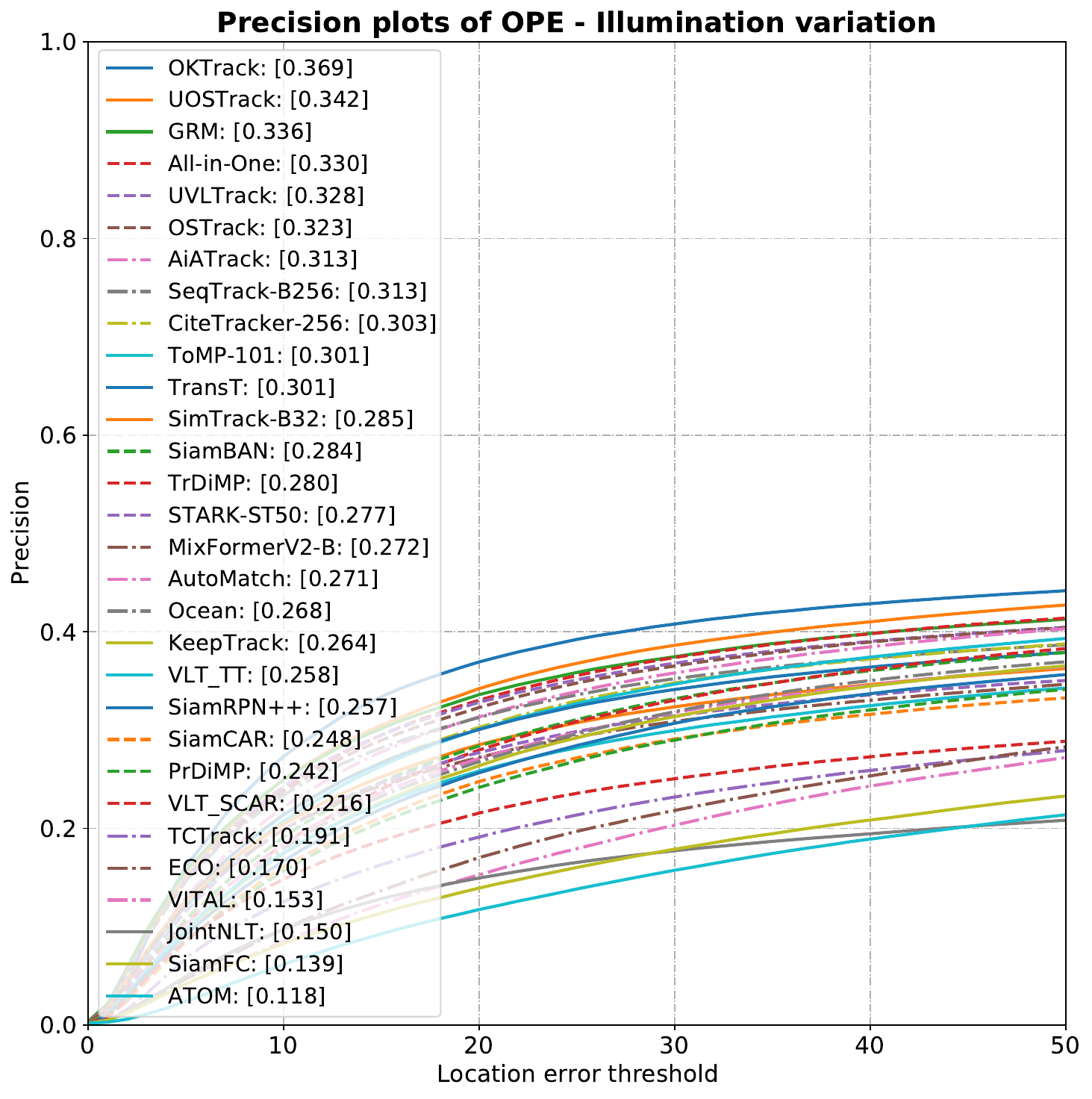}}

\subfloat{\includegraphics[width =0.25\columnwidth]{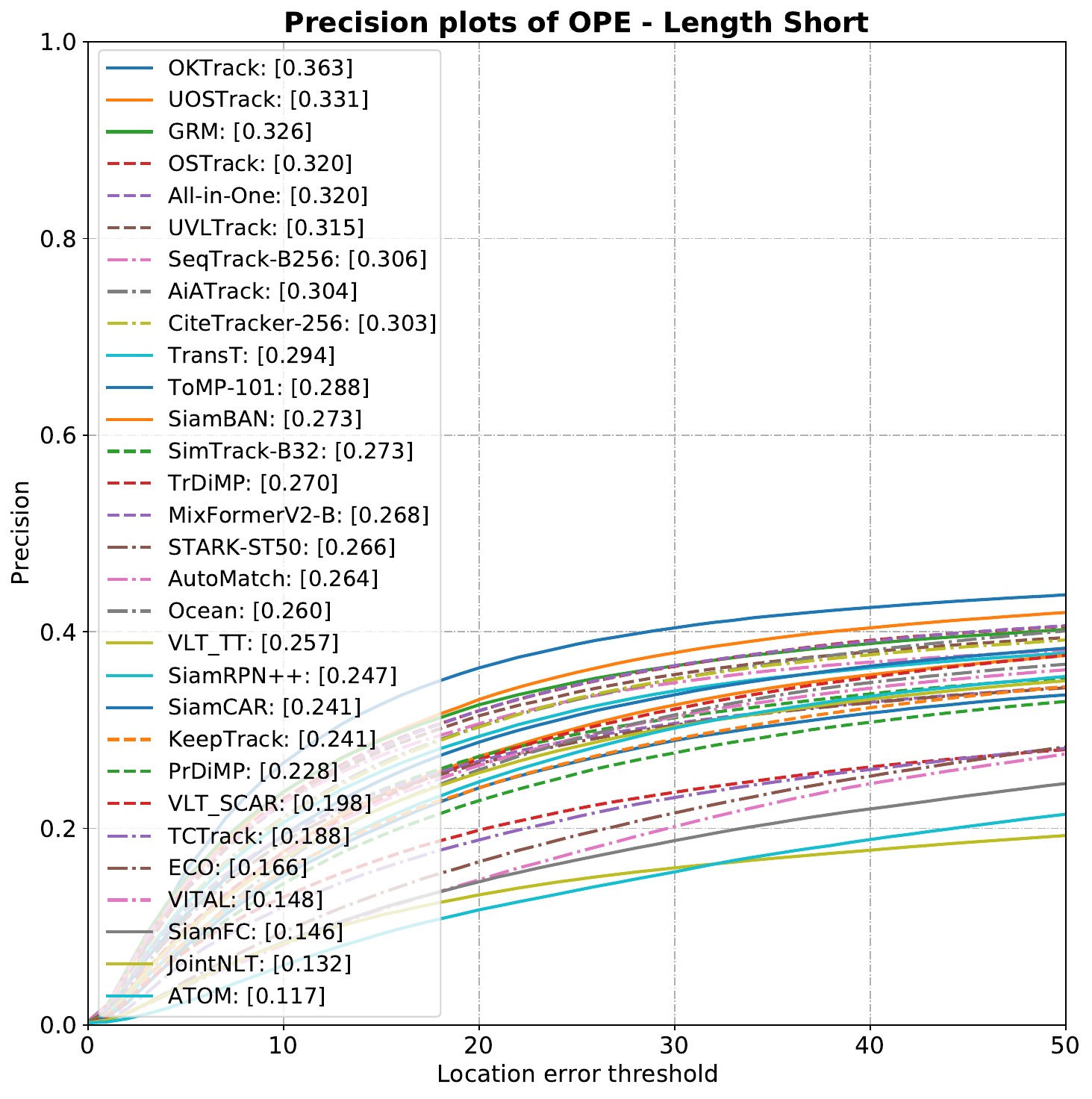}}
\subfloat{\includegraphics[width =0.25\columnwidth]{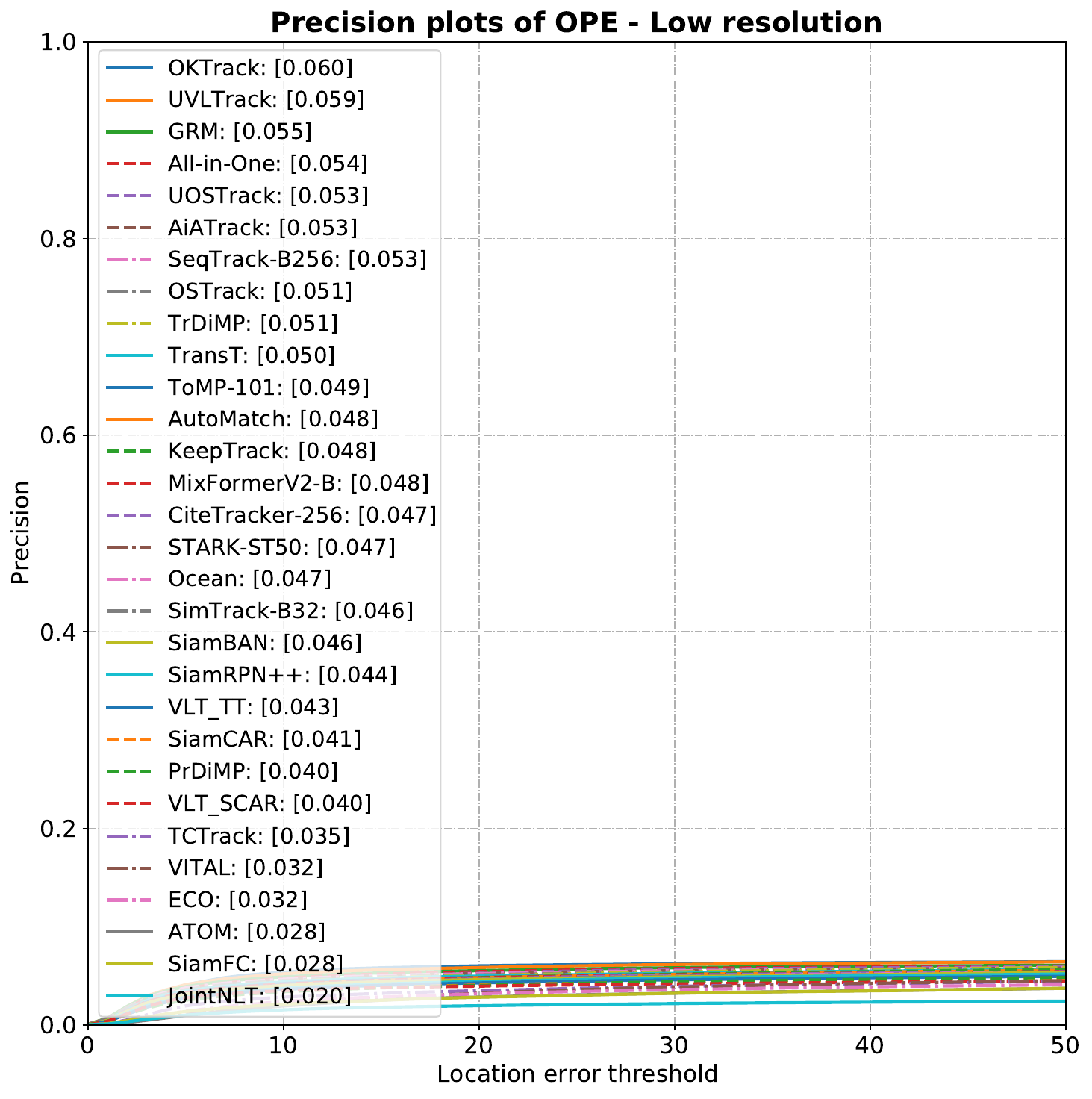}}
\subfloat{\includegraphics[width =0.25\columnwidth]{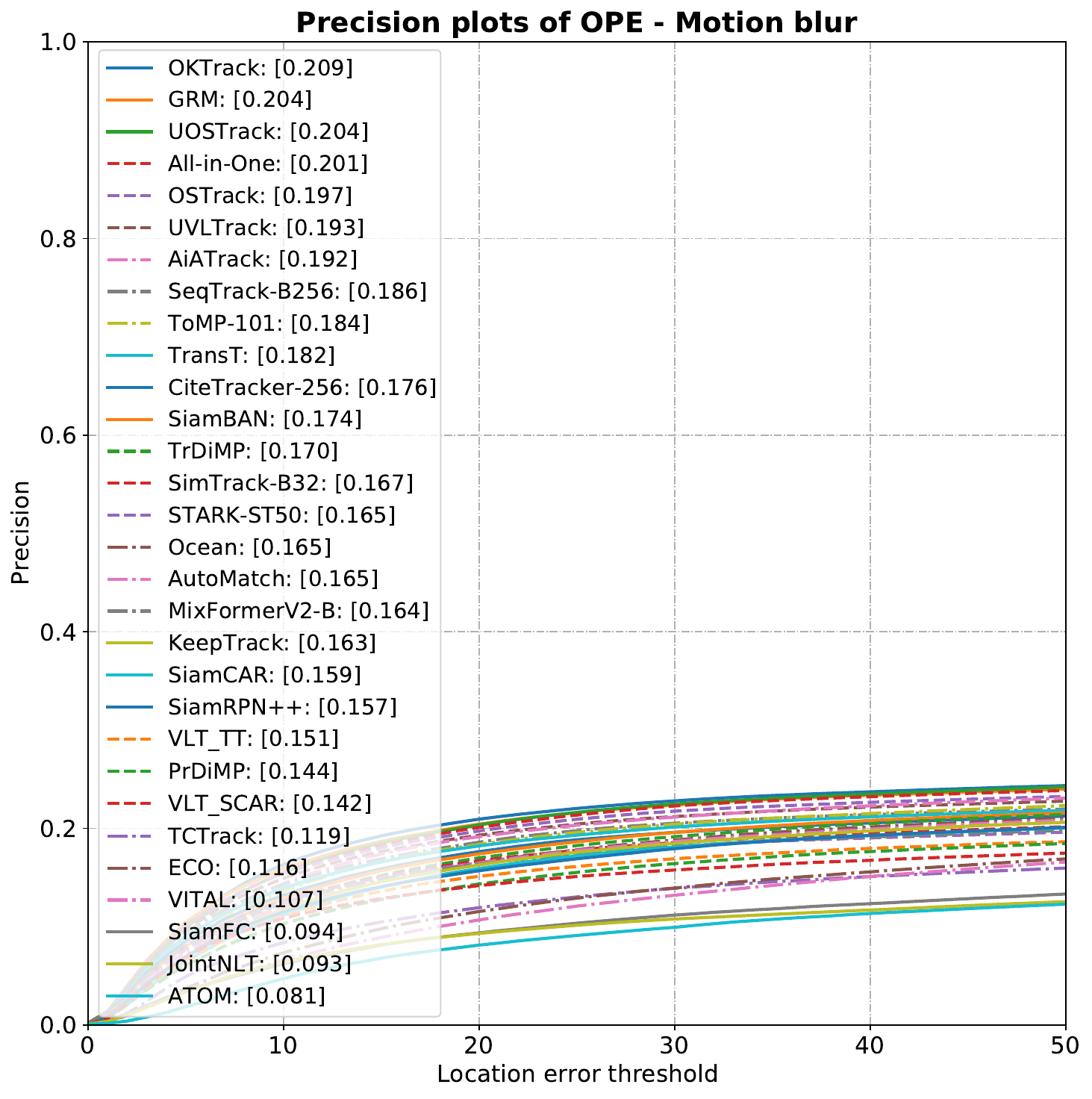}}
\subfloat{\includegraphics[width =0.25\columnwidth]{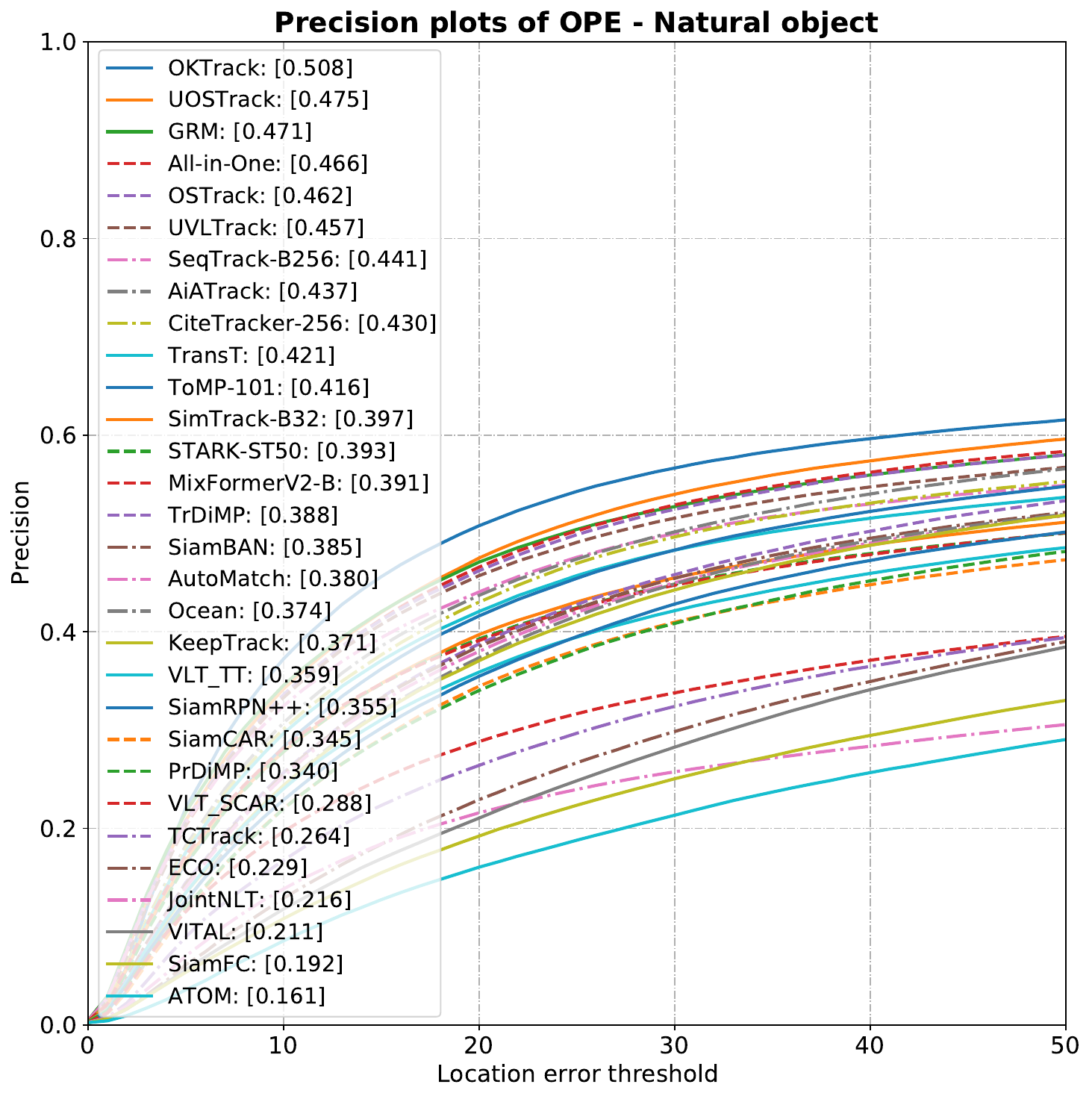}}

\subfloat{\includegraphics[width =0.25\columnwidth]{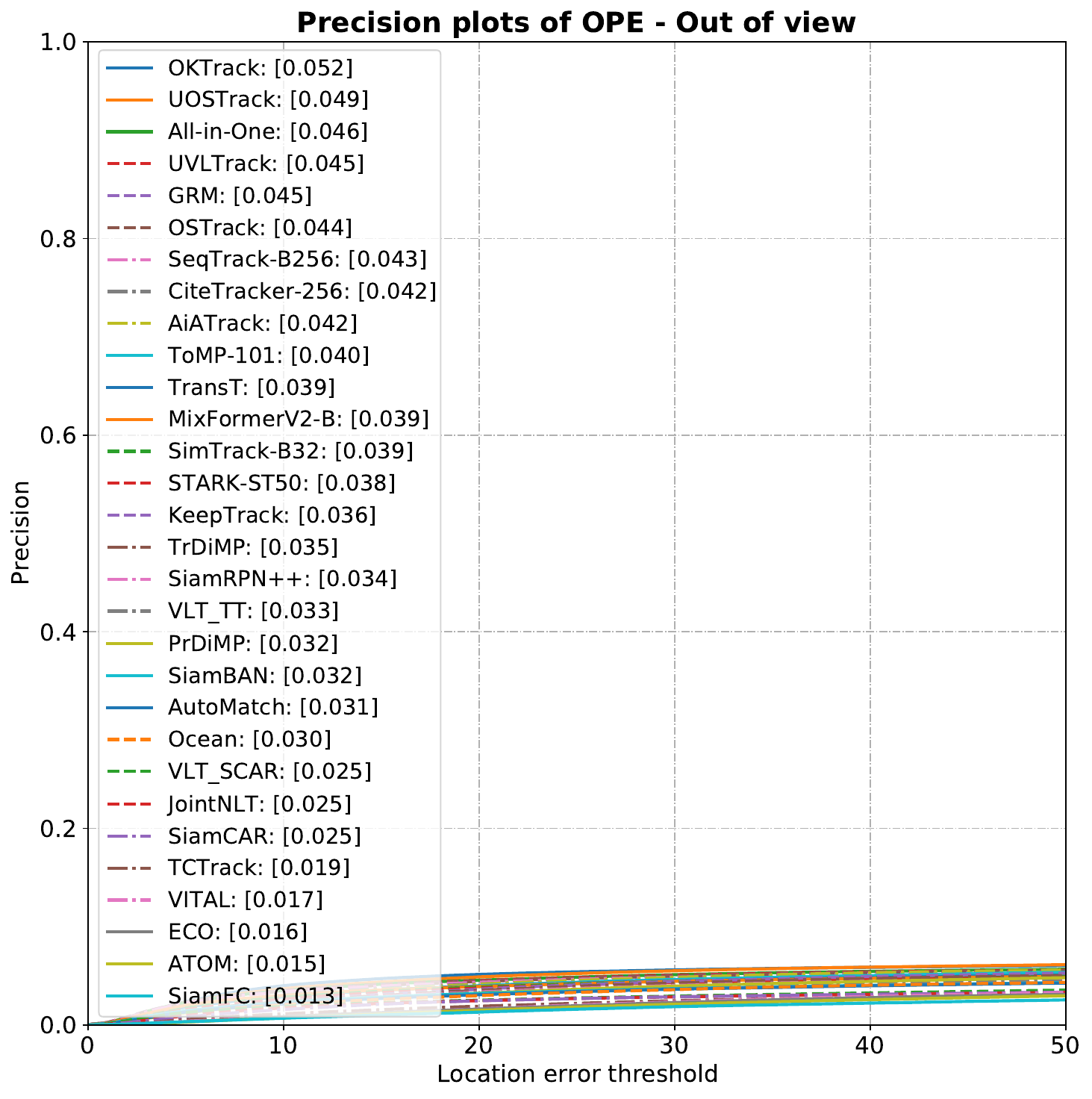}}
\subfloat{\includegraphics[width =0.25\columnwidth]{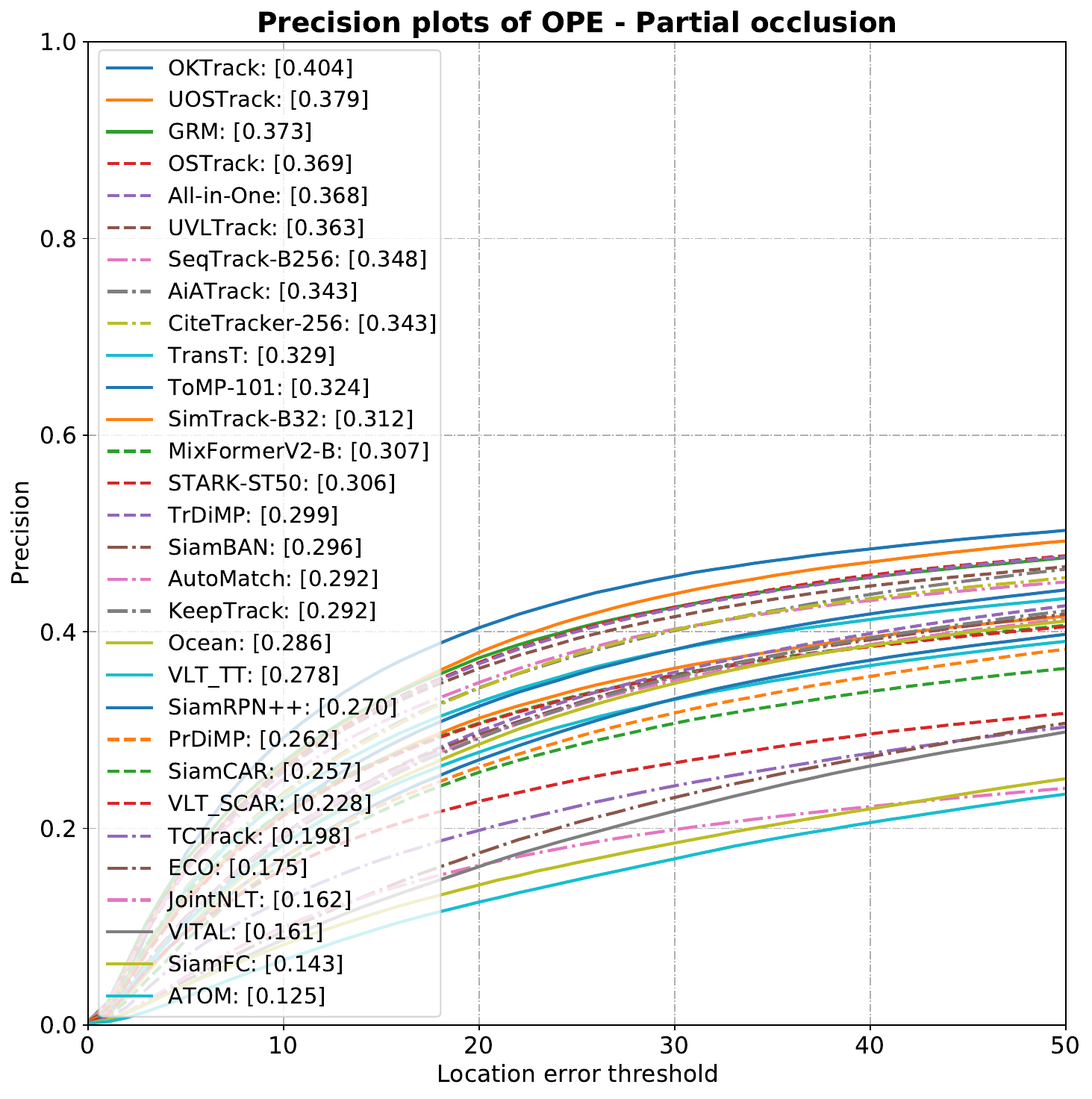}}
\subfloat{\includegraphics[width =0.25\columnwidth]{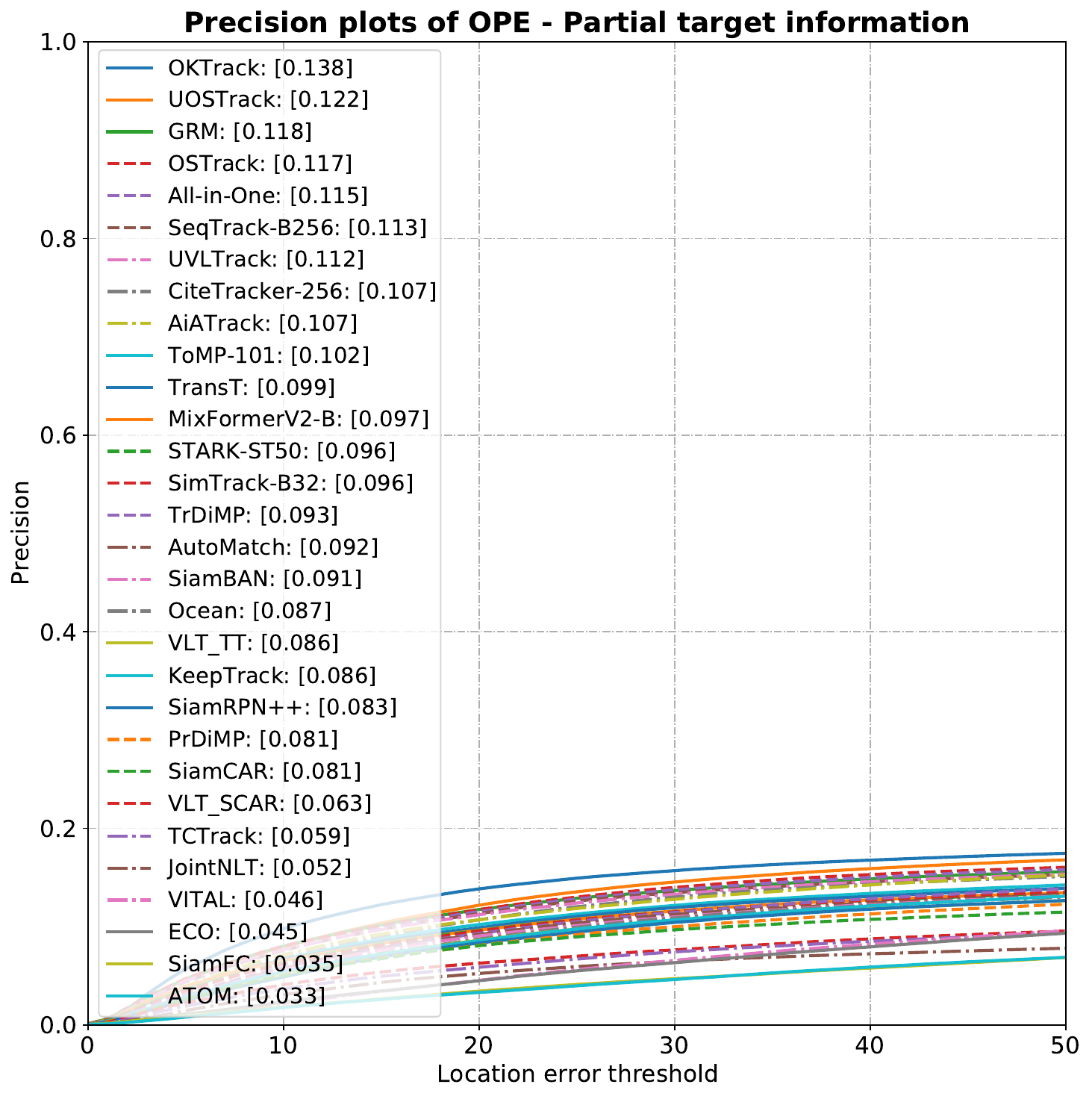}}
\subfloat{\includegraphics[width =0.25\columnwidth]{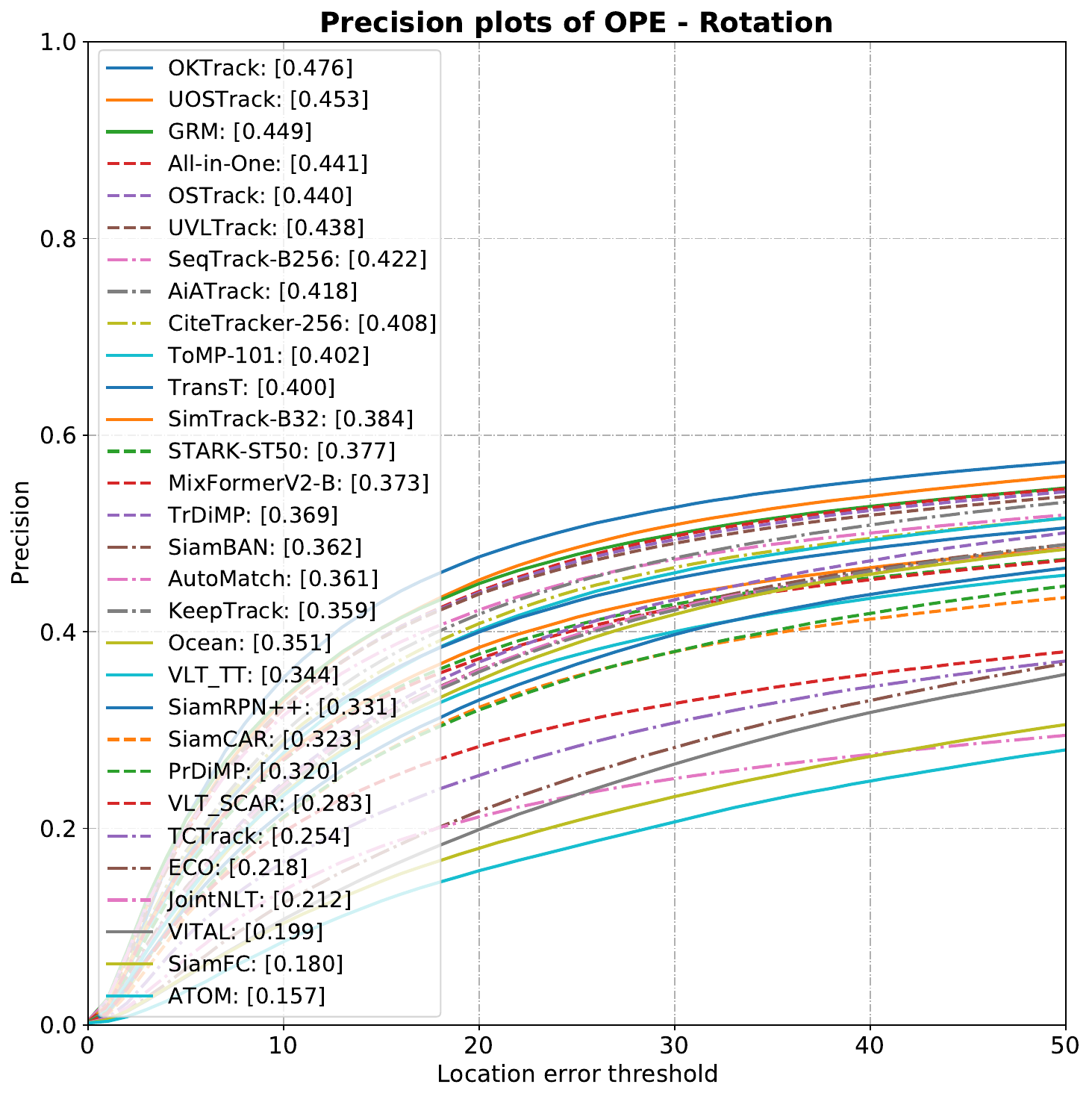}}

\subfloat{\includegraphics[width =0.25\columnwidth]{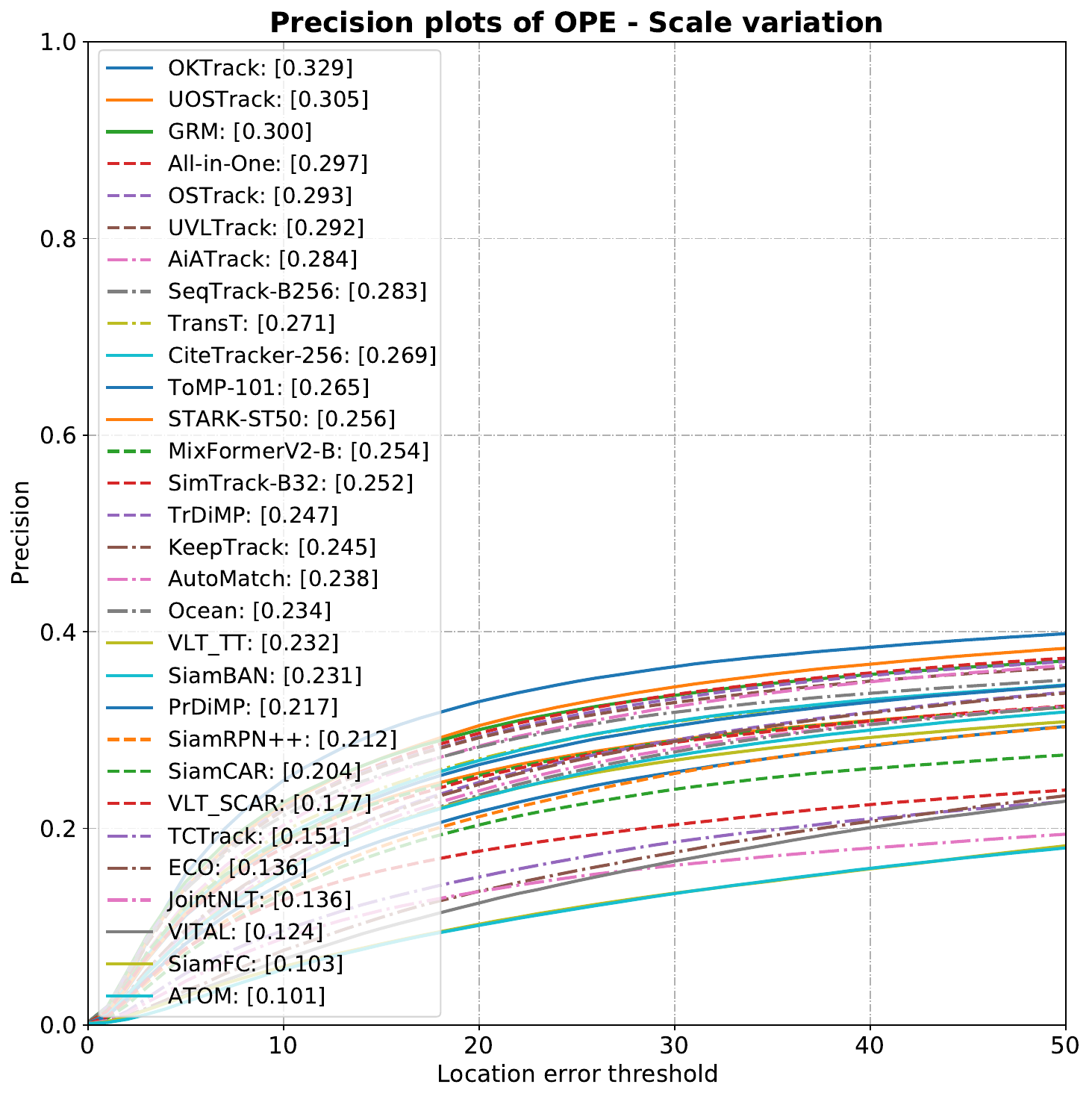}}
\subfloat{\includegraphics[width =0.25\columnwidth]{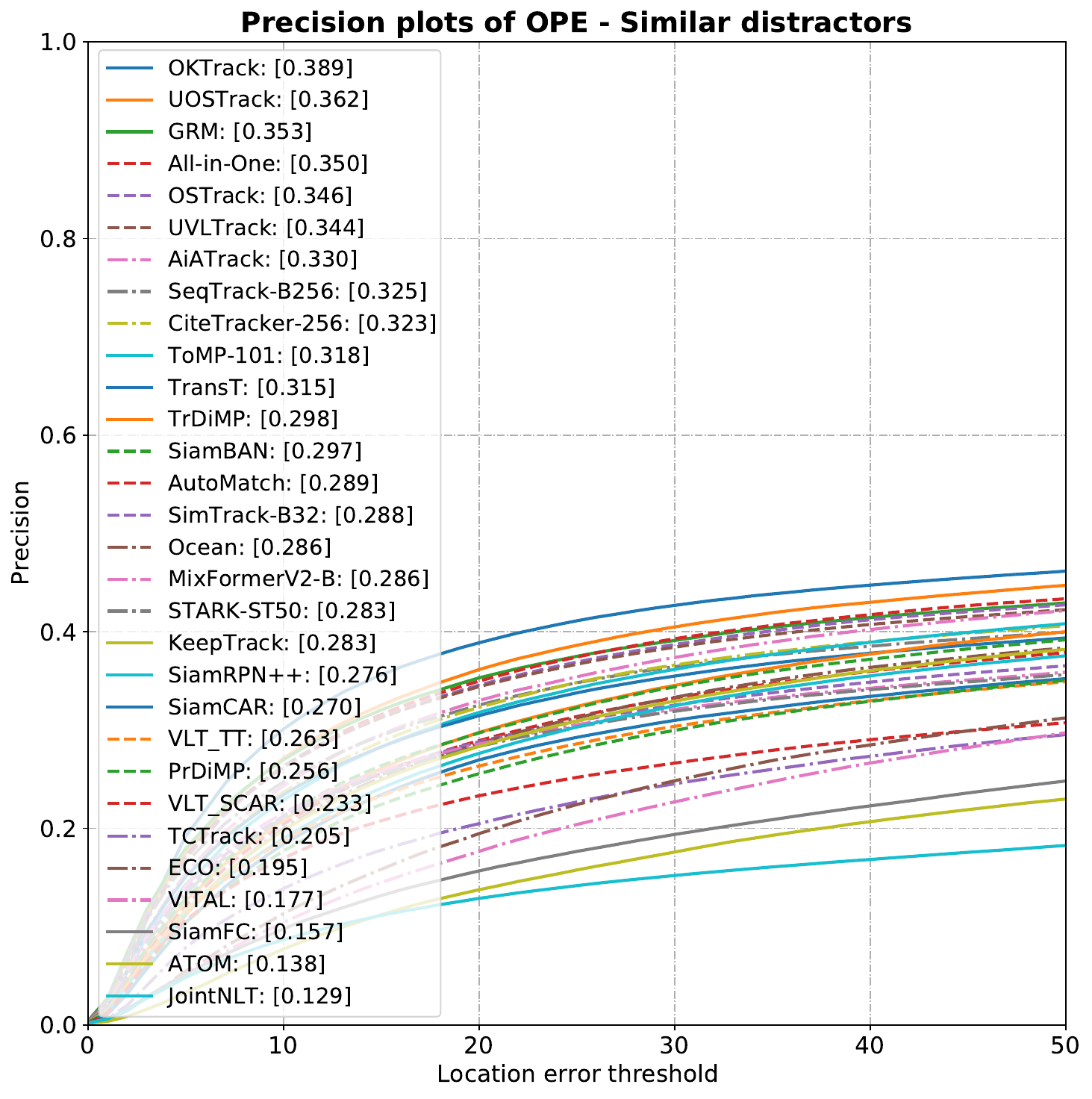}}
\subfloat{\includegraphics[width =0.25\columnwidth]{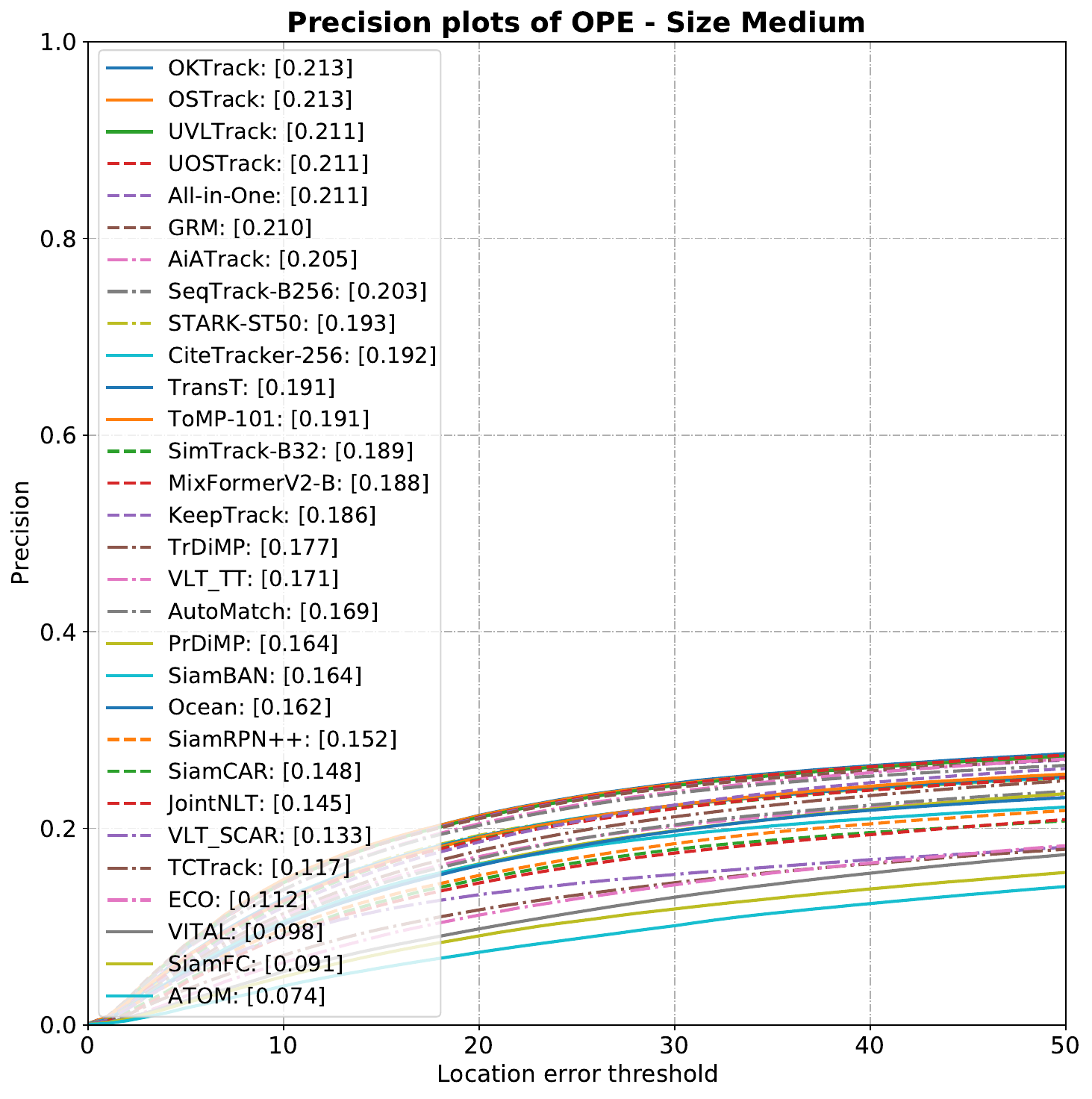}}
\subfloat{\includegraphics[width =0.25\columnwidth]{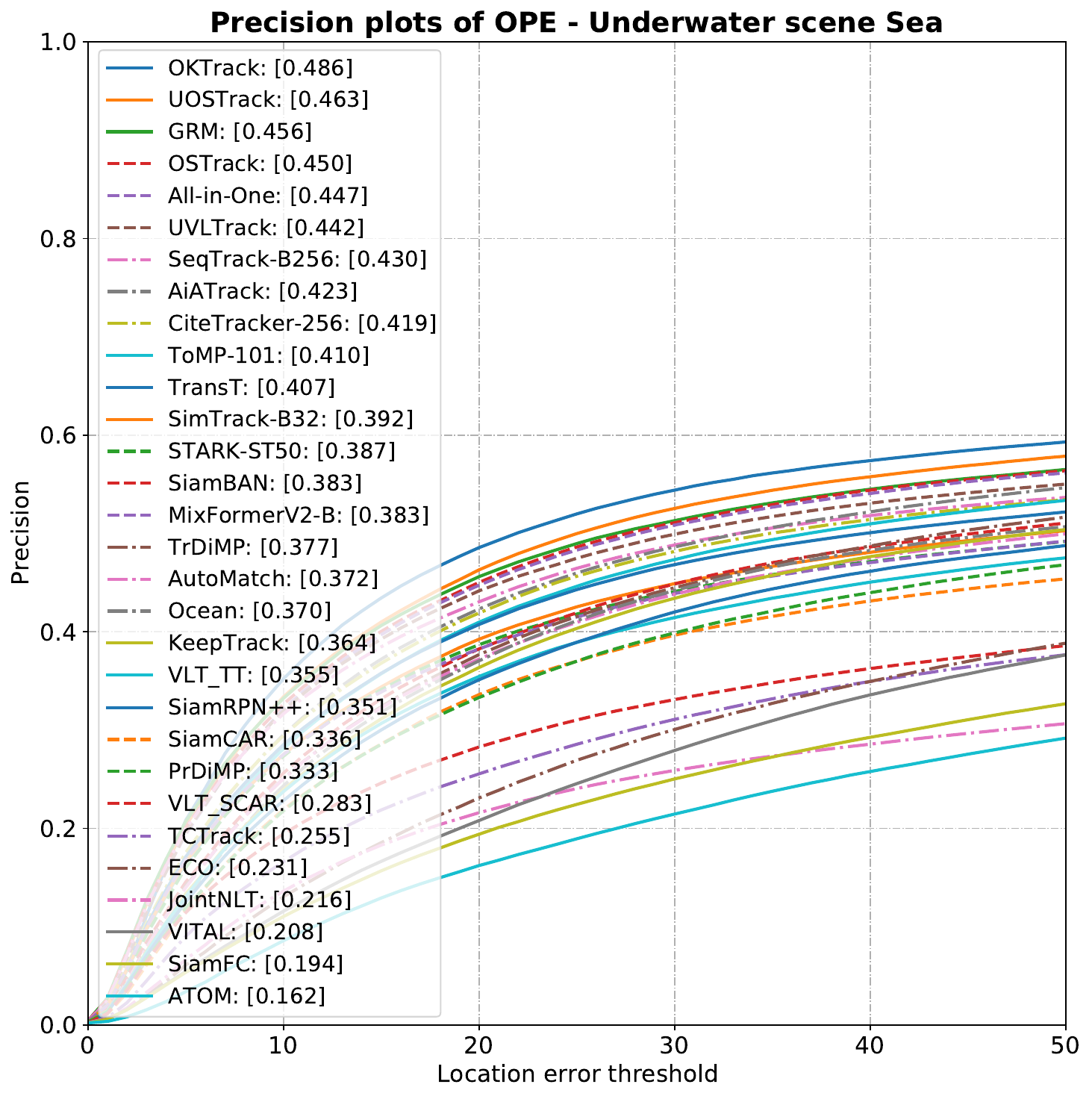}}

\subfloat{\includegraphics[width =0.25\columnwidth]{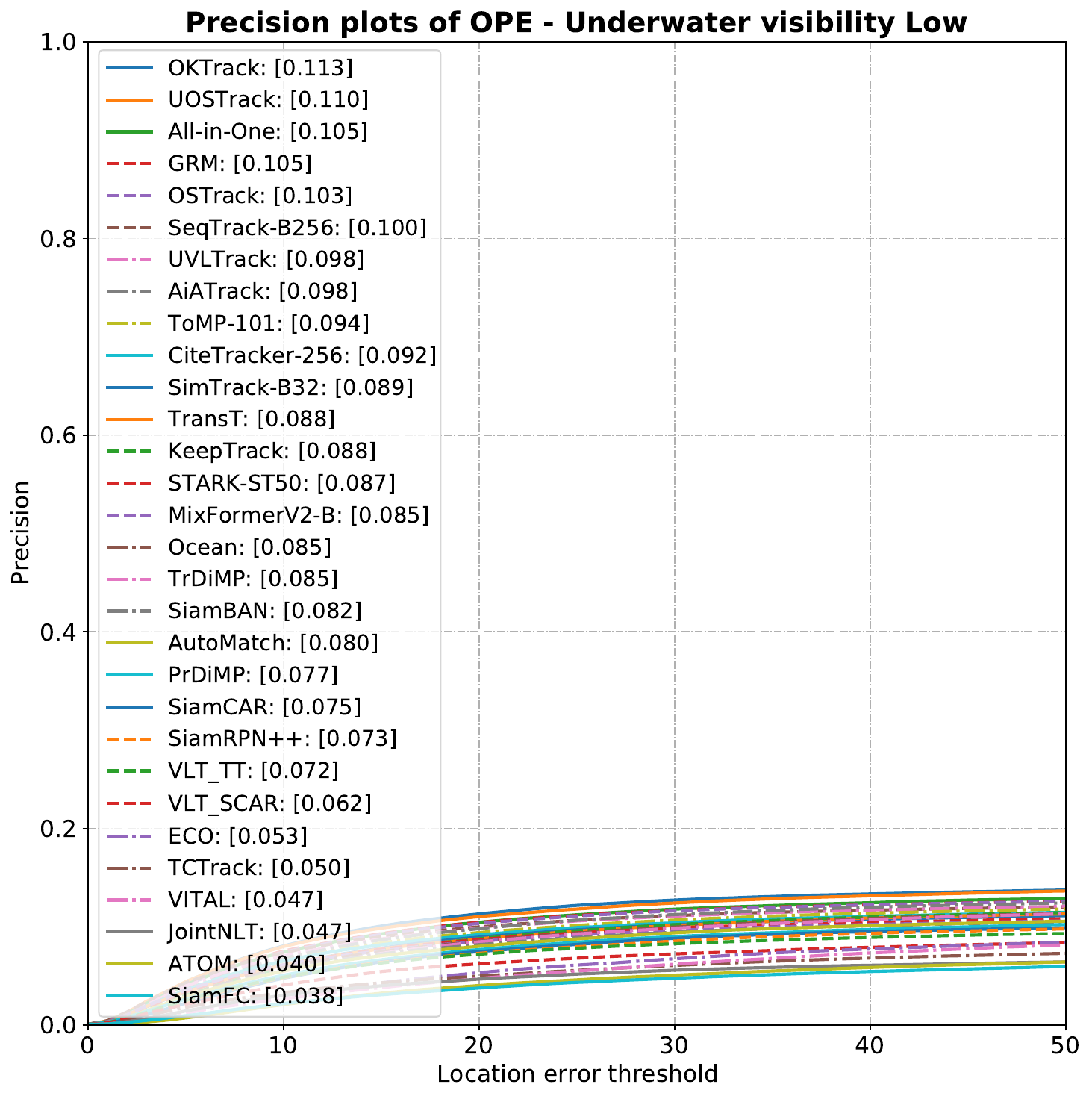}}
\subfloat{\includegraphics[width =0.25\columnwidth]{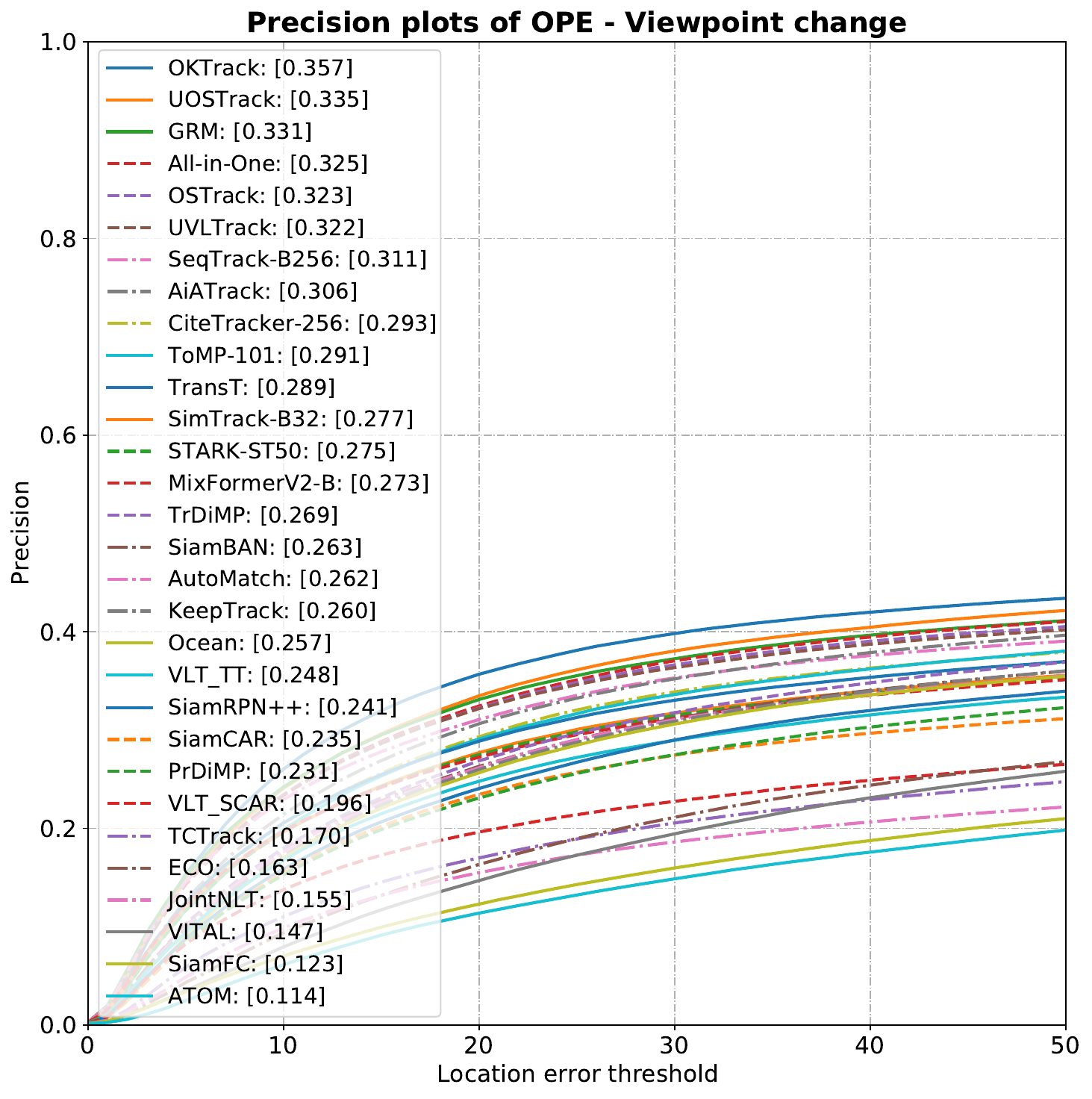}}
\subfloat{\includegraphics[width =0.25\columnwidth]{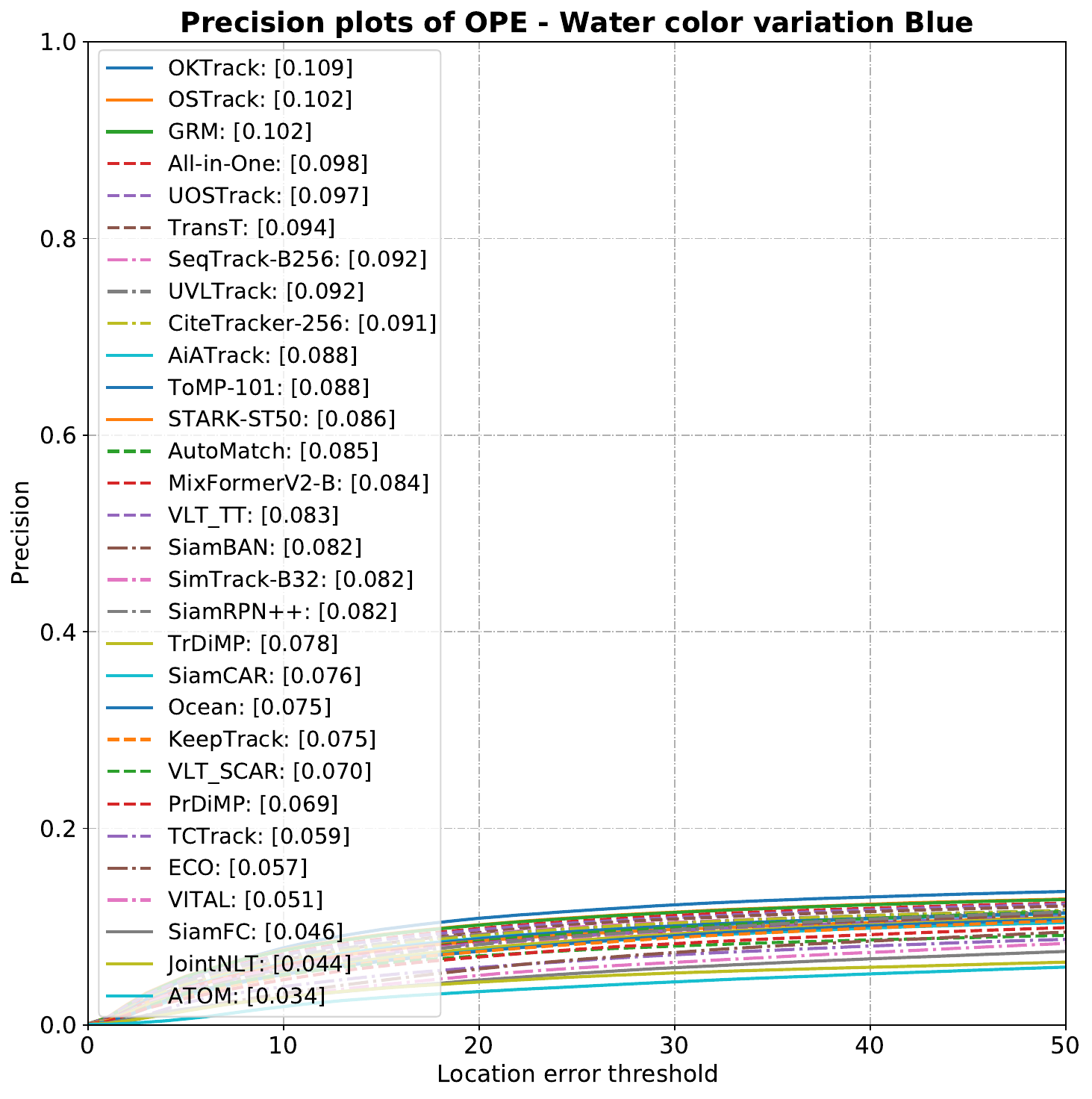}}
\subfloat{\includegraphics[width =0.25\columnwidth]{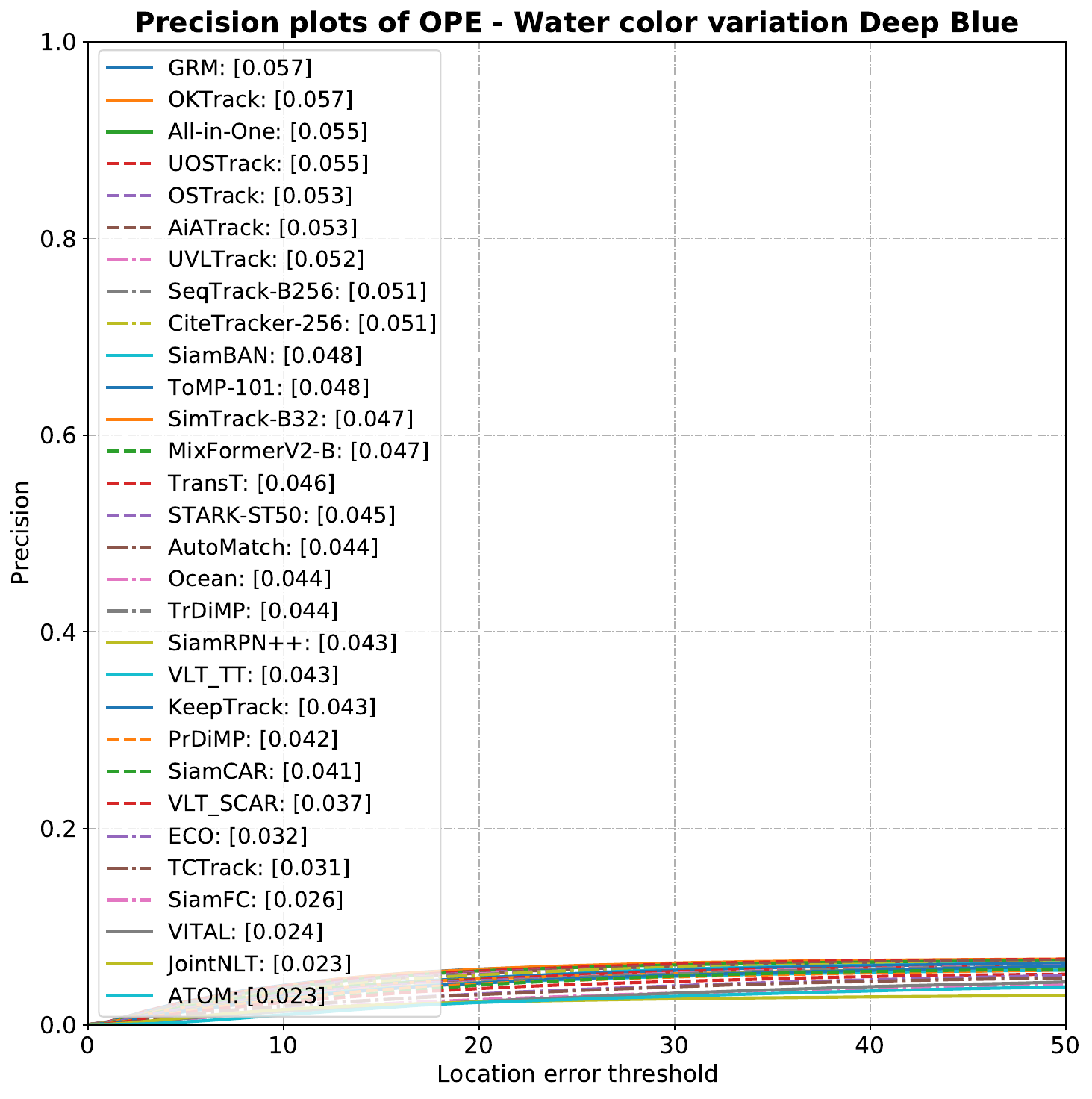}}

  \caption{Performances of baseline trackers on the WebOUT-1M test set of different attributes using \textbf{Pre} scores. Best viewed by zooming in.}
  \label{fig:attribute_results_Pre}
\end{figure*}

\begin{figure*}[t]
\vspace{-0.6cm}
  \centering

\subfloat{\includegraphics[width =0.25\columnwidth]{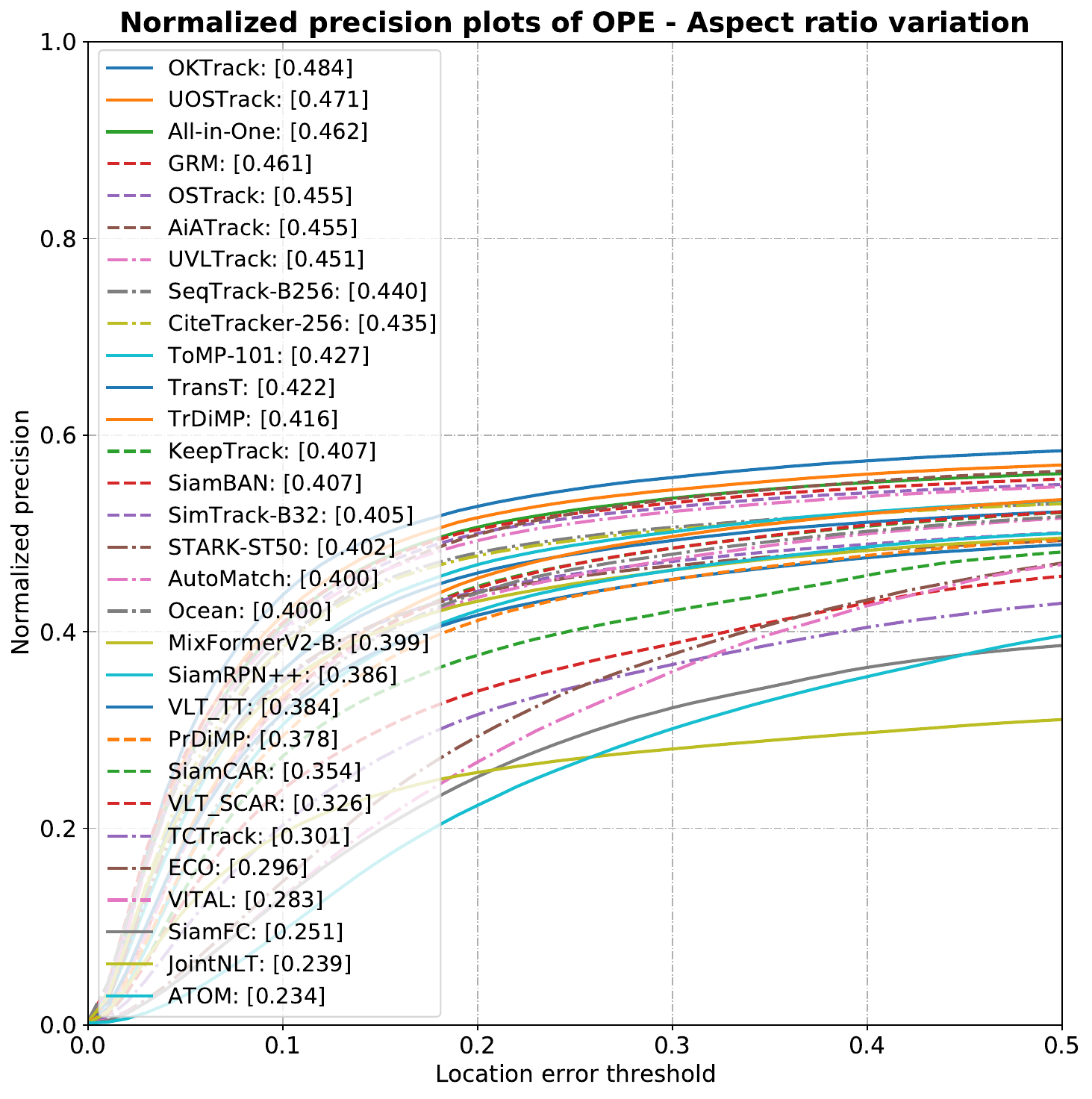}}
\subfloat{\includegraphics[width =0.25\columnwidth]{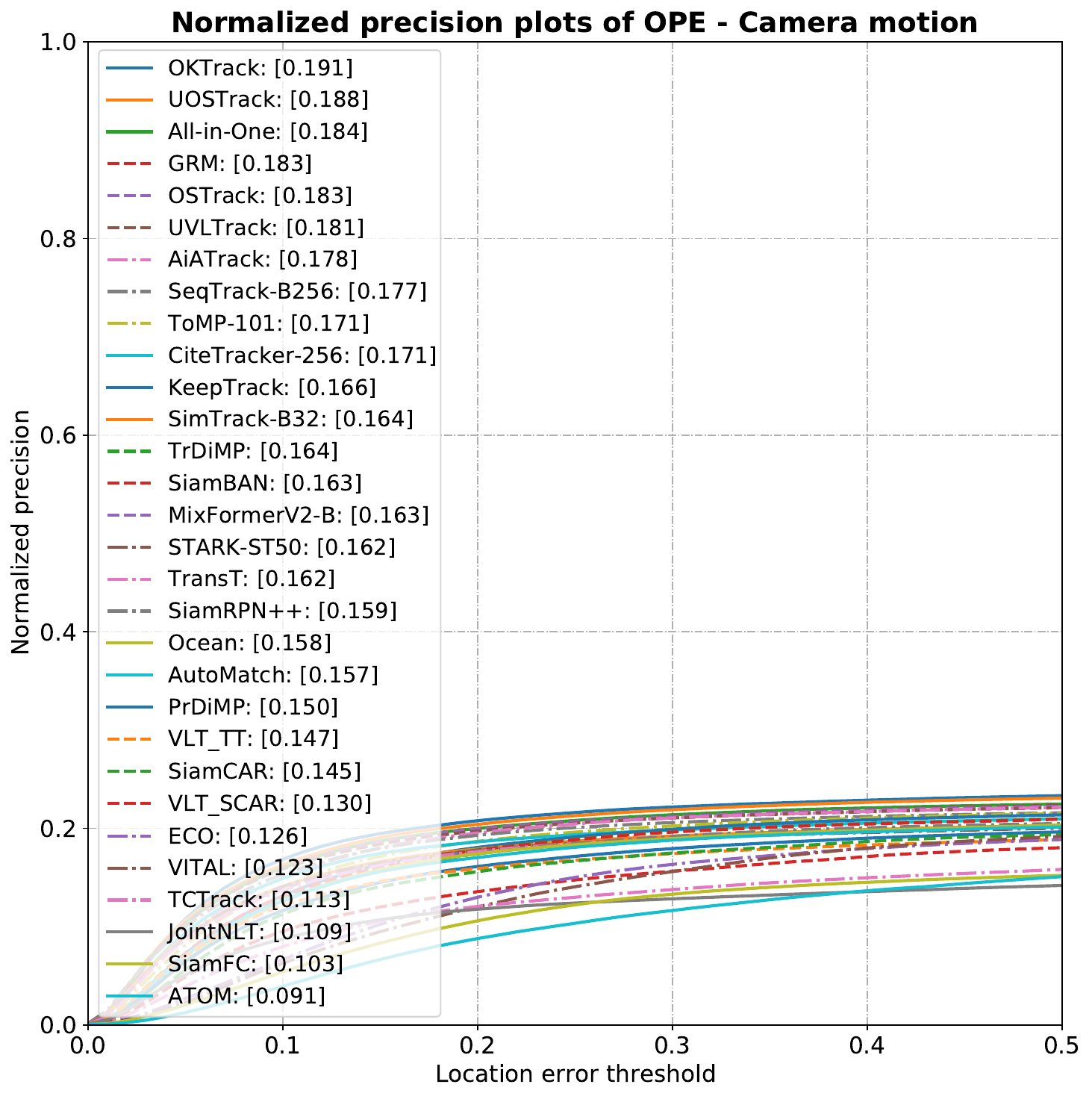}}
\subfloat{\includegraphics[width =0.25\columnwidth]{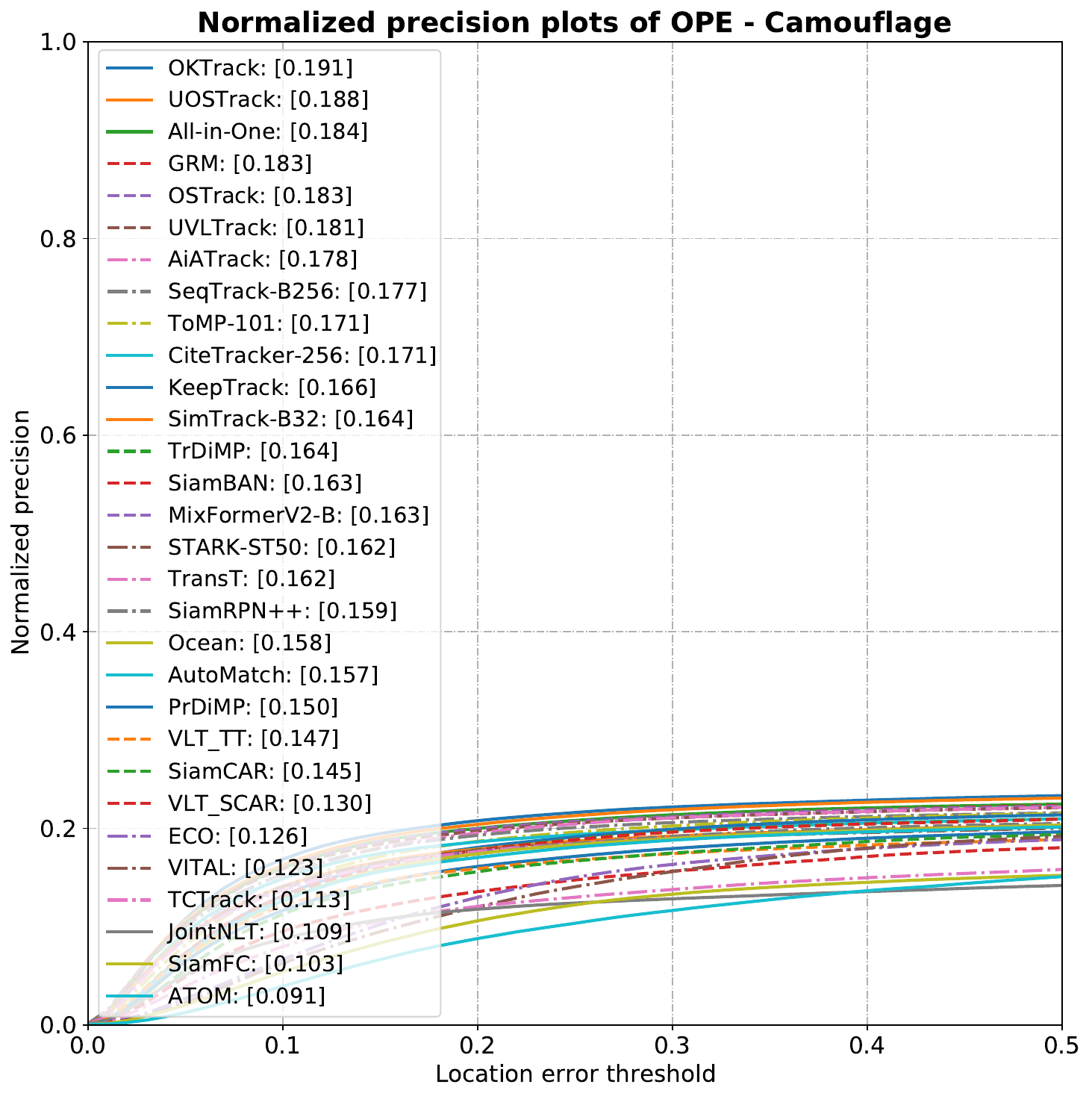}}
\subfloat{\includegraphics[width =0.25\columnwidth]{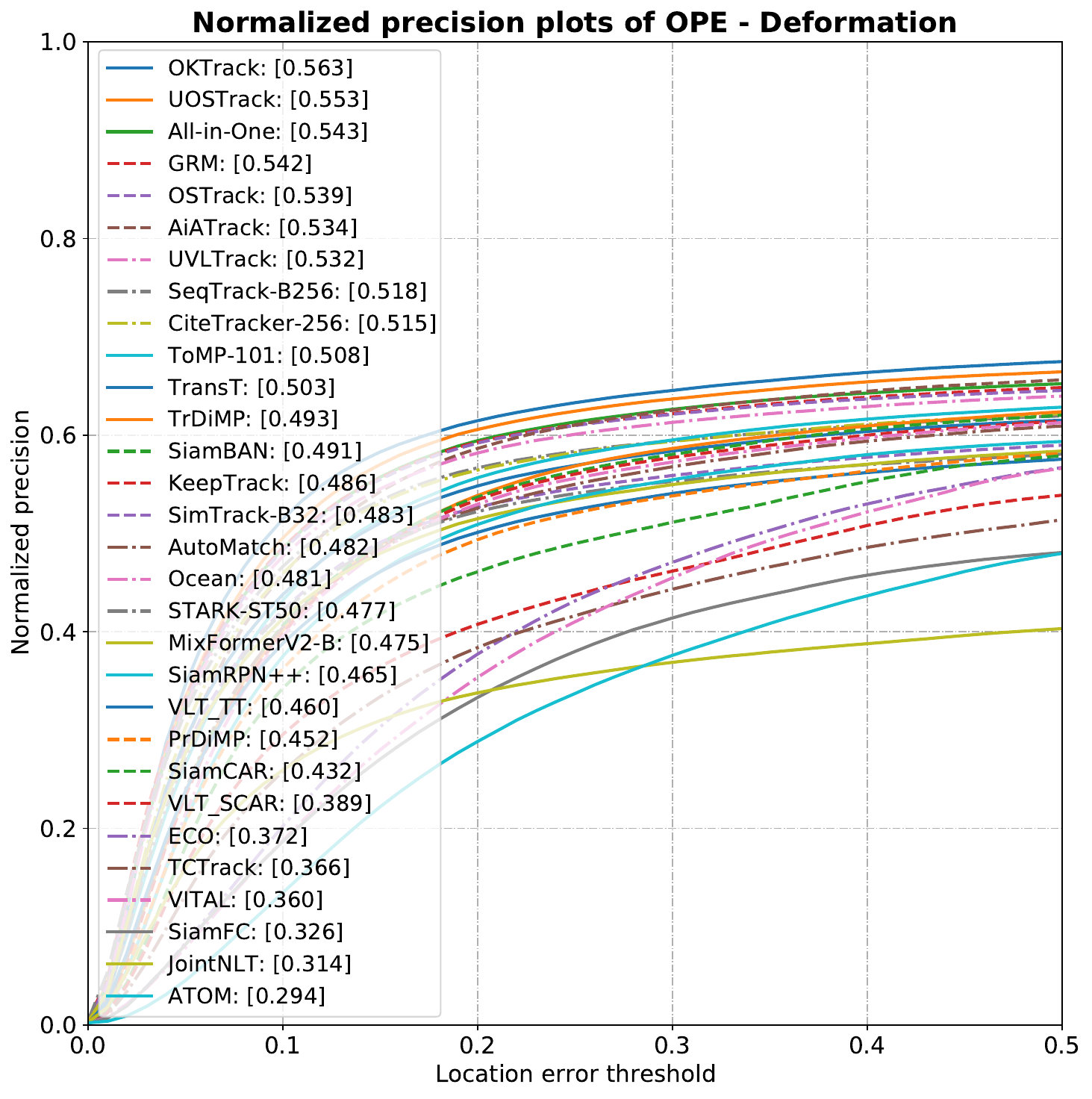}}

\subfloat{\includegraphics[width =0.25\columnwidth]{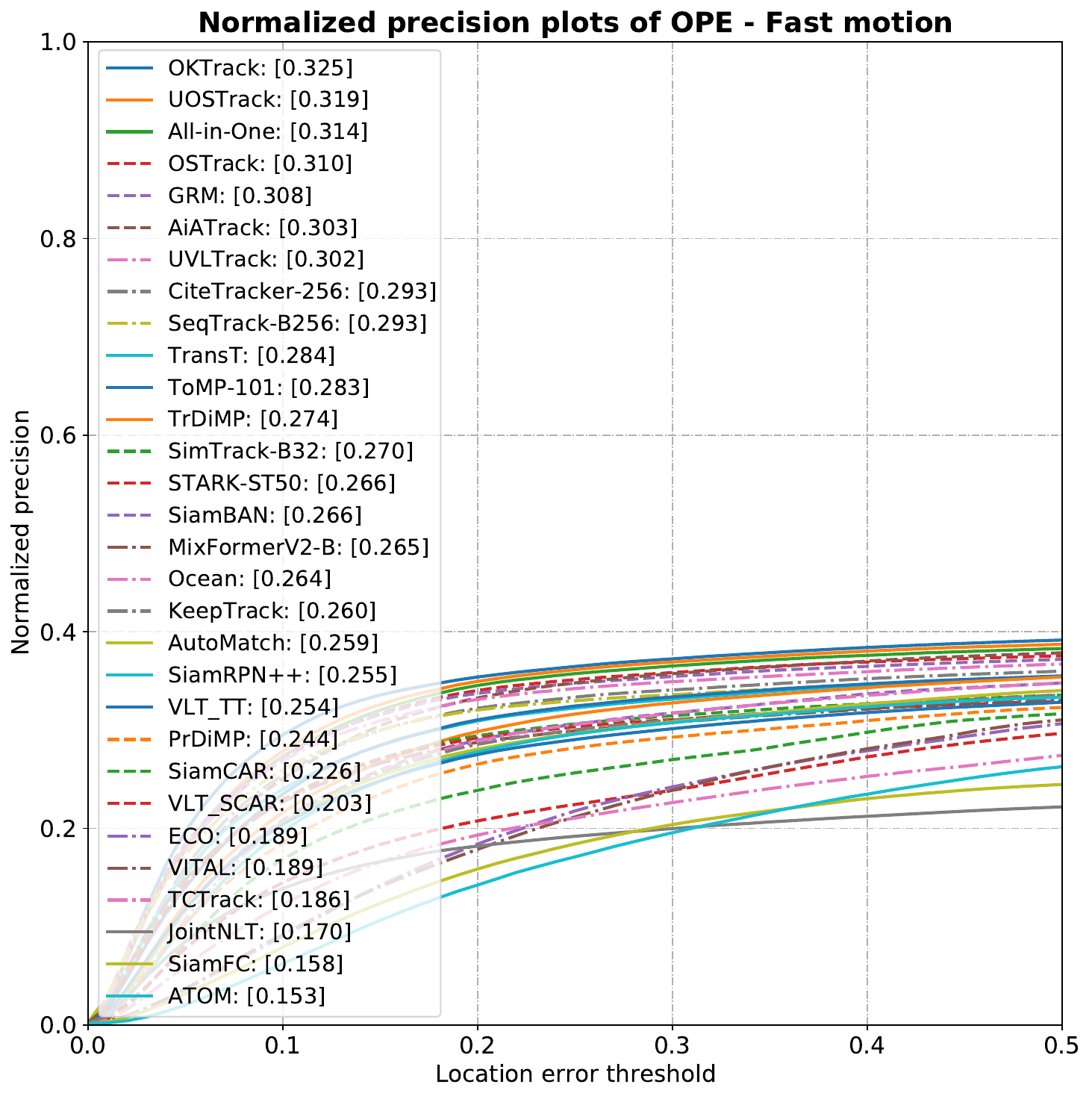}}
\subfloat{\includegraphics[width =0.25\columnwidth]{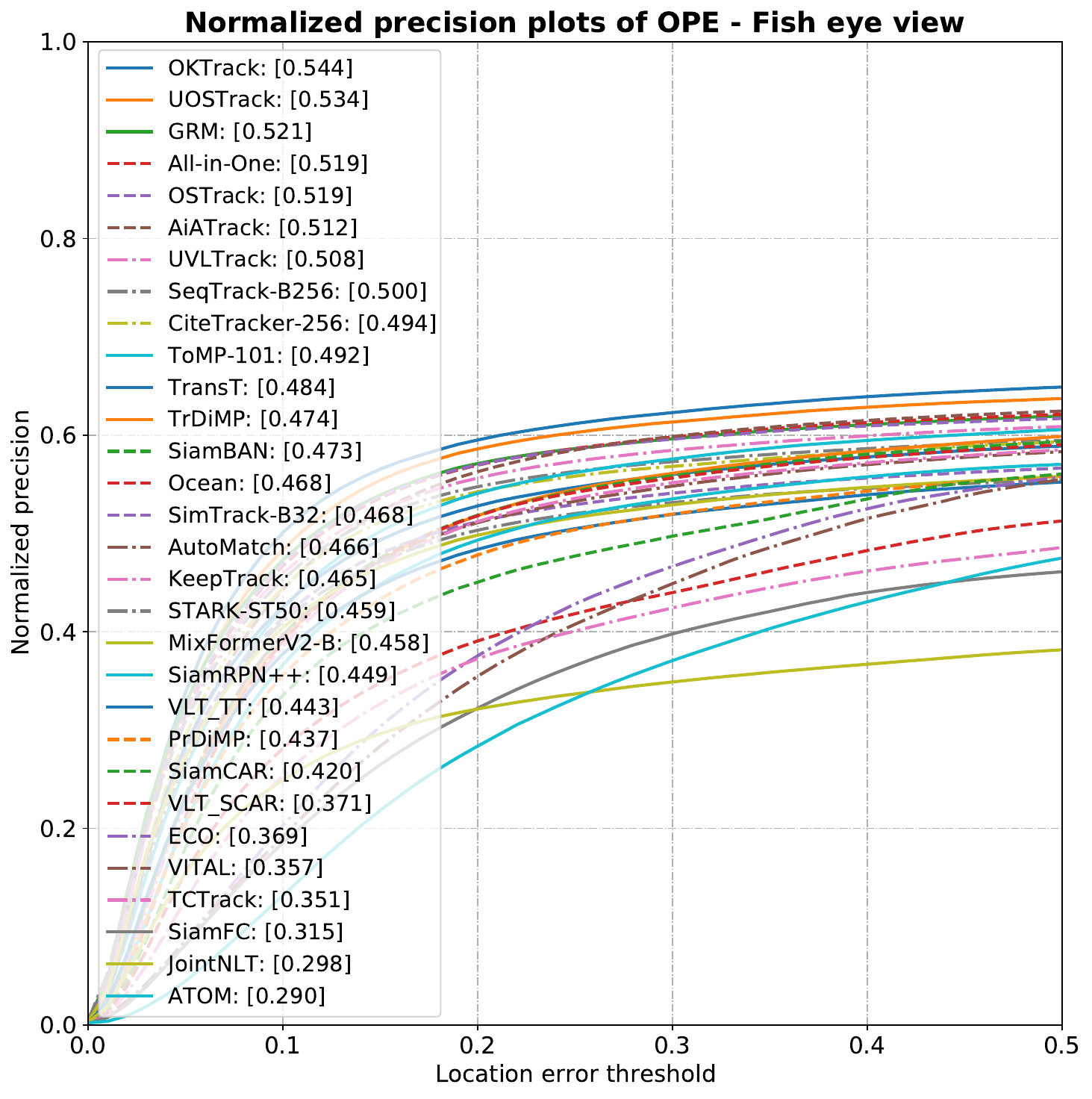}}
\subfloat{\includegraphics[width =0.25\columnwidth]{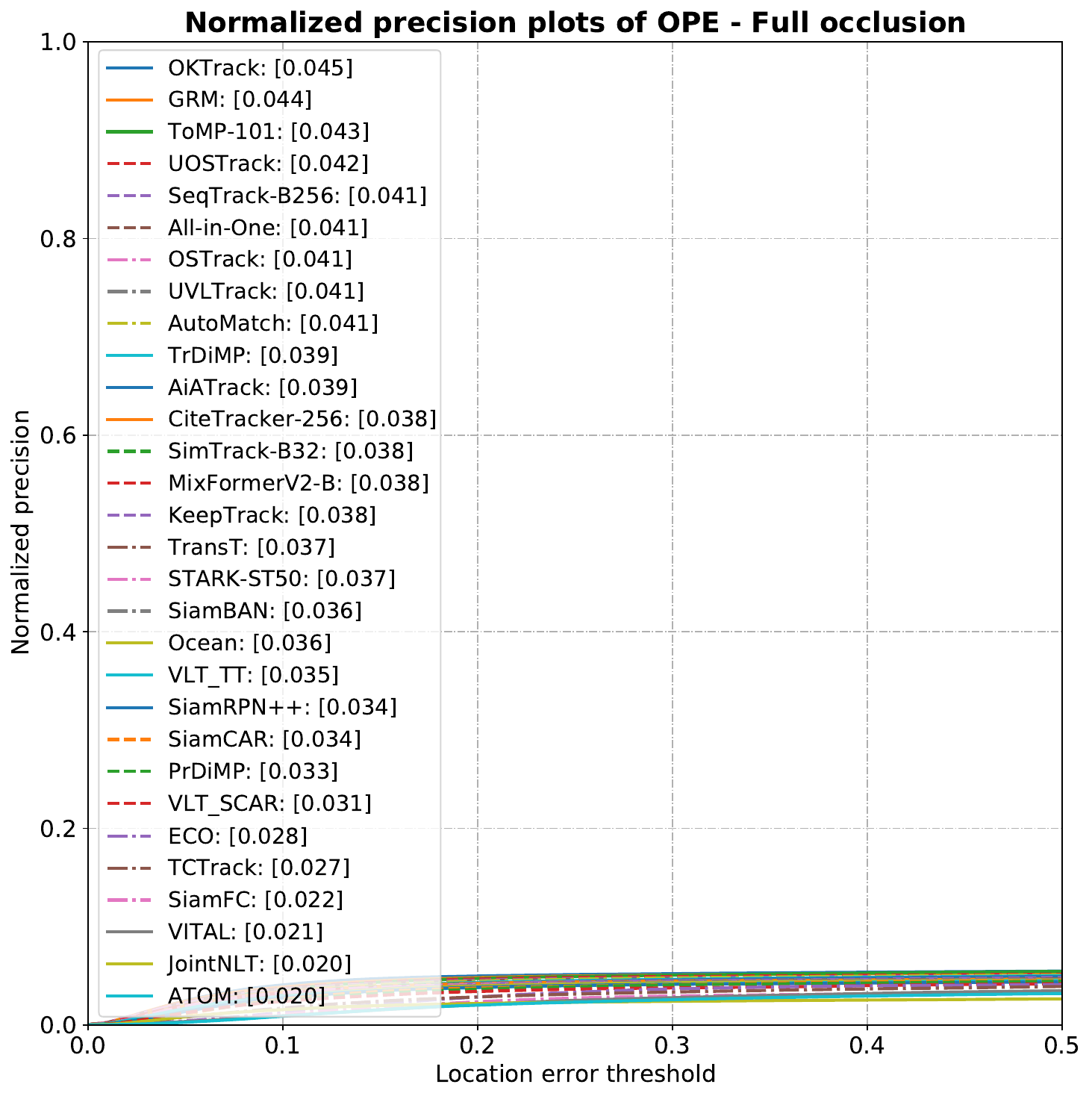}}
\subfloat{\includegraphics[width =0.25\columnwidth]{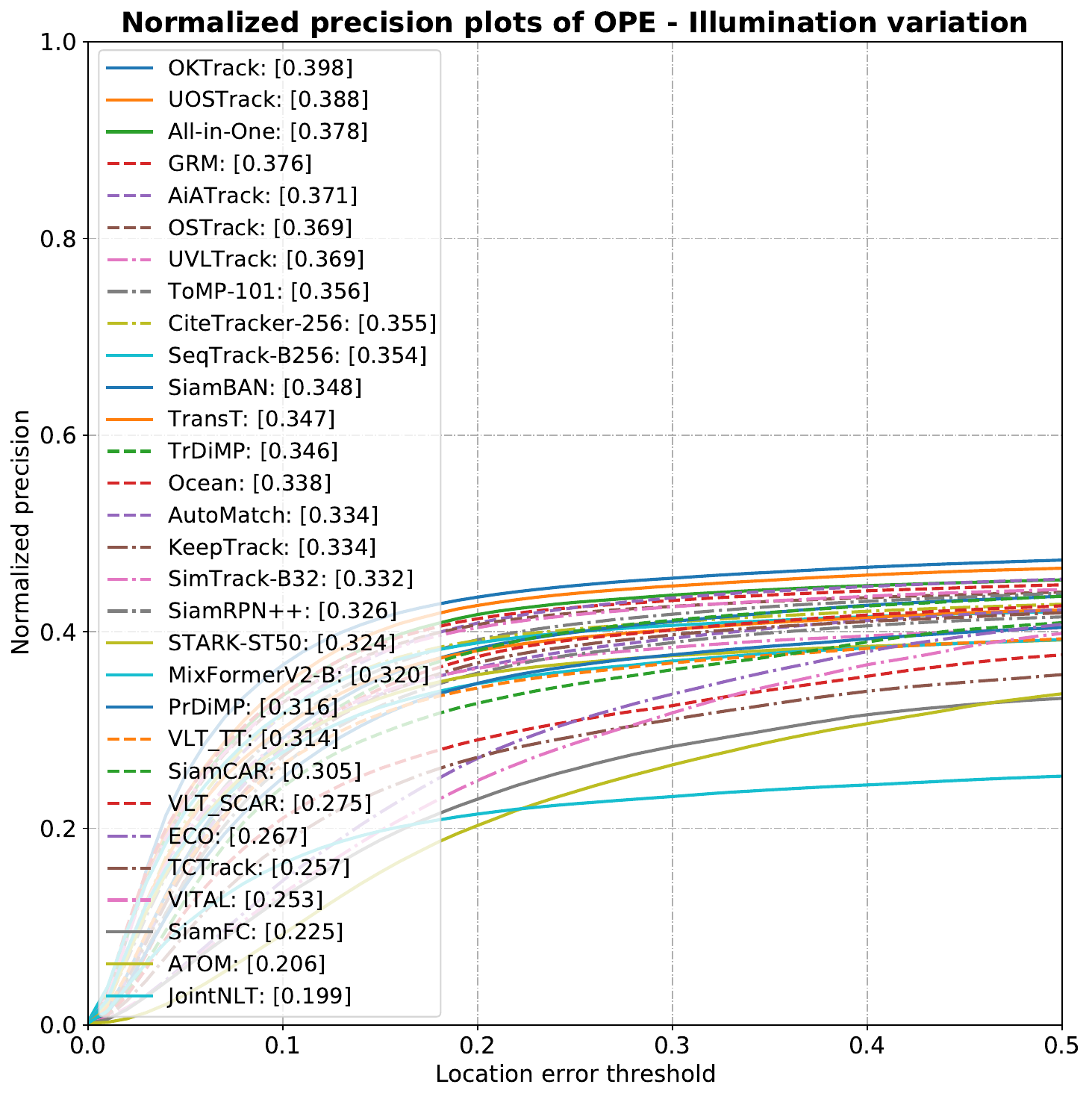}}

\subfloat{\includegraphics[width =0.25\columnwidth]{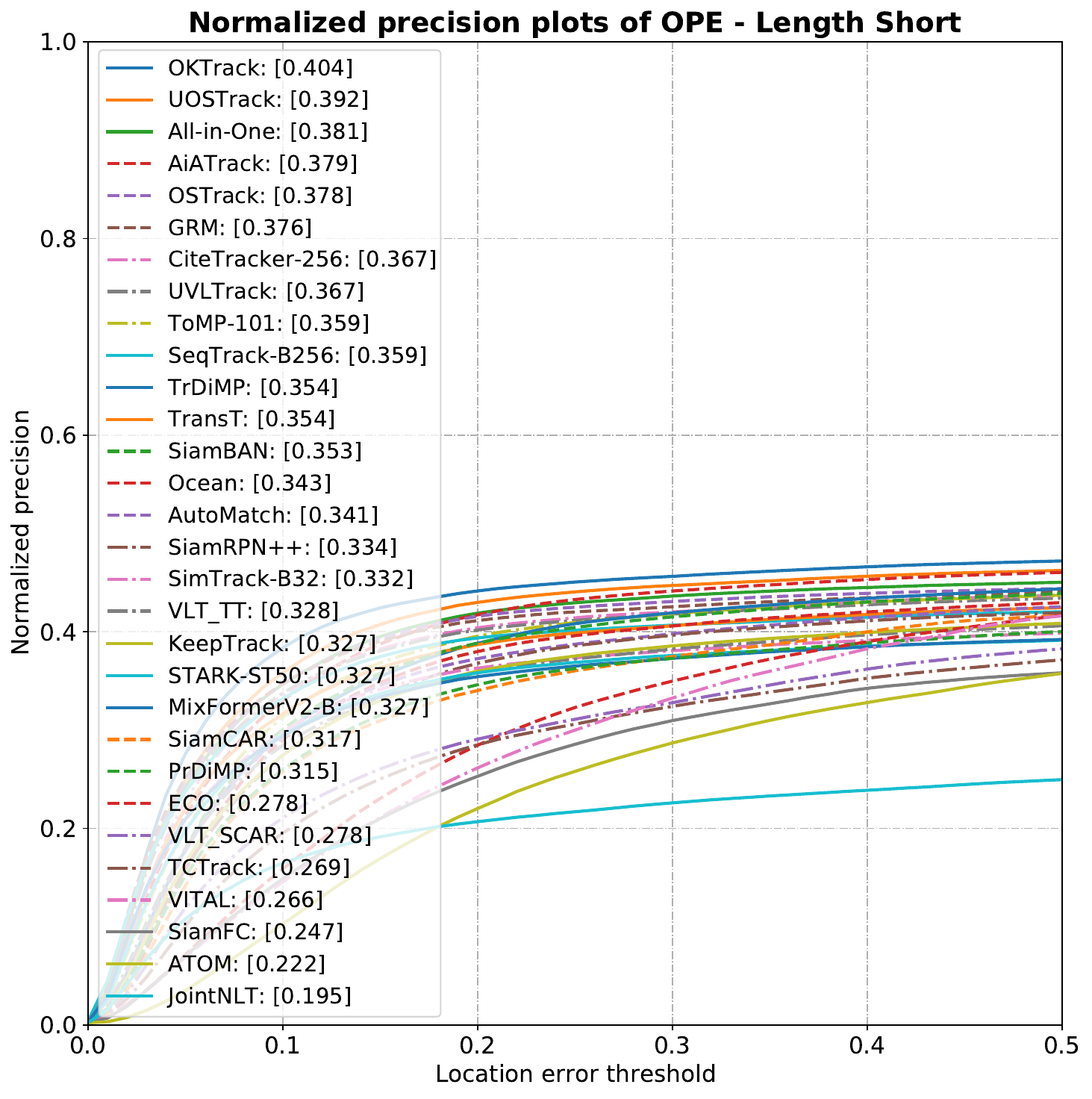}}
\subfloat{\includegraphics[width =0.25\columnwidth]{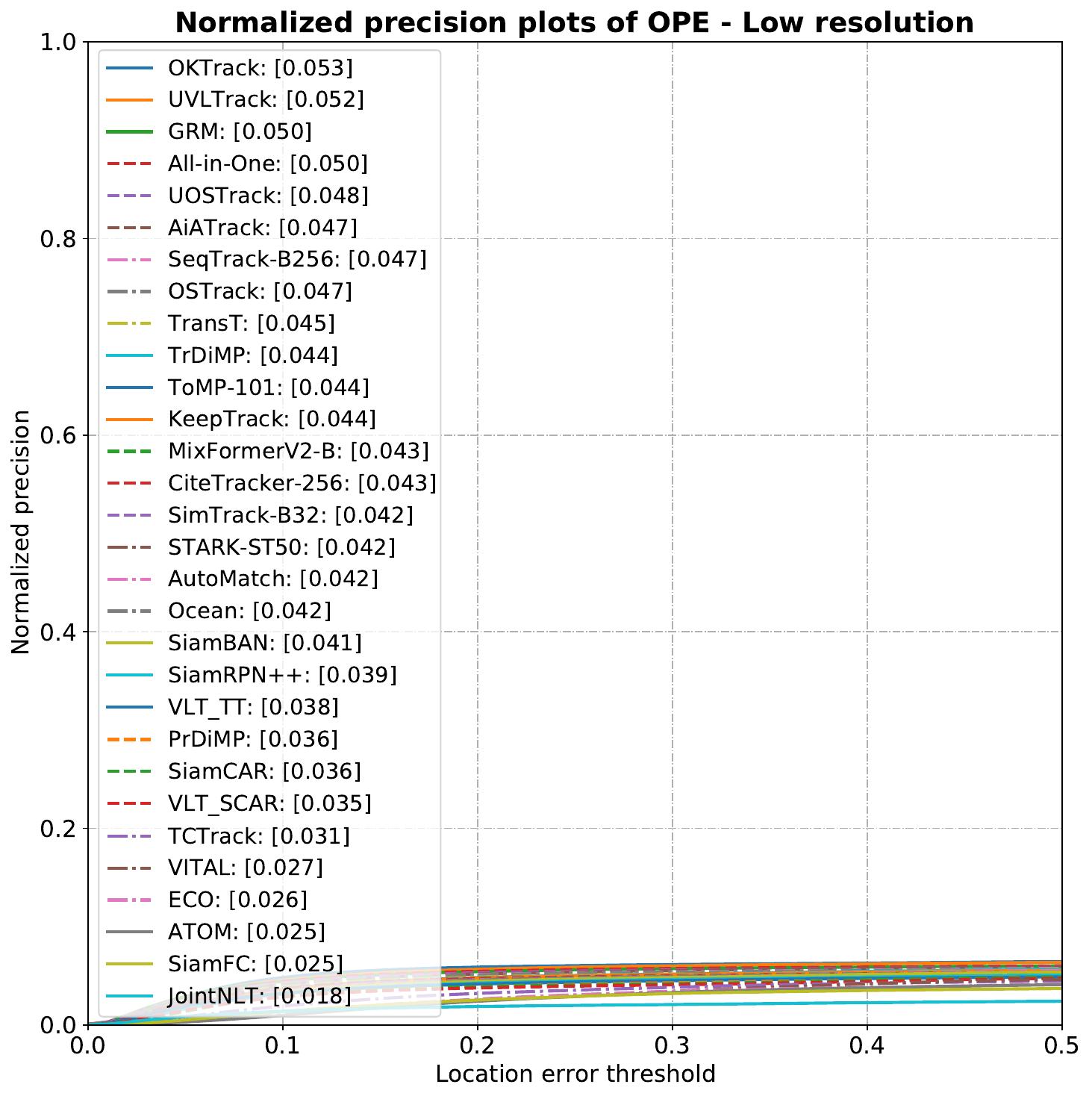}}
\subfloat{\includegraphics[width =0.25\columnwidth]{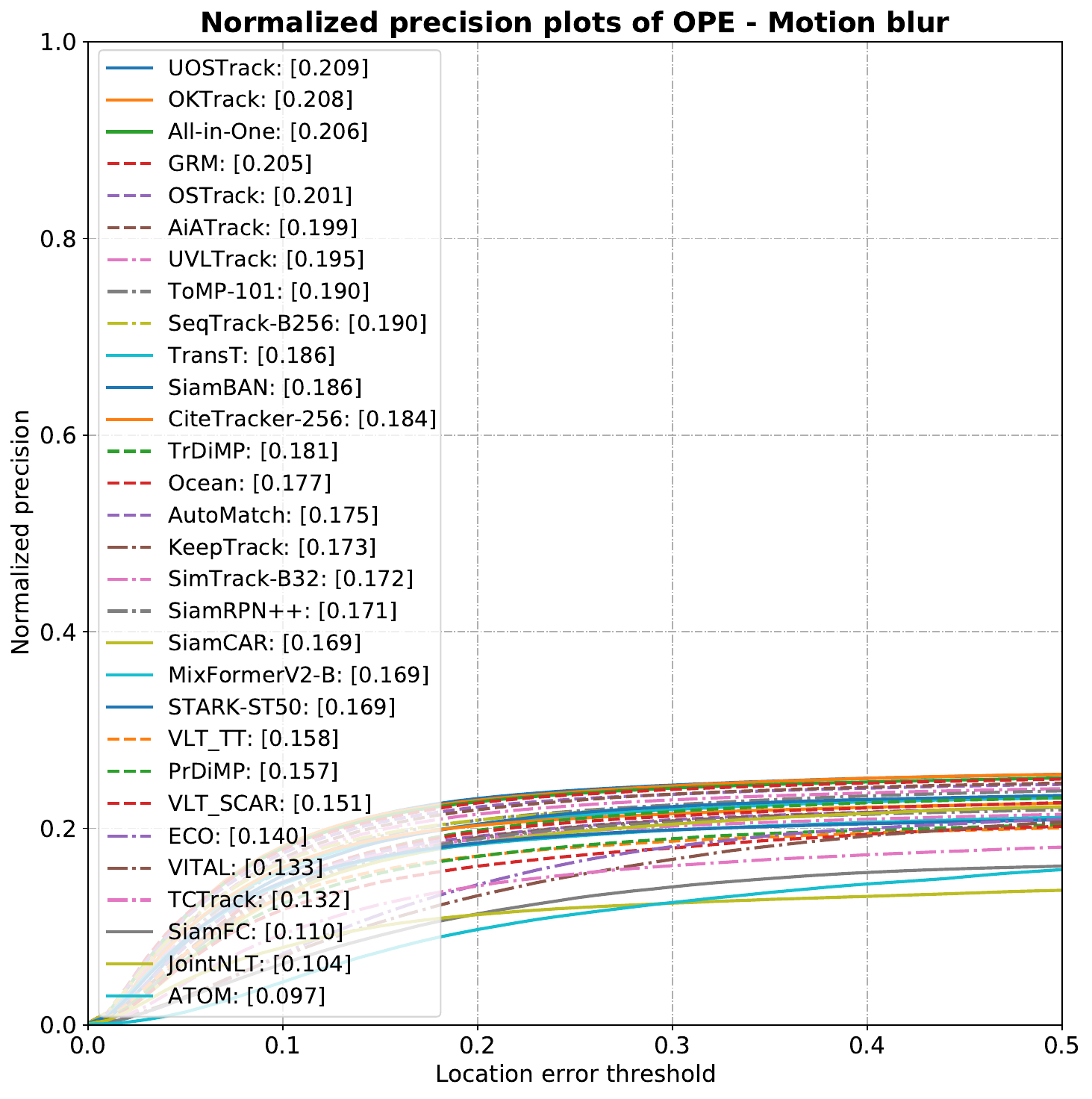}}
\subfloat{\includegraphics[width =0.25\columnwidth]{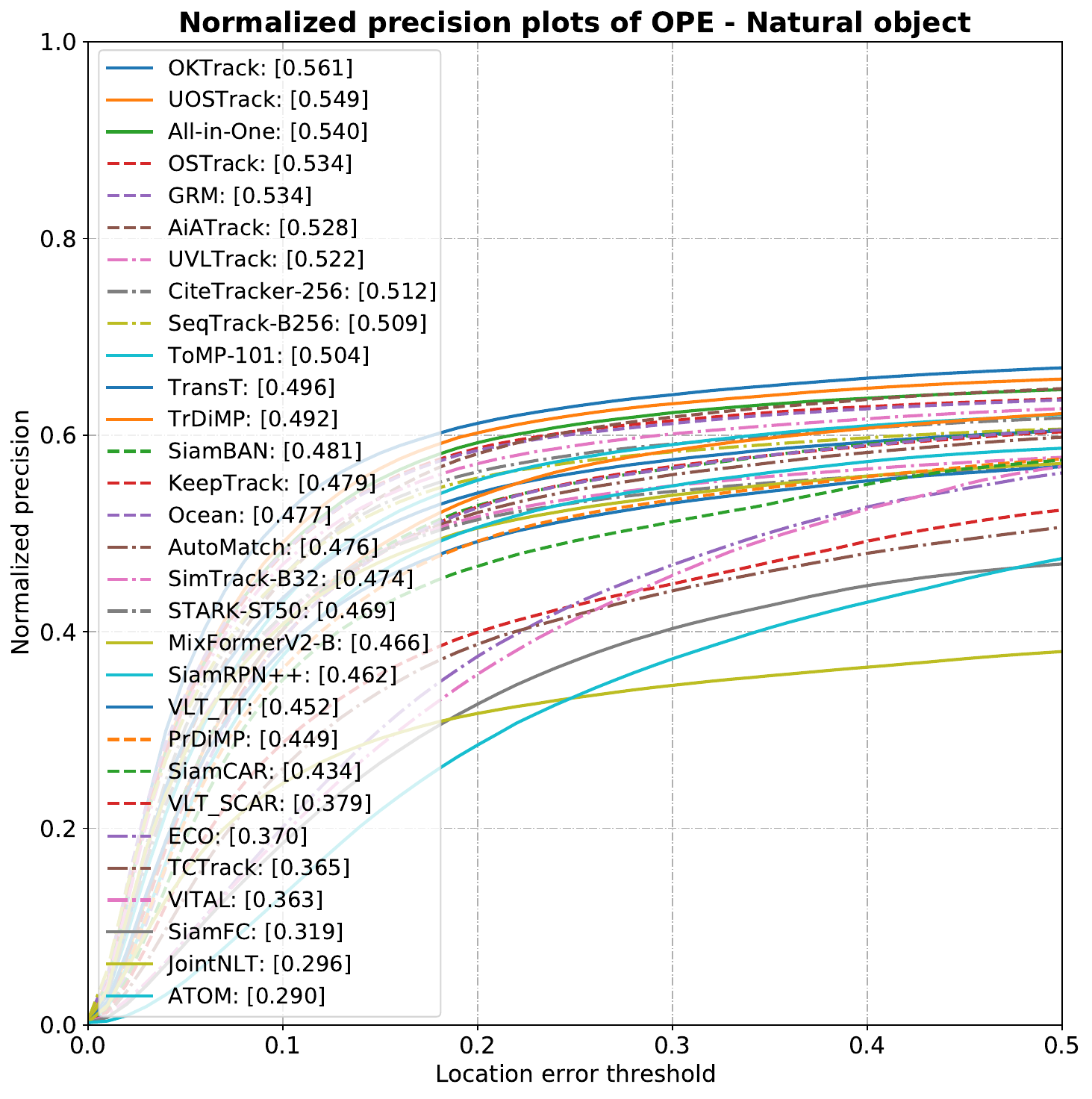}}

\subfloat{\includegraphics[width =0.25\columnwidth]{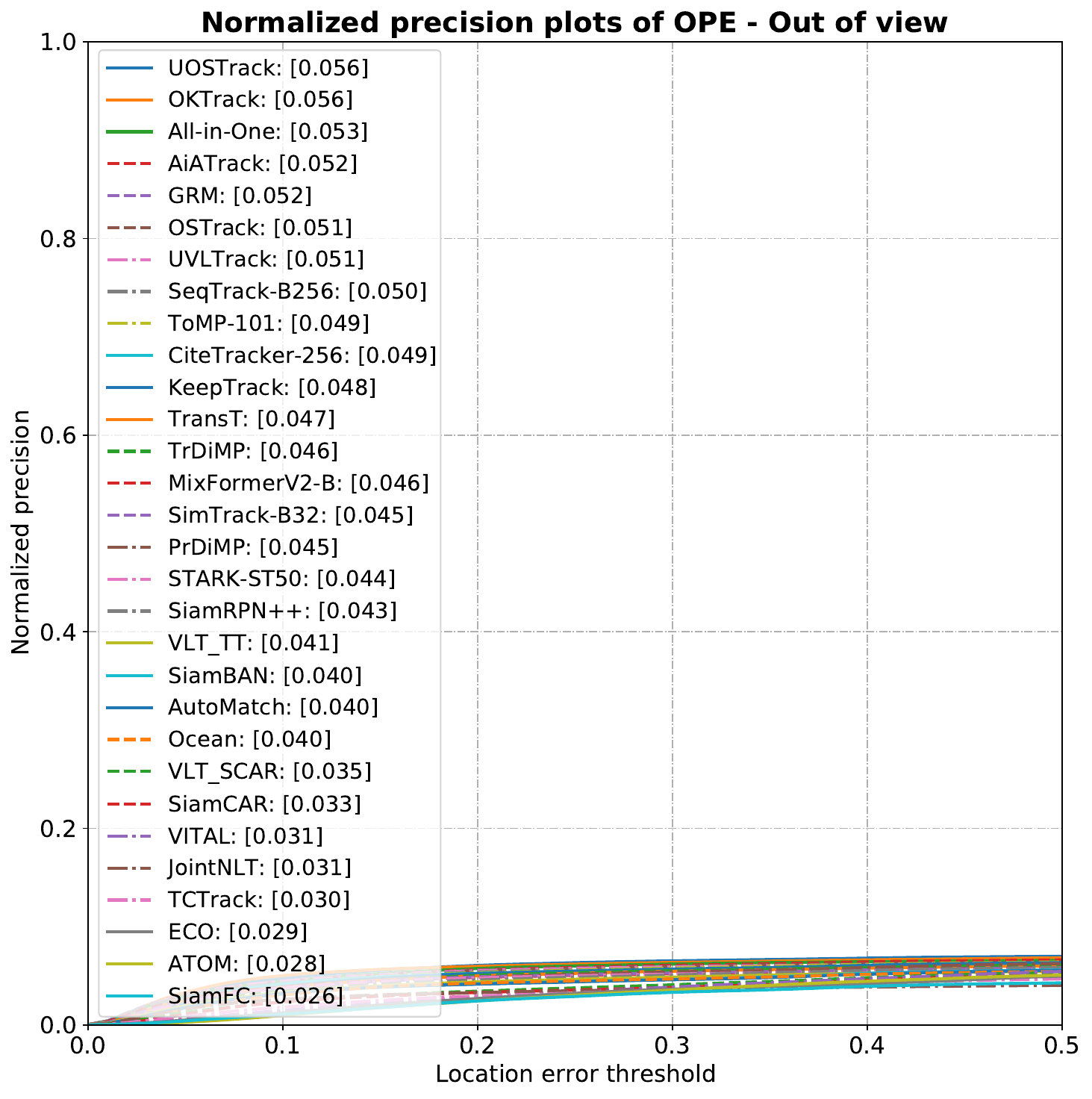}}
\subfloat{\includegraphics[width =0.25\columnwidth]{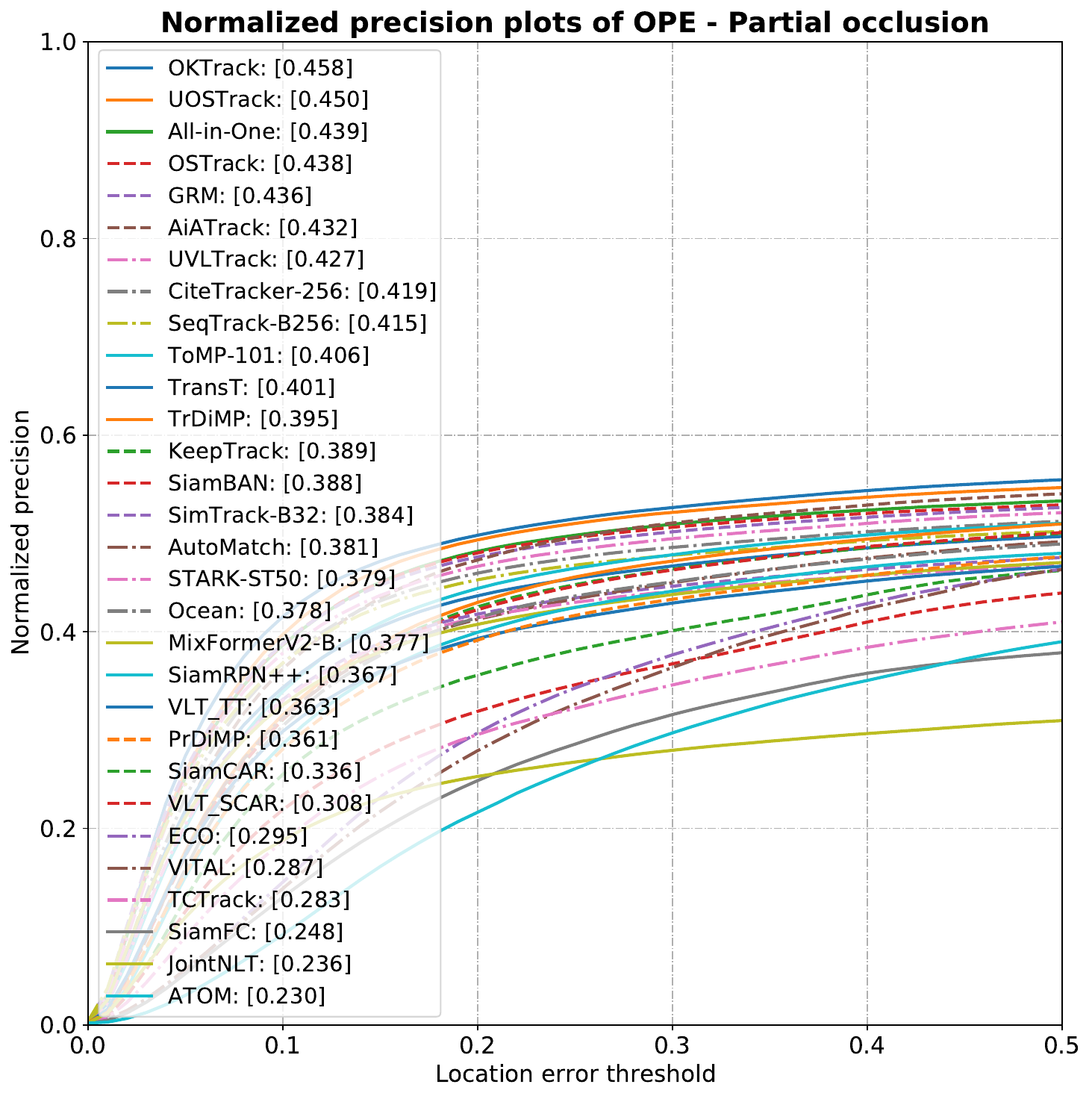}}
\subfloat{\includegraphics[width =0.25\columnwidth]{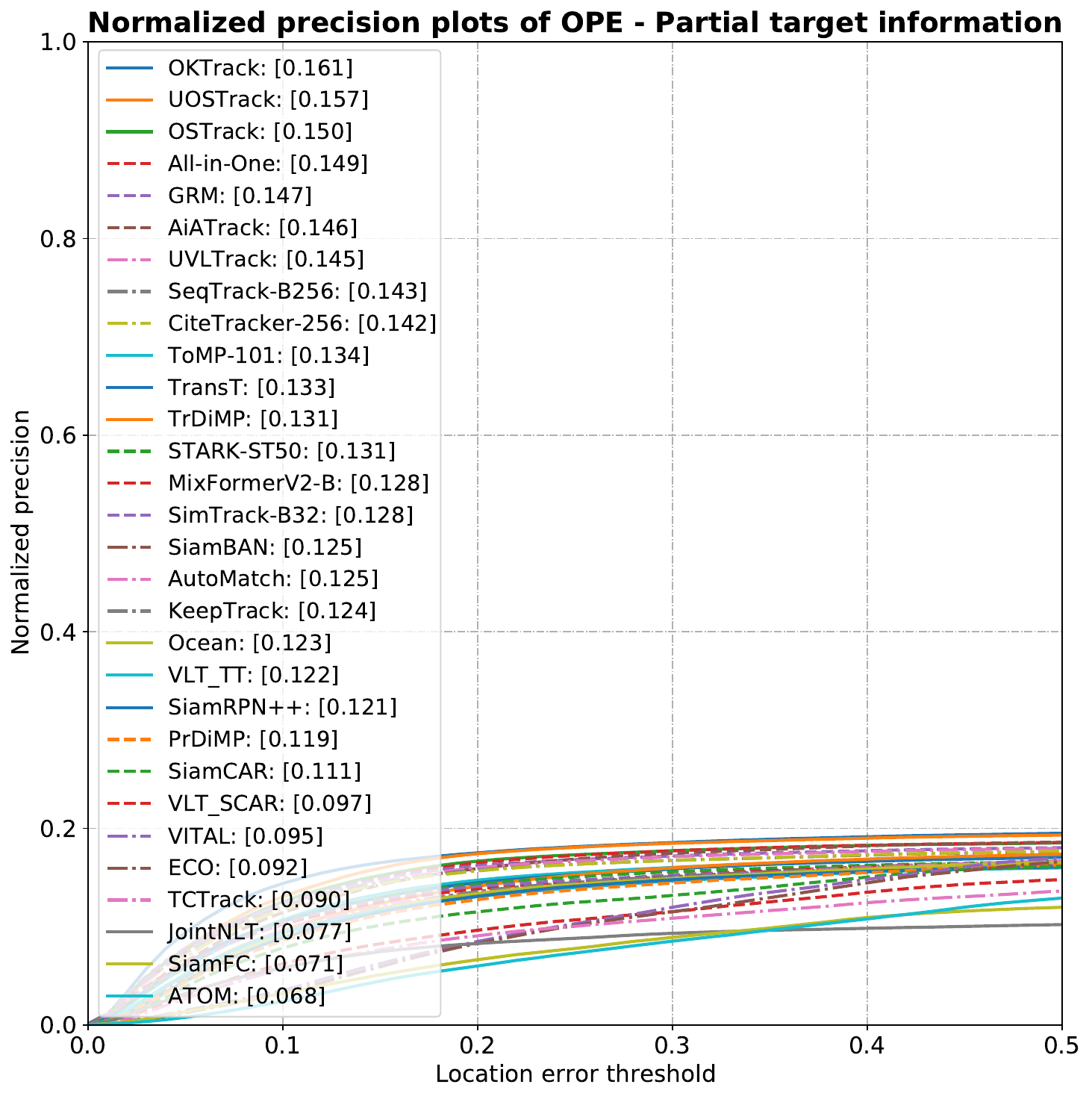}}
\subfloat{\includegraphics[width =0.25\columnwidth]{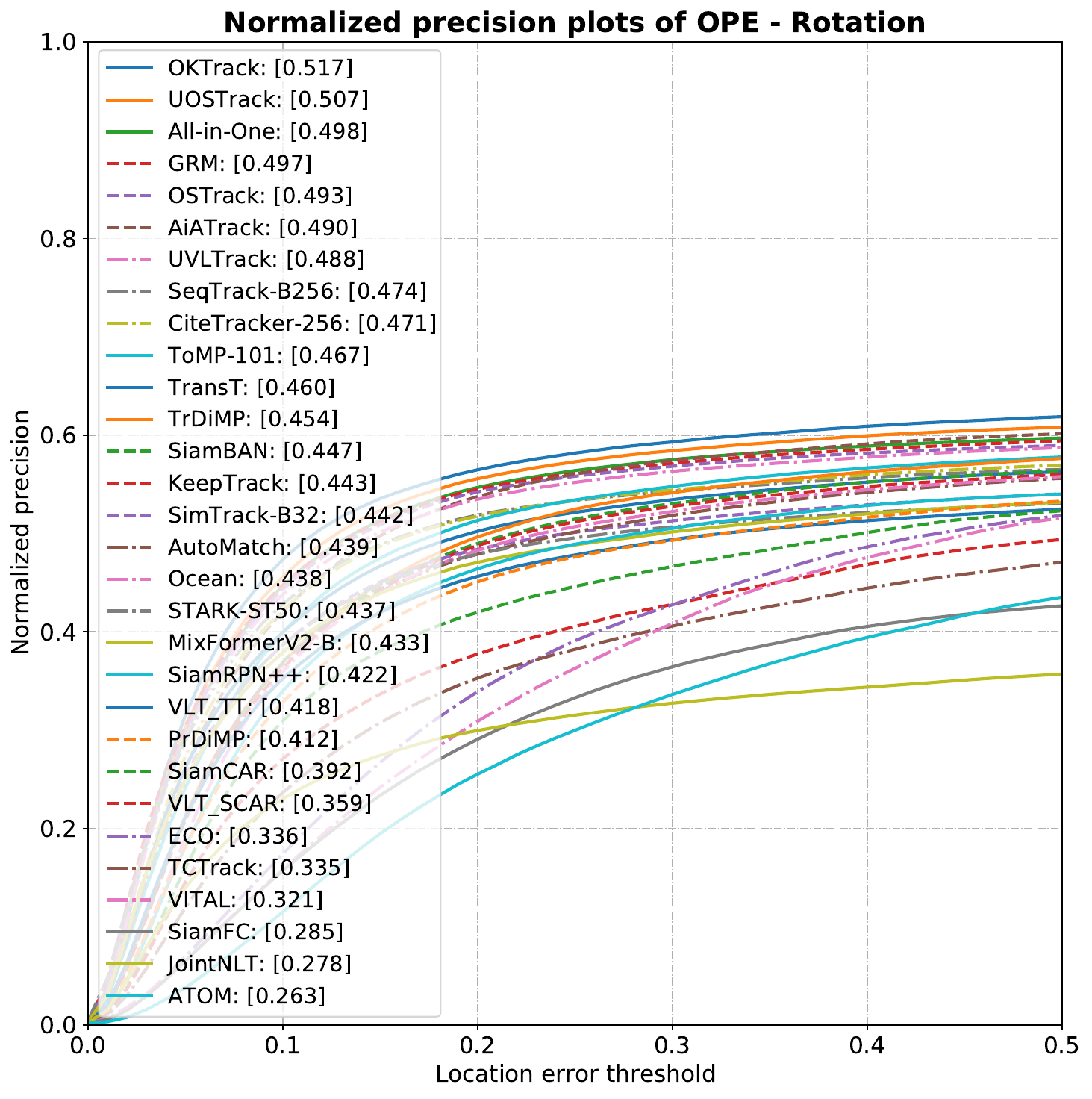}}

\subfloat{\includegraphics[width =0.25\columnwidth]{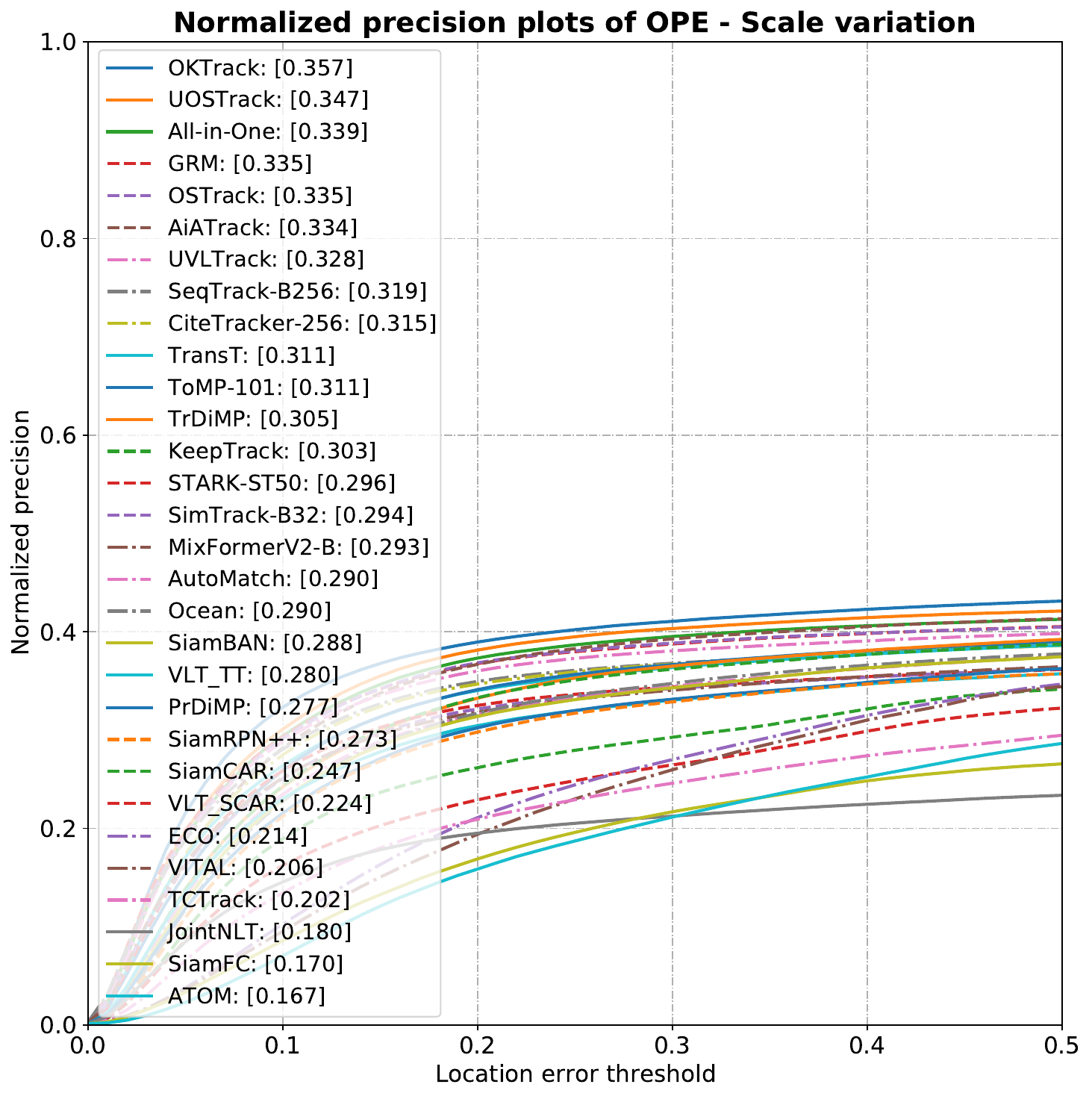}}
\subfloat{\includegraphics[width =0.25\columnwidth]{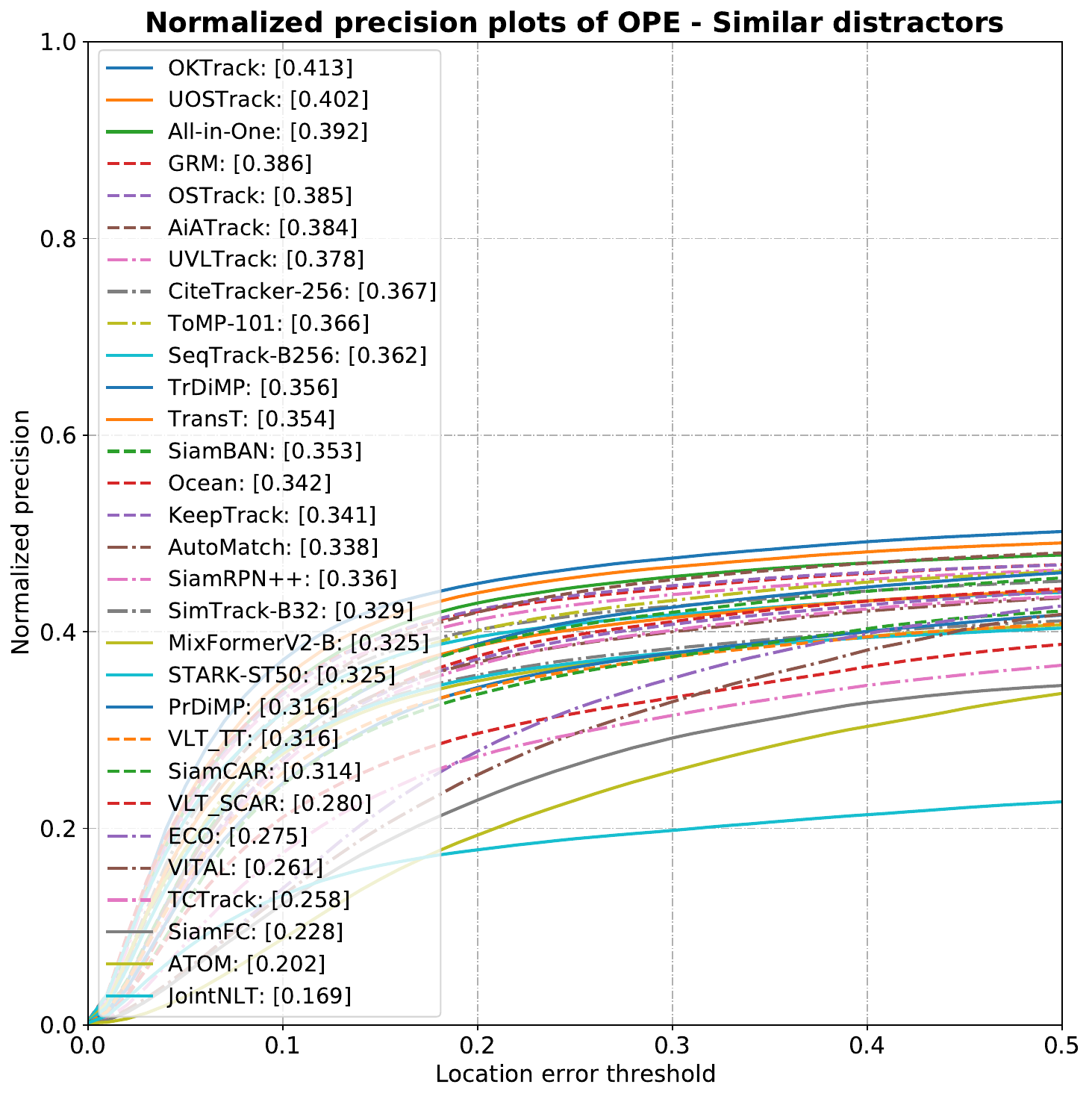}}
\subfloat{\includegraphics[width =0.25\columnwidth]{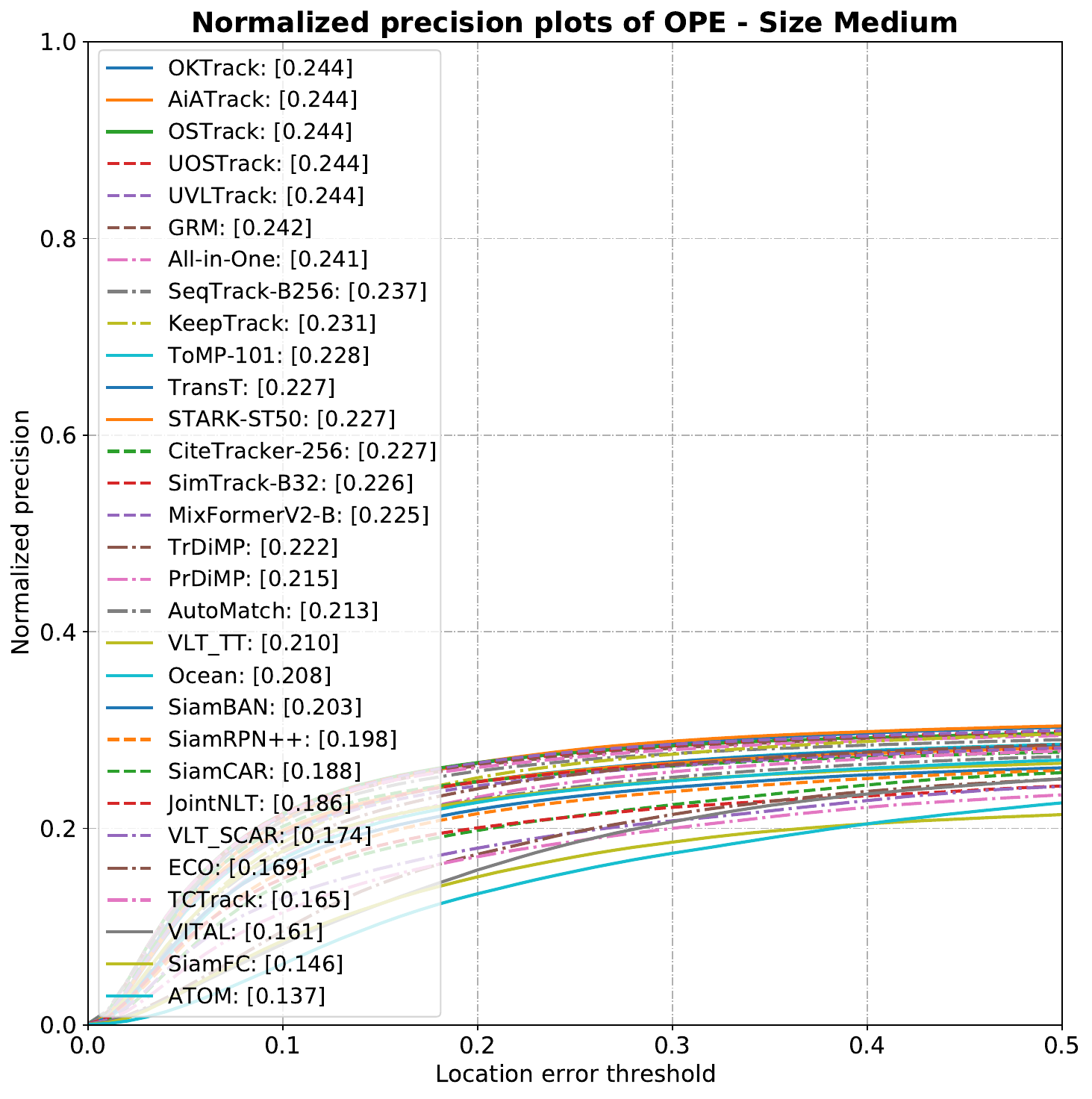}}
\subfloat{\includegraphics[width =0.25\columnwidth]{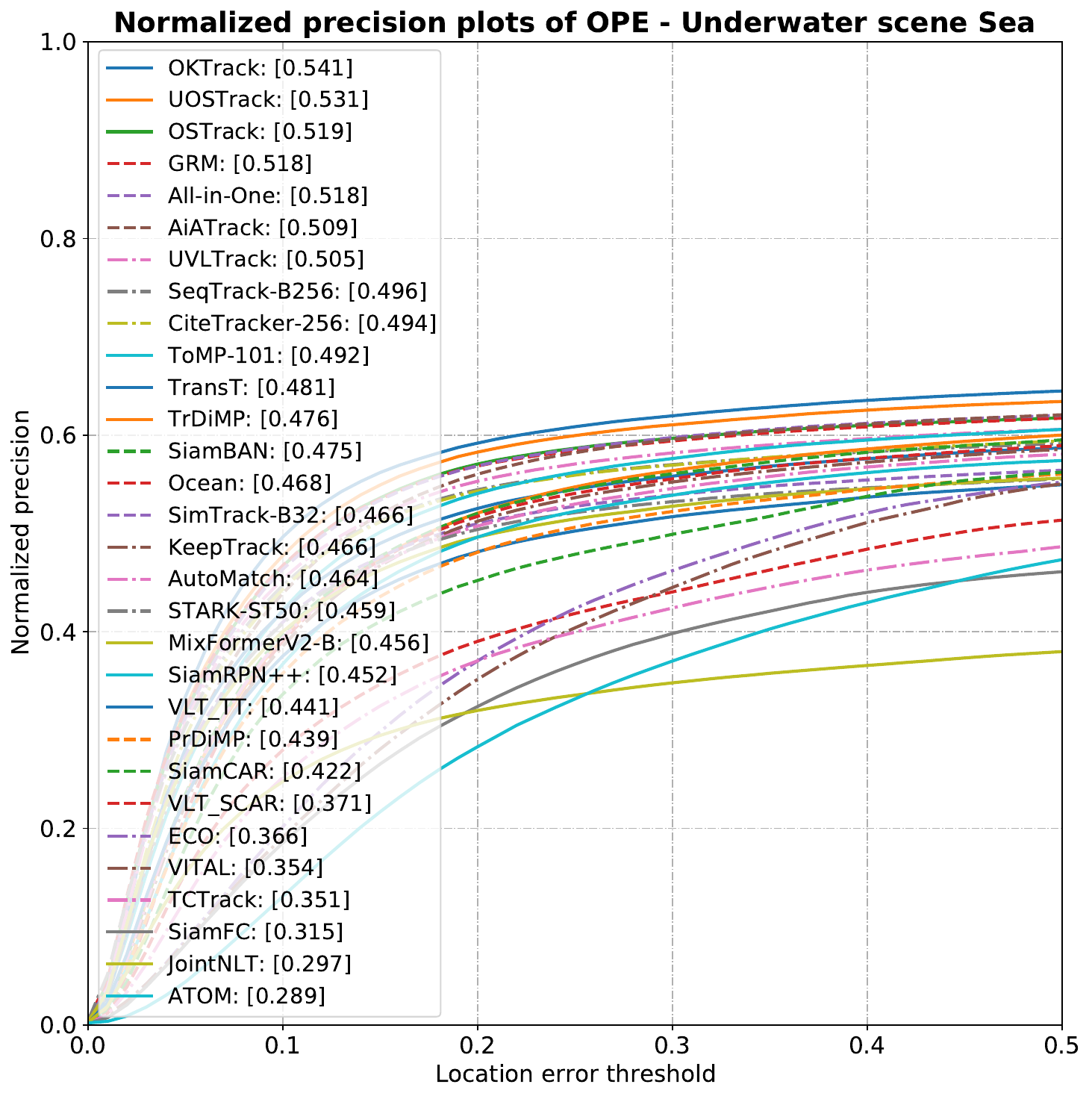}}

\subfloat{\includegraphics[width =0.25\columnwidth]{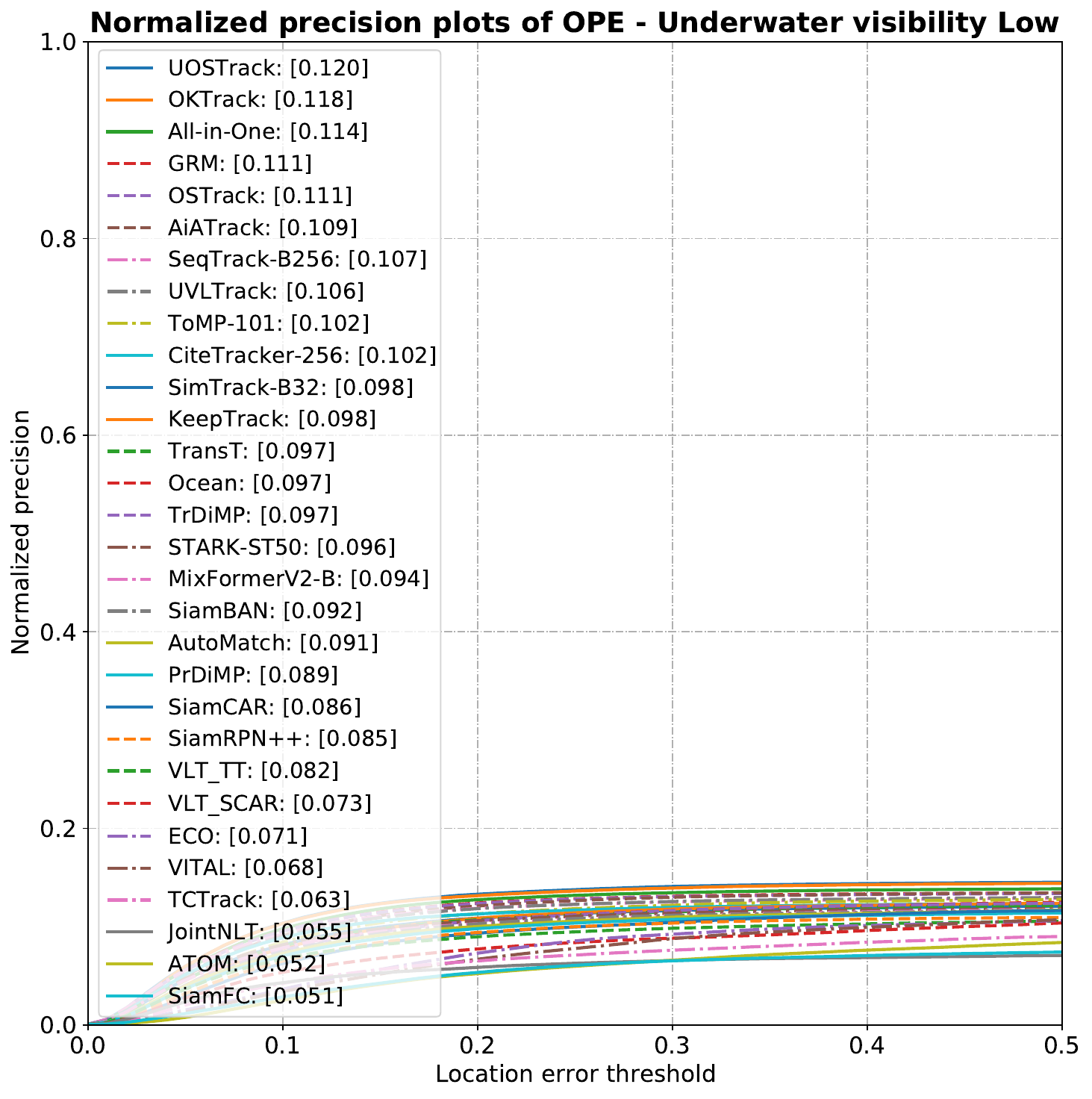}}
\subfloat{\includegraphics[width =0.25\columnwidth]{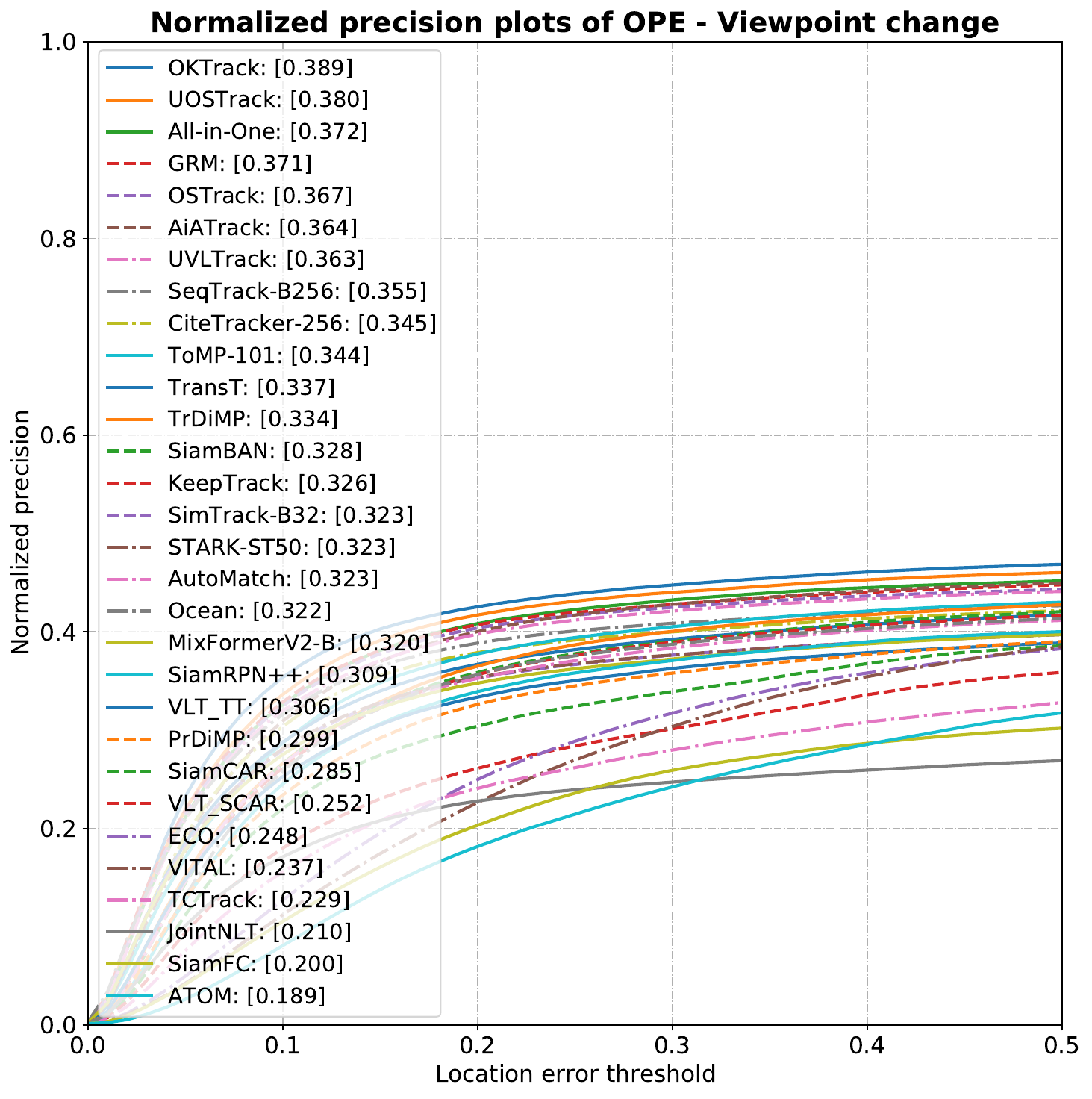}}
\subfloat{\includegraphics[width =0.25\columnwidth]{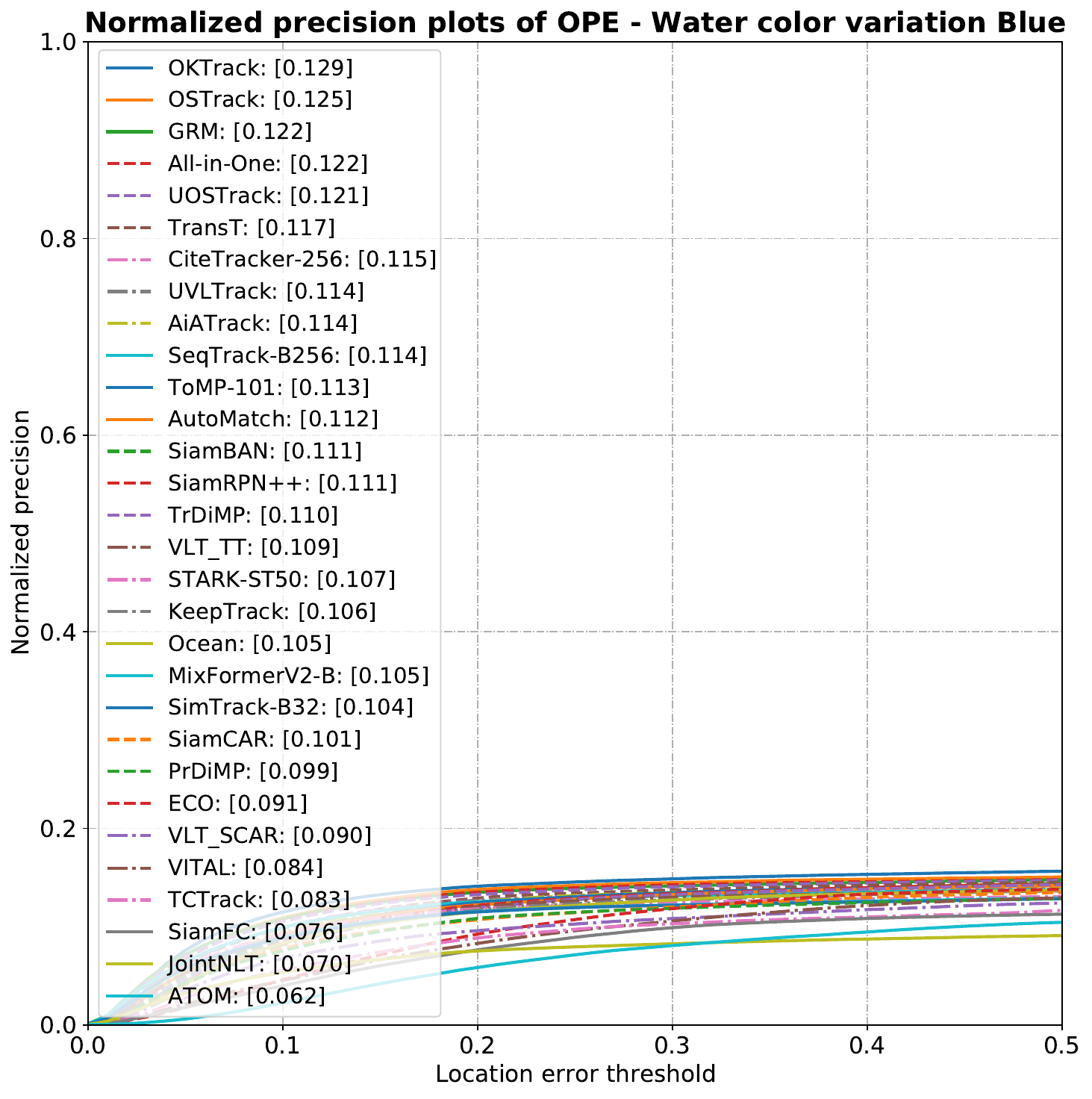}}
\subfloat{\includegraphics[width =0.25\columnwidth]{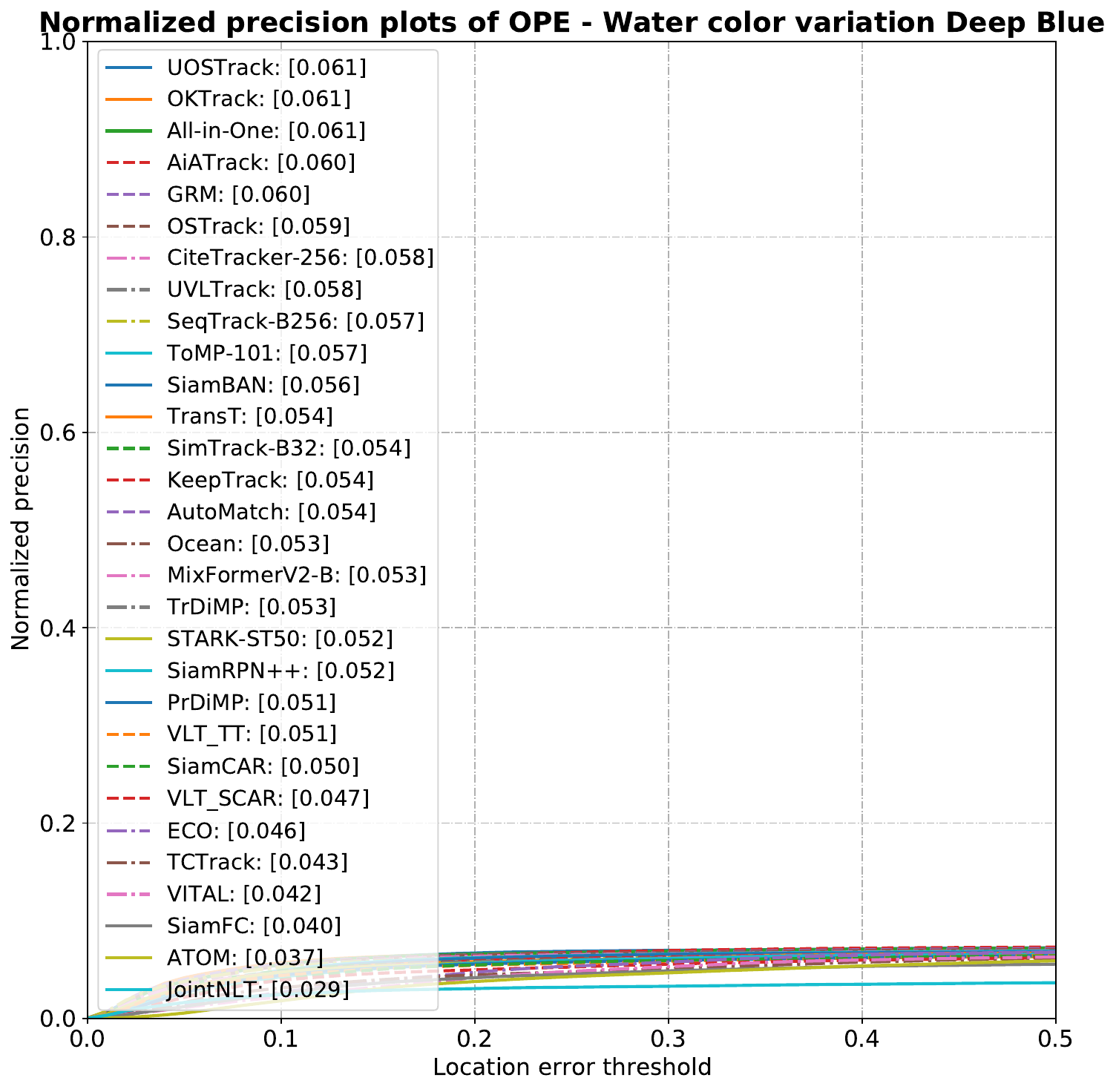}}

  \caption{Performances of baseline trackers on the WebOUT-1M test set of different attributes using \textbf{nPre} scores. Best viewed by zooming in.}
  \label{fig:attribute_results_nPre}
\end{figure*}

\begin{figure*}[t]
\vspace{-0.6cm}
  \centering

\subfloat{\includegraphics[width =0.25\columnwidth]{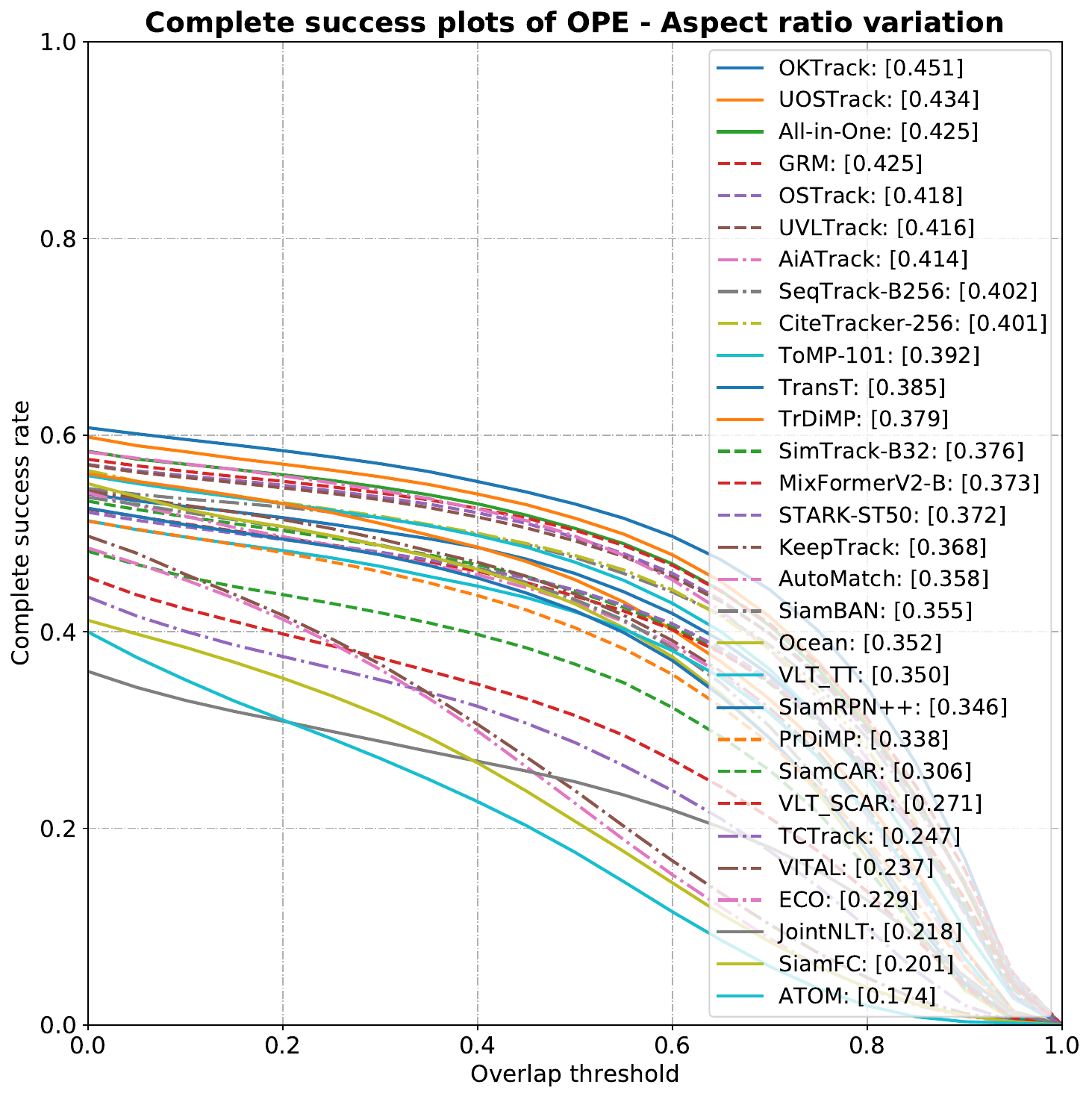}}
\subfloat{\includegraphics[width =0.25\columnwidth]{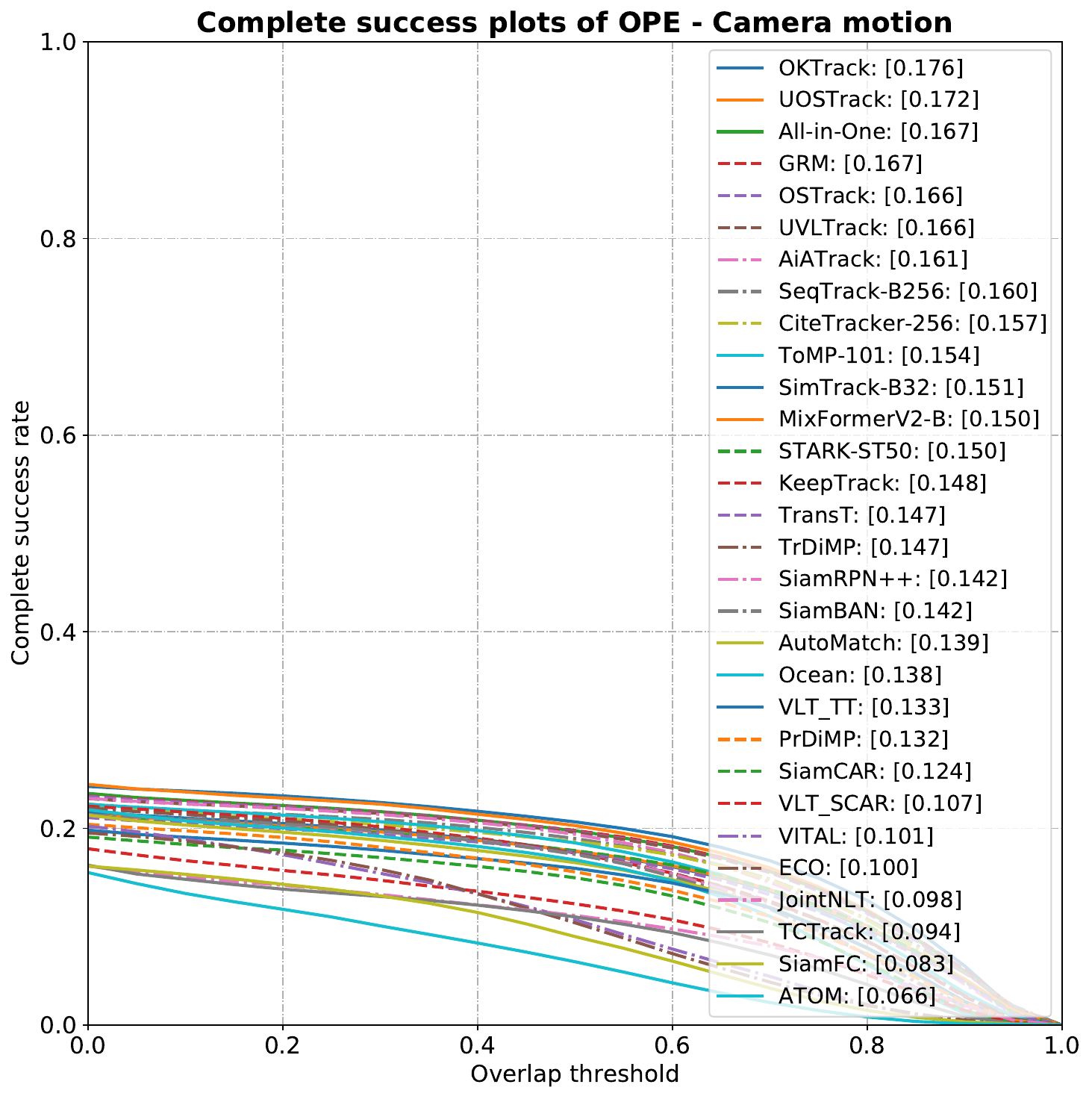}}
\subfloat{\includegraphics[width =0.25\columnwidth]{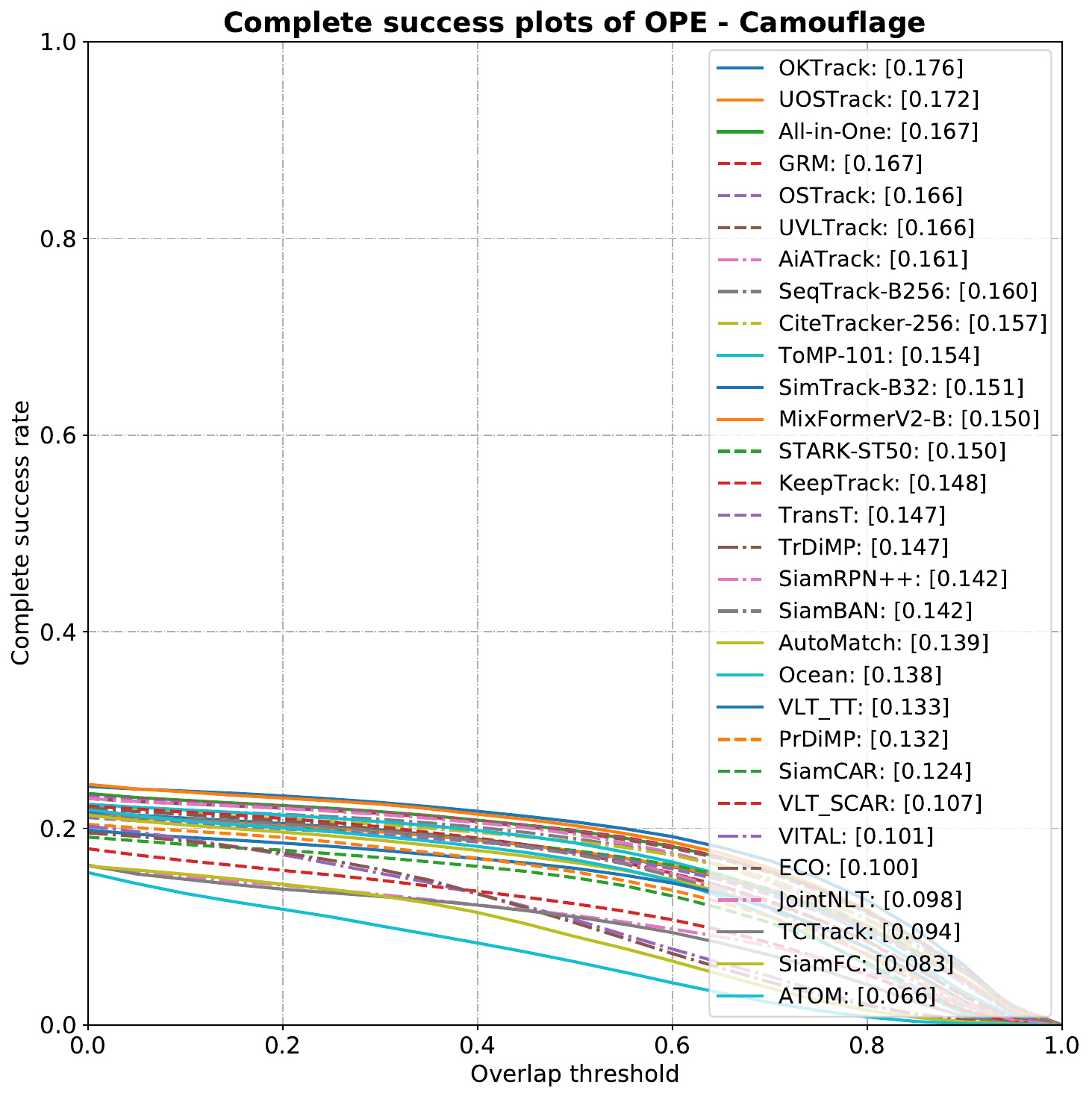}}
\subfloat{\includegraphics[width =0.25\columnwidth]{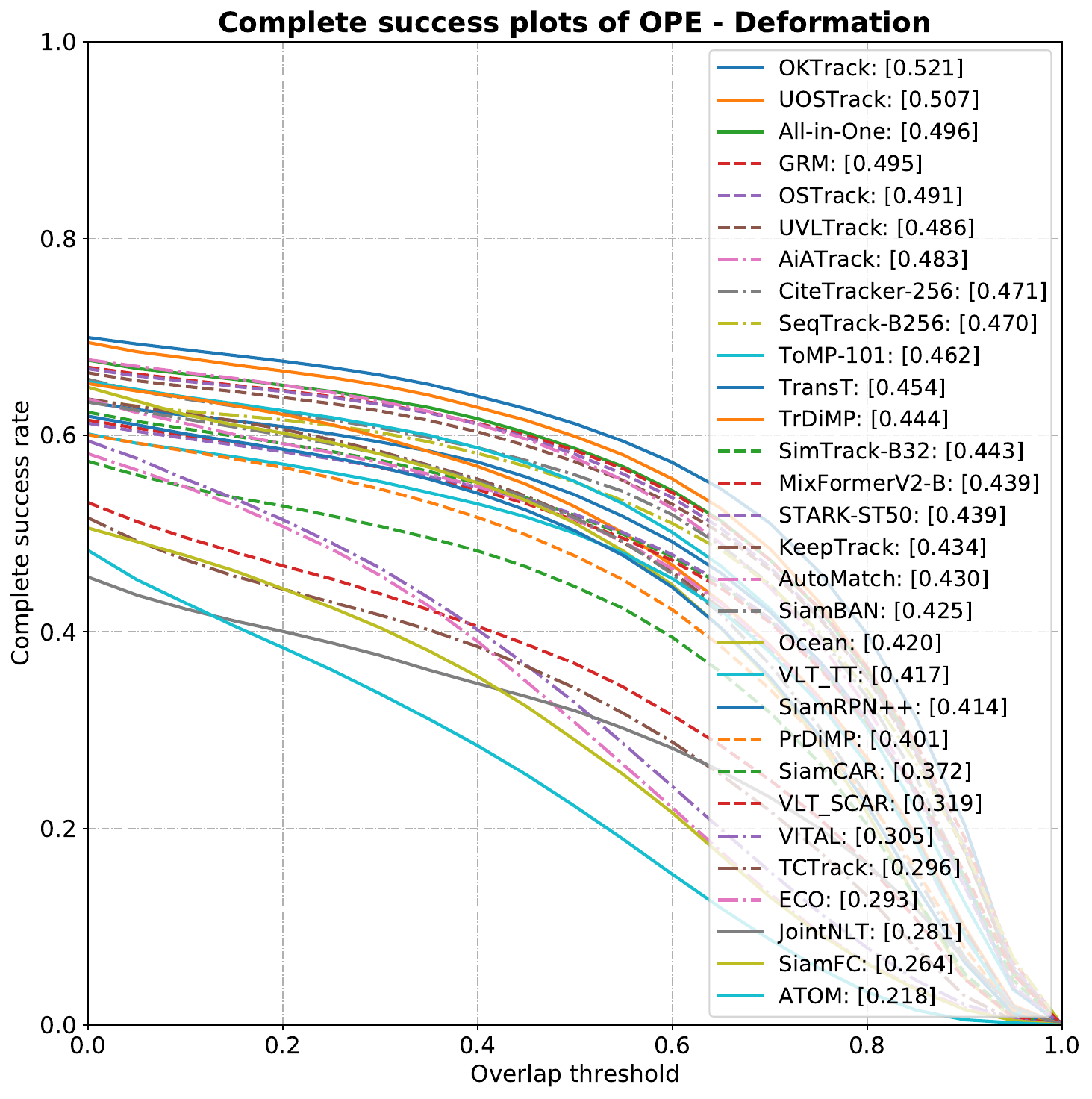}}

\subfloat{\includegraphics[width =0.25\columnwidth]{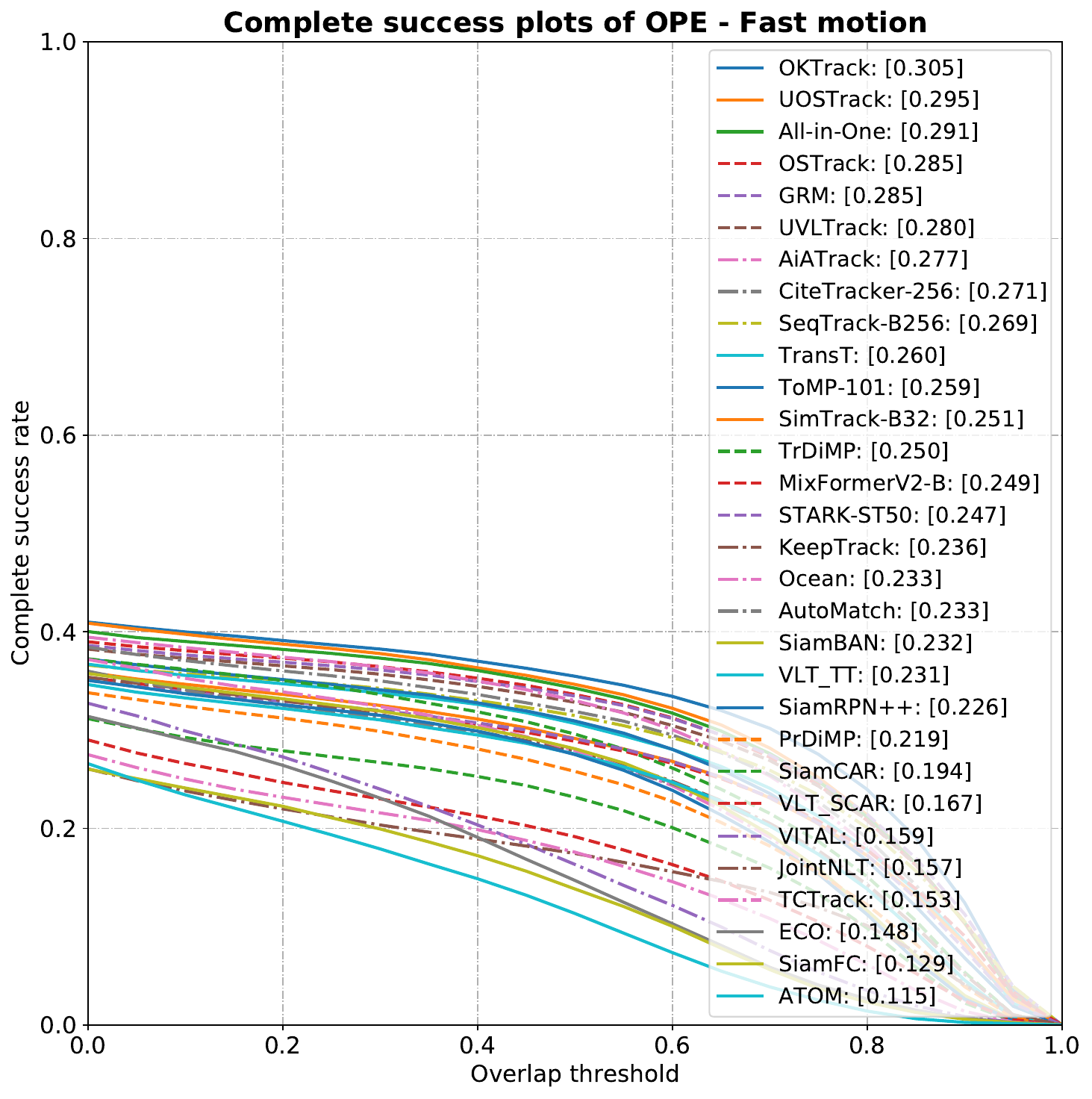}}
\subfloat{\includegraphics[width =0.25\columnwidth]{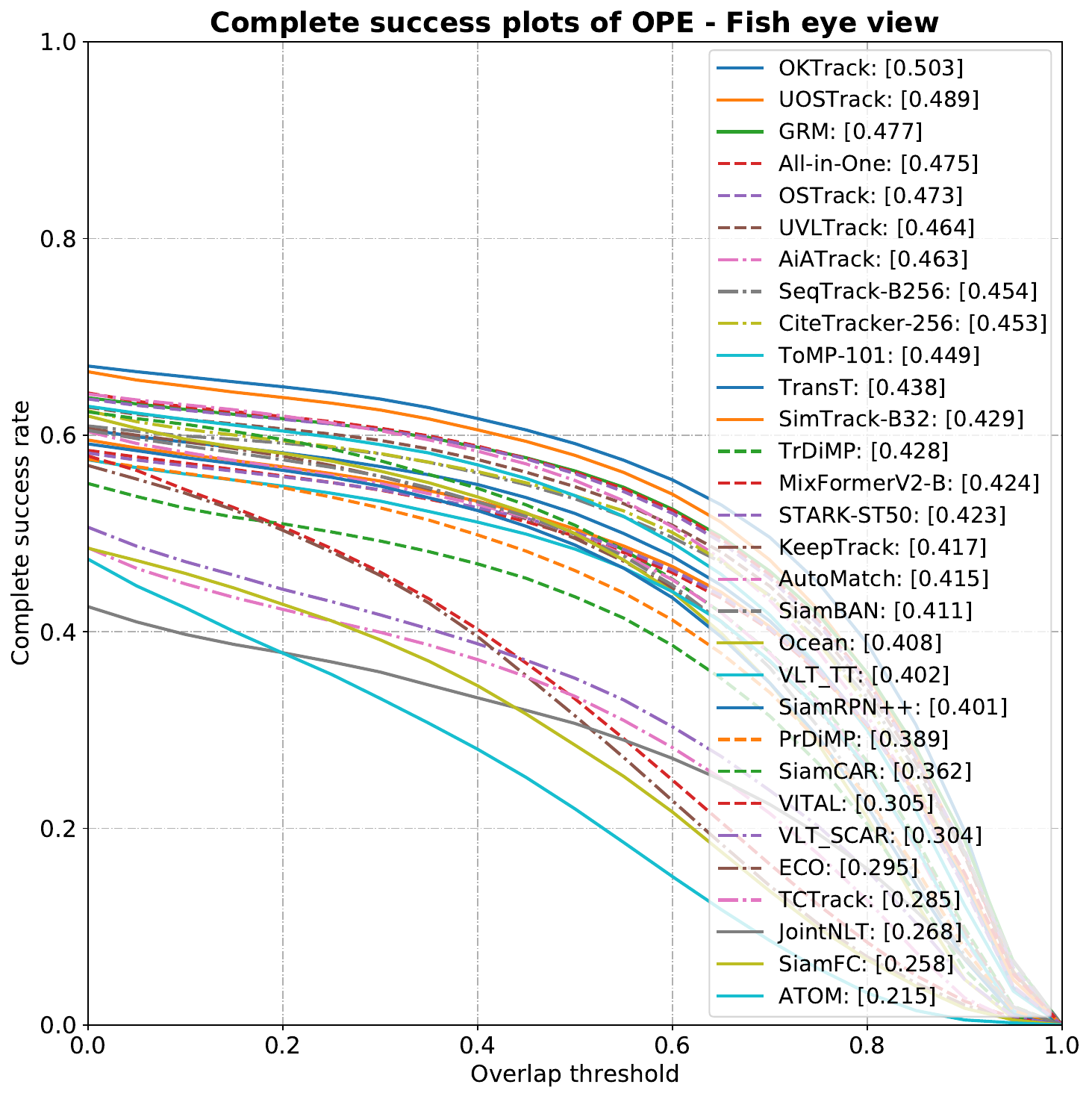}}
\subfloat{\includegraphics[width =0.25\columnwidth]{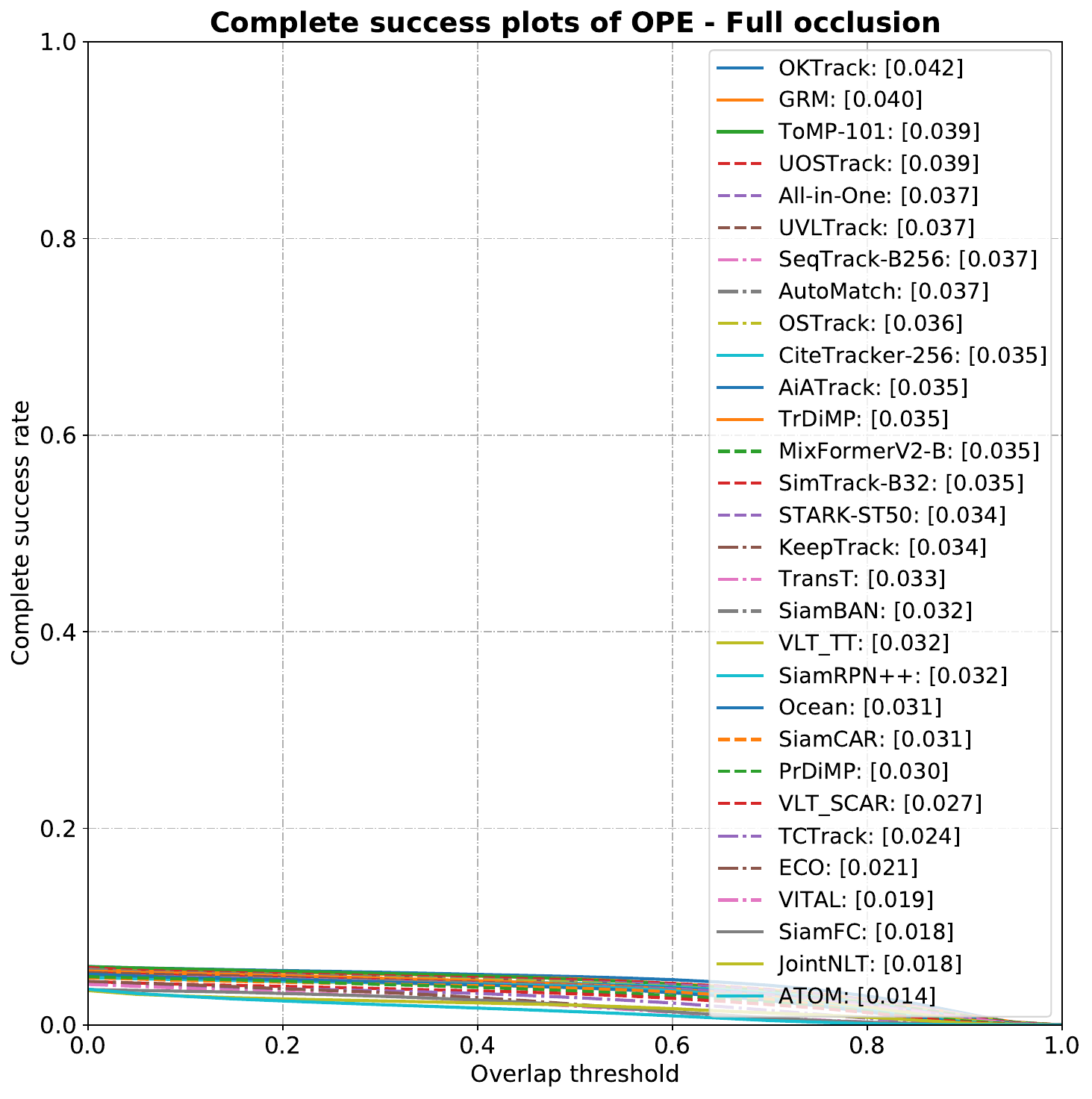}}
\subfloat{\includegraphics[width =0.25\columnwidth]{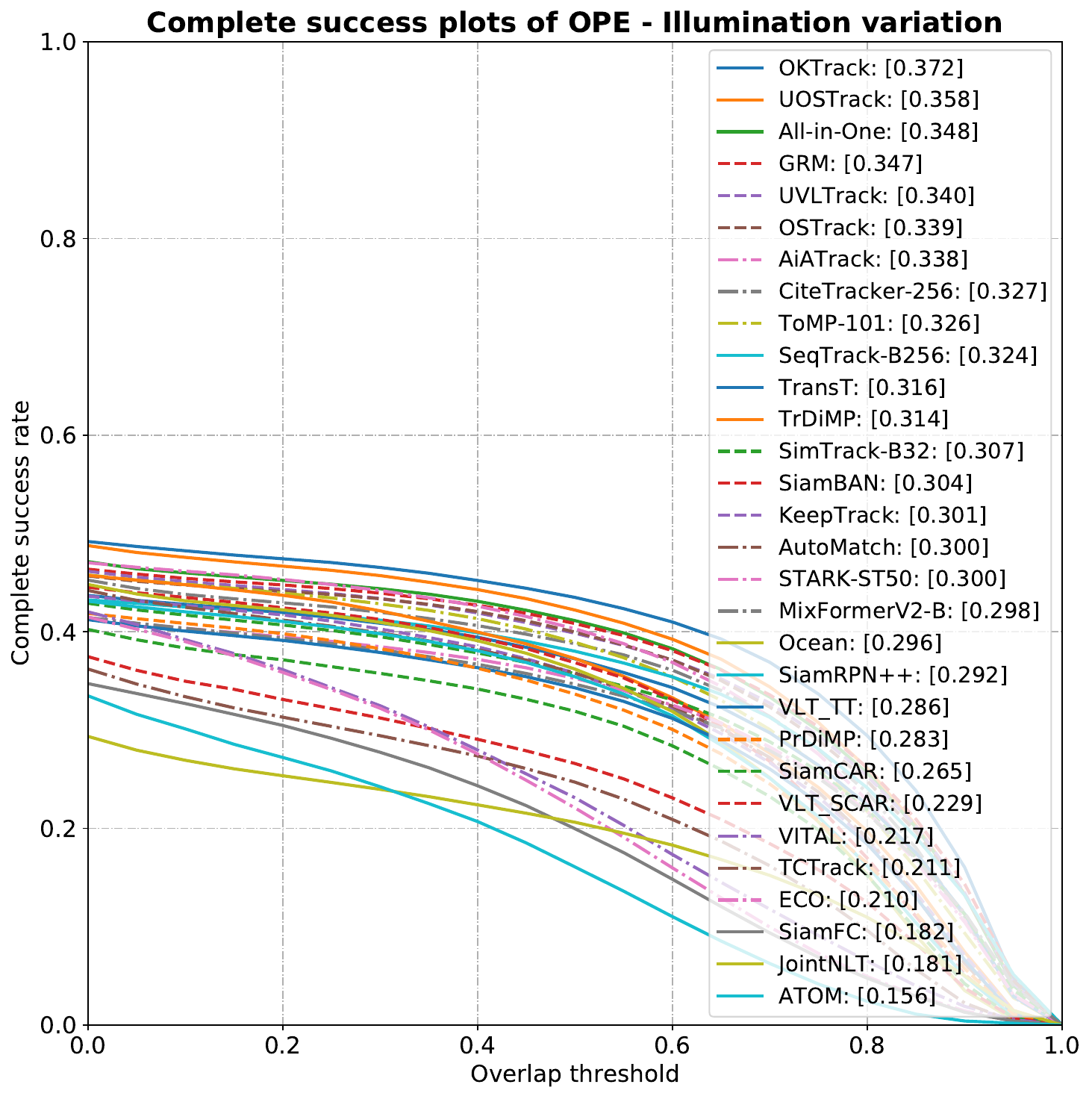}}

\subfloat{\includegraphics[width =0.25\columnwidth]{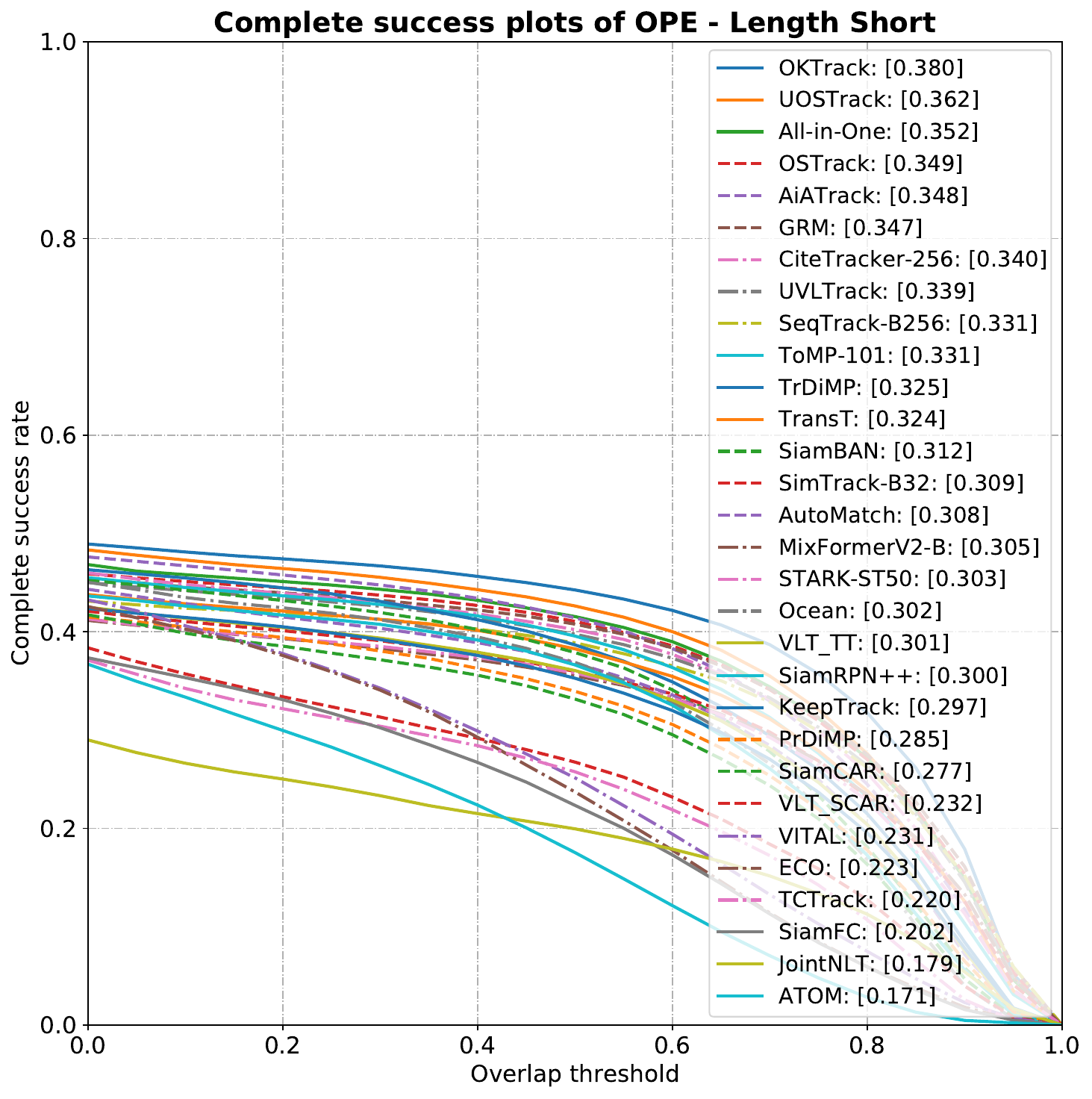}}
\subfloat{\includegraphics[width =0.25\columnwidth]{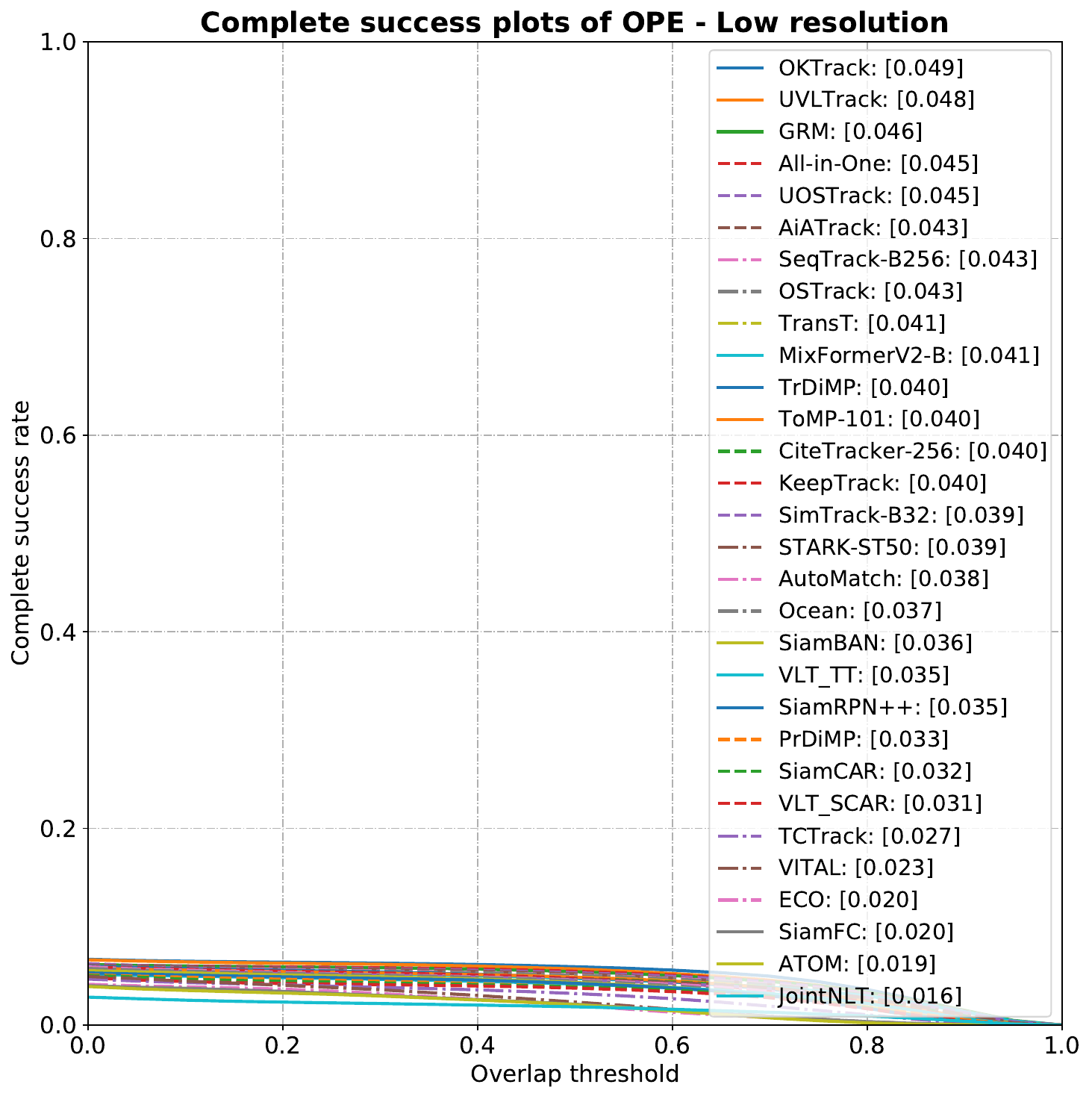}}
\subfloat{\includegraphics[width =0.25\columnwidth]{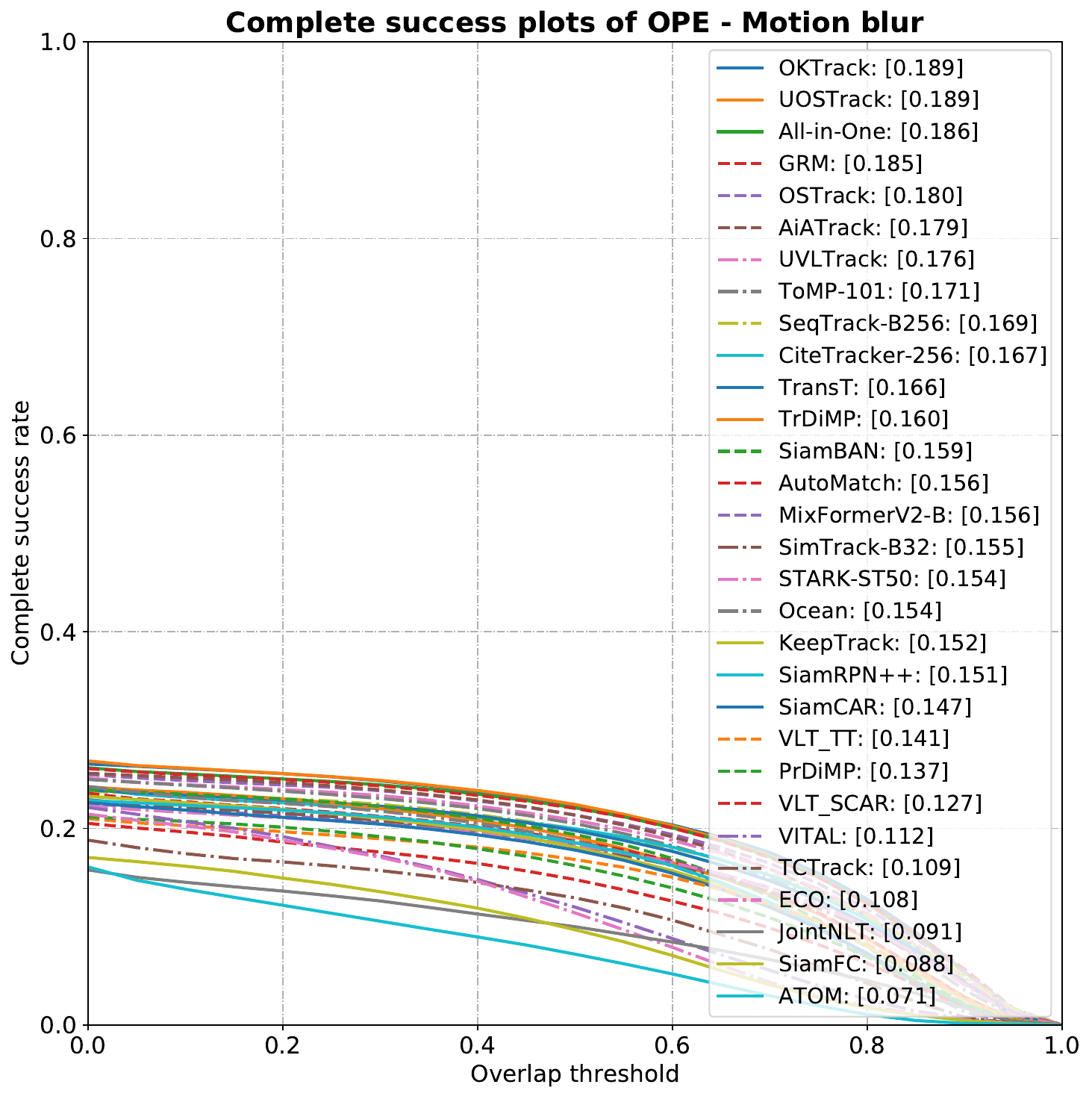}}
\subfloat{\includegraphics[width =0.25\columnwidth]{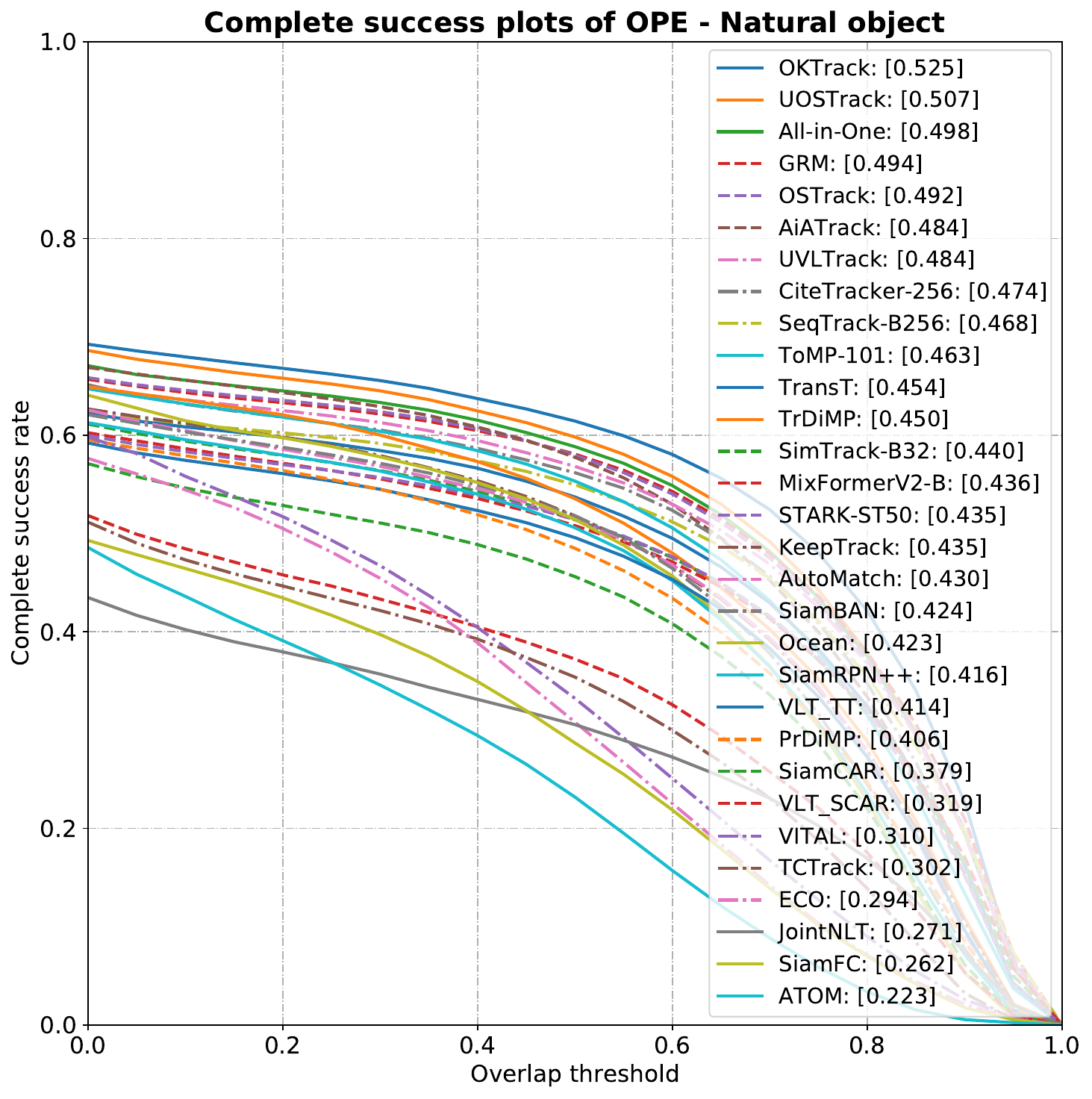}}

\subfloat{\includegraphics[width =0.25\columnwidth]{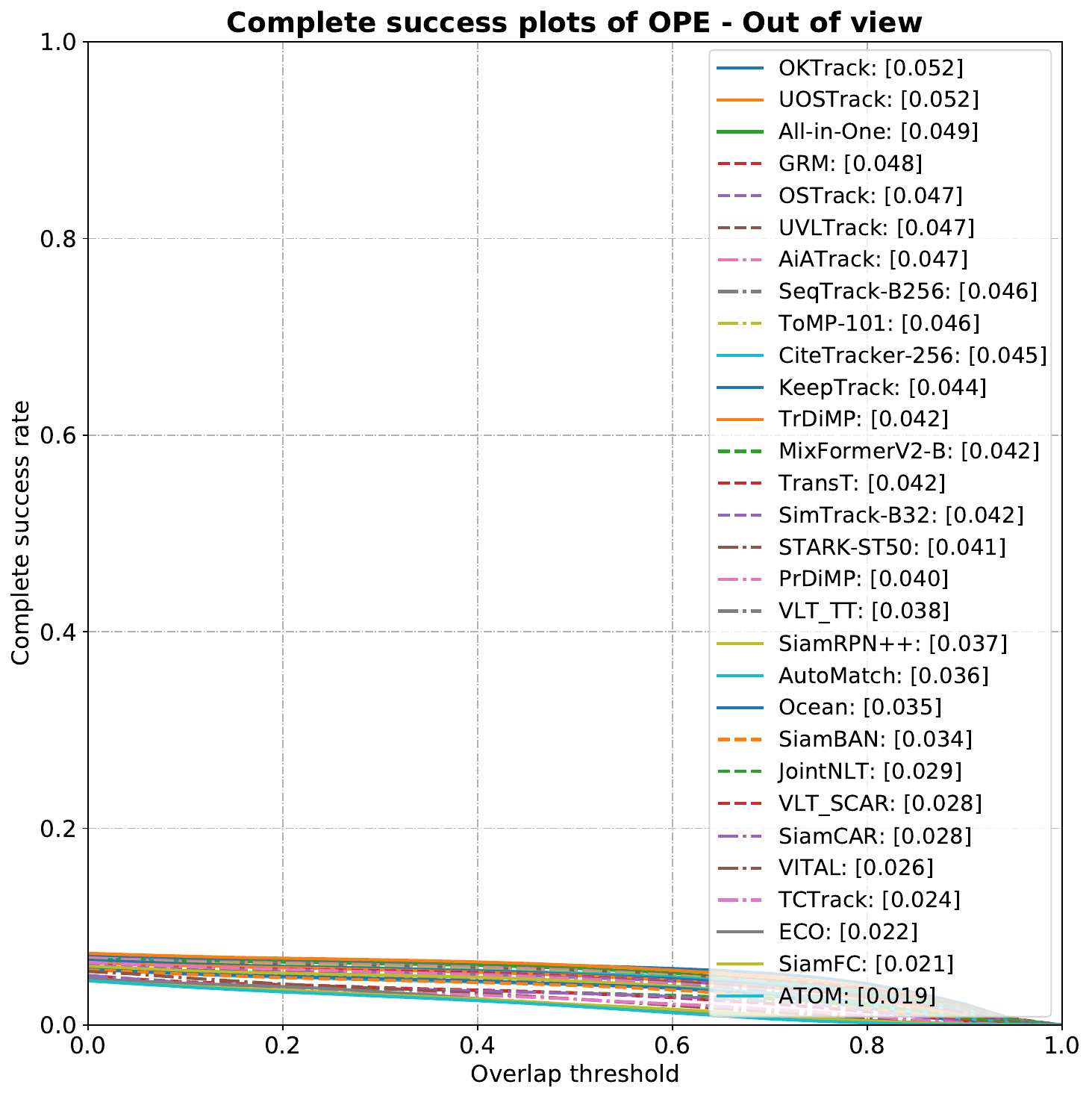}}
\subfloat{\includegraphics[width =0.25\columnwidth]{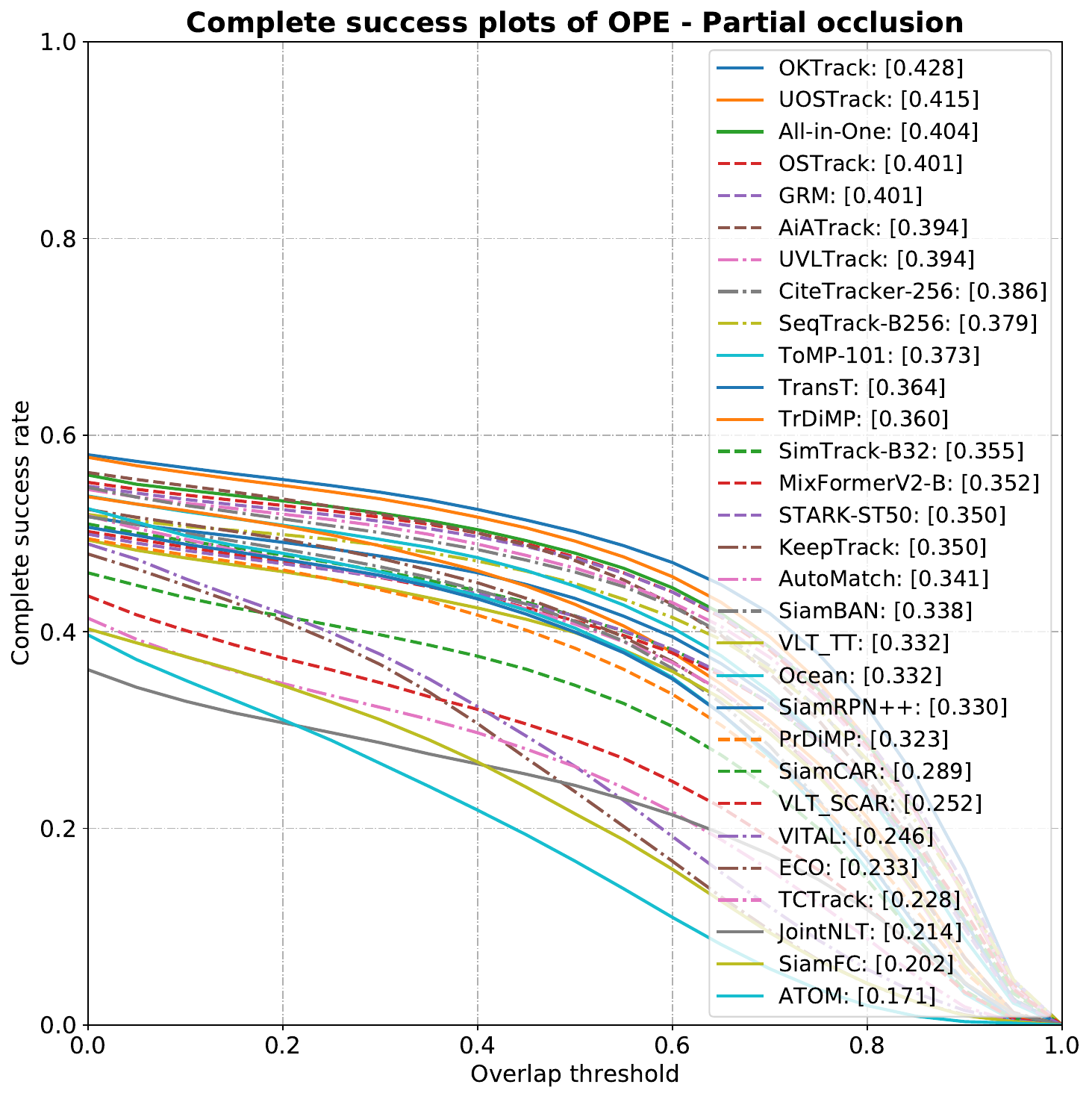}}
\subfloat{\includegraphics[width =0.25\columnwidth]{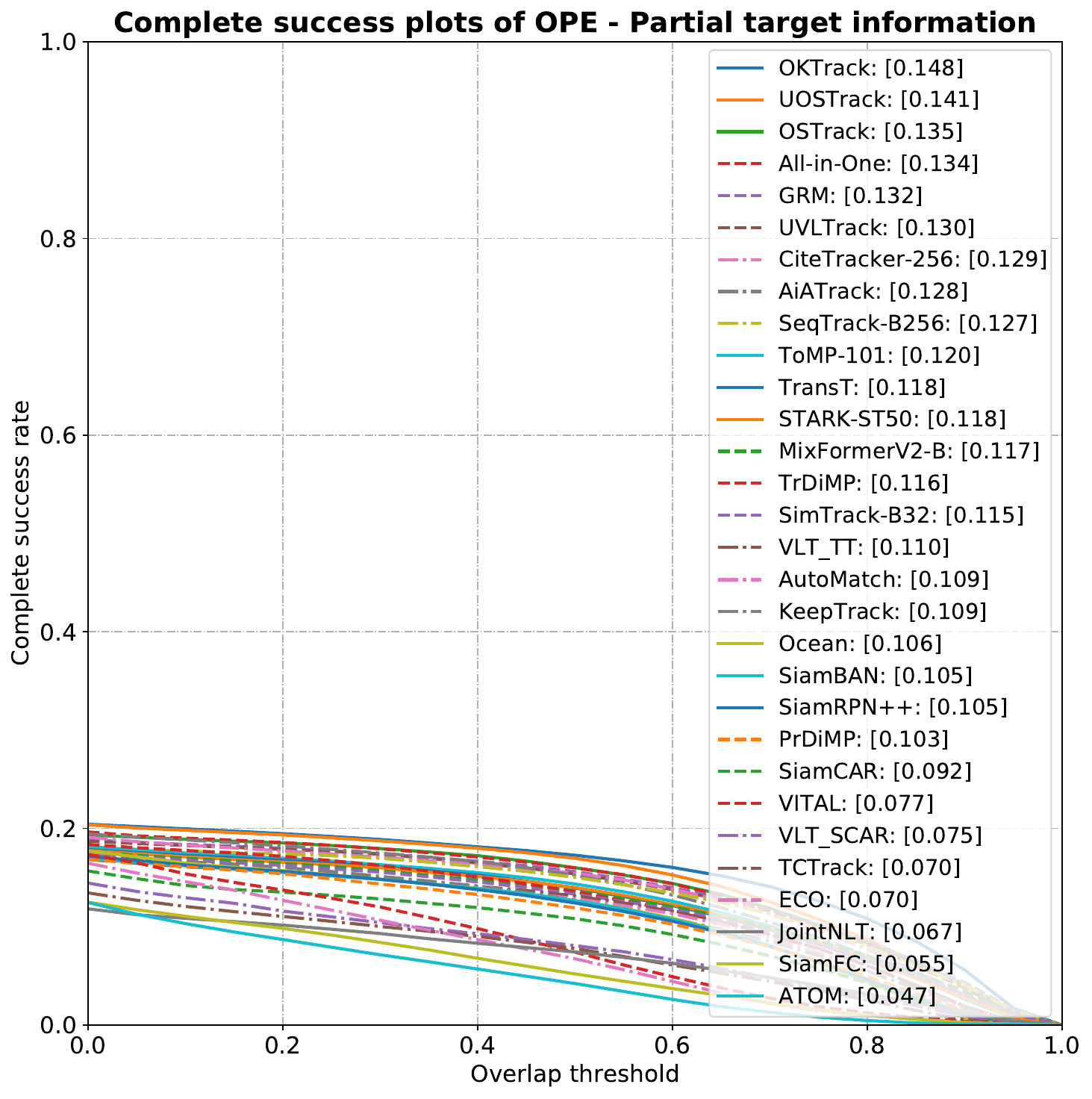}}
\subfloat{\includegraphics[width =0.25\columnwidth]{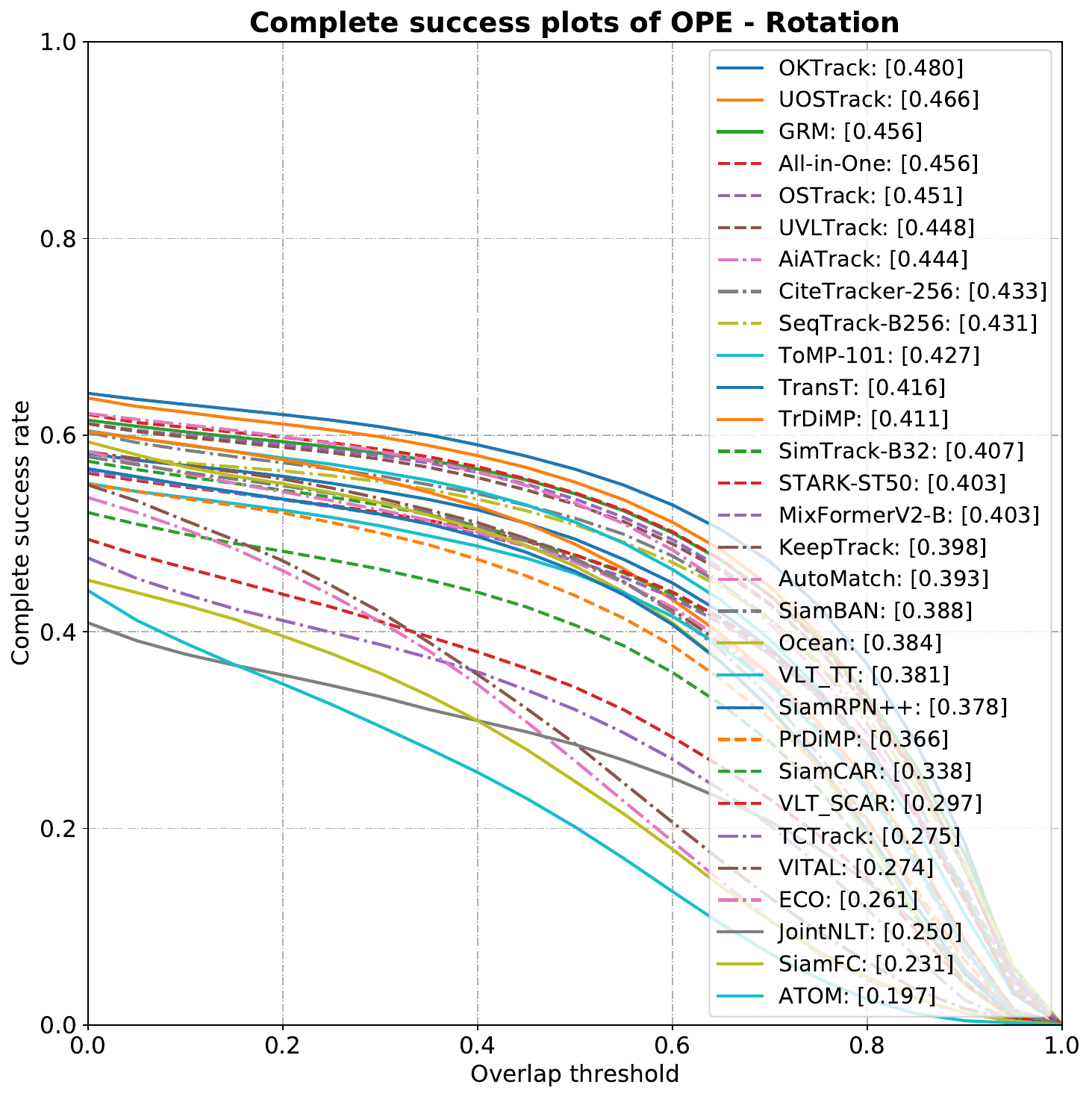}}

\subfloat{\includegraphics[width =0.25\columnwidth]{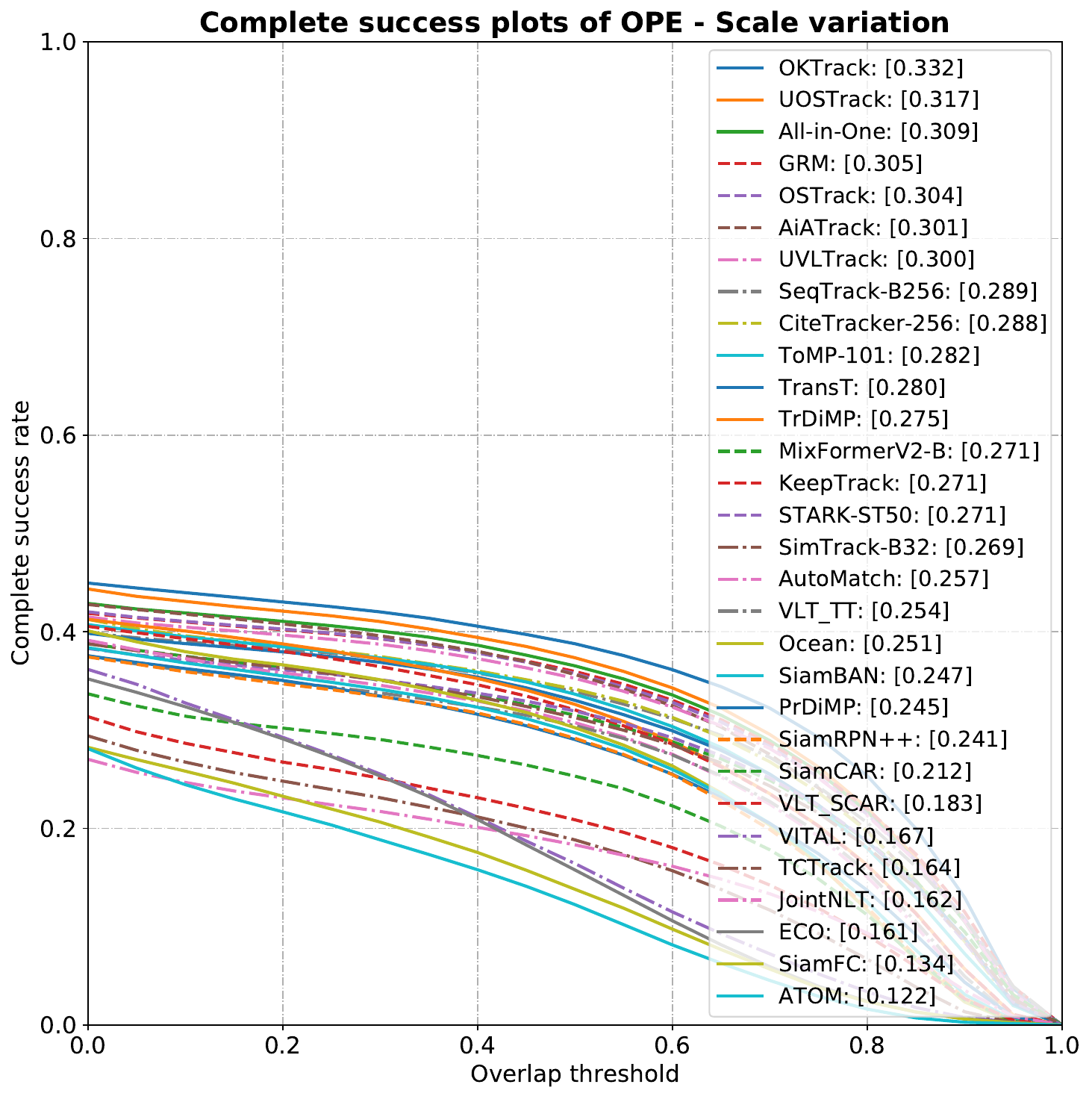}}
\subfloat{\includegraphics[width =0.25\columnwidth]{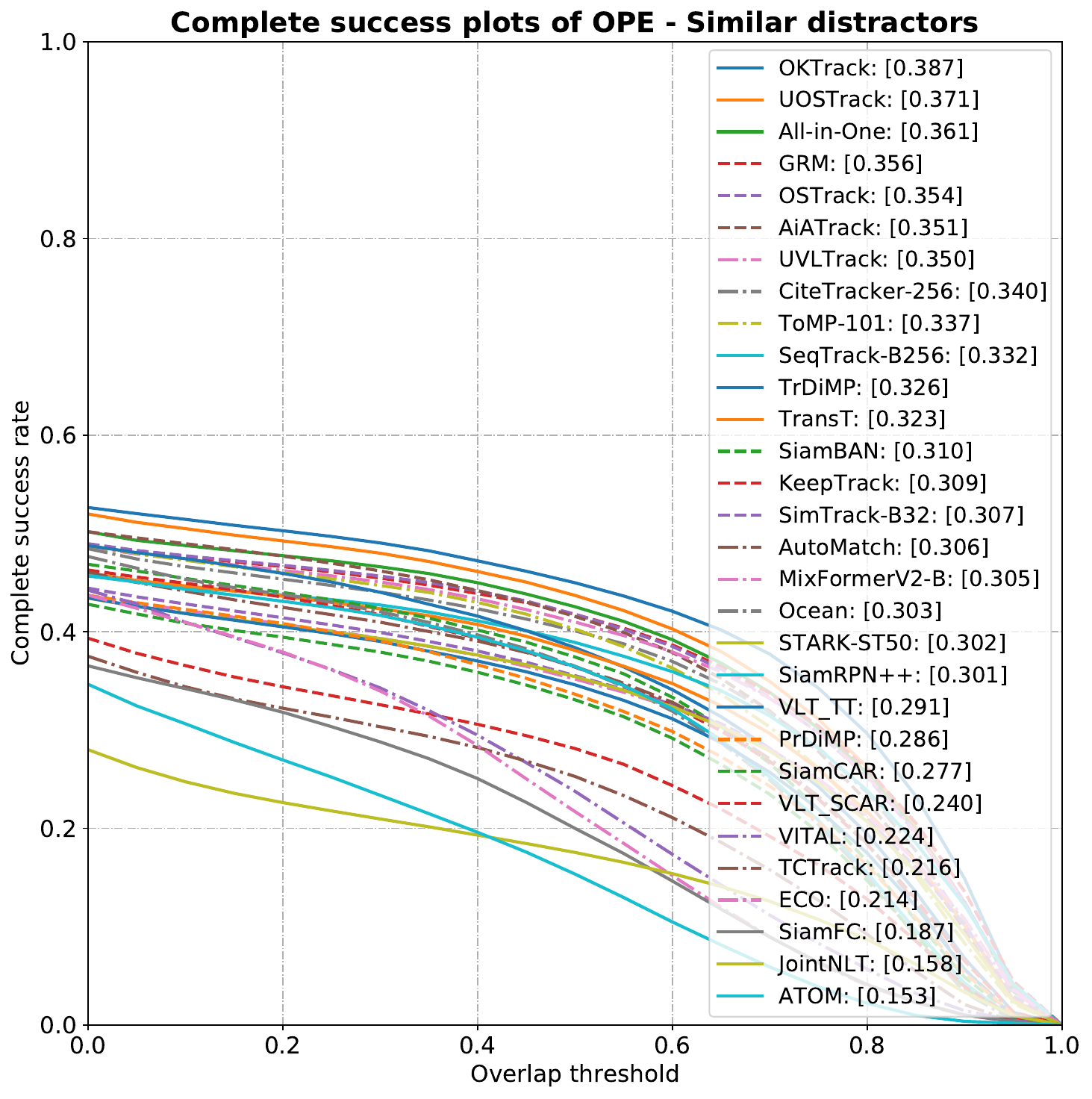}}
\subfloat{\includegraphics[width =0.25\columnwidth]{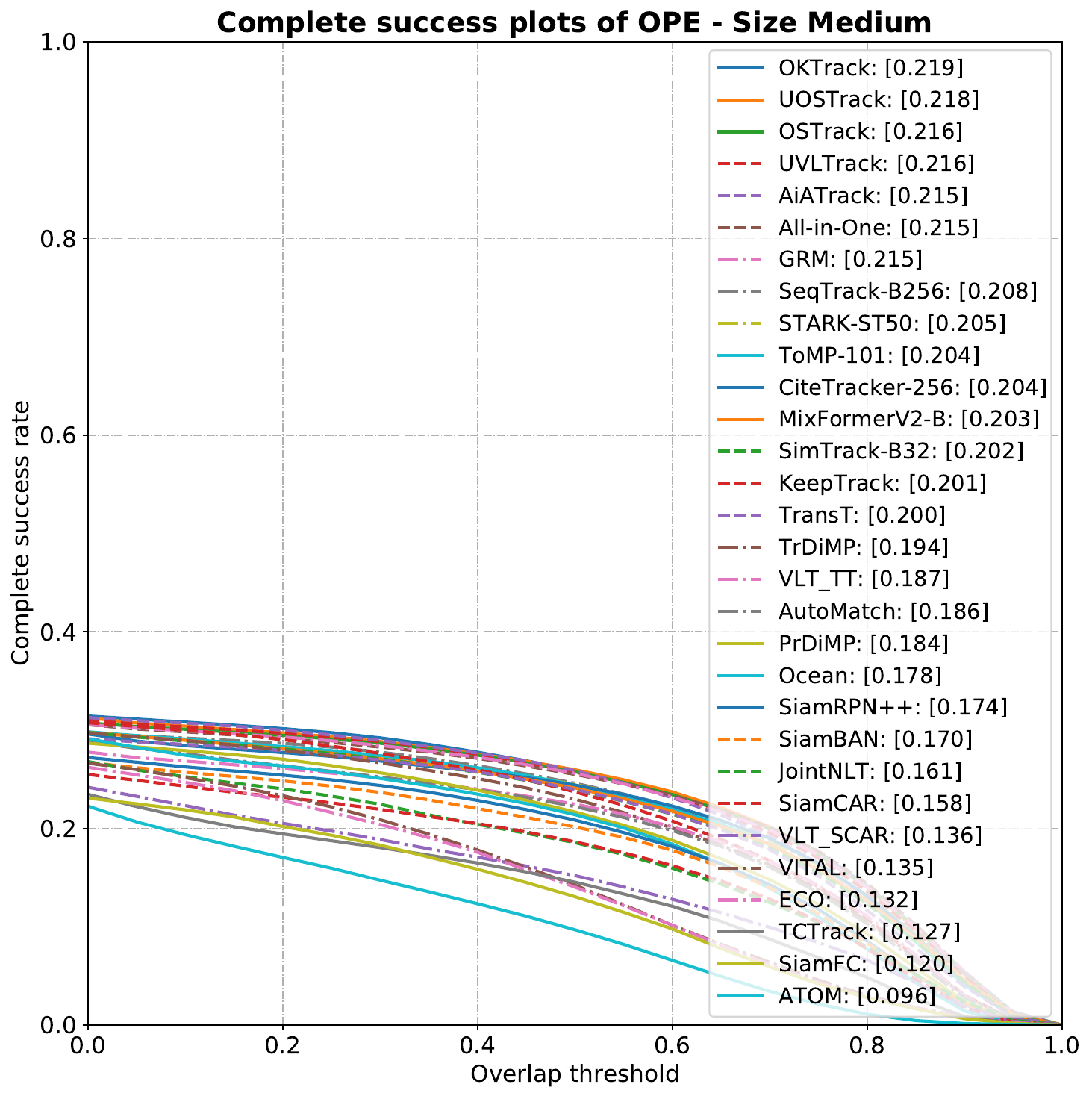}}
\subfloat{\includegraphics[width =0.25\columnwidth]{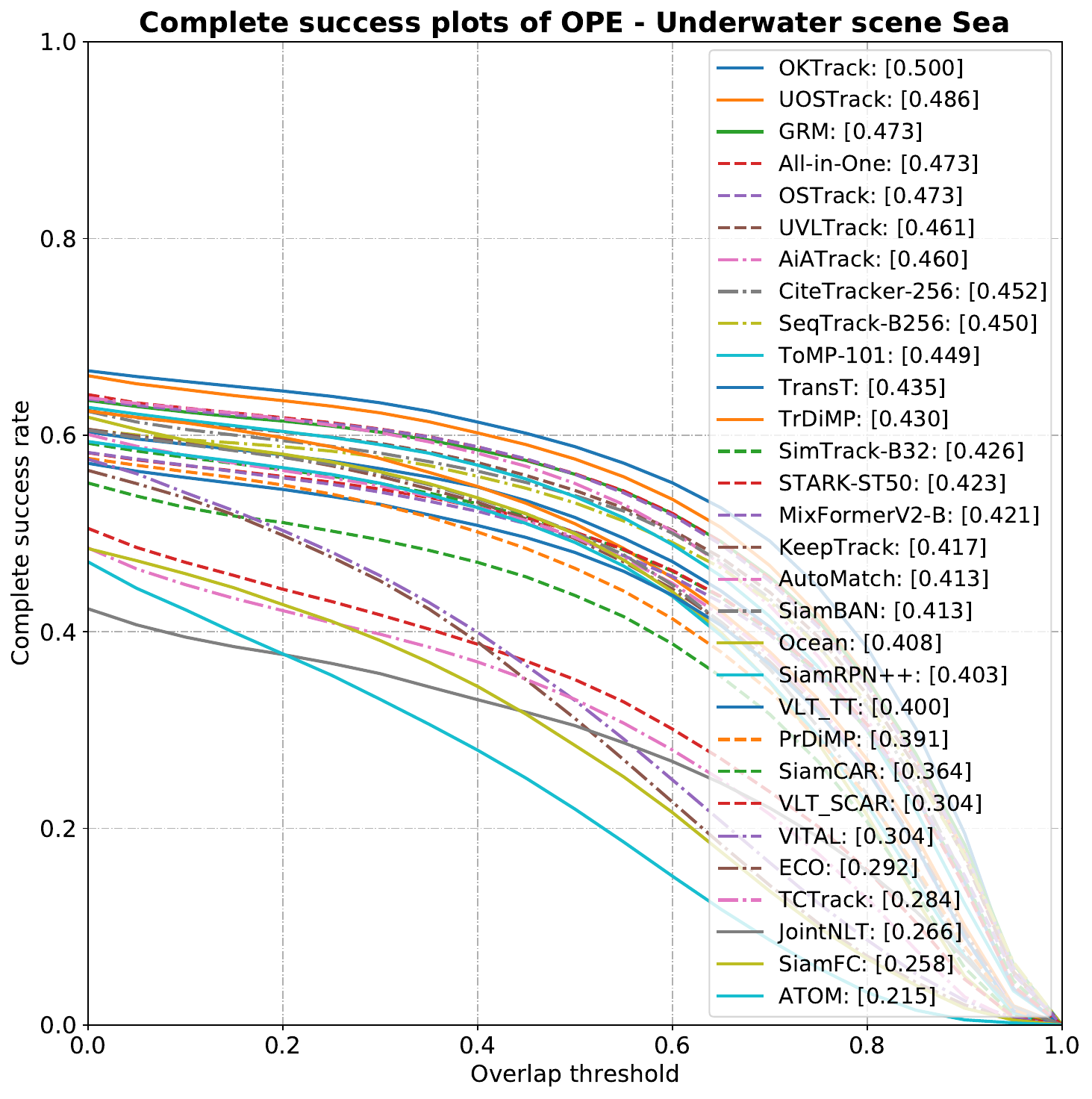}}

\subfloat{\includegraphics[width =0.25\columnwidth]{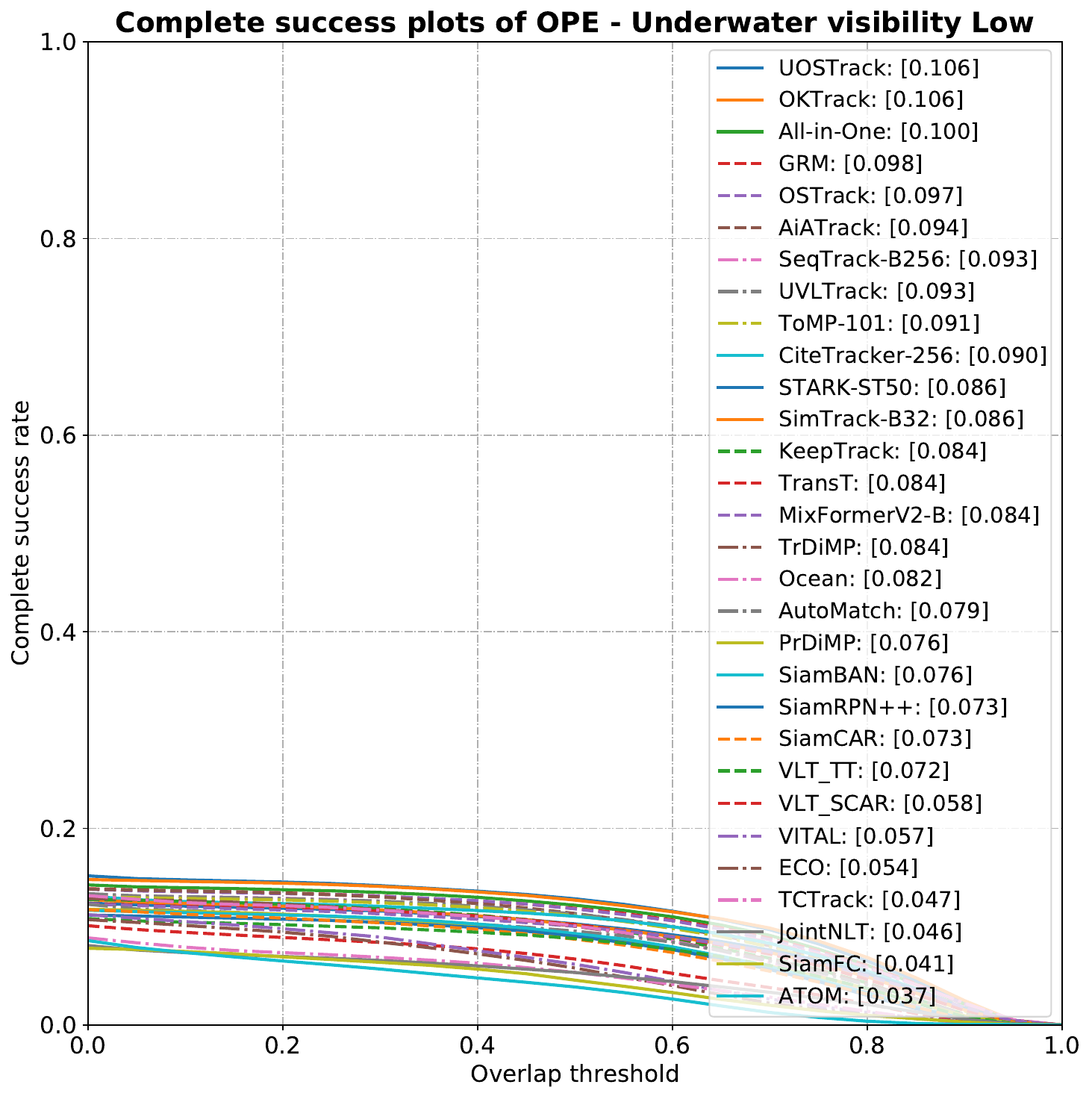}}
\subfloat{\includegraphics[width =0.25\columnwidth]{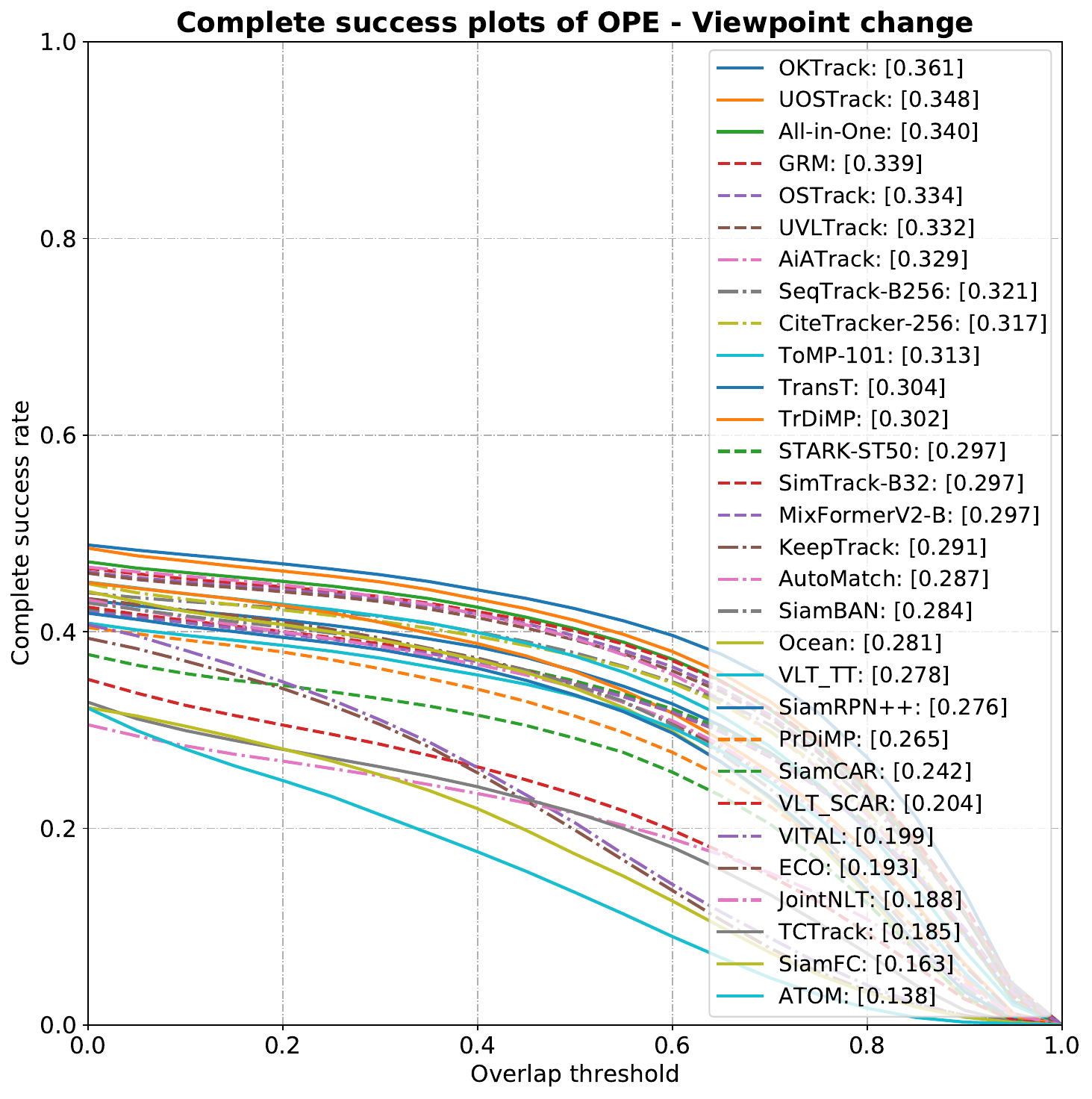}}
\subfloat{\includegraphics[width =0.25\columnwidth]{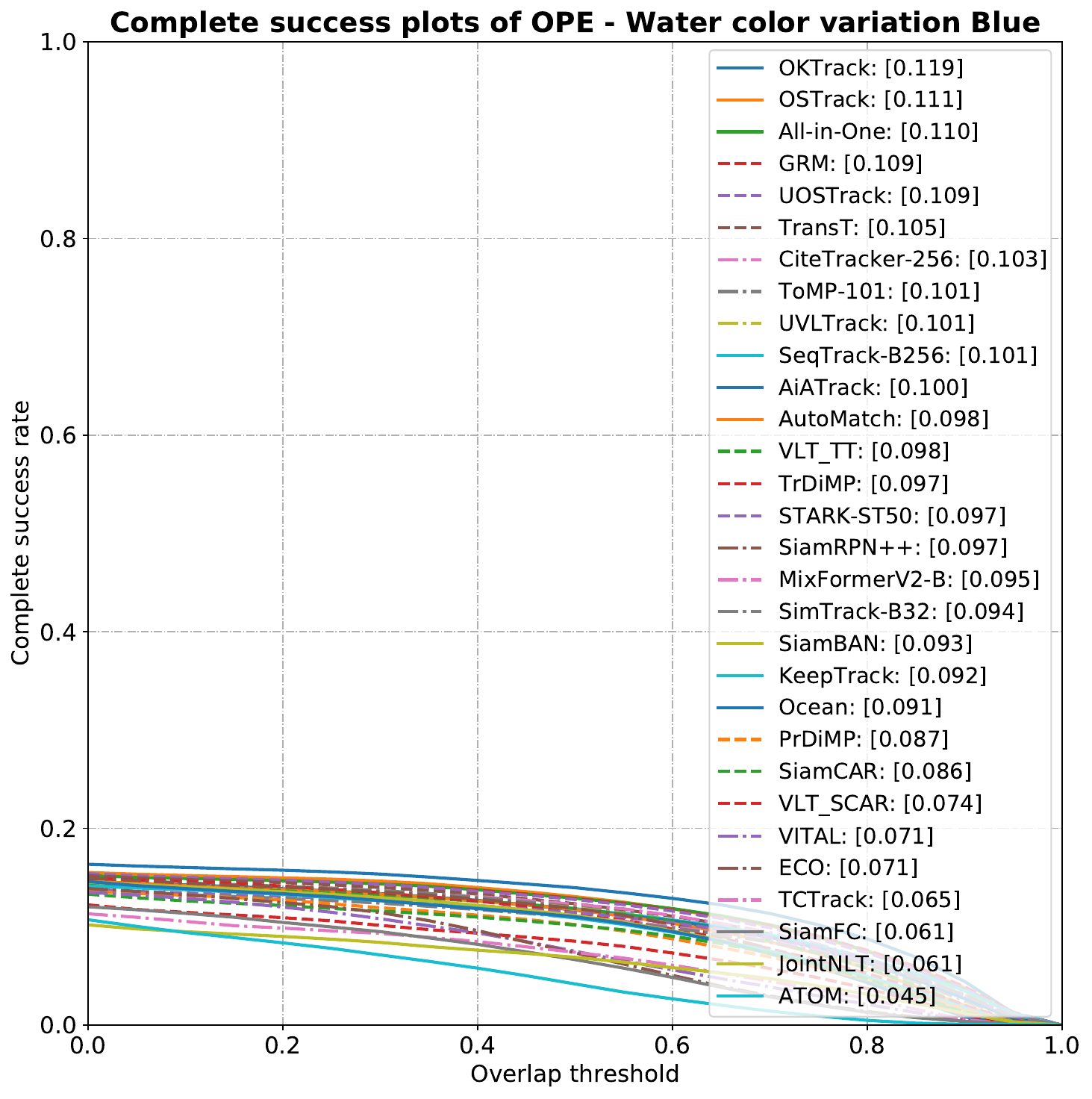}}
\subfloat{\includegraphics[width =0.25\columnwidth]{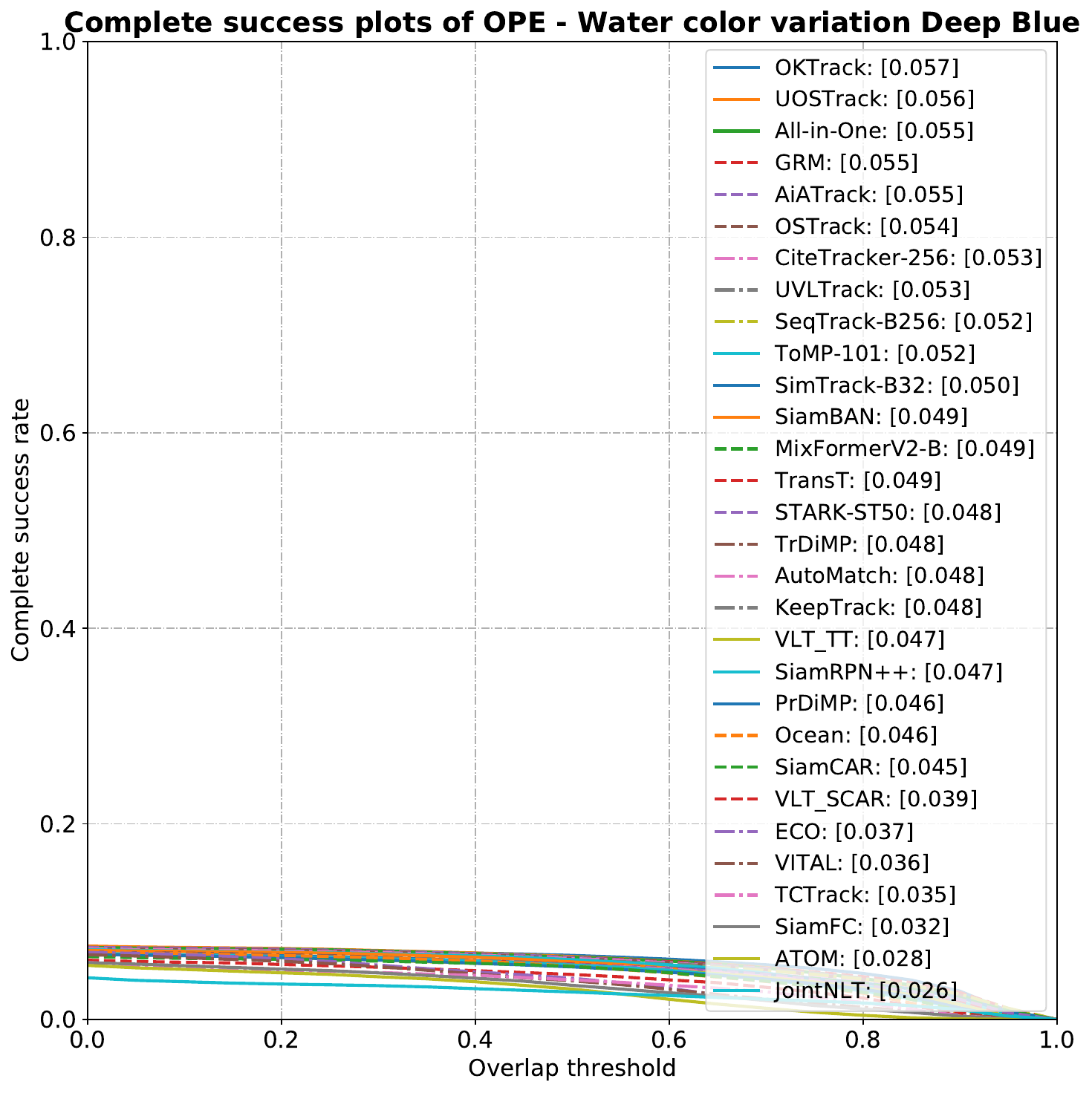}}

  \caption{Performances of baseline trackers on the WebOUT-1M test set of different attributes using \textbf{cAUC} scores. Best viewed by zooming in.}
  \label{fig:attribute_results_cAUC}
\end{figure*}

We present more attribute-based results on the WebOUT-1M test set. Fig.~\ref{fig:attribute_results_AUC} shows the performance of 30 deep trackers on the WebOUT-1M test set of different attributes using \textbf{AUC} scores. Fig.~\ref{fig:attribute_results_Pre} shows the performance of 30 deep trackers on the WebOUT-1M test set of different attributes using \textbf{Pre} scores. Fig.~\ref{fig:attribute_results_nPre} shows the performance of 30 deep trackers on the WebOUT-1M test set of different attributes using \textbf{nPre} scores. Fig.~\ref{fig:attribute_results_cAUC} shows the performance of 30 deep trackers on the WebOUT-1M test set of different attributes using \textbf{cAUC} scores.

\section{Datasheet}
\label{sec:Datasheet}

Following~\cite{gebru2021datasheets}, we provide the datasheet for our WebUOT-1M dataset:

\subsection{Motivation}

\begin{enumerate}[(1)]
    \item \textbf{For what purpose was the dataset created?} Was there a specific task in mind? Was there a specific gap that needed to be filled? Please provide a description.

    A1: The WebUOT-1M dataset was created to facilitate the development and evaluation of video-based underwater object tracking. 1) Previous UOT datasets suffer from limitations in scale, diversity of target categories, and scenarios covered, hindering the training and evaluation of modern tracking algorithms. 2) Existing UOT datasets only provide bounding box annotations, which do not support multi-modal underwater object tracking. 3) How to effectively transfer knowledge from large-scale open-air data to underwater tracking models has yet to be explored. To fill these gaps, we propose WebUOT-1M, \ie, the largest and most diverse underwater tracking dataset in terms of target categories and underwater scenarios. Based on the established WebUOT-1M dataset, we further propose a simple yet effective omni-knowledge distillation tracking framework, called OKTrack, for the community. We believe that WebUOT-1M can contribute a valuable benchmark to the community for developing more general tracking models for UOT and broader fields.
    
    \item \textbf{Who created the dataset (\eg, which team, research group) and on behalf of which entity (\eg, company, institution, organization)?}

    A2: This dataset was created by Chunhui Zhang, Li Liu, Guanjie Huang, Hao Wen, Xi Zhou, and Yanfeng Wang. Chunhui Zhang and Yanfeng Wang are from Shanghai Jiao Tong University, Li Liu and Guanjie Huang are from the Hong Kong University of Science and Technology (Guangzhou), and Hao Wen and Xi Zhou are from CloudWalk Technology. At the time of creation, Chunhui Zhang was a visiting Ph.D. student at the Hong Kong University of Science and Technology (Guangzhou).

    \item \textbf{Who funded the creation of the dataset?} If there is an associated grant, please provide the name of the grantor and the grant name and number.

    A3: The project was funded by the National Natural Science Foundation of China (No. 62101351), and the Key Research and Development Program of Chongqing (cstc2021jscx-gksbX0032).

    \item \textbf{Any other comments?}

    A4: None.
    
\end{enumerate}

\subsection{Composition}

\begin{enumerate}[(1)]
    \item \textbf{What do the instances that comprise the dataset represent (\eg, documents, photos, people, countries)? } Are there multiple types of instances (\eg, movies, users, and ratings; people and interactions between them; nodes and edges)? Please provide a description.

    A1: WebUOT-1M comprises 1.1 million frames with precise bounding box annotations across 1,500 underwater videos and 408 highly diverse target categories. These targets are further classed into 12 superclasses with reference to WordNet to facilitate the evaluation of the cross-superclass generalization ability of tracking models. We annotate the dataset with 23 tracking attributes and annotate a language prompt for each video.

    \item \textbf{How many instances are there in total (of each type, if appropriate)?}

    A2: The dataset consists of 1,500 videos, totaling 1.1 million frames, and 10.5 hours.

    \item \textbf{Does the dataset contain all possible instances or is it a sample (not necessarily random) of instances from a larger set?} If the dataset is a sample, then what is the larger set? Is the sample representative of the larger set (\eg, geographic coverage)? If so, please describe how this representativeness was validated/verified. If it is not representative of the larger set, please describe why not (\eg, to cover a more diverse range of instances, because instances were withheld or unavailable).

    A3: Despite our best efforts to collect as many target categories as possible, due to the vast diversity of underwater targets in the real world, we were unable to include all underwater targets in a single dataset. We will continue to increase the diversity and volume of WebUOT-1M in future work.
    
    \item \textbf{What data does each instance consist of?} “Raw” data (\eg, unprocessed text or images) or features? In either case, please provide a description.

    A4: Each instance includes diverse annotations (underwater images, bounding boxes, absent labels, language prompt, category name, and superclass name).
    
    \item \textbf{Is there a label or target associated with each instance?} If so, please provide a description. 

    A5: Yes. See I.2 (1)-(4).

    \item \textbf{Is any information missing from individual instances?} If so, please provide a description, explaining why this information is missing (\eg, because it was unavailable). This does not include intentionally removed information, but might include, \eg, redacted text.

    A6: No.

    \item \textbf{Are relationships between individual instances made explicit
(\eg, users’ movie ratings, social network links)?} If so, please describe how these relationships are made explicit.

    A7: Yes. We only annotate one instance per video and use a bounding box to represent their motion trajectory.

    \item \textbf{Are there recommended data splits (\eg, training, development/validation,
testing)?} If so, please provide a description of these splits, explaining
the rationale behind them.

    A8: The dataset is divided into a training set and a test set. Please refer to appendix~\ref{sec:MoreStatisticsaboutWebUOT1M} for details.

    \item \textbf{Are there any errors, sources of noise, or redundancies in the dataset?} If so, please provide a description.

    A9: Despite our multiple rounds of careful annotation checks, there may still be some inaccuracies due to occlusion or blurring caused by the movement of underwater targets, such as slight shifts of bounding boxes. Manually annotated language prompts might not adequately describe the movement and appearance changes of targets over long video sequences. In the future, we plan to use large multi-modal models to further improve the accuracy and scientific quality of the language annotations.

    \item \textbf{Is the dataset self-contained, or does it link to or otherwise rely on external resources (\eg, websites, tweets, other datasets)?} If it links to or relies on external resources, a) are there guarantees that they will exist, and remain constant, over time; b) are there official archival versions of the complete dataset (\ie, including the external resources as they existed at the time the dataset was created); c) are there any restrictions (\eg, licenses, fees) associated with any of the external resources that might apply to a dataset consumer? Please provide descriptions of all external resources and any restrictions associated with them, as well as links or other access points, as appropriate.

    A10: Yes, the dataset is self-contained.

    \item \textbf{Does the dataset contain data that might be considered confidential (\eg, data that is protected by legal privilege or by doctor–patient confidentiality, data that includes the content of individuals’ non-public communications)?} If so, please provide a description.

    A11: No.

    \item \textbf{Does the dataset contain data that, if viewed directly, might be offensive, insulting, threatening, or might otherwise cause anxiety?} If so, please describe why.

    A12: No.

\end{enumerate}

\subsection{Collection Process}

\begin{enumerate}[(1)]
    \item \textbf{How was the data associated with each instance acquired?} Was the data directly observable (\eg, raw text, movie ratings), reported by subjects (\eg, survey responses), or indirectly inferred/derived from other data (\eg, part-of-speech tags, model-based guesses for age or language)? If the data was reported by subjects or indirectly inferred/derived from other data, was the data validated/verified? If so, please describe how.

    A1: We assembled a professional annotation team from a qualified data labeling company. The author team conducted the last data verification to ensure high-quality annotations. Specifically, in each frame of the video, the visual bounding box $[x, y, w, h]$ is used as the ground truth for the target, where $(x, y)$, $w$, and $h$ represent the target's top-left corner, width, and height, respectively. A sentence of language prompt describing the color, behavior, attributes, and surroundings of the target is given for each video sequence to encourage the exploration of multi-modal UOT.

    \item \textbf{What mechanisms or procedures were used to collect the data (\eg, hardware apparatuses or sensors, manual human curation, software programs, software APIs)?} How were these mechanisms or procedures validated?

    A2: Most of the video clips are collected from YouTube and BiliBili with careful filtering. We manually selected videos suitable for underwater object tracking and randomly chose targets to increase the diversity of our dataset. A professional data annotation team conducted multiple rounds of manual annotations, and the author team performed the final data verification to ensure high-quality annotations.

    \item \textbf{If the dataset is a sample from a larger set, what was the sampling strategy (\eg, deterministic, probabilistic with specific sampling probabilities)?}

    A3: N/A.

    \item \textbf{Who was involved in the data collection process (\eg, students, crowdworkers, contractors) and how were they compensated (\eg, how much were crowdworkers paid)?}

    A4: The authors of the paper.
    
    \item \textbf{Over what timeframe was the data collected?} Does this timeframe match the creation timeframe of the data associated with the instances (\eg, recent crawl of old news articles)? If not, please describe the timeframe in which the data associated with the instances was created.

    A5: Collecting the data took about one month, and completing the data cleaning, organization, annotation, and verification took approximately six months.

    \item \textbf{Were any ethical review processes conducted (\eg, by an institutional review board)?} If so, please provide a description of these review processes, including the outcomes, as well as a link or other access point to any supporting documentation.

    A6: N/A.
    
\end{enumerate}

\subsection{Preprocessing/cleaning/labeling}

\begin{enumerate}[(1)]
    \item \textbf{Was any preprocessing/cleaning/labeling of the data done (\eg, discretization or bucketing, tokenization, part-of-speech tagging, SIFT feature extraction, removal of instances, processing of missing values)?} If so, please provide a description. If not, you may skip the remaining questions in this section.

    A1: We manually collected and cleaned data from YouTube and BiliBili to ensure high-quality videos in WebUOT-1M. We discarded videos that are not suitable for tracking, such as repeated scenes, long-term static targets, and incomplete trajectories.

    \item \textbf{Was the “raw” data saved in addition to the preprocessed/cleaned/labeled data (\eg, to support unanticipated future uses)?} If so, please provide a link or other access point to the “raw” data.

    A2: No, we only provide the community with cleaned and annotated video sequences.
    
    \item \textbf{Is the software that was used to preprocess/clean/label the data available?} If so, please provide a link or other access point.


    A3: We use the standard Python library Beautiful Soup\footnote{https://beautiful-soup-4.readthedocs.io/en/latest/} to crawl videos from online websites. The dataset is manually annotated using an in-house annotation tool of the data labeling company.

    \item \textbf{Any other comments?} 

    A4: None.

\end{enumerate}

\subsection{Uses}

\begin{enumerate}[(1)]
    \item \textbf{Has the dataset been used for any tasks already?} If so, please provide a description.

    A1: No, this dataset is newly proposed.
    
    \item \textbf{Is there a repository that links to any or all papers or systems that use the dataset?}  If so, please provide a link or other access point.

    A2: N/A.

    \item \textbf{What (other) tasks could the dataset be used for?}

    A3: WebUOT-1M can be used for underwater object tracking and underwater vision-language tracking. Additionally, it can be utilized for underwater vision understanding, marine environmental monitoring, and marine animal conservation.
    
    \item \textbf{Is there anything about the composition of the dataset or the way it was collected and preprocessed/cleaned/labeled that might impact future uses?} For example, is there anything that a dataset consumer might need to know to avoid uses that could result in unfair treatment of individuals or groups (\eg, stereotyping, quality of service issues) or other risks or harms (\eg, legal risks, financial harms)? If so, please provide a description. Is there anything a dataset consumer could do to mitigate these risks or harms?

    A4: No.

    \item \textbf{Are there tasks for which the dataset should not be used?} If so, please provide a description.

    A5: No.

    \item \textbf{Any other comments?}

    A6: None.
\end{enumerate}

\subsection{Distribution}

\begin{enumerate}[(1)]
    \item \textbf{Will the dataset be distributed to third parties outside of the entity (\eg, company, institution, organization) on behalf of which the dataset was created?} If so, please provide a description.

    A1: Yes, the WebUOT-1M dataset will be made publicly available to the community.
    
    \item \textbf{How will the dataset will be distributed (\eg, tarball on website, API, GitHub)?} Does the dataset have a digital object identifier (DOI)?

    
    A2: The WebUOT-1M dataset will be publicly released on the GitHub project\footnote{https://github.com/983632847/Awesome-Multimodal-Object-Tracking}.
    
    \item \textbf{When will the dataset be distributed?}

    A3: The WebUOT-1M dataset will be distributed once the paper is accepted after peer review.
    
    \item \textbf{Will the dataset be distributed under a copyright or other intellectual property (IP) license, and/or under applicable terms of use (ToU)?} If so, please describe this license and/or ToU, and provide a link or other access point to, or otherwise reproduce, any relevant licensing terms or ToU, as well as any fees associated with these restrictions.

    A4: We release our dataset and benchmark under Creative Commons licenses\footnote{https://creativecommons.org/licenses/}.

    \item \textbf{Have any third parties imposed IP-based or other restrictions on the data associated with the instances?} If so, please describe these restrictions, and provide a link or other access point to, or otherwise reproduce, any relevant licensing terms, as well as any fees associated with these restrictions.

    A5: No.
    
    \item \textbf{Do any export controls or other regulatory restrictions apply to the dataset or to individual instances?} If so, please describe these restrictions, and provide a link or other access point to, or otherwise reproduce, any supporting documentation.

    A6: No.

    \item \textbf{Any other comments?}

    A7: None.

\end{enumerate}

\subsection{Maintenance}

\begin{enumerate}[(1)]
    \item \textbf{Who will be supporting/hosting/maintaining the dataset?}

    A1: The authors of the paper.

    \item \textbf{How can the owner/curator/manager of the dataset be contacted (\eg, email address)?}

    A2: You can contact them via email on the GitHub project.

    \item \textbf{Is there an erratum? If so, please provide a link or other access point.}

    A3: No.
    
    \item \textbf{Will the dataset be updated (\eg, to correct labeling errors, add new instances, delete instances)?} If so, please describe how often, by whom, and how updates will be communicated to dataset consumers (\eg, mailing list, GitHub)?

    A4: To ensure the accuracy of the dataset, if the author identifies any errors or other researchers notify us of labeling errors in the data, we will promptly review and update the dataset.

    \item \textbf{If the dataset relates to people, are there applicable limits on the retention of the data associated with the instances (\eg, were the individuals in question told that their data would be retained for a fixed period of time and then deleted)?} If so, please describe these limits and explain how they will be enforced.

    A5: N/A.

    \item \textbf{Will older versions of the dataset continue to be supported/hosted/maintained?} If so, please describe how. If not, please describe how its obsolescence will be communicated to dataset consumers.

    A6: No, we maintain the latest version of this dataset for the community on our GitHub project.
    
    \item \textbf{If others want to extend/augment/build on/contribute to the dataset, is there a mechanism for them to do so?} If so, please provide a description. Will these contributions be validated/verified? If so, please describe how. If not, why not? Is there a process for communicating/distributing these contributions to dataset consumers? If so, please provide a description.
    
    A7: N/A. \textbf{\color{magenta}All researchers are welcome to collaboratively develop and extend WebUOT-1M.}

    \item \textbf{Any other comments?}

    A8: None.

\end{enumerate}

\end{document}